%% file: main_iclr.tex
\documentclass[nohyperref]{article}

\usepackage{microtype}
\usepackage{graphicx}
\usepackage{subfigure}
\usepackage{booktabs} % for professional tables
\usepackage{adjustbox}

\usepackage{algorithm, algorithmic}

\usepackage{iclr2024_conference,times}
\usepackage[utf8]{inputenc} % allow utf-8 input
\usepackage[T1]{fontenc}    % use 8-bit T1 fonts
\usepackage[pagebackref,breaklinks,colorlinks]{hyperref}

\usepackage{graphicx}
\usepackage{amsmath}
\usepackage{amssymb}
\usepackage{booktabs}
\usepackage[utf8]{inputenc} 

\usepackage{adjustbox}

\usepackage[nameinlink]{cleveref}
\crefname{section}{Sec.}{Secs.}
\Crefname{section}{Section}{Sections}
\Crefname{table}{Table}{Tables}
\crefname{table}{Tab.}{Tabs.}

\usepackage{pifont}

\usepackage{url}

\usepackage{tikz}
\usepackage{comment}
\usepackage{amsfonts}       % blackboard math symbols
\usepackage{color}

\usepackage{amsmath}
\usepackage{amssymb}
\usepackage{bbm, dsfont}
\usepackage{mathtools}
\usepackage{amsthm}
\usepackage[normalem]{ulem}
\usepackage{xcolor,colortbl}
\usepackage{multirow}
\usepackage{cutwin}
\usepackage{arydshln}
\usepackage{enumitem}
\usepackage{wrapfig}
\usepackage{microtype}
\usepackage{graphicx}
\usepackage{bm}
\usepackage{eqparbox}
\usepackage{makecell}
\usepackage{threeparttable}

\newcommand{\eat}[1]{}

\newcommand{\newor}{%
  \mathbin{%
    {\vee}\mspace{-2.9mu}
  }%
}

\definecolor{Red}{rgb}{0.6,0,0}
\definecolor{Blue}{rgb}{0,0,0.8}
\definecolor{Green}{rgb}{0,0.7,0.3}
\definecolor{airforceblue}{rgb}{0.36, 0.54, 0.66}
\definecolor{ao(english)}{rgb}{0.0, 0.5, 0.0}
\definecolor{azure(colorwheel)}{rgb}{0.0, 0.5, 1.0}
\definecolor{crimson}{rgb}{0.86, 0.08, 0.24}
\definecolor{darkcerulean}{rgb}{0.03, 0.27, 0.49}
\definecolor{cobalt}{rgb}{0.0, 0.28, 0.67}
\definecolor{rosegold}{rgb}{0.72, 0.43, 0.47}
\definecolor{orange-red}{rgb}{1.0, 0.27, 0.0}
\definecolor{mountainmeadow}{rgb}{0.19, 0.73, 0.56}
\definecolor{malachite}{rgb}{0.04, 0.85, 0.32}
\definecolor{darkblue}{rgb}{0.0, 0.0, 0.55}
\definecolor{customblue}{rgb}{0.2, 0.35, 0.8}
\definecolor{gg}{gray}{0.92}
\newcolumntype{a}{>{\columncolor{gg}}c}

\definecolor{gg}{gray}{0.9}
\hypersetup{colorlinks=true}
\hypersetup{linktoc=all}
\hypersetup{citecolor=customblue}
\hypersetup{linkcolor=crimson}
\hypersetup{urlcolor=darkblue}
\usepackage[all]{hypcap}

\creflabelformat{equation}{#2\textup{#1}#3}  

\newcommand{\bsy}{\boldsymbol}
\usepackage{svg}
\DeclareMathOperator*{\minimize}{minimize}

\title{Progressive Fourier Neural Representation \\ for Sequential Video Compilation}

\author{Haeyong Kang,   \;
         Jaehong Yoon, \;
         DaHyun Kim, \;
        Sung Ju Hwang, \;
        and Chang D Yoo\thanks{Corresponding author} \\
         Korea Advanced Institute of Science and Technology (KAIST) \\
         {\{haeyong.kang, jaehong.yoon, dahyun.kim, sjhwang82, cdyoo\}@kaist.ac.kr}
}

\iclrfinalcopy

\begin{document}

\maketitle

\input{1_abstract}

\input{2_intro}

\input{3_related_work}

\input{4_approach}

\input{5_experiment}
\input{6_conclusion}

\bibliography{reference}
\bibliographystyle{iclr2024_conference}

%%%%%%%%%%%%%%%%%%%%%%%%%%%%%%%%%%%%%%%%%%%%%%%%%%%%%%%%%%%%

\appendix
\input{7_appendix}

%\bibliography{reference}
%\bibliographystyle{iclr2024_conference}

%Optionally include extra information (complete proofs, additional experiments and plots) in the appendix.
%This section will often be part of the supplemental material.

\end{document}

%% file: 1_abstract.tex
\begin{abstract}
Neural Implicit Representation (NIR) has recently gained significant attention due to its remarkable ability to encode complex and high-dimensional data into representation space and easily reconstruct it through a trainable mapping function. However, NIR methods assume a one-to-one mapping between the target data and representation models regardless of data relevancy or similarity. This results in poor generalization over multiple complex data and limits their efficiency and scalability. Motivated by continual learning, this work investigates how to accumulate and transfer neural implicit representations for multiple complex video data over sequential encoding sessions. To overcome the limitation of NIR, we propose a novel method, \textit{Progressive Fourier Neural Representation (PFNR)}, that aims to find an adaptive and compact sub-module in Fourier space to encode videos in each training session. This sparsified neural encoding allows the neural network to hold free weights, enabling an improved adaptation for future videos. In addition, when learning a representation for a new video, PFNR transfers the representation of previous videos with frozen weights. This design allows the model to continuously accumulate high-quality neural representations for multiple videos while ensuring lossless decoding that perfectly preserves the learned representations for previous videos. We validate our PFNR method on the UVG8/17 and DAVIS50 video sequence benchmarks and achieve impressive performance gains over strong continual learning baselines. The PFNR code is available at \url{https://github.com/ihaeyong/PFNR.git}.
\end{abstract}

%% file: 2_intro.tex
\section{Introduction}
Neural Implicit Representation (NIR)~\citep{chen2021nerv, li2022nerv, chen2023hnerv, mehta2021modulated} is a research field that aims to represent complex data, such as videos or 3D objects, as continuous functions learned by neural networks. Instead of explicitly describing data points, NIR models compress high-dimensional data from a low-dimensional embedding space. This process enables efficient data storage, compression, and synthesis. However, there's a challenge: when compressing multiple pieces of data, each high-dimensional data needs to be encoded in the neural network, increasing linear memory requirements. To address this, Neural Video Representation proposes a solution, as explored in studies by \cite{chen2021nerv,chen2022cnerv}. This approach combines different videos into a single video format and then reduces the model size through techniques like weight pruning and quantization post-training. While this method is effective for current data compression, it has a significant limitation: it restricts the model's ability to adapt to new videos as they are added. To overcome this limitation, inspired by incremental knowledge transfer and expansion in continual learning, we investigate a practical implicit representation learning scenario with video data, which aims to accumulate neural implicit representations for multiple videos into a single model under the condition that videos are incoming sequentially.

Continual Learning (CL)~\citep{ThrunS1995, rusu2016progressive, zenke2017continual, hassabis2017neuroscience} is a learning paradigm where a model learns over multiple sequential sessions. It seeks to mimic human cognition, characterized by the ability to learn new concepts incrementally throughout a lifetime without the degeneration of previously acquired functionality. Yet, incremental training of NIR is a challenging problem since the model detrimentally loses the learned implicit representations of past session videos while encoding newly arrived ones, a phenomenon known as \textit{catastrophic forgetting}~\citep{McCloskey1989}. This issue particularly matters as neural representation methods for videos encode and reconstruct the target data stream conditioned to its frame indices. Then, the model more easily ruins its generation ability while learning to continuously encode new videos due to the distributional disparities in \textit{holistic videos} and their \textit{individual frames}. Furthermore, the \textit{compression phase} of neural representation makes it wayward to transfer the model to future tasks. Various approaches have been proposed to address catastrophic forgetting during continual learning, which are often conventionally classified as follows: (1) \emph{Regularization-based methods}~\citep{Kirkpatrick2017, chaudhry2020continual, Jung2020, titsias2019functional, mirzadeh2020linear} aim to keep the learned information of past sessions during continual training aided by sophisticatedly designed regularization terms, (2) \emph{Architecture-based methods}~\citep{YoonJ2018iclr, mallya2018piggyback, Serra2018, wortsman2020supermasks, kang2022forget, kang2022soft} propose to minimize the inter-task interference via newly designed architectural components, and (3) \emph{Rehearsal-based methods}~\citep{rebuffi2017icarl, chaudhry2019continual, Saha2021, yoon2022online, sarfraz2023error} involves replaying real or synthesized data from previous sessions. However, these methods are less suitable for video data in CL due to the substantial memory and computational costs required to store and revisit high-dimensional samples. While conventional architecture-based methods offer solutions to prevent forgetting, they are unsuited for sequential complex video processing as they reuse a few or all adaptive parameters without finely discretized operations.

To enhance neural representation incrementally on complex sequential videos, we propose a novel sequential video compilation method, coined \textbf{P}rogressive \textbf{F}ourier \textbf{N}eural \textbf{R}epresentation (\textbf{PFNR}) to identify and utilize Lottery tickets (i.e., the weights of complicated oscillatory signals) in frequency space. To achieve this, we define \textbf{F}ourier \textbf{S}ubnetwork \textbf{O}perator (\textbf{FSO}), which breaks down a neural implicit representation into its sine and cosine components (real and imaginary parts) and then selectively, identifies the most effective \emph{Lottery tickets} for representing complex periodic signals. In practice, given a backbone and FSO architecture, our method continuously learns to identify input-adaptive subnetwork modules and encode each new video into the corresponding module during sequential training sessions. Our approach draws inspiration from the \emph{Lottery Ticket Hypothesis (LTH)}~\citep{frankle2018lottery}, which suggests that sparse subnetworks can maintain the performance of a dense network and from the  \emph{Fourier Neural Operator} concepts developed in studies by ~\cite{li2020fourier, li2020neural, kovachki2021neural, tran2021factorized}. A challenge in this domain is the inefficiency of continually searching for optimal subnetworks in Fourier space. This process typically requires iterative training steps with repeated pruning and retraining for each new task. To address this, PFNR introduces a parametric score function. This function learns to produce binary masks for the real and imaginary components, enabling the identification of adaptive substructures for video encoding in each training session by selecting the top-percentage weights based on their ranking scores. This allows PFNR to discover the optimal subnetwork during training, through joint training of weights and structure, thus avoiding the laborious processes of iterative retraining, pruning, and rewinding inherent in \emph{LTH}. Crucially, PFNR permits overlapping subnetworks with those from previous sessions during training. This overlap allows the transfer of learned representations from earlier videos when relevant while keeping the weights for these earlier sessions fixed. As a result, our model can continuously expand its representation space across successive video sessions, ensuring that it maintains the encoding and generation quality of previous videos without any degradation (i.e., remaining forgetting-free). This is achieved without needing a replay buffer to store multiple high-dimensional frames, a significant advancement in the field.

%However, searching for optimal subnetworks in Fourier space during continual learning is also inefficient since they require iterative training steps with repetitive pruning and retraining in each arriving task. To this end, our proposed PFNR introduces a parametric score function that learns to generate binary masks of real and imaginary parts to find adaptive substructures for video encoding in each training session by directly choosing the top-$\%$ percent in weight ranking scores. We emphasize that PFNR can discover the optimal subnetwork online through joint training of the weights and structure and bypass arduous procedures in LTH, such as iterative retraining, pruning, and rewinding. Our PFNR allows overlapping subnetworks with previous sessions during training to transfer the learned representation of previous videos when relevant but keeps the weights for previous video sessions frozen. Consequently, we enable the model to expand its representation space throughout consecutive video sessions continuously, ensuring maintaining the encoding and generation quality of the previous video intact (i.e., forgetting-free) even without resorting to a replay buffer to store multiple high-dimensional frames.

\noindent
Our contributions can be summarized as follows:
\begin{itemize}[leftmargin=*]
    \item We suggest a practical learning scenario for neural implicit representation where the model encodes multiple videos continually in successive training sessions. Earlier NIR methods suffered from poor transferability to new videos due to the distributional shift of holistic video and frames.
    
    \item We propose a cutting-edge method referred to as the Progressive Fourier Neural Representation for a complex sequential video compilation. The proposed method continuously learns a compact subnetwork for each video session given a supernet backbone while preserving the generative quality of previous videos flawlessly in Fourier space.
    
    \item We demonstrate the effectiveness of our method on multiple sequential video sessions by achieving superior performance over conventional baselines in average PSNR and MS-SSIM without any quantitative or qualitative degeneration in reconstructing previously encoded videos during sequential video compilation.
\end{itemize}

%% file: 3_related_work.tex
\section{Related Works}

\paragraph{Neural Implicit Representation (NIR).} Neural Implicit Representations (NIR)~\citep{mehta2021modulated} are neural network architectures for parameterizing continuous, differentiable signals. Based on coordinate information, they provide a way to represent complex, high-dimensional data with a small set of learnable parameters that can be used for various tasks such as image reconstruction~\citep{sitzmann2020implicit, tancik2020fourier}, shape regression~\citep{chen2019learning, park2019deepsdf}, and 3D view synthesis ~\citep{mildenhall2021nerf, schwarz2020graf}. Instead of using coordinate-based methods, NeRV~\citep{chen2021nerv} proposes an image-wise implicit representation that takes frame indices as inputs, enabling fast and accurate video compression. NeRV has inspired further improvements in video regression by CNeRV~\citep{chen2022cnerv}, DNeRV~\citep{he2023towards}, E-NeRV~\citep{li2022nerv}, and NIRVANA~\citep{maiya2022nirvana}, and HNeRV~\citep{chen2023hnerv}. A few recent works have explored video continual learning (VCL) scenarios for the NIR. To tackle non-physical environments, Continual Predictive Learning (CPL)~\citep{chen2022continual} learns a mixture world model via predictive experience replay and performs test-time adaptation using non-parametric task inference. PIVOT~\citep{villa2022pivot} leverages the past knowledge present in pre-trained models from the image domain to reduce the number of trainable parameters and mitigate forgetting. CPL needs memory to replay, while PIVOT needs pre-training and fine-tuning steps. In contrast, along with the conventional progressive training techniques \citep{rusu2016progressive, cho2022streamable}, we introduce a novel neural video representation referred to as \textit{"Progressive Fourier Neural Representation (PFNR)"}, which utilizes the Lottery Ticket Hypothesis (LTH) to identify an adaptive substructure within the dense networks that are tailored to the specific video input index. Our PFNR doesn't use memory, a pre-trained model, or fine-tuning for a sequential video representation compilation.

\paragraph{Continual Learning.} Most continual learning approaches introduce extra memory like additional model capacity~\citep{li2019learn, yoon2020} or a replay buffer~\citep{riemer2018learning, chaudhry2018efficient, buzzega2020dark, arani2022learning, sarfraz2023error}. However, several works have focused on building memory-efficient continual learners using pruning-based constraints to exploit initial model capability more compactly. CLNP~\citep{golkar2019continual} selects important neurons for a given task using $\ell_1$ regularization to induce sparsity and freezes them to maintain performance. And pruned neurons are reinitialized for future task training. Piggyback \citep{mallya2018piggyback} trains task-specific binary masks on the weights given a pre-trained model. However, it does not allow for knowledge transfer among tasks, so the performance highly depends on the quality of the backbone model. HAT~\citep{Serra2018} proposes task-specific learnable attention vectors to identify significant weights per task. The masks are formulated to layerwise cumulative attention vectors during continual learning. LL-Tickets~\citep{Chen2021lifelonglottery} recently suggests sparse subnetworks called lifelong tickets that perform well on all tasks during continual learning. The method searches for more prominent tickets from current ones if the obtained tickets cannot sufficiently learn the new task while maintaining performance on past tasks. However, LL-Tickets require external data to maximize knowledge distillation with learned models for prior tasks, and the ticket expansion process involves retraining and pruning steps. As a strong architecture-based baseline, WSN~\citep{kang2022forget} jointly learns the model weights and task-adaptive binary masks during continual learning. It prevents catastrophic forgetting of previous tasks by keeping the model weights selected, called winning tickets, intact at the end of each training. However, WSN is inappropriate for sequential complex video compilation since it reuses a few adaptive but sparse learnable parameters. To overcome the weakness of WSN, our PFNR explores more appropriate forget-free weights for representing complex video in Fourier space \citep{li2020fourier, li2020neural, kovachki2021neural, tran2021factorized} using a newly proposed Fourier Subnetwork Operator (FSO).

%% file: 4_approach.tex
\input{materials/outline}

\section{Progressive Fourier Neural Representation}
This section presents our proposed continual neural implicit representation method, named \textit{Progressive Fourier Neural Representation (PFNR)}. Given a supernet backbone, where we follow the NeRV~\citep{chen2021nerv} architecture for video embedding and decoding, PFNR aims to expand its representation space continuously by sequentially encoding multiple videos within the Fourier space. As new videos arrive in the model, PFNR jointly updates the binary masks (including real and imaginary parts) with neural network weights, searching for the adaptive subnetwork to encode given videos. Once a video session is completed, we 'freeze' the weights of the chosen subnetwork. This approach ensures that the quality of previously learned representations and generated outputs remains unaffected by future training sessions, even if the new subnetwork structure shares some weights with videos encoded earlier. While the weights learned in earlier video sessions are frozen, we enable our PFNR to transfer prior knowledge to future video tasks (i.e., forward transfer). This makes the model adapt new videos effectively by leveraging the representation of past videos (Please see \Cref{fig:concept_CVRNet}).
%After training each video session, we freeze the weights of the selected subnetwork so that future training does not hurt the quality of the learned representation and the generated output, even though the new subnetwork structure contains some weights already encoded in the previous video. 

% \noindent 
\textbf{Problem Statement.} 
Let a video at $s_{th}$ session $\bm{V}_s =\{\bm{v}^s_t\}^{T_s}_{t=1} \in \mathbb{R}^{T_s \times H \times W \times 3}$ be represented by a function with the trainable parameter $\bm\theta$, $f_{\bm\theta}: \mathbb{R} \rightarrow \mathbb{R}^{H \times W \times 3}$, during Video Continual Learning (VCL), where $T_s$ denotes the number of frames in a video at session $s$, and $s \in \{1 \dots, |S|\}$. Given a session and frame index $s$ and $t$, respectively, the neural implicit representation aims to predict a corresponding RGB image $\bm{v}^s_t \in \mathbb{R}^{H \times W \times 3}$ by fitting an encoding function to a neural network: $\bm{v}^s_t = f_{\bm\theta}([s;t], H_s)$ where $H_s$ is $s_{th}$ head. For the sake of simplicity, we omit $H_s$ in the following equations. Let's consider a real-world learning scenario in which $|\mathcal{S}|=N$ or more sessions arrive in the model sequentially. We denote that $\mathcal{D}_s=\{\bm{e}_{s,t}, \bm{v}_{s,t}\}_{t=1}^{T_s}$ is the dataset of session $s$, composed of $T_s$ pairs of raw embeddings $\bm{e}_{s,t} = \left[\bm{e}_s; \bm{e}_t\right] \in \mathbb{R}^{1 \times 160}$ and corresponding frames $\bm{v}^s_t$. Here, we assume that $\mathcal{D}_s$ for session $s$ is only accessible when learning session $s$ due to the limited hardware memory and privacy-preserving issues, and session identity is given in the training and testing stages. The primary training objective in this sequence of $N$ video sessions is to minimize the following optimization problem:
\begin{equation}
\bm{\theta}^{\ast}=\minimize_{\bsy\theta} \frac{1}{N} \frac{1}{T_s}\sum^{N}_{s=1}\sum^{T_s}_{t=1}\mathcal{L}(f(\bm{e}_{s,t};\bsy\theta), \bm{v}^s_t),
\label{eq:sess_loss}
\end{equation}
where the loss function $\mathcal{L}(\bm{v}^s_t)$ is composed of $\ell_1$ loss and \textit{SSIM loss}. The former minimizes the pixel-wise RGB gap with the original input frames evenly, and the latter maximizes the similarity between the two entire frames based on luminance, contrast, and structure, as follows: 
\begin{equation}
\mathcal{L}(\bm{V}_s) = {1\over T_s} \sum_{t=1}^{T_s} \alpha ||\bm{v}^s_t-\hat{\bm{v}}^s_t||_1 + (1-\alpha)(1-\textbf{SSIM}(\bm{v}^s_t, \hat{\bm{v}}^s_t)),
\label{eq:l1_loss}
\end{equation} 
where %T_s$ is the number of frames for session $s$, 
%$\bm{v}^s_t$ is the $s_{th}$ session ground truth frame, and
$\hat{\bm{v}}^s_t$ is the output generated by the model $f$. For all experiments, we set the hyperparameter $\alpha$ to $0.7$, and we adapt PixelShuffle~\citep{shi2016real} for session and time positional embedding. 

Continual learners $f$ frequently use over-parameterized deep neural networks to ensure enough capacity for learning future tasks. This approach often leads to the discovery of subnetworks that perform as well as or better than the original network. Given the neural network parameters $\bm\theta$, the binary attention mask $\bm{m}^*_s$ that describes the optimal subnetwork for session $s$ such that $|\bm{m}^*_s$| is less than the model capacity $c$ follows as:
\begin{equation}
\begin{split}
    \bm{m}^*_s = \underset{\bm{m}_s\in\{0,1\}^{|\bm\theta|}}{\minimize} \frac{1}{T_s}\sum^{T_s}_{t=1}\mathcal{L}\big(f(\bm{e}_{s,t};\bm{\theta}\odot \bm{m}_s), \bm{v}^s_t \big) - \mathcal{J},
    ~~~~~\text{subject to~}|\bm{m}^*_s|\leq c,
\end{split}
\label{eq:subnetwork}
\end{equation}
where session loss $\mathcal{J}=\mathcal{L}(\bm{v}^s_t)$ and $c\ll|\bm\theta|$ (used as the selected proportion $\%$ of model parameters in the following section). A robust model adhering to this condition is known as WSN~\citep{kang2022forget}. However, WSN falls short in handling sequential complex video compilation due to its reliance on a limited set of adaptable yet sparse parameters in convolutional operators. In the following section, we introduce a novel Fourier Subnetwork Operator (FSO) to address this limitation.

\subsection{Fourier SubNueral Operator (FSO)} 
Conventional continual learner (i.e., WSN) only uses a few learnable parameters in convolutional operations to represent complex sequential image streams. To capture more parameter-efficient, forget-free NIRs, the NIR model requires fine discretization and video-specific sub-parameters. This motivation leads us to propose a novel subnetwork operator in Fourier space, which provides it with various bandwidths. Following the previous definition of Fourier convolutional operator~\citep{li2020fourier}, we adapt and redefine this definition to better fit the needs of the NIR framework. We use the symbol $\mathcal{F}$ to represent the Fourier transform of a function $f$, which maps from an embedding space of dimension $d_{\bm{e}}=1 \times 160$ to a frame size denoted as $d_{\bm{v}}$. The inverse of this transformation is represented by $\mathcal{F}^{-1}$. In this context, we introduce our \textbf{F}ourier-integral \textbf{S}ubneural \textbf{O}perator (\textbf{FSO}), symbolized as $\mathcal{K}$, which is tailored to enhance the capabilities of our NIR system (see \Cref{app:fso}):
\begin{equation}
\left( \mathcal{K}(\phi) \tilde{\bm{v}}_t^s \right)(\bm{e}_{s,t}) = \mathcal{F}^{-1}(R_{\bm \phi} \cdot (\mathcal{F} \tilde{\bm{v}}_t^s))(\bm{e}_{s,t}),
%\left( \mathcal{K}(\phi) \bm{e}_{s,t} \right)(\tilde{\bm{v}}_t^s) = \mathcal{F}^{-1}(R_{\bm \phi} \cdot (\mathcal{F} \bm{e}_{s,t}))(\tilde{\bm{v}}_t^s),
\label{eq:FSOper}
\end{equation}
where $\tilde{\bm{v}}_t^s$ is a hidden representation; $R_{\bm \phi}$ is the Fourier transform of a periodic subnetwork function which is parameterized by its subnetwork's parameters of real $(\bm{\theta}^{real} \odot \bm{m}_s^{real})$ and imaginary $(\bm{\theta}^{imag} \odot \bm{m}_s^{imag})$. We thus parameterize $R_{\bm \phi}$ separately as complex-valued tensors of real and imaginary $\bm{\phi}_{FSO} \in \{ \bm{\theta}^{real}, \bm{\theta}^{imag} \}$. One key aspect of the FSO is that its parameters grow with the depth of the layer and the input/output size. However, through careful layer-wise inspection and adjustments for sparsity, we can find a balance that allows the FSO to describe neural implicit representations efficiently. In the experimental section, we will showcase the most efficient FSO structure and its performance. \Cref{fig:concept_CVRNet} shows one possible PFNR structure of a single FSO. We describe the optimization in the following section.

%In the optimization \Cref{sub_sec:wsn}, we describe how to obtain $\bm{m}^*_s$ using a single learnable weight score $\bm{\rho}$ subject to updates while minimizing task loss jointly for each video session. 

%Since the parameters of FSO depend on the input/output size, the deeper the FSO layer, the larger the parameters increase. Nevertheless, we find the most parameter-efficient FSO through layer-wise inspection and various sparsity to describe the best neural implicit representations in the following experimental section. \Cref{fig:concept_CVRNet} shows one possible PFNR structure of a single FSO. 

\input{materials/algorithm}

\subsection{Sequential Video Representational Subnetworks}\label{sub_sec:wsn}
Let each weight $\bm{\theta}_\ast=\{\bm{\theta}, \bm{\phi}_{FSO}\}$ be associated with a learnable parameter we call \textit{weight score} $\bm{\rho}_\ast=\{\bm{\rho}, \bm{\rho}_{FSO}\}$, which numerically determines the importance of the weight associated with it; that is, a weight with a higher weight score is seen as more important. We find a sparse subnetwork $\hat{\bsy\theta}_s$ of the neural network and assign it as a solver of the current session $s$. We use subnetworks instead of the dense network as solvers for two reasons: (1) Lottery Ticket Hypothesis~\citep{frankle2018lottery} demonstrates the existence of a competitive subnetwork that is comparable with the dense network, and (2) the subnetwork requires less capacity than dense networks, and therefore it inherently reduces the size of the expansion of the solver.

Motivated by such benefits, we propose a novel PFNR, the joint-training method for sequential video representation compilation, as shown in \Cref{alg:algorithm}. The pseudo-code explains how to acquire subnetworks within a dense network. We find $\hat{\bm\theta}_s = \bm{\theta}_\ast \odot \bm{m}_s$ by selecting the top-$c$\% weights from the weight scores $\bm{\rho}_\ast$, where $c$ is the target layer-wise capacity ratio in \%; $\mathbf{m}_s$ is a session-dependent binary mask. Formally, $\mathbf{m}_s$ is obtained by applying a indicator function $\mathbbm{1}_c$ on $\bm{\rho}_\ast$ where $\mathbbm{1}_c(\rho_\ast)=1$ if $\rho_\ast$ belongs to top-$c\%$ scores and $0$ otherwise. Therefore, the subnetworks $\{\hat{\bsy\theta}_s\}_{s=1}^{N}$ for all video session $\mathcal{S}$ are obtained by $\hat{\bm{\theta}}_s = \bm{\theta}_\ast \odot \mathbf{m}_s$. Straight-through estimator~\citep{Bengio2013, Hinton2012, Ramanujan2020} is used to update $\bm{\rho}_\ast$, which ignores the derivative of the indicator function and passes on the incoming gradient as if the indicator function were an identity function.

%% file: materials/outline.tex
\begin{figure*}
    \small
    \centering
    \vspace{-0.3in}
    \setlength{\tabcolsep}{-2pt}{% left, bottom, right, top
    \includegraphics[height=5.8cm, trim={1.2cm 0.3cm 0.3cm 0.3cm}, clip]{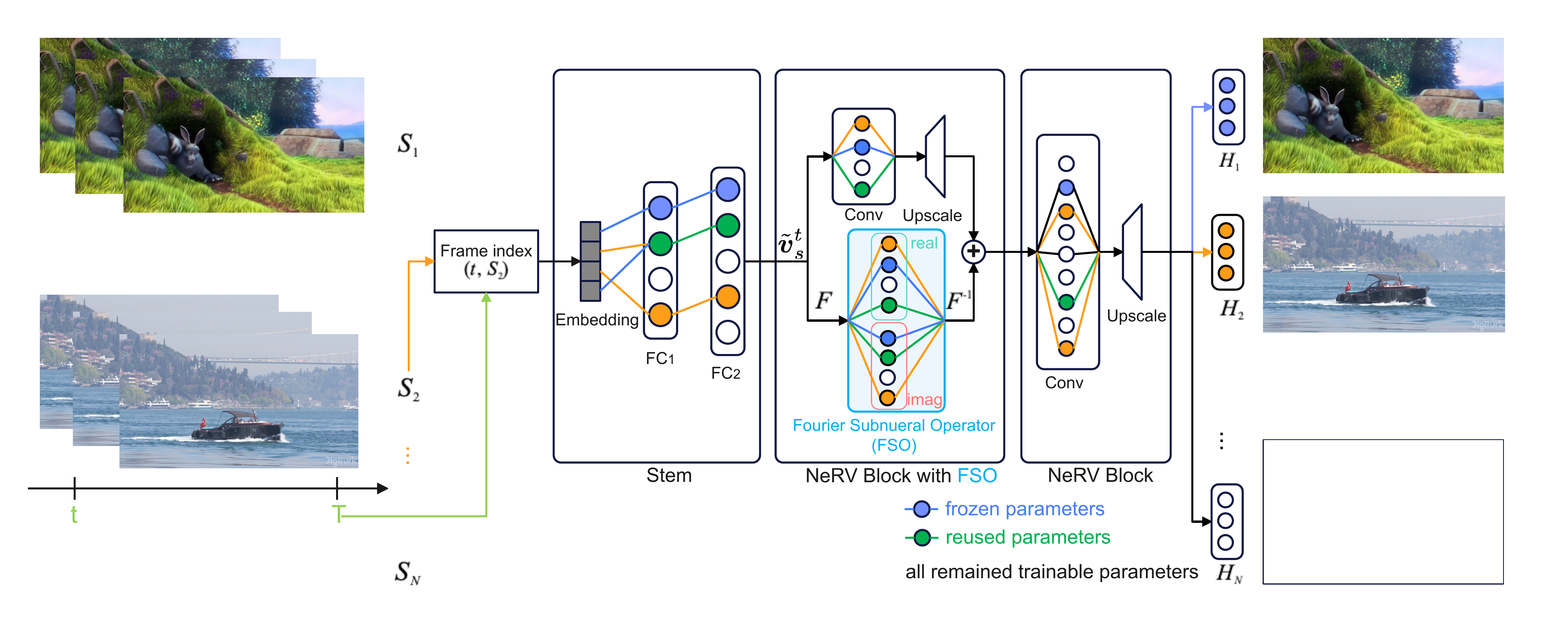}
    }
    \vspace{-0.3in}
    \caption{\small \textbf{Progressive Fourier Neural Representation (PFNR)}: PFNR takes time and video (session) indices as input and uses a sparse Stem + NeRV Blocks with \textit{Fourier Subneural Operator} (FSO) to output the whole image through multi-heads $H_N$ where $\tilde{\bm{v}}_s^t$ is a hidden representation. We denote frozen, reused, and trainable parameters in training at session 2. Note that each video representation is colored. In inference, we only need indices of session $s$ and frame $t$ and session mask (subnetwork).}
    \label{fig:concept_CVRNet}
    \vspace{-0.15in}
\end{figure*}

%% file: materials/algorithm.tex
\begin{algorithm}[ht]
  %\vspace{-0.1in}
  \caption{Progressive Fourier Neural Representation (PFNR) for VCL}\label{alg:algorithm}
  \small
  \hspace*{\algorithmicindent} \textbf{input}: $\{\mathcal{D}_s\}_{s=1}^{N}$, model weights of FSO $\bm{\theta}_{\ast}=\{\bm{\theta}, \bm{\phi_{FSO}}\}$, score weights of FSO $\bm{\rho}_{\ast} = \{\bm{\rho}, \bm{\rho}_{FSO} \}$,  \\ 
  \hspace*{\algorithmicindent}~~~~~~~~~ binary mask $\mathbf{M}_0=\{\bm{0}^{|\bm{\theta}|}$, $\bm{0}^{|\bm{\theta_{FSO}}|}$\}, and layer-wise capacity $c \%$. 
  \begin{algorithmic}[1]   
    \STATE randomly initialize $\bm{\theta}_\ast$ and $\bm{\rho}_\ast$.
    \FOR{session $s = 1, \cdots, |\mathcal{S}|$}
        \IF{ s > 1 }
            \STATE randomly re-initialize $\bm{\rho}_{\ast}$.
        \ENDIF
        \FOR{batch $\mathbf{b}_t\sim\mathcal{D}_s$} 
            \STATE obtain mask $\mathbf{m}_s$ of the top-$c\%$ scores $\bm{\rho}_{\ast}$ at each layer %\COMMENT{Weight ranking}
            \STATE compute $\mathcal{L}\left( f(\bm{e}_{s,t};\bm{\theta}_{\ast} \odot \mathbf{m}_s), \mathbf{b}_t \right)$, where input embedding, $\bm{e}_{s,t} = [\bm{e}_s; \bm{e}_t]$.
            \STATE $\bm{\theta}_{\ast} \leftarrow \bm{\theta}_{\ast} - \eta \left(\frac{\partial \mathcal{L}}{\partial \bsy\theta_{\ast}} \odot (\mathbf{1}-\mathbf{M}_{s-1})\right)$ \COMMENT{trainable weight update}
            
            \STATE $\bm{\rho}_{\ast} \leftarrow \bm{\rho}_{\ast} - \eta(\frac{\partial \mathcal{L}}{\partial \bm{\rho}_{\ast}})$ \COMMENT{weight score update}
        \ENDFOR
        \STATE $\hat{\bm{\theta}}_s = \bm{\theta}_{\ast} \odot \bm{m}_s$
        \STATE $\mathbf{M}_{s} \leftarrow \mathbf{M}_{s-1} \newor \mathbf{m}_s$ \COMMENT{accumulate binary mask}
    \ENDFOR
  \end{algorithmic}
  \hspace*{\algorithmicindent} \textbf{output}: $\{\hat{\bm{\theta}}_s\}_{s=1}^N$
  %\vspace{-0.1in}
\end{algorithm}

%% file: 5_experiment.tex
\section{Experiments}
Our method is validated on benchmark datasets for Video Task-incremental Learning (VTL) and compared against various continual learning baselines. In all experiments conducted for this paper, we utilize a multi-head configuration for continual video representation learning. This means the session identifier, denoted as $s$, is provided during the training and inference phases. Our experimental setups align with NeRV~\citep{chen2021nerv} and HNeRV~\citep{chen2023hnerv}.
%We validate our method on video benchmark datasets against continual learning baselines on Video Task-incremental Learning (VTL). We consider continual video representation learning with a multi-head configuration (session id, i.e., $s$ is given in training and inference) for all experiments in the paper. We follow the experimental setups in NeRV~\citep{chen2021nerv} and HNeRV~\citep{chen2023hnerv}. 

%\noindent
%\textcolor{blue}{multi-head setting: describes the need for neural implicit representation to emphasize our experimental settings. }

\noindent 
\textbf{Datasets.} 
\textit{1) UVG of 8 Video Sessions}: We experiment on eight sequential videos to validate our PFNR. The eight videos consist of one from the scikit-video and seven from the UVG dataset. The category index and order in UVG8 are as follows: \textit{1.bunny, 2.beauty, 3.bosphorus, 4.bee, 5.jockey, 6.setgo, 7.shake, 8.yacht}. 

\noindent 
\textit{2) UVG of 17 Video Sessions}:
We conducted an extended experiment on 17 video sessions by adding 9 more videos to the UVG of 8 video sessions. The category index and order in UVG17 are as follows: \textit{1.bunny, 2.city, 3.beauty, 4.focus, 5.bosphorus, 6.kids, 7.bee, 8.pan, 9.jockey, 10.lips, 11.setgo, 12.race, 13.shake, 14.river, 15.yacht, 16.sunbath, 17.twilight}. Please refer to the supplementary material.

\noindent 
\textbf{Architecture.} We employ NeRV as our baseline architecture and follow its details for a fair comparison. After the positional encoding, we apply 2 sparse MLP layers on the output of the positional encoding layer, followed by five sparse NeRV blocks with upscale factors of 5, 2, 2, 2, 2. These sparse NeRV blocks decode 1280$\times$720 frames from the 16$\times$9 feature map obtained after the sparse MLP layers. For the upscaling method in the sparse NeRV blocks, we also adopt PixelShuffle~\citep{shi2016real}. Fourier Subneural Operator (FSO) is used at the NeRV2 or NeRV3 layer, as depicted in \Cref{table:architecture_detail}. The positional encoding for the video index $s$ and frame index $t$ is as follows:
\begin{equation}
\begin{split}
\bm{\Gamma}(s, t) = &[~\sin(b^0\pi s), \cos(b^0\pi s), \cdots ,\sin(b^{l-1}\pi s), \cos(b^{l-1}\pi s), \\ & ~~\sin(b^0\pi t), \cos(b^0\pi t), \cdots ,\sin(b^{l-1}\pi t), \cos(b^{l-1}\pi t)~],
\end{split}
\label{eq:pos_embedding}
\end{equation}
where the hyperparameters are set to $b=1.25$ and $l=80$ such that $\bm{\Gamma}(s, t) \in \mathbb{R}^{1 \times 160}$. As differences from the previous NeRV model, the first layer of the MLP has its input size expanded from 80 to 160 to incorporate both frame and video indices, and distinct head layers after the NeRV block are utilized for each video. For the loss objective in \Cref{eq:l1_loss}, $\alpha$ is set to $0.7$. We evaluate the video quality, average video session quality, and backward transfer with two metrics: PSNR and MS-SSIM~\citep{wang2003multiscale}. We implement our model in PyTorch and train it in full precision (FP32). All experiments are run with NVIDIA RTX8000. %Please refer to the supplementary material for more experimental details.

\input{materials/main_table_uvg8_psnr}
\input{materials/main_table_uvb8_msssim}

\noindent 
\textbf{Baselines.} To show the effectiveness, we compare our PFNR with strong CL baselines: Single-Task Learning (STL), which trains on single tasks independently, EWC~\citep{Kirkpatrick2017}, which is a regularized baseline, iCaRL~\citep{rebuffi2017icarl}, and ESMER~\citep{sarfraz2023error} which are current strong rehearsal-based baseline, WSN~\citep{kang2022forget} which is a current strong architecture-based baseline, and Multi-Task Learning (MTL) which trains on multiple video sessions simultaneously, showing the upper-bound of WSN. Except for STL, all models are trained and evaluated on multi-head settings where a video session and time $(s, t)$ indices are provided.

\noindent 
\textbf{Training.} In all experiments, we follow the same experimental settings as NeRV~\citep{chen2023hnerv} and HNeRV~\citep{chen2023hnerv} for fair comparisons. We train WSN, PFNR, NeRV (STL), and MTL using Adam optimizer with a learning rate 5e-4. For the ablation study on UVG8 and UVG17, we use a cosine annealing learning rate schedule~\citep{loshchilov2016sgdr}, batch size of 1, training epochs of 150, and warmup epochs of 30 unless otherwise denoted.

%% Performance Metrics
\noindent 
\textbf{VCL's performance metrics of PSNR} \& \textbf{MS-SSIM.} We evaluate all methods based on the following continual learning metrics: 
\begin{enumerate}[itemsep=0em, topsep=-1ex, itemindent=0em, leftmargin=1.2em, partopsep=0em]
\item \small {\textit{Average PSNR or MS-SSIM (i.e., Ave. PSNR)}} measures the average of the final performances on all video sessions: $\mathrm{PSNR} \text{ or } \mathrm{MS}\text{-}{SSIM}=\frac{1}{N} \sum_{s=1}^{N} A_{N, s}$, where $A_{N,s}$ is the test PSNR or MS-SSIM for session $s$ after training on the final video session $S$.   

\item \small {\textit{Backward Transfer (BWT) of PSNR or MS-SSIM}} measures the video representation forgetting during continual learning. Negative BWT means that learning new video sessions causes the video representation forgetting of past sessions: $\mathrm{BWT}=\frac{1}{N-1}\sum_{s=1}^{N-1} A_{N, s}-A_{s, s}$.

%\item \small {\textit{Capacity (CAP)}} measures the total percentage of non-zero weights for all video sessions as follows: $\mathrm{CAP}= ((1-\mathcal{S}) + \mathcal{B}) \times 100.0$ where $\mathcal{S}$ is the sparsity of the accumulate binary mask $\mathbf{M}_{T}$; $\mathcal{B}$ are obtained by ASCII-code based Hoffman compression $\mathcal{B}=(1-\alpha)*N/32$ where compression rate $\alpha$. However, the binary capacity could be ignored because FP16 (see \Cref{fig:main_psnr_bit_bpp}) does not affect PSNR and MS-SSIM.

%$(7bit * N/7) * (1-\alpha) / 32bit$ where compression rate $\alpha \approx 0.78$, i.e., $\mathcal{B} = 5.5 \%$ for UVG8 and $11.6 \%$ for UVG17, respectively. The binary capacity could be ignored because FP16 (see \Cref{fig:main_psnr_bit_bpp}) does not affect PSNR and MS-SSIM.

\end{enumerate}

\input{materials/main_table_uvg17_psnr}
\input{materials/main_table_uvg17_msssim}

\subsection{Comparisons with Baselines}

\noindent 
\textbf{PSNR} \& \textbf{MS-SSIM.}
To compare PFNR with conventional representative continual learning methods such as EWC, iCaRL, ESMER, and WSN, we prepare the reproduced results, as shown in \Cref{table:uvg8_result}, \ref{table:uvg8_fso_msssim}, \ref{table:uvg17_fso_psnr}, and \ref{table:uvg17_fso_msssim}. The architecture-based WSN outperformed the regularized method and replay method. The sparseness of WSN does not significantly affect sequential video representation results on two sequential benchmark datasets. Our PFNR outperforms all conventional baselines including WSN and MLT (upper-bound of WSN) on the UVG8 and UVG17 benchmark datasets. Moreover, our performances of PFNR with $f$-NeRV3 are better than those of $f$-NeRV2 since $f$-NeRV3 tends to represent local textures, stated in the following \Cref{sec:video_reps}. Note that the number of parameters of MLT is precisely the same as those of WSN.

%\input{materials/main_table_larger_psnr}
%\input{materials/main_table_larger_msssim}

% Input Embedding
%\noindent 
%\textbf{Input Embedding.} We observe that the input embedding resolutions affect video representation as shown in \Cref{table:uvg8_psnr} and \Cref{table:uvg8_msssim}. Even though the video sessions are the same, the performances of PSRN and MS-SSIM decrease by 0.8, 0.2, depending on input embedding resolution determined by the maximum number of input index (m-IDX). The results with m-IDX=17 are reported by the longer sequence learning in \Cref{table:uvg17_psnr} and \Cref{table:uvg17_msssim}. From this observation, we can expect more precise video representation if we use more discriminative input embedding for PNR. Here, we do not care about the video's contextual information.

%\input{materials/plot_transfer_matrix}
% Transfer Matrix
%\noindent 
%\textbf{Transfer Matrix.} 
%We prepare the transfer matrix to prove our PFNR's forget-freeness and to show video correlation among other videos, as shown in \Cref{fig:transf_matrix} on the UVG17 dataset; lower triangular estimated by each session subnetwork denotes that our PNR is a forget-free method and upper triangular calculated by current session subnetwork denotes the video similarity between source and target. The reinitialized PFNR proves the effectiveness from the lower triangular of \Cref{fig:transf_matrix} (a) and (b). Nothing special is observable from the upper triangular since they are not correlated. 

\noindent 
\textbf{PFNR's Compression.} 
We follow NeRV's video quantization and compression pipeline~\citep{chen2021nerv}, except for the model pruning step, to evaluate performance drops and backward transfer in the video sequential learning, as shown in \Cref{fig:uvg17_fso_psnr}. Once sequential training is done, our PFNR doesn't need any extra prune and finetune steps, unlike NeRV. This point is our key advantage of PFNR over NeRV. \Cref{fig:uvg17_fso_psnr} (a) shows the results of various sparsity and bit-quantization on the UVG17 datasets: the 8bit PFNR's performances are comparable with 32bit ones without a significant video quality drop. From our observations, the 8-bit subnetwork seems to be enough for video implicit representation. \Cref{fig:uvg17_fso_psnr} (b) shows the rate-distortion curves. We compare PFNR with WSN and NeRV (STL). For a fair comparison, we take steps of pruning, fine-tuning, quantizing, and encoding NeRV. Our PFNR outperforms all baselines.

\input{materials/plot_main_psnr_bpp}

\noindent 
\textbf{Performance and Capacity.}
Our PFNR outperforms WSN and MTL, as stated in \Cref{fig:psnr_cap} (a). This result might suggest that properly selected weights in Fourier space lead to generalization more than others in VCL. Moreover, to show the behavior of PSNR, We prepare a progressive PSNR's capacity and investigate how PFNR reuses weights over sequential video sessions, as shown in \Cref{fig:psnr_cap} (b). PFNR tends to progressively transfer weights used for a prior session to weights for new ones, but the proposition of reused weights gets smaller as video sessions increase.

%compared with others, i.e., WSN. Since the reinitialized PFNR explores more new weights than PFNR, PFNR with \textcolor{red}{reinit} outperforms PFNR, as stated in \Cref{fig:psnr_cap} (a), leading to comparable with MTL. This result might suggest that properly reused weights lead to generalization more than others in VCL with low video contextual similarity.

\input{materials/plot_capacity}

\subsection{PFNR's Video Representations}\label{sec:video_reps}
We prepare the results of video generation as shown in \Cref{fig:video_reinit_mtl}. We demonstrate that a sparse solution (PFNR with $c=30.0 \%$, $f$-NeRV3) generates video representations sequentially without significant performance drops. Compared with WSN, PFNR provides more precise representations. 
To find out the results, we inspect the layer-wise representations as shown in \Cref{fig:fmap_uvg17}, which provides essential observations that PFNR tends to capture local textures broadly at the NeRV3 layer while WSN focuses on local objects. This behavior of PFNR could lead to more generalized performances. Moreover, we conduct an ablation study to inspect the best sparsity of $f$-NeRV3 while holding the remaining parameters' sparsity (c=50.0 \%), as shown in \Cref{fig:fmap_sparsity_uvg17}. Please refer to the supplementary materials for comparisons with baselines.

%Please refer to the supplementary materials for more visualizations, layer-wise representations, and comparisons with baselines. 

%\input{materials/plot_video}
\input{materials/plot_main_fmap_uvg17}
\input{materials/plot_main_video_mtl}

%% file: materials/main_table_uvg8_psnr.tex
\begin{table*}[!ht]
\small
\centering
\vspace{-0.3in}
\caption{\small PSNR results with Fourier Subnueral Operator (FSO) layer (\textcolor{red}{$f$-NeRV$\ast$}) (detailed in \Cref{table:architecture_detail}) on UVG8 Video Sessions with average PSNR and Backward Transfer (BWT). Note that $\ast$ denotes our reproduced results.}
\resizebox{0.83\textwidth}{!}{
\begin{tabular}{lcccccccccc}
\toprule 

\multicolumn{1}{c}{\multirow{2}{*}{\textbf{Method}}} & \multicolumn{8}{c}{\textbf{Video Sessions}} & \multirow{2}{*}{\thead{\textbf{Avg. PSNR / } \\ \textbf{BWT}}} \\ % & \multirow{2}{*}{\textbf{CAP}}  \\

\cline{2-9}
& \textbf{1} & \textbf{2} & \textbf{3} & \textbf{4} & \textbf{5} & \textbf{6} & \textbf{7} & \textbf{8} &  \\ \midrule 
%STL, NeRV~\cite{chen2023hnerv} & 39.63 & 36.06 & 37.35 & 41.23 & 38.14 & 31.86 & 37.22 & 32.45 & 36.74 / ~~-~~ \\ % & ~~800.00 \% \\

STL, NeRV~\cite{chen2021nerv}$^{\ast}$  & 39.66 & 36.28 & 38.14 & 42.03 & 36.58 & 29.22 & 37.27 & 31.45 & 36.33 / ~~-~~ \\ %& ~~800.00 \% \\ 
\midrule 

EWC~\cite{Kirkpatrick2017}$^{\ast}$    & 10.19 & 11.15  & 14.47 & 8.39 & 12.21 & 10.27 & 9.97 & 23.98 & 12.58 / -17.59 \\ %& ~~100.00 \% \\  
iCaRL~\cite{rebuffi2017icarl}$^{\ast}$ & 30.84 & 26.30  & 27.28 & 34.48 & 20.90 & 17.28 & 30.33 & 24.64 & 26.51 / ~-3.90 \\ %& ~~100.00 \% \\ 
ESMER~\cite{sarfraz2023error}$^{\ast}$ & 31.71 & 23.09  & 24.15 & 28.03 & 17.30  & 13.81 &  12.45 &  24.57 & 21.92  / ~-9.99  \\ % & ~~100.00 \% \\
\midrule

% init 
%WSN$^{\ast}$, c = 10.0 \% & 27.81 & 30.66 & 29.30 & 33.06 & 22.16 & 18.40 & 27.81 & 22.97 & 26.52 / 0.0 \\ 
%WSN$^{\ast}$, c = 30.0 \% & 31.37 & 32.19  & 29.92 & {33.62} & {22.82}  & {18.96} & {28.43} & 23.40 & 27.59 / 0.0 \\ 
WSN$^{\ast}$, c = 50.0 \% & 34.05 & {32.28}  & {29.98} & 32.88 & 22.15  & 18.61 & 27.68 & {23.64} & {27.66} / 0.0 \\ 
%WSN$^{\ast}$, c = 70.0 \% & {35.62} & 32.08 & 29.46 & 31.37 & 21.60  & 18.13 & 27.33 & 22.61 & 27.28 / 0.0 \\ 

% re-init
%WSN, c = 10.0 \% & 28.12 & 31.31 & 29.89 & 34.83 & 23.82 & 19.56 & 29.46 & 24.58 & 27.72 / 0.0 \\ % & ~~~~30.00 \%\\ 
%WSN, c = 30.0 \% & 31.36 & 32.91 & 31.42 & 36.39 & 24.93 & 20.58 & 30.78 & 25.25 & 29.20 / 0.0 \\ % & ~~~~68.00 \% \\ 
%WSN, c = 50.0 \% & 34.10 & 33.45 & 31.77 & 36.09 & 24.82 & 20.25 & 30.02 & 25.01 & 29.44 / 0.0 \\ % & ~~~~91.00 \% \\ 
%WSN, c = 70.0 \% & 35.55 & 33.04 & 30.44 & 32.11 & 23.00 & 19.02 & 28.09 & 23.52 & 28.10 / 0.0 \\ % & ~~~~99.00 \% \\ 
%\midrule 

% real
%PFNR, c = 10.0 \%, \textcolor{red}{$f$-NeRV2} & 28.49 & 32.30 & 30.30 & 35.12 & 24.10 & 19.82 & 29.89 & 24.76 & 28.10 / 0.0  \\ % &  ~~~~31.91 \% \\ 
%PFNR, c = 30.0 \%, \textcolor{red}{$f$-NeRV2} & 31.99 & 33.56 & 31.82 & 36.61 & 25.28 & 20.97 & 31.07 & 25.73 & 29.63 / 0.0  \\ % &  ~~~~72.35 \% \\
PFNR~, c = 50.0 \%, \textcolor{red}{$f$-NeRV2} & 34.46 & 33.91 & 32.17 & 36.43 & 25.26 & 20.74 & 30.18 & 25.45 & 29.82 / 0.0  \\ % &  ~~96.82 \% \\ 
%PFNR, c = 70.0 \%, \textcolor{red}{$f$-NeRV2} & 36.04 & 33.46 & 31.05 & 32.57 & 23.40 & 19.41 & 28.31 & 24.31 & 28.57 / 0.0  \\ % &  ~~104.26 \% \\
% real-image
%PFNR, c = 10.0 \%, \textcolor{red}{$f$-NeRV2} & 28.49 & 32.30 & 30.30 & 35.12 & 24.10 & 19.82 & 29.89 & 24.76 & 28.10 / 0.0  &  ~~~~33.83 \% \\ 
%PFNR, c = 30.0 \%, \textcolor{red}{$f$-NeRV2} & 31.99 & 33.56 & 31.82 & 36.61 & 25.28 & 20.97 & 31.07 & 25.73 & 29.63 / 0.0  &  ~~~~76.76 \% \\
%PFNR, c = 50.0 \%, \textcolor{red}{$f$-NeRV2} & 34.46 & 33.91 & 32.17 & 36.43 & 25.26 & 20.74 & 30.18 & 25.45 & 29.82 / 0.0  &  ~~102.64 \% \\ 
%PFNR, c = 70.0 \%, \textcolor{red}{$f$-NeRV2} & 36.04 & 33.46 & 31.05 & 32.57 & 23.40 & 19.41 & 28.31 & 24.31 & 28.57 / 0.0  &  ~~111.66 \% \\
%\midrule 

% real
%PFNR, c = 10.0 \%, \textcolor{red}{$f$-NeRV3} & 29.78 & 33.30 & 32.29 & 37.29 & 24.65 & 20.82 & 31.94 & 26.19 & 29.53 / 0.0 \\ % & ~~482.38 \% \\ 
%PFNR, c = 30.0 \%, \textcolor{red}{$f$-NeRV3} & 33.69 & 34.76 & 34.57 & 38.50 & 27.09 & 23.16 & 33.10 & 27.94 & 31.36 / 0.0 \\ % & 1093.04 \% \\ 
PFNR~, c = 50.0 \%, \textcolor{red}{$f$-NeRV3} & \textbf{36.45} & \textbf{35.15} & \textbf{35.10} & \textbf{38.57} & \textbf{28.07} & \textbf{23.06} & \textbf{32.83} & \textbf{27.70} & \textbf{32.12} / \textbf{0.0} \\ % & 1463.60 \% \\ 
%PFNR, c = 70.0 \%, \textcolor{red}{$f$-NeRV3} & 38.15 & 34.91 & 34.05 & 35.94 & 24.32 & 20.37 & 30.58 & 25.69 & 30.50 / 0.0 \\ % & 1575.00 \% \\ 
% real-image
%PFNR, c = 10.0 \%, \textcolor{red}{$f$-NeRV3} & 29.78 & 33.30 & 32.29 & 37.29 & 24.65 & 20.82 & 31.94 & 26.19 & 29.53 / 0.0 & ~~933.17 \% \\ 
%PFNR, c = 30.0 \%, \textcolor{red}{$f$-NeRV3} & 33.69 & 34.76 & 34.57 & 38.50 & 27.09 & 23.16 & 33.10 & 27.94 & 31.36 / 0.0 & 2770.04 \% \\ 
%PFNR, c = 50.0 \%, \textcolor{red}{$f$-NeRV3} & 36.45 & 35.15 & 35.10 & 38.57 & 28.07 & 23.06 & 32.83 & 27.70 & 32.12 / 0.0 & 2116.60 \% \\ 
%PFNR, c = 70.0 \%, \textcolor{red}{$f$-NeRV3} & 38.15 & 34.91 & 34.05 & 35.94 & 24.32 & 20.37 & 30.58 & 25.69 & 30.50 / 0.0 & 2832.00 \% \\ 
\midrule 

%PFNR, c = 50.0 \%, \textcolor{red}{$f$-NeRV3-FP32} & 36.45 & 35.15 & 35.10 & 38.57 & 28.07 & 23.06 & 32.83 & 27.70 & 32.12 / ~~0.0~~ & 1463.60 \% \\
%PFNR, c = 50.0 \%, \textcolor{red}{$f$-NeRV3-FP16} & 36.45 &	35.15 &	35.10 &	38.57 &	28.07 &	23.06 &	32.83 &	27.70 & 32.12 / ~~0.0~~ & ~~777.11 \% \\ 
%PFNR, c = 50.0 \%, \textcolor{red}{$f$-NeRV3-FP8}  & 36.43 & 35.15 &	35.09 &	38.56 &	28.07 &	23.06 &	32.83 &	27.70 &	32.11 / -0.01 & ~~434.58 \% \\ 
%PFNR, c = 50.0 \%, \textcolor{red}{$f$-NeRV3-FP4}  & 32.79 & 34.27 & 34.30 & 37.70 & 27.74 & 22.76 & 32.40 & 27.37 & 31.17 / -0.60 & ~~262.52 \% \\  \midrule 

MTL (upper-bound) & 34.22 & 32.79 & 32.34 & 38.33 & 25.30 & 22.44 & 33.73 & 27.05 & 30.78 / -~~~~ \\ % &  ~~100.00 \% \\ 
\bottomrule
\end{tabular}
}
\label{table:uvg8_result}
\vspace{-0.12in}
\end{table*}

%% file: materials/main_table_uvb8_msssim.tex
\begin{table}[ht]
\small\centering
\caption{\small MS-SSIM results with Fourier Subnueral Operator (FSO) layer (\textcolor{red}{$f$-NeRV$\ast$}) (detailed in \Cref{table:architecture_detail}) on UVG8 Video Sessions with average MS-SSIM, Backward Transfer (BTW) of MS-SSIM. Note that $\ast$ denotes our reproduced results.}
\resizebox{0.8\textwidth}{!}{
\begin{tabular}{lcccccccccc}
\toprule 

\multicolumn{1}{c}{\multirow{2}{*}{\textbf{Method}}} & \multicolumn{8}{c}{\textbf{Video Sessions}} & \multirow{2}{*}{\thead{\textbf{Avg. MS-SSIM / } \\ \textbf{BWT}}}  \\

\cline{2-9}

& \textbf{1} & \textbf{2} & \textbf{3} & \textbf{4} & \textbf{5} & \textbf{6} & \textbf{7} & \textbf{8} &  \\ \midrule 
%NeRV~\cite{chen2023hnerv} & 39.63 & 36.06 & 37.35 & 41.23 & 38.14 & 31.86 & 37.22 & 32.45 & 36.74/- \\

%NeRV$^{\ast}$             & 39.66 & 36.28 & 38.14 & 42.03 & 36.59 & 29.23 & 37.27 & 31.45 & 36.33/- \\ \midrule 
STL, NeRV~\cite{chen2021nerv}$^{\ast}$   & 0.99 & 0.95 & 0.98 & 0.99 & 0.97 & 0.96 & 0.98 & 0.96 & 0.97 / - ~~ \\ \midrule 

EWC~\cite{Kirkpatrick2017}$^{\ast}$ & 0.22 & 0.23  & 0.35 & 0.10 & 0.27 & 0.19 & 0.21 & 0.79 & 0.30 / -0.62 \\ 
iCaRL~\cite{rebuffi2017icarl}$^{\ast}$ & 0.94 & 0.80  & 0.82 & 0.97 & 0.59 & 0.57 & 0.92 & 0.81 & 0.80 / -0.11 \\ 
ESMER~\cite{sarfraz2023error}$^{\ast}$ & 0.88 & 0.65  & 0.68 & 0.90 & 0.42 & 0.32 & 0.19 & 0.81 & 0.61 / -0.33 \\
\midrule 

% init
% WSN$^{\ast}$, c = 10.0 \% & 0.91 & 0.89 & 0.89 & 0.97 & 0.73 & 0.61 & 0.88 & 0.77 & 0.83 / 0.0 \\ 
% WSN$^{\ast}$, c = 30.0 \% & 0.96 & 0.91 & 0.90 & {0.98} & {0.76} & {0.65} & {0.89} & {0.78} & 0.85 / 0.0 \\ 
WSN$^{\ast}$, c = 50.0 \% & 0.98 & {0.91} & {0.90} & 0.97 & 0.74 & 0.62 & 0.88 & 0.77 & {0.85} / 0.0 \\ 
%WSN$^{\ast}$, c = 70.0 \% & {0.98} & 0.91 & 0.89 & 0.96 & 0.71 & 0.59 & 0.87 & 0.74 & 0.83 / 0.0 \\ 

% re-init
%WSN, c = 10.0 \% & 0.91 & 0.90 & 0.90 & 0.98 & 0.78 & 0.68 & 0.91 & 0.81 & 0.86 / 0.0 \\ 
%WSN, c = 30.0 \% & 0.96 & 0.99 & 0.92 & 0.99 & 0.82 & 0.74 & 0.93 & 0.84 & 0.89 / 0.0 \\ 
%WSN, c = 50.0 \% & 0.98 & 0.92 & 0.93 & 0.98 & 0.82 & 0.72 & 0.92 & 0.83 & 0.89 / 0.0 \\ 
%WSN, c = 70.0 \% & 0.98 & 0.92 & 0.91 & 0.97 & 0.76 & 0.65 & 0.89 & 0.79 & 0.86 / 0.0 \\ 
%\midrule 
%PFNR, c = 10.0 \%, \textcolor{red}{$f$-NeRV2} & 0.92 & 0.91 & 0.91 &	0.98 & 0.79 & 0.70 & 0.92 &	0.82 & 0.87 / 0.0 \\ 
%PFNR, c = 30.0 \%, \textcolor{red}{$f$-NeRV2} & 0.97	& 0.92 & 0.93 &	0.99 & 0.83 & 0.76 & 0.93 &	0.85 & 0.90	/ 0.0 \\
PFNR~, c = 50.0 \%, \textcolor{red}{$f$-NeRV2} & 0.98	& 0.93 & 0.93 & 0.99 & 0.83 & 0.75 & 0.92 & 0.84 & 0.90	/ 0.0 \\ 
%PFNR, c = 70.0 \%, \textcolor{red}{$f$-NeRV2} & 0.99 & 0.92 & 0.92 & 0.97 & 0.78 & 0.67 & 0.89 & 0.81 & 0.87 / 0.0 \\
%\midrule 

%PFNR, c = 10.0 \%, \textcolor{red}{$f$-NeRV3} & 0.95	& 0.92 & 0.94 &	0.99 & 0.82 & 0.76 & 0.94 & 0.87 & 0.90 / 0.0  \\ 
%PFNR, c = 30.0 \%, \textcolor{red}{$f$-NeRV3} & 0.98	& 0.93 & 0.96 &	0.99 & 0.87 & 0.85 & 0.95 & 0.91 & 0.93 / 0.0 \\ 
PFNR~, c = 50.0 \%, \textcolor{red}{$f$-NeRV3} & \textbf{0.99} & \textbf{0.94} & \textbf{0.97} & \textbf{0.99} & \textbf{0.88} & \textbf{0.84} & \textbf{0.95} & \textbf{0.90} & \textbf{0.93} / \textbf{0.0} \\ 
%PFNR, c = 70.0 \%, \textcolor{red}{$f$-NeRV3} & 0.99 & 0.92 & 0.92 & 0.97 & 0.78 & 0.67 & 0.89 & 0.81 & 0.87 / 0.0  \\

\midrule 

MTL (upper-bound) & 0.98 & 0.91  & 0.93 & 0.99 & 0.84  & 0.82 & 0.95 & 0.89 & 0.91 / - ~~~ \\ 
\bottomrule
\end{tabular}
}
\vspace{-0.1in}
\label{table:uvg8_fso_msssim}
\end{table}

%% file: materials/main_table_uvg17_psnr.tex
\begin{table}[ht]
\small\centering
\caption{\small PSNR results with Fourier Subnueral Operator (FSO) layer (\textcolor{red}{$f$-NeRV$\ast$}) (detailed in \Cref{table:architecture_detail}) on UVG17 Video Sessions with average PSNR and Backward Transfer (BWT) of PSNR. Note that $\ast$ denotes our reproduced results.}
\resizebox{\textwidth}{!}{
\renewcommand{\arraystretch}{1.2}
\begin{tabular}{lccccccccccccccccccc}
\toprule 

\multicolumn{1}{c}{\multirow{2}{*}{\textbf{Method}}} & \multicolumn{17}{c}{\textbf{Video Sessions}} & \multirow{2}{*}{\thead{\textbf{Avg. PSNR} \\ \textbf{BWT}}} \\ % & \multirow{2}{*}{\thead{\textbf{CAP}}} \\

\cline{2-18}

& \textbf{1} & \textbf{2} & \textbf{3} & \textbf{4} & \textbf{5} & \textbf{6} & \textbf{7} & \textbf{8} & \textbf{9} & \textbf{10} & \textbf{11} & \textbf{12} & \textbf{13} & \textbf{14} & \textbf{15} & \textbf{16} & \textbf{17} \\ \midrule 
%STL, NeRV~\cite{chen2023hnerv} & 39.63 & - & 36.06 & - & 37.35 & - & 41.23 & - & 38.14 & - & 31.86 & - & 37.22 & - & 32.45 & - & - & - / - \\ % & 1700.00 \% \\

STL, NeRV~\cite{chen2021nerv}$^{\ast}$ & 39.66 & 44.89 & 36.28 & 41.13 & 38.14 & 31.53 & 42.03 & 34.74 & 36.58 & 36.85 & 29.22 & 31.81 & 37.27 & 34.18 & 31.45 & 38.41 & 43.86 & 36.94 / - \\ % & 1700.00 \% \\ 
\midrule 

EWC~\cite{Kirkpatrick2017}$^{\ast}$ & 11.15 & 9.21  & 12.71 & 11.40 & 15.58  & 9.25 & 7.06 & 12.96 & 6.34 & 10.31 & 9.55 & 13.39 & 5.76 & 8.67 & 10.93 & 10.92 & 28.29 & 11.38 / -16.13 \\ % & 100.00 \% \\
iCaRL~\cite{rebuffi2017icarl}$^{\ast}$ & 24.31 & 28.25  & 22.19 & 22.74 & 22.84  & 16.55 & 29.37 & 17.92 & 16.65 & 27.43 & 13.64 & 16.42 & 24.02 & 21.60 & 19.40 & 18.60 & 26.46 & 21.67 / ~~-6.23 \\ % & 100.00 \% \\ 
ESMER~\cite{sarfraz2023error}$^{\ast}$ &  30.77 & 26.33 & 22.79 & 21.35 & 23.76 & 13.64 & 28.25 & 15.22 & 16.71 & 23.78 &	13.35 & 15.23 & 18.21 & 19.22 & 24.59 & 20.61 & 22.42 &  20.95 / -15.23 \\ % &  100.00 \% \\
\midrule

%WSN$^{\ast}$, c = 10.0 \% & 27.68 & 31.31 & 30.29 & 31.63 & 28.66 & 22.57 & 31.62 & 22.04 & 21.05 & 32.71 & 17.85 & 20.09 & 27.07 & 23.84 & 22.98 & 20.50 & 28.56 & 25.91 / ~0.0 \\ 
WSN$^{\ast}$, c = 30.0 \% & 31.50 & 34.37 & 31.00 & {{32.38}} & {{29.26}} & {{23.08}} & {{31.96}} & {{22.64}} & {{22.07}} & {{33.48}} & {{18.34}} & {{20.45}} & {{27.21}} & {{24.33}} & {{23.09}} & {{21.23}} & {{29.13}} & {{26.80}} / ~0.0 \\ 
%WSN$^{\ast}$, c = 50.0 \% & {{34.02}} & {{34.93}} & {{31.04}} & 31.74 & 28.95 & 23.07 & 31.26 & 22.32 & 21.93 & 33.35 & 18.22 & 20.34 & 26.88 & 24.22 & 22.72 & 21.30 & 28.86 & 26.77 / ~0.0 \\ 
%WSN$^{\ast}$, c = 70.0 \% & 35.64 & 34.36 & 30.26 & 30.27 & 27.99  & 22.55 & 29.88 & 21.46 & 20.79 & 32.37 & 17.63 & 20.00 & 26.68 & 23.79 & 22.34 & 20.69 & 28.68 &  26.20 / ~0.0 \\ 

% reinit
%WSN, c = 10.0 \% & 28.02 & 32.68 & 31.38 & 33.13 & 29.54 & 23.75 & 33.91 & 24.49 & 23.63 & 34.91 & 19.42 & 21.71 & 29.09  & 26.14 & 24.47 & 23.57 & 31.34 &  27.72 / ~~0.0 \\ % & 56.00 \% \\ 
%WSN, c = 30.0 \% & 31.47 & 35.42 & 32.51 & 34.73 & {{30.70}} & {{24.53}} & {35.63} & {{25.49}} & {{24.50}} & {35.59} & {{20.24}} & {{22.49}} & {30.22} & {{27.03}} & {{25.14}} & {24.86} & {{32.16}} &  {{28.98}} / ~~0.0 \\ % & 94.00 \% \\ 
%WSN, c = 50.0 \% & {{34.05}} & {{36.61}} & {{32.73}} & {{34.99}} & 30.43 & 24.37 & 33.98 & 24.53 & 24.31 & 35.51 & 19.99 & 22.31 & 29.65 & 26.77 & 24.94 & 24.76 & 31.84 & 28.93 / ~~0.0 \\ % & 99.00 \% \\ 
%WSN, c = 70.0 \% & 35.57 & 35.93 & 31.61 & 31.39 & 28.32 & 23.19 & 30.40 & 22.69 & 22.70 & 34.69 & 18.82 & 21.09 & 27.75 & 25.43 & 23.47 &  23.24 & 30.30 &  27.45 / ~~0.0 \\ % & 99.00 \% \\  

%\midrule

% real
%PFNR, c = 10.0 \%, \textcolor{red}{$f$-NeRV2} &  28.31 & 33.57 & 31.92 & 33.67 & 29.98 & 23.99 &	34.39 & 24.8 & 23.94 & 35.08 & 19.70 & 22.03 & 29.56 & 26.57 & 24.79 & 24.10 &	31.35 & 28.10 / ~0.0 \\ % & 59.58 \% \\
PFNR~, c = 30.0 \%, \textcolor{red}{$f$-NeRV2} & 32.01 &	35.84 &	32.97 &	35.17 &	31.24 &	24.82 &	36.01 &	25.85 &	24.83 &	35.76 &	20.50 &	22.79 &	30.40 &	27.37 &	25.52 &	25.40 &	32.70 &	29.36 / ~0.0 \\ % & 100.01 \% \\
%PFNR, c = 50.0 \%, \textcolor{red}{$f$-NeRV2} & 34.49 &	37.13 &	33.21 &	35.50 &	30.87 &	24.72 &	34.36 &	24.79 &	24.73 &	35.65 &	20.33 &	22.65 &	29.78 &	27.05 &	25.18 &	25.18 &	32.39 &	29.29 / ~0.0 \\ % & 105.33 \% \\
%PFNR, c = 70.0 \%, \textcolor{red}{$f$-NeRV2} & 36.02 &	36.50 &	32.09 &	32.15 &	28.67 &	23.35 &	30.63 &	22.86 &	23.18 &	34.90 &	19.08 &	21.30 &	27.87 &	25.86 &	24.12 &	23.47 &	30.34 &	27.79 / ~0.0 \\ % & 105.33 \% \\

% real-imag
%PFNR, c = 10.0 \%, \textcolor{red}{$f$-NeRV2} &  28.31 & 33.57 & 31.92 & 33.67 & 29.98 & 23.99 &	34.39 & 24.8 & 23.94 & 35.08 & 19.70 & 22.03 & 29.56 & 26.57 & 24.79 & 24.10 &	31.35 & 28.10 / ~~0.0 & 56.53 \% \\
%PFNR, c = 30.0 \%, \textcolor{red}{$f$-NeRV2} & 32.01 &	35.84 &	32.97 &	35.17 &	31.24 &	24.82 &	36.01 &	25.85 &	24.83 &	35.76 &	20.50 &	22.79 &	30.40 &	27.37 &	25.52 &	25.40 &	32.70 &	29.36 / ~~0.0 & 94.43 \% \\
%PFNR, c = 50.0 \%, \textcolor{red}{$f$-NeRV2} & 34.49 &	37.13 &	33.21 &	35.50 &	30.87 &	24.72 &	34.36 &	24.79 &	24.73 &	35.65 &	20.33 &	22.65 &	29.78 &	27.05 &	25.18 &	25.18 &	32.39 &	29.29 / ~~0.0 & 99.54 \% \\
%PFNR, c = 70.0 \%, \textcolor{red}{$f$-NeRV2} & 36.02 &	36.50 &	32.09 &	32.15 &	28.67 &	23.35 &	30.63 &	22.86 &	23.18 &	34.90 &	19.08 &	21.30 &	27.87 &	25.86 &	24.12 &	23.47 &	30.34 &	27.79 / ~~0.0 & 99.65 \% \\
%\midrule

% real
%PFNR, c = 10.0 \%, \textcolor{red}{$f$-NeRV3} & 30.40 &	36.94 &	32.92 &	35.64 &	31.90 &	25.25 &	36.86 &	28.36 &	24.46 &	35.68 &	20.84 &	22.86 &	31.61 &	28.8 &	26.07 &	26.31 &	33.68 & 29.92 / 0.0 \\ % & 900.45 \% \\
PFNR~, c = 30.0 \%, \textcolor{red}{$f$-NeRV3} &  \textbf{33.64} & \textbf{39.24} & \textbf{34.21} & \textbf{37.79} & \textbf{34.05} & \textbf{27.17} & \textbf{38.17} & \textbf{29.79} & \textbf{26.56} & \textbf{36.18} & \textbf{22.97} & \textbf{24.36} & \textbf{32.50} & \textbf{30.22} &	\textbf{27.62} & \textbf{29.15} & \textbf{35.68} &	\textbf{31.72} / ~\textbf{0.0} \\ % & 1511.47 \%\\
%PFNR, c = 50.0 \%, \textcolor{red}{$f$-NeRV3} &  36.59 & 39.95 & 34.46 & 38.01 & 34.21 & 27.26 & 37.47 & 29.07 & 26.72 & 36.17 & 22.65 & 24.46 & 32.43	& 30.07 &	27.62 &	29.30 &	35.26 &	31.65 / 0.0 \\ % & 1591.00 \%\\
%PFNR, c = 70.0 \%, \textcolor{red}{$f$-NeRV3} &  38.24 & 39.47 & 33.72 & 35.18 & 31.58 & 25.65 & 33.64 & 26.23 & 23.92 & 35.75 & 20.81 & 22.73 & 30.54 &	28.55 & 26.17 & 26.84 & 33.15 & 30.13 / 0.00 \\ % & 1591.87 \% \\

% real-imag
%PFNR, c = 10.0 \%, \textcolor{red}{$f$-NeRV3} & 30.40 &	36.94 &	32.92 &	35.64 &	31.90 &	25.25 &	36.86 &	28.36 &	24.46 &	35.68 &	20.84 &	22.86 &	31.61 &	28.8 &	26.07 &	26.31 &	33.68 &	29.92  / 0.0 & 1743.17 \% \\
%PFNR, c = 30.0 \%, \textcolor{red}{$f$-NeRV3} &  33.64 & 39.24 & 34.21 & 37.79 & 34.05 & 27.17 & 38.17 & 29.79 & 26.56 & 36.18 & 22.97 & 24.36 & 32.50 &	30.22 &	27.62 &	29.15 &	35.68 &	31.72 / 0.0 & 2926.00 \%\\
%PFNR, c = 50.0 \%, \textcolor{red}{$f$-NeRV3} &  36.59 & 39.95 & 34.46 & 38.01 & 34.21 & 27.26 & 37.47 & 29.07 & 26.72 & 36.17 & 22.65 & 24.46 & 32.43	& 30.07 &	27.62 &	29.30 &	35.26 &	31.65 / 0.0 & 3081.00 \%\\
%PFNR, c = 70.0 \%, \textcolor{red}{$f$-NeRV3} &  38.24 & 39.47 & 33.72 & 35.18 & 31.58 & 25.65 & 33.64 & 26.23 & 23.92 & 35.75 & 20.81 & 22.73 & 30.54 &	28.55 & 26.17 & 26.84 & 33.15 & 30.13 / 0.00 & 3082.00 \% \\
\midrule

MTL (upper-bound) & 32.39 & 34.35  & 31.45 & 34.03 & 30.70  & 24.53 & 37.13 & 27.83 & 23.80 & 34.69 & 20.77 & 22.37 & 32.71 & 28.00 & 25.89 & 26.40 & 33.16 & 29.42 / - ~~~ \\ % & 100.00 \% \\ 
\bottomrule
\end{tabular}
}
\vspace{-0.2in}
\label{table:uvg17_fso_psnr}
\end{table}

%% file: materials/main_table_uvg17_msssim.tex
\begin{table}[ht]
\small\centering
\vspace{-0.3in}
\caption{\small MS-SSIM results with Fourier Subnueral Operator (FSO) layer (\textcolor{red}{$f$-NeRV$\ast$}) (detailed in \Cref{table:architecture_detail}) on UVG17 Video Sessions with average MS-SSIM and Backward Transfer (BWT) of MS-SSIM. Note that $\ast$ denotes our reproduced results.}
\resizebox{\textwidth}{!}{
\renewcommand{\arraystretch}{1.2}
\begin{tabular}{lccccccccccccccccccc}
\toprule 

\multicolumn{1}{c}{\multirow{2}{*}{\textbf{Method}}} & \multicolumn{17}{c}{\textbf{Video Sessions}} & \multirow{2}{*}{\thead{\textbf{Avg. MS-SSIM} \\ \textbf{BWT}}} \\

\cline{2-18}

& \textbf{1} & \textbf{2} & \textbf{3} & \textbf{4} & \textbf{5} & \textbf{6} & \textbf{7} & \textbf{8} & \textbf{9} & \textbf{10} & \textbf{11} & \textbf{12} & \textbf{13} & \textbf{14} & \textbf{15} & \textbf{16} & \textbf{17} \\ \midrule 
STL, NeRV~\cite{chen2021nerv}$^{\ast}$   & 0.99 & 0.99 & 0.95 & 0.98 & 0.98 & 0.96 & 0.99 & 0.98 & 0.97 & 0.95 & 0.96 & 0.96 & 0.98 & 0.98 & 0.96 & 0.99 & 0.99 & 0.97 / - ~~~~ \\ \midrule 

%& 99.61 & 98.21 & 96.10 & 98.37 & 94.57 & 96.27 & 98.03 & 99.13 & 99.37 & 97.74/- \\ \midrule 

EWC~\cite{Kirkpatrick2017}$^{\ast}$ & 0.26 & 0.24  & 0.44 & 0.24 & 0.40  & 0.29 & 0.15 & 0.17 & 0.26 & 0.26 & 0.17 & 0.34 & 0.04 & 0.30 & 0.33 & 0.31 & 0.91 & 0.30 / -0.55  \\

iCaRL~\cite{rebuffi2017icarl}$^{\ast}$ & 0.74 & 0.88  & 0.67 & 0.67 & 0.64  & 0.48 & 0.91 & 0.53 & 0.37 & 0.82 & 0.35 & 0.53 & 0.75 & 0.70 & 0.61 & 0.60 & 0.87 & 0.65 / -0.20  \\ 
ESMER~\cite{sarfraz2023error}$^{\ast}$ &  0.85 & 0.86 & 0.64 & 0.63 & 0.66 & 0.46 &	0.89 & 0.51 & 0.42 & 0.79 &	0.30 & 0.51 &	0.43 &	0.68 &  0.82 & 0.6 & 0.63 & 0.62 / -0.37  \\
\midrule

%WSN$^{\ast}$, c = 10.0 \% & 0.90 & 0.94 & 0.88 & 0.92 & 0.87 & 0.75 & 0.96 & 0.74 & 0.69 & 0.91 & 0.57 & 0.72 & 0.86 & 0.80 & 0.74 & 0.69 & 0.92 & 0.82 / 0.0  \\ 
WSN$^{\ast}$, c = 30.0 \% & 0.96 & 0.97 & 0.89 & {{0.93}} & {{0.88}} & {{0.77}} & {{0.97}} & {{0.77}} & {{0.73}} & {{0.91}} & {{0.60}} & {{0.74}} & {{0.86}} & {{0.81}} & {{0.76}} & {{0.72}} & {{0.93}} & {{0.84}} / 0.0  \\ 
%WSN$^{\ast}$, c = 50.0 \% & {{0.98}} & {{0.97}} & {{0.89}} & 0.92 & 0.88 & 0.77 & 0.96 & 0.75 & 0.73 & 0.91 & 0.60 & 0.74 & 0.85 & 0.80 & 0.75 & 0.73 & 0.92 & 0.83 / 0.0  \\ 
%WSN$^{\ast}$, c = 70.0 \% & 0.98 & 0.97 & 0.88 & 0.91 & 0.85 & 0.75 & 0.95 & 0.70 & 0.68 & 0.91 & 0.55 & 0.72 & 0.85 & 0.80 & 0.73 & 0.70 & 0.92 & 0.82 / 0.0  \\ 

% reinit
% WSN, c = 10.0 \%  & 0.91 & 0.96 & 0.90 & 0.94 & 0.89 & 0.80 & 0.98 & 0.84 & 0.78 & 0.93 & 0.68 & 0.78 & 0.91 & 0.86 & 0.81 & 0.82 & 0.96 & 0.87 / 0.0 \\ 
% WSN, c = 30.0 \%  & 0.96 & 0.98 & 0.91 & 0.95 & 0.91 & 0.83 & 0.98 & 0.87 & 0.81 & 0.93 & 0.72 & 0.81 & 0.92 & 0.88 & 0.83 & 0.86 & 0.97 & 0.89 / 0.0 \\ 
% WSN, c = 50.0 \% & 0.98 & 0.98 & 0.91 & 0.95 & 0.91 & 0.82 & 0.98 & 0.85 & 0.80 & 0.93 & 0.71 & 0.80 & 0.91 &  0.88 & 0.83 & 0.86 & 0.96 & 0.89 / 0.0  \\ 
% WSN, c = 70.0 \% & 0.98 & 0.98 & 0.90 & 0.92 & 0.87 & 0.78 & 0.95 & 0.77 & 0.76 & 0.92 & 0.64 & 0.77 & 0.88 & 0.84 & 0.79 & 0.81 & 0.95 &  0.85 / 0.0  \\ 
%\midrule

%PFNR, c = 10.0 \%, \textcolor{red}{$f$-NeRV2} & 0.92 &	0.97 &	0.90 &	0.94 &	0.90 &	0.81 &	0.98 &	0.86 &	0.79 &	0.93 &	0.69 &	0.79 &	0.91 &	0.87 &	0.82 &	0.84 &	0.96 &	0.88  / 0.0  \\
PFNR~, c = 30.0 \%, \textcolor{red}{$f$-NeRV2} & 0.97 &	0.98 &	0.92 &	0.95 &	0.92 &	0.84 &	0.98 &	0.88 &	0.82 &	0.93 &	0.73 &	0.82 &	0.92 &	0.89 &	0.84 &	0.87 &	0.97 &	0.90 / 0.0 \\
%PFNR, c = 50.0 \%, \textcolor{red}{$f$-NeRV2} & 0.98 &	0.99 &	0.92 &	0.95 &	0.92 &	0.83 &	0.98 &	0.86 &	0.81 &	0.93 &	0.72 &	0.82 &	 0.91 &	0.88 &	0.84 &	0.87 &	0.97 &	0.89 / 0.0 \\
%PFNR, c = 70.0 \%, \textcolor{red}{$f$-NeRV2} & 0.99 & 0.98 & 0.91 & 0.93 & 0.87 & 0.78 & 0.96 & 0.77 & 0.77 & 0.93 & 0.66	& 0.77 & 0.89 &	0.85 & 0.80 &	0.82 & 0.95 & 0.86 / 0.0  \\
%\midrule

%PFNR, c = 10.0 \%, \textcolor{red}{$f$-NeRV3} & 0.96 &	0.99 &	0.92 &	0.96 &	0.94 &	0.86 &	0.99 &	0.94 &	0.82 &	0.93 &	0.76 &	0.83 &	0.93 &	0.92 &	0.87 &	0.90 &	0.98 &	0.91 / 0.0 \\
PFNR~, c = 30.0 \%, \textcolor{red}{$f$-NeRV3} & \textbf{0.98} &	\textbf{0.99} &	\textbf{0.93} &	\textbf{0.97} &	\textbf{0.96} &	\textbf{0.91} &	\textbf{0.99} &	\textbf{0.96} &	\textbf{0.87} &	\textbf{0.94} &	\textbf{0.84} &	\textbf{0.87} &	\textbf{0.94} &	\textbf{0.94} &	\textbf{0.90} & 	\textbf{0.94} &	\textbf{0.98} &  \textbf{0.94}  / \textbf{0.0} \\
%PFNR, c = 50.0 \%, \textcolor{red}{$f$-NeRV3} & 0.99 &	0.99 &	0.93 &	0.97 &	0.96 &	0.91 &	0.99 &	0.95 &	0.87 &	0.94 &	0.83 &	0.88 &	 0.94 &	0.94 &	0.9 &	0.95 &	0.98 &	0.94 / 0.0 \\
%PFNR, c = 70.0 \%, \textcolor{red}{$f$-NeRV3} & 0.99 & 0.99 & 0.93 & 0.96 & 0.93 & 0.87 & 0.98 & 0.90 & 0.81 & 0.93	 & 0.76 & 0.83 & 0.93 & 0.92 & 0.87 &	0.91  & 0.97 & 0.91 / 0.0 \\
\midrule

MTL (upper-bound) & 0.97 & 0.97  & 0.90 & 0.94 & 0.91  & 0.82 & 0.99 & 0.92 & 0.80 & 0.92 & 0.75 & 0.81 & 0.94 & 0.90 & 0.85 & 0.89 & 0.97 & 0.90 / - ~~~  \\ 
\bottomrule
\end{tabular}
}
\vspace{-0.12in}
\label{table:uvg17_fso_msssim}
\end{table}

%% file: materials/plot_main_psnr_bpp.tex
\begin{figure}[h]
    \centering
    \small
    \vspace{-0.1in}
    \setlength{\tabcolsep}{0pt}{%
    \begin{tabular}{cc}
    \includegraphics[width=0.5\columnwidth]{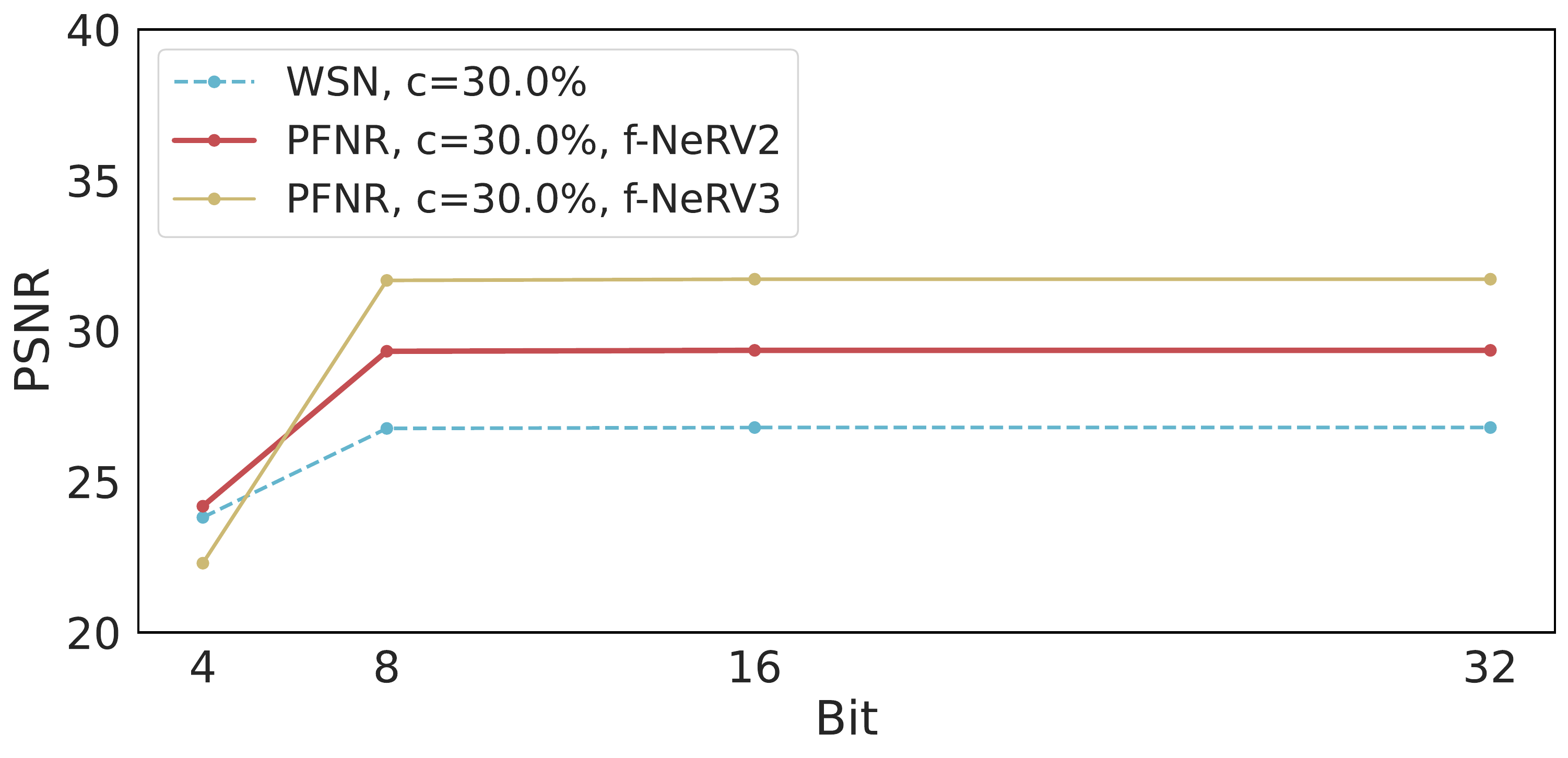} &
    \includegraphics[width=0.5\columnwidth]{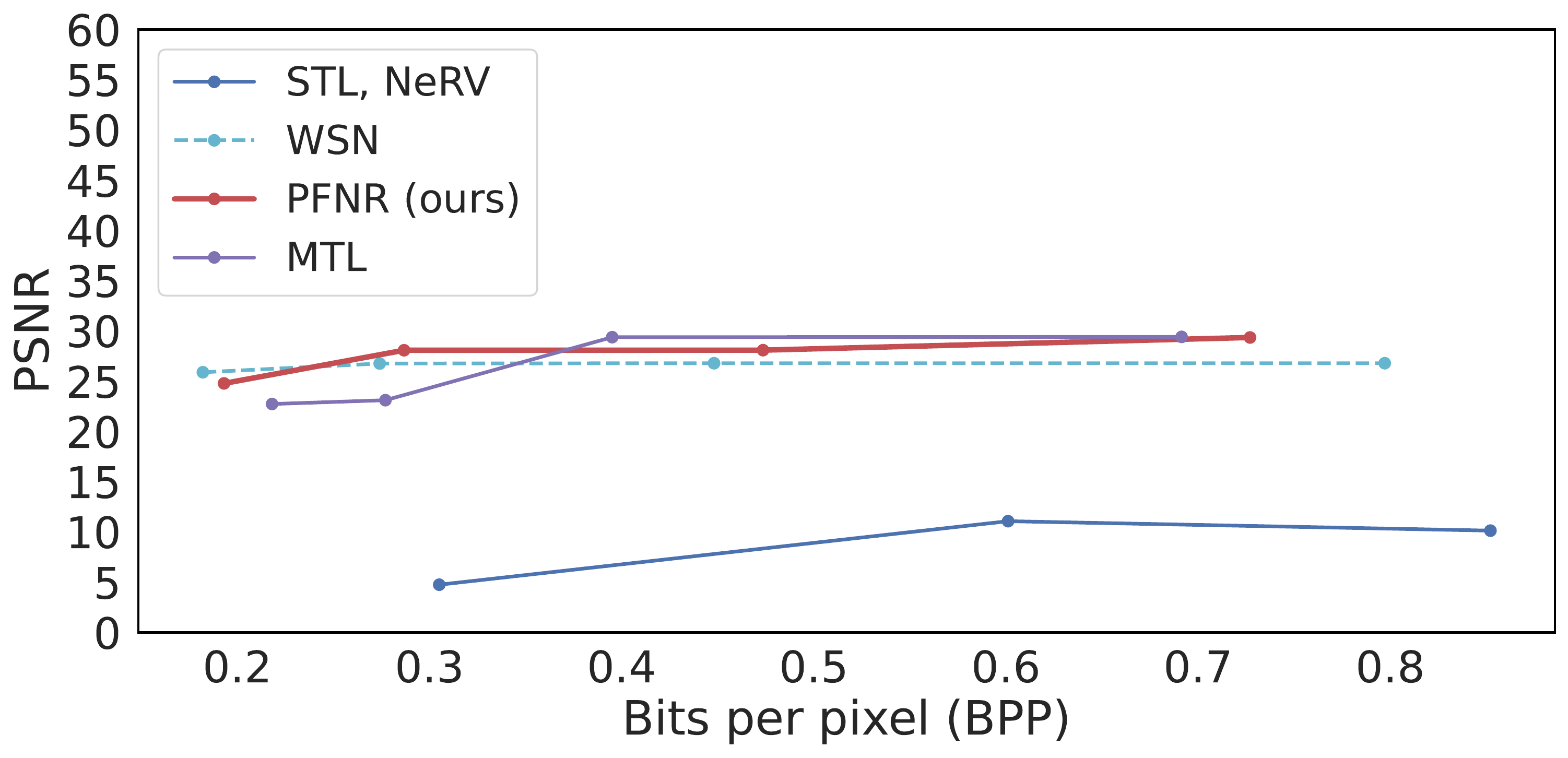} \\
    \small (a) Quantization and Compression of PFNR & \small (b) PSNR v.s. BPP
    \end{tabular}
    }
    \vspace{-0.1in}
    \caption{\small PSNR v.s. Bits-per-pixel (BPP) on the UVG17 datasets }
    \label{fig:uvg17_fso_psnr}
    \vspace{-0.12in}
\end{figure}

%% file: materials/plot_capacity.tex
\begin{figure}[ht]
    \centering
    \small
    %\vspace{-0.1in}
    \setlength{\tabcolsep}{0pt}{%
    \begin{tabular}{cc}
    % we prepare the figure 
    \includegraphics[width=0.5\columnwidth]{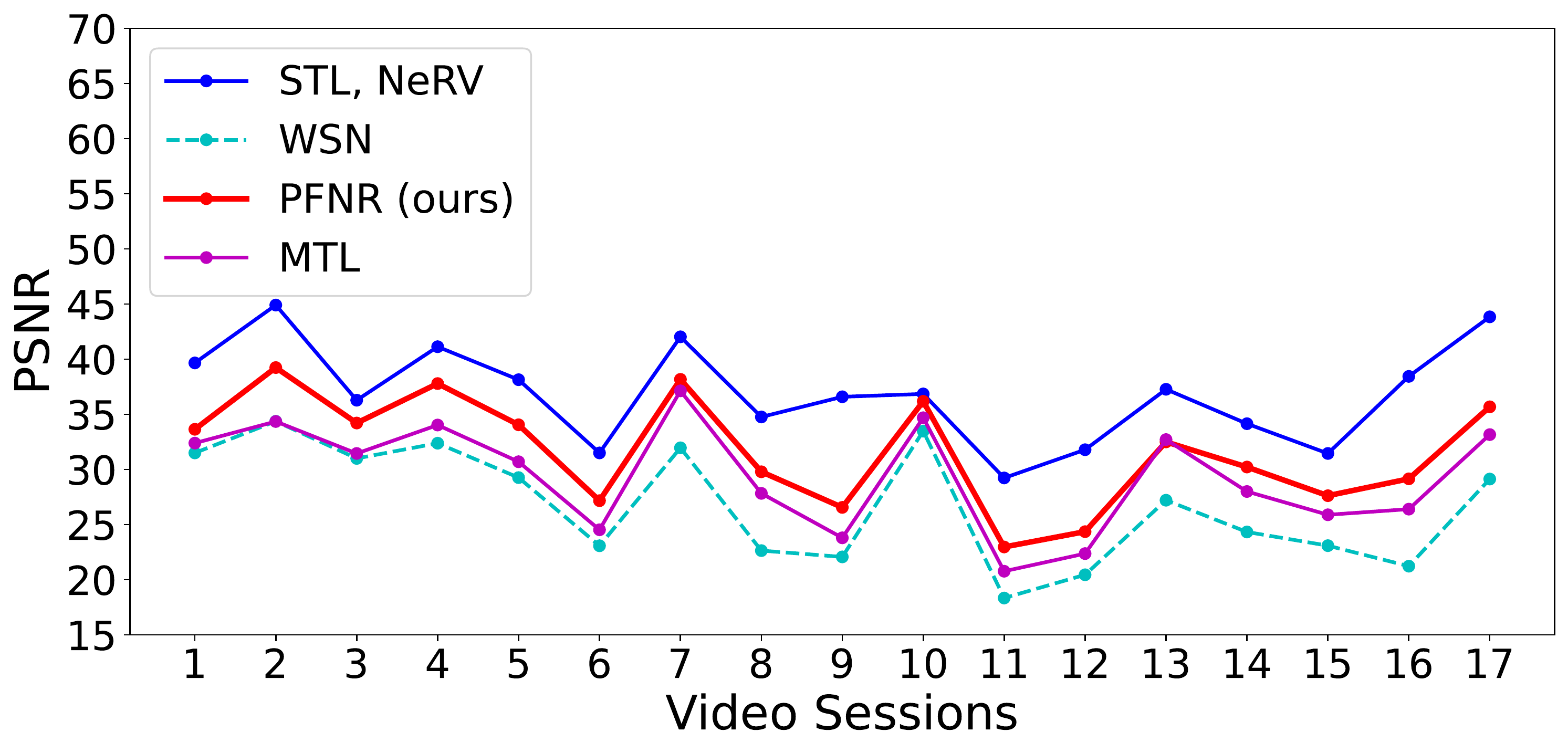} &
    \includegraphics[width=0.5\columnwidth]{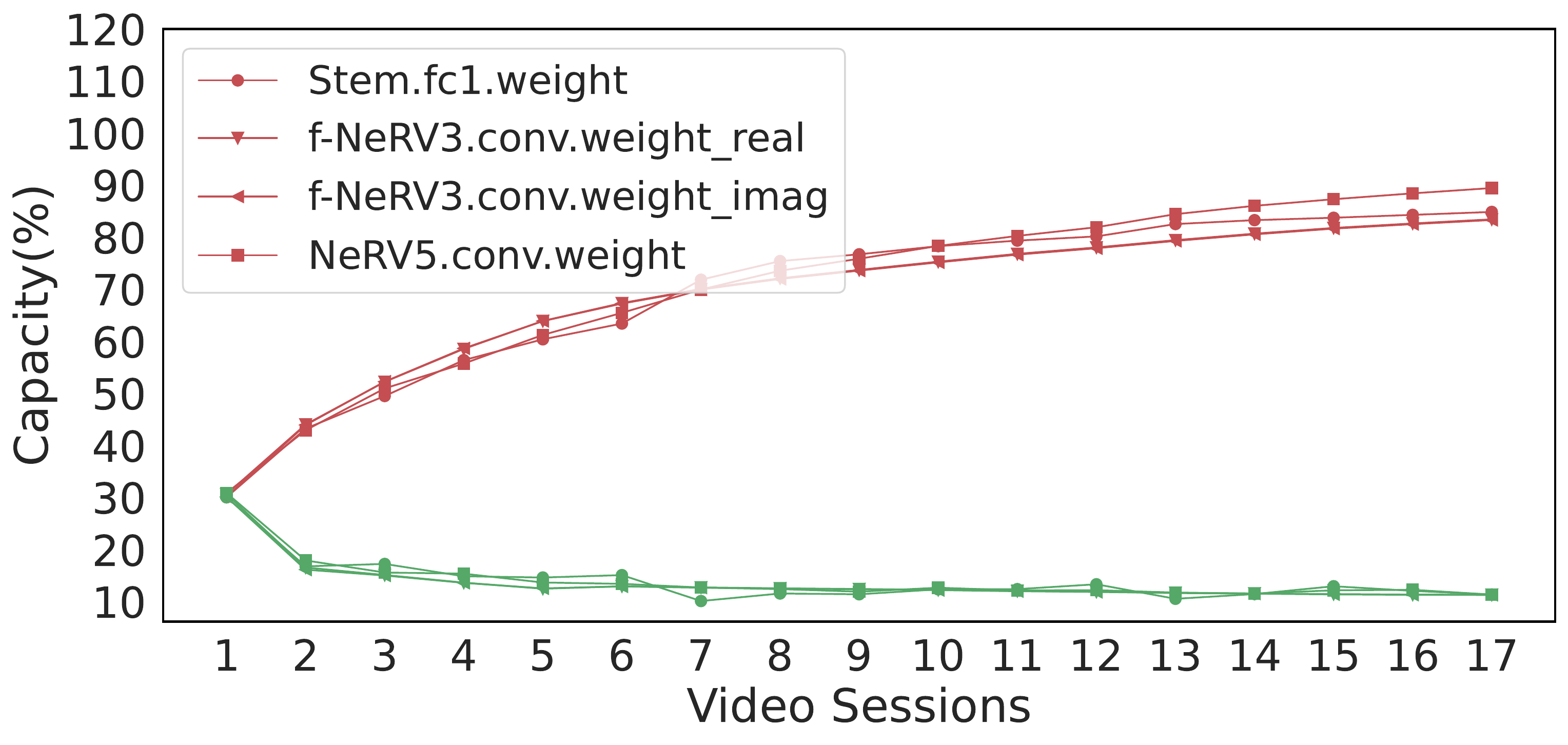} \\
    \small (a) PFNR's performance with $c = 30.0\%$ & \small (b) PFNR'capacity with $c = 30.0\%$ \\
    \end{tabular}
    }
    \vspace{-0.1in}
    \caption{\small PFNR's Comparison of PSNR with others and layer-wise accumulated capacities on the UVG17 dataset. Note that, in (b), \textcolor{Green}{green} represents the percentage of reused subnetwork's parameters of Stem, $f$-NeRV3, and NeRV5 at the current session (s) obtained at the past (s-1) video sessions}
    \label{fig:psnr_cap}
    \vspace{-0.12in}
\end{figure}

%% file: materials/plot_main_fmap_uvg17.tex
\begin{figure*}[h]
    \centering 
    \vspace{-0.1in}
    \setlength{\tabcolsep}{0pt}{%
    \begin{tabular}{ccccc}
    
    % 2.city 
    %\textit{2.city} & \textit{3.beauty} & \textit{7.bee} \\ 
     NeRV3 & NeRV4 & NeRV5 & NeRV6 & Head \\ 

    %\makecell{\small WSN \\ \tiny c=50\% } & 
    %\includegraphics[width=0.2\columnwidth]{images/fmaps/5/nerv3/gt_0.png} &  
    %\includegraphics[width=0.3\columnwidth]{images/fmaps/5/wsn/pred_0_l1.pdf} & % NeRV2
    \includegraphics[width=0.20\columnwidth, trim={0.1cm 0.4cm 0.1cm 0.1cm}, clip]{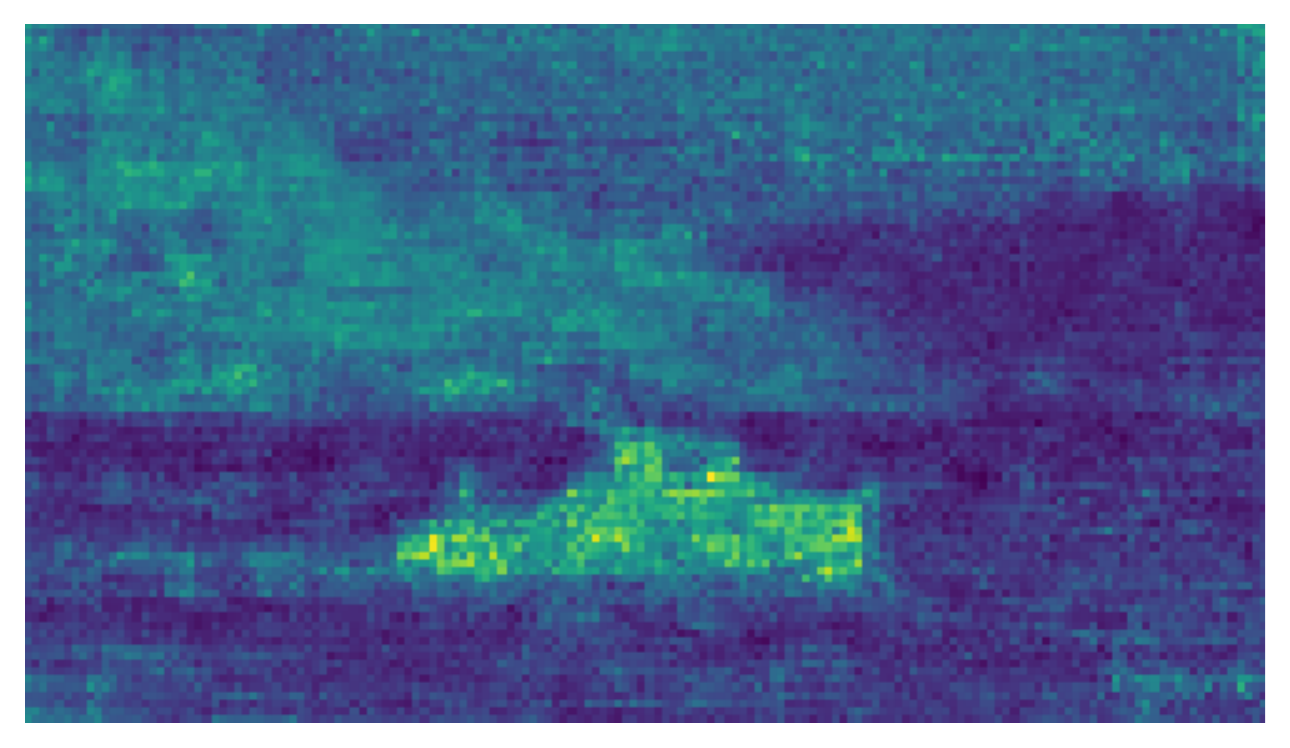} & % NeRV3
    \includegraphics[width=0.20\columnwidth, trim={0.1cm 0.4cm 0.1cm 0.1cm}, clip]{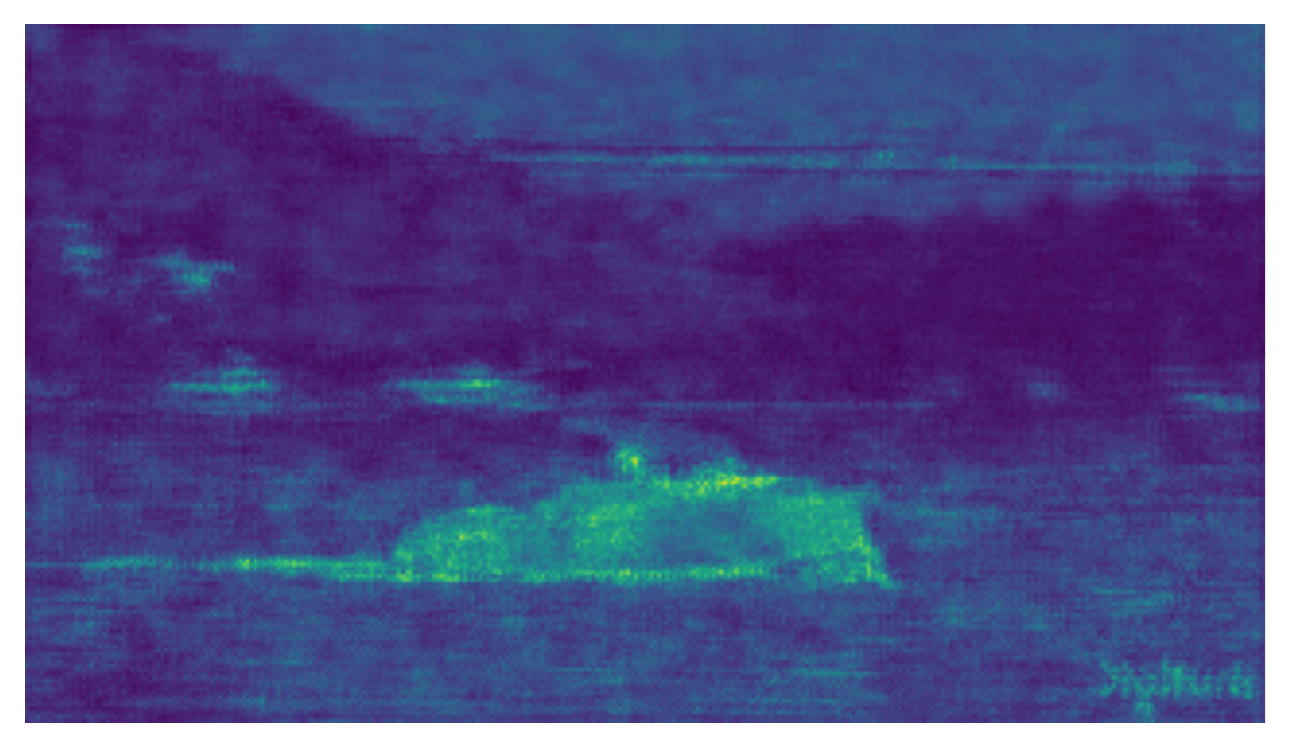} & % NeRV4
    \includegraphics[width=0.20\columnwidth, trim={0.1cm 0.4cm 0.1cm 0.1cm}, clip]{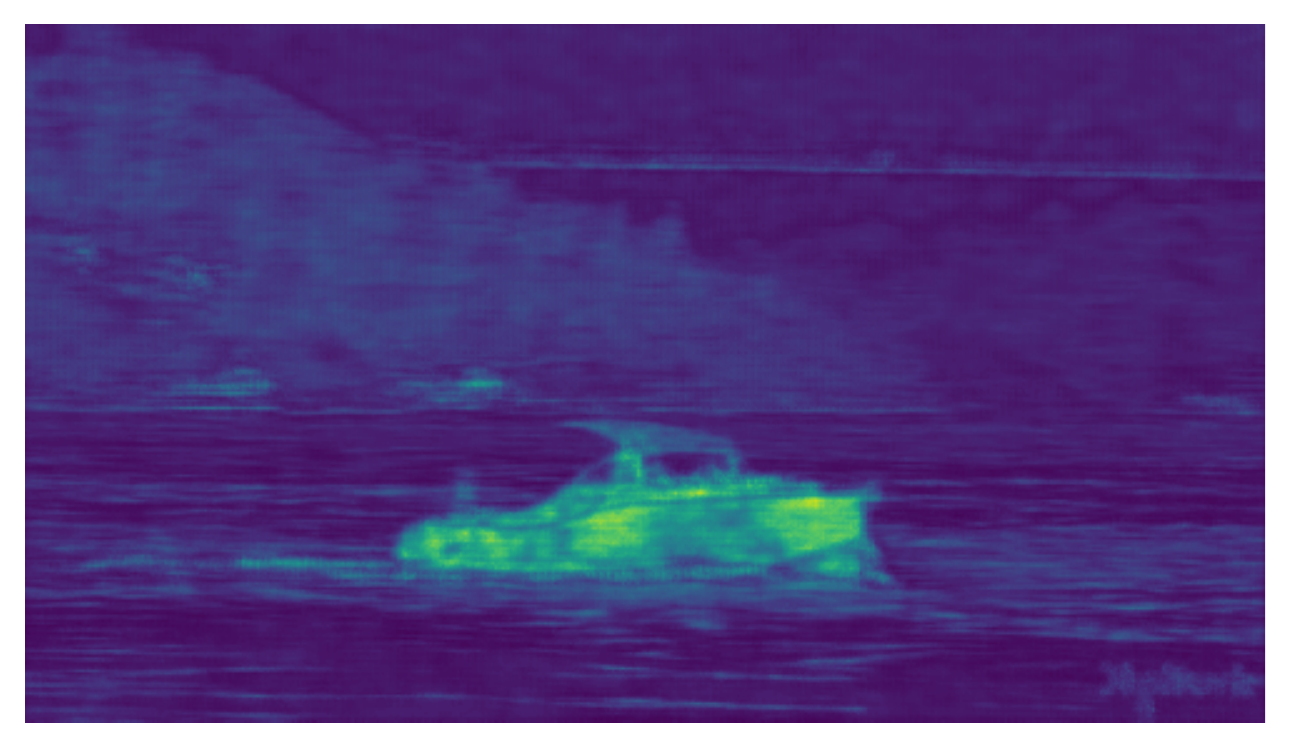} & % NeRV5
    \includegraphics[width=0.20\columnwidth, trim={0.1cm 0.4cm 0.1cm 0.1cm}, clip]{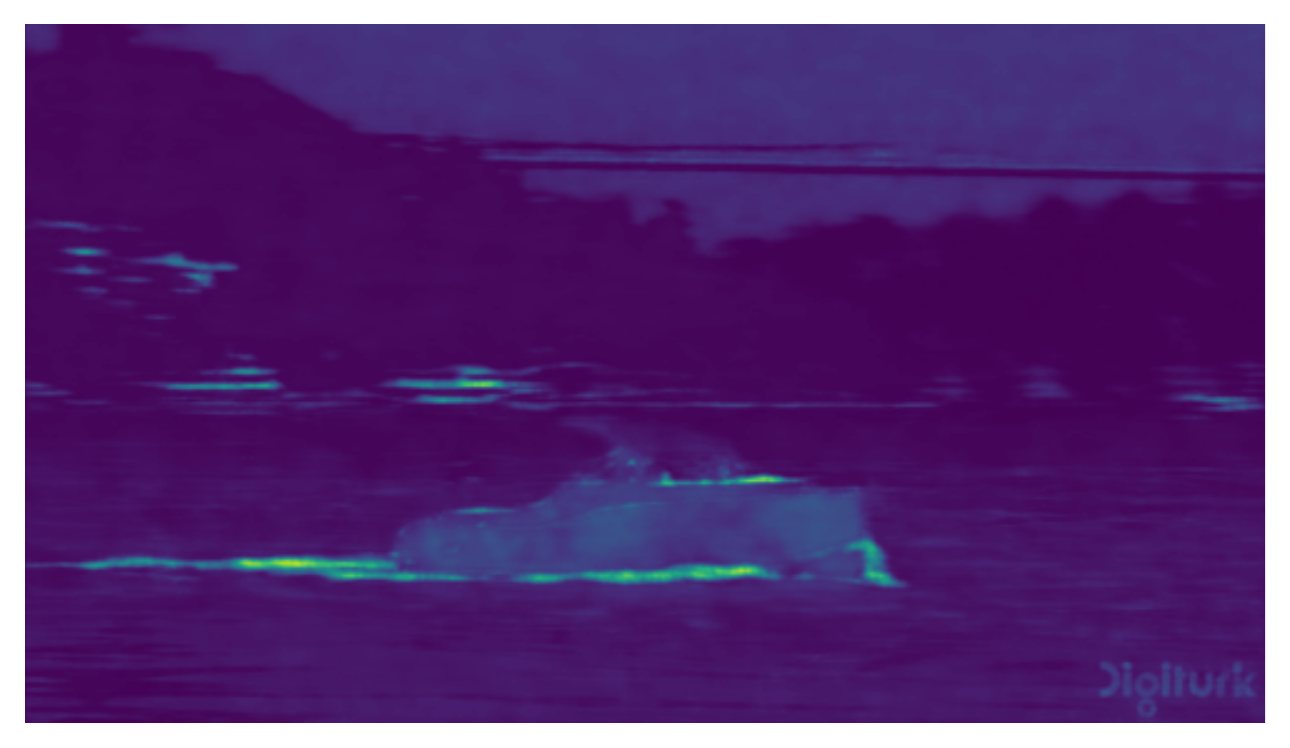} & % NeRV6
    \includegraphics[width=0.19\columnwidth]{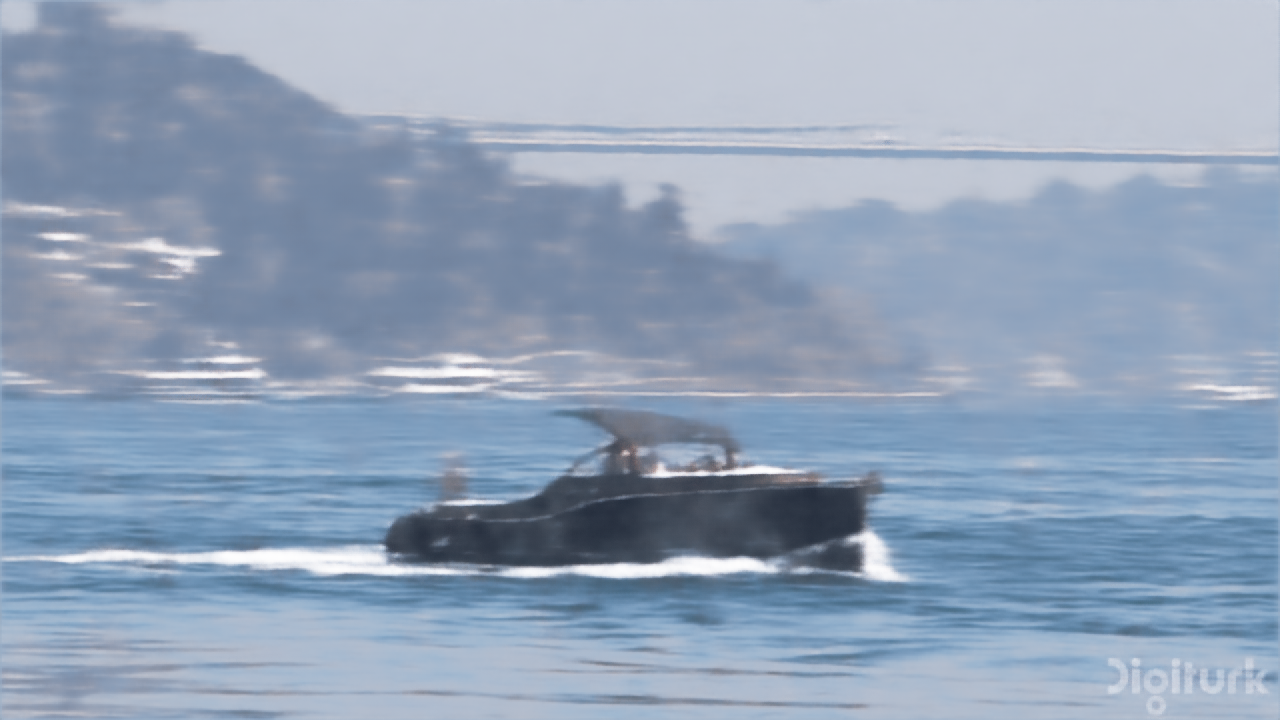} \\

    \multicolumn{5}{l}{\small WSN, c=50.0\% (28.95, PSNR)} \\
    
    %\makecell{\small PFNR \\ \tiny c=50\% \\ \tiny $f$-NeRV2} & 
    %\includegraphics[width=0.2\columnwidth]{images/fmaps/5/nerv3/gt_0.png} &  
    %\includegraphics[width=0.3\columnwidth]{images/fmaps/5/nerv3/pred_0_l1.pdf} & % NeRV2
    \includegraphics[width=0.20\columnwidth, trim={0.1cm 0.4cm 0.1cm 0.1cm}, clip]{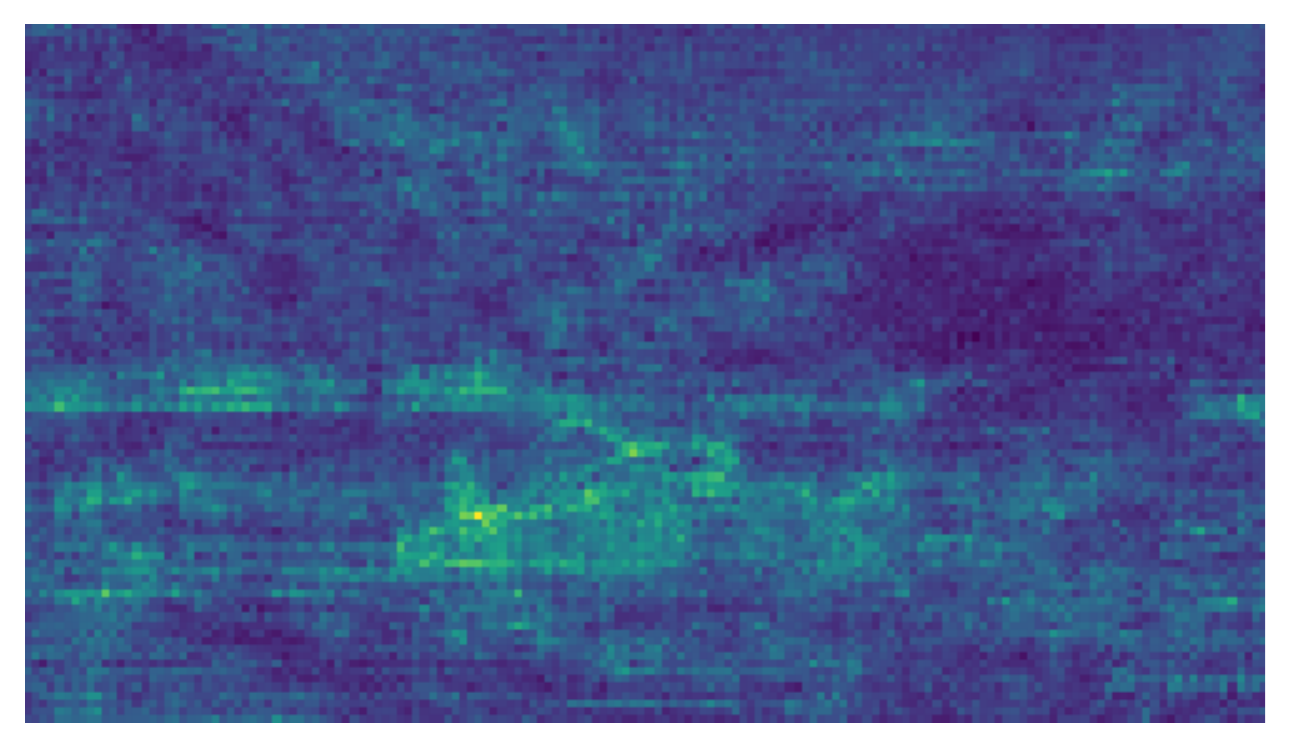} & % NeRV3
    \includegraphics[width=0.20\columnwidth, trim={0.1cm 0.4cm 0.1cm 0.1cm}, clip]{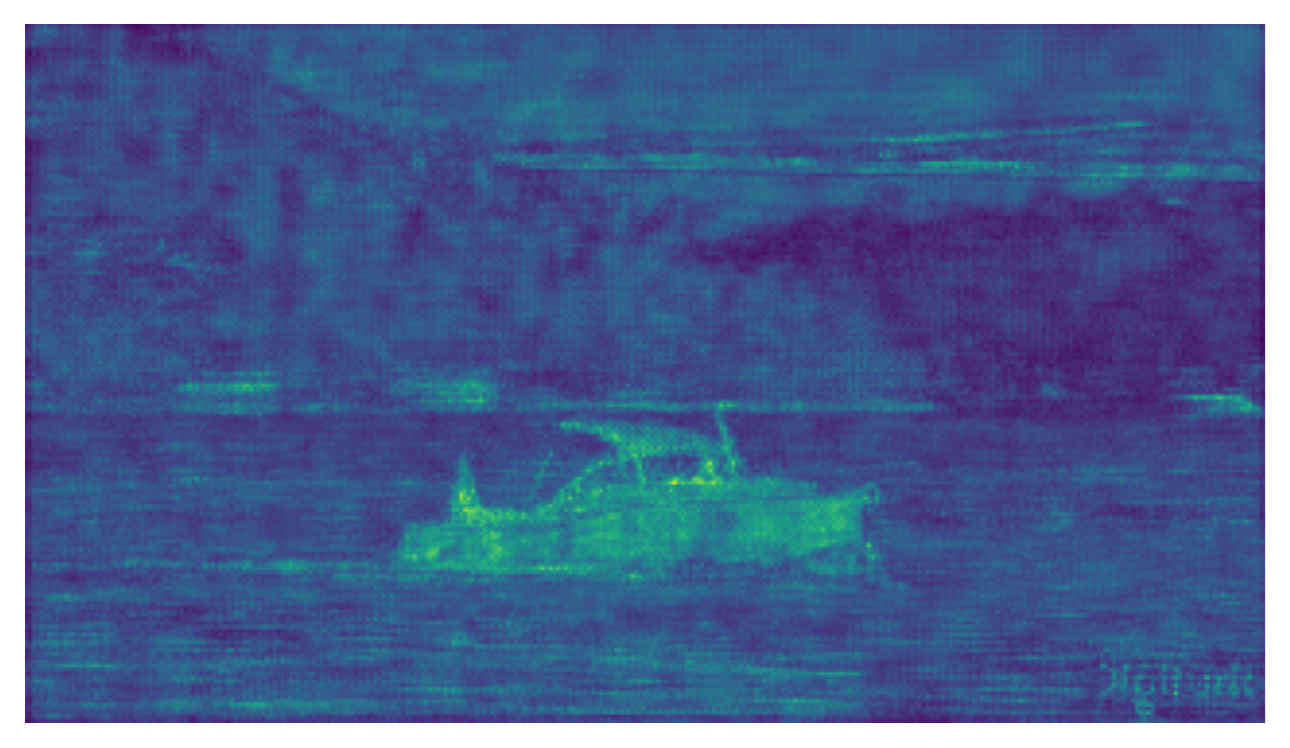} & % NeRV4
    \includegraphics[width=0.20\columnwidth, trim={0.1cm 0.4cm 0.1cm 0.1cm}, clip]{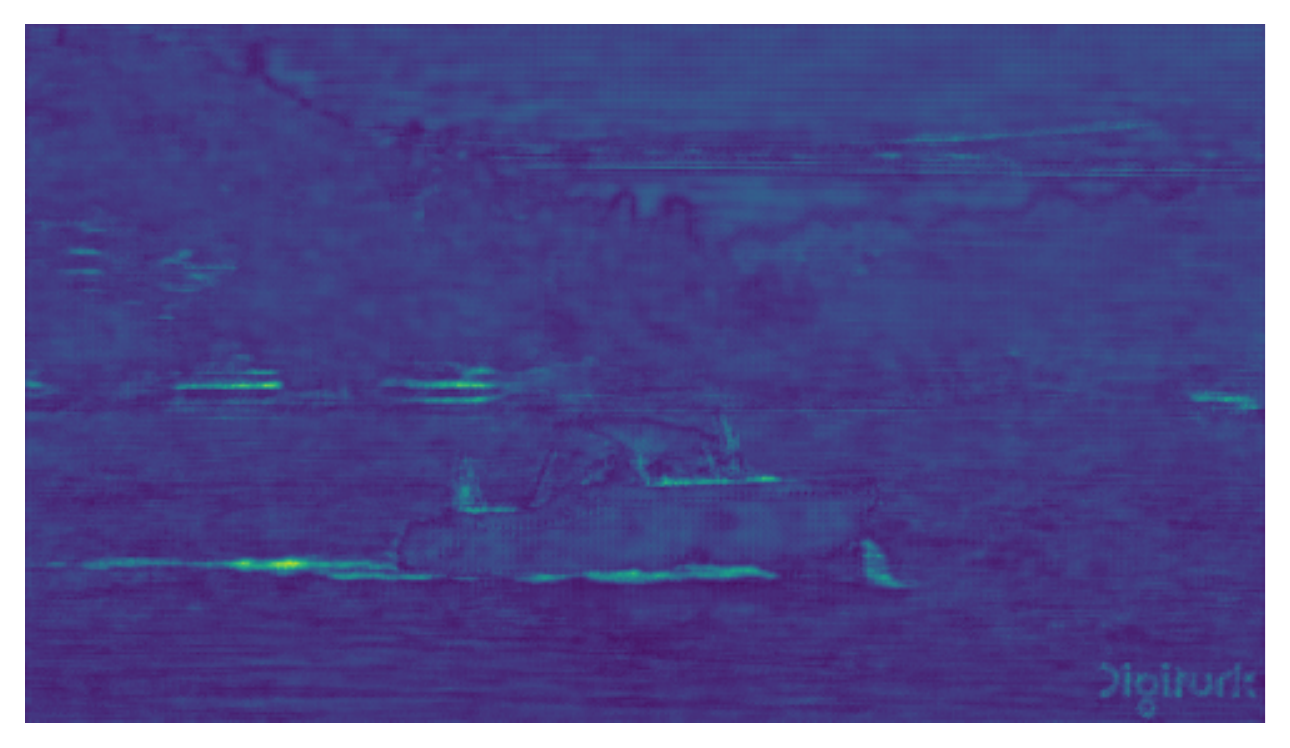} & % NeRV5
    \includegraphics[width=0.20\columnwidth, trim={0.1cm 0.4cm 0.1cm 0.1cm}, clip]{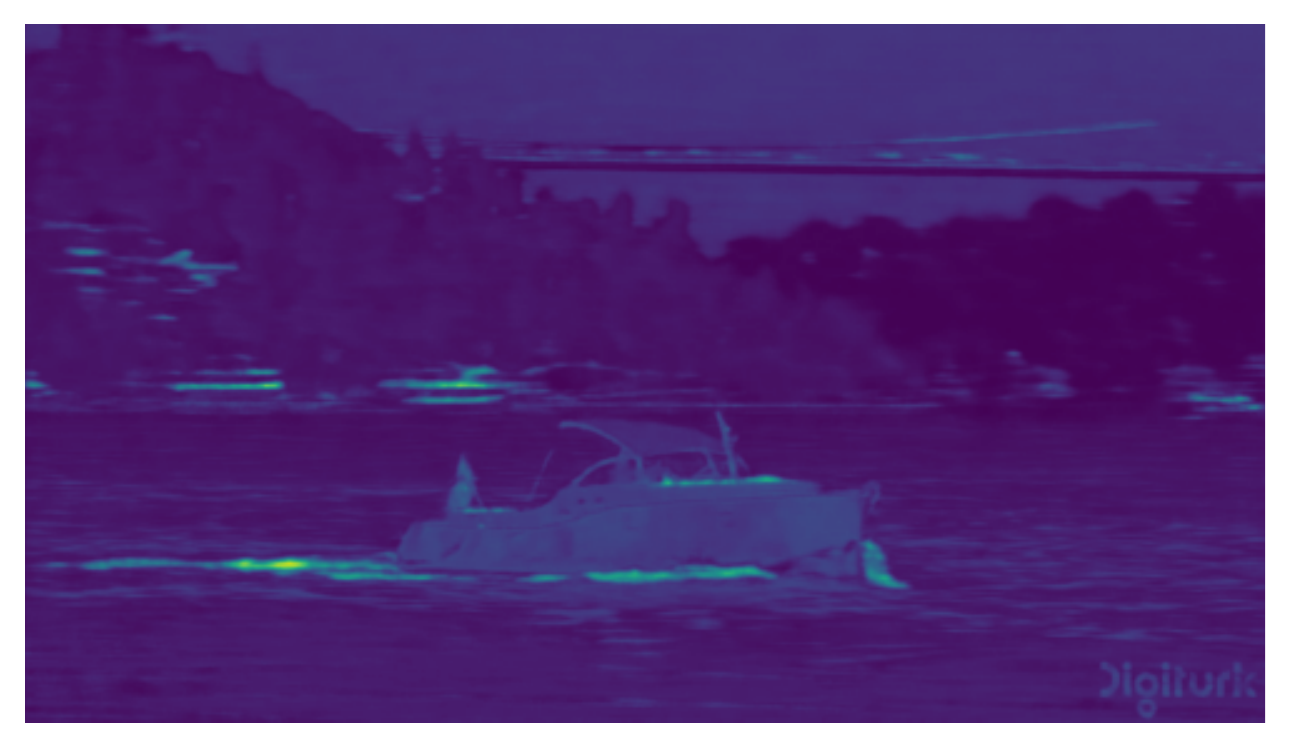} & % NeRV6
    \includegraphics[width=0.19\columnwidth]{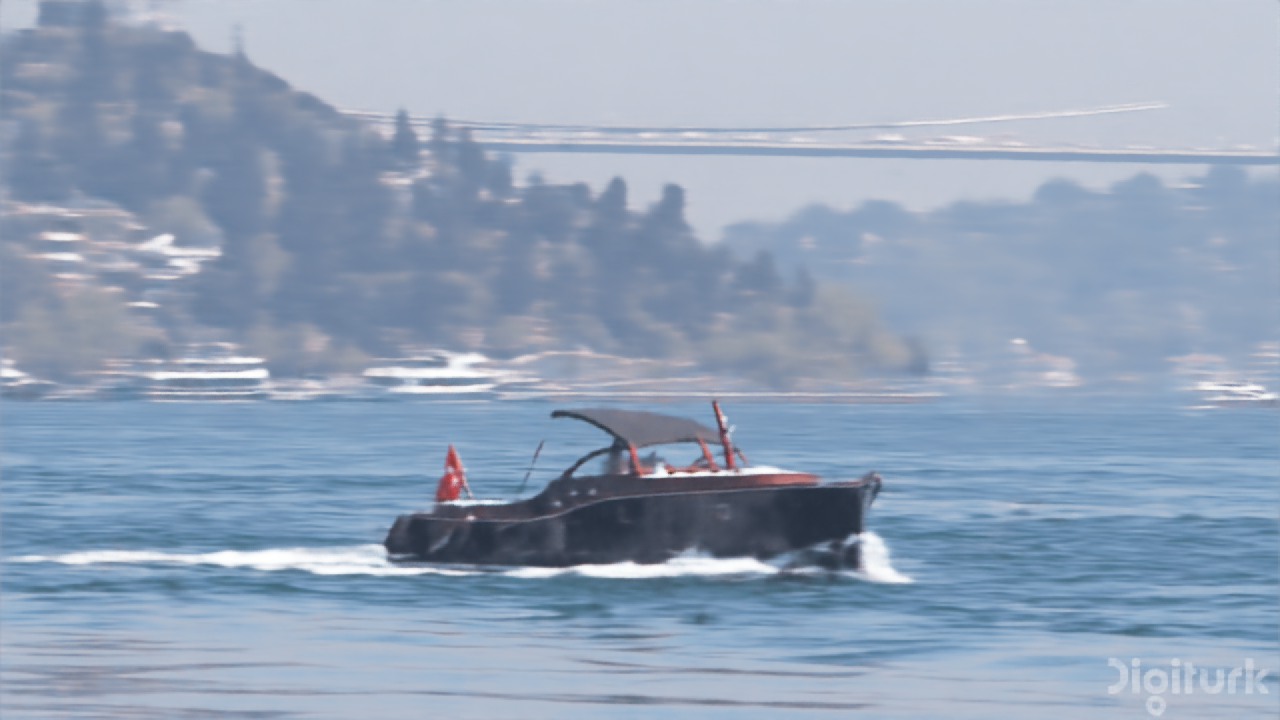} \\

    \multicolumn{5}{l}{\small PFNR, c=50.0\%, \textcolor{red}{$f$-NeRV2} (31.24, PSNR)} \\

    %\makecell{\small PFNR \\ \tiny c=50\% \\ \tiny $f$-NeRV3} & 
    %\includegraphics[width=0.2\columnwidth]{images/fmaps/5/nerv3/gt_0.png} &  
    %\includegraphics[width=0.3\columnwidth]{images/fmaps/5/nerv3/pred_0_l1.pdf} & % NeRV2
    \includegraphics[width=0.20\columnwidth, trim={0.1cm 0.4cm 0.1cm 0.1cm}, clip]{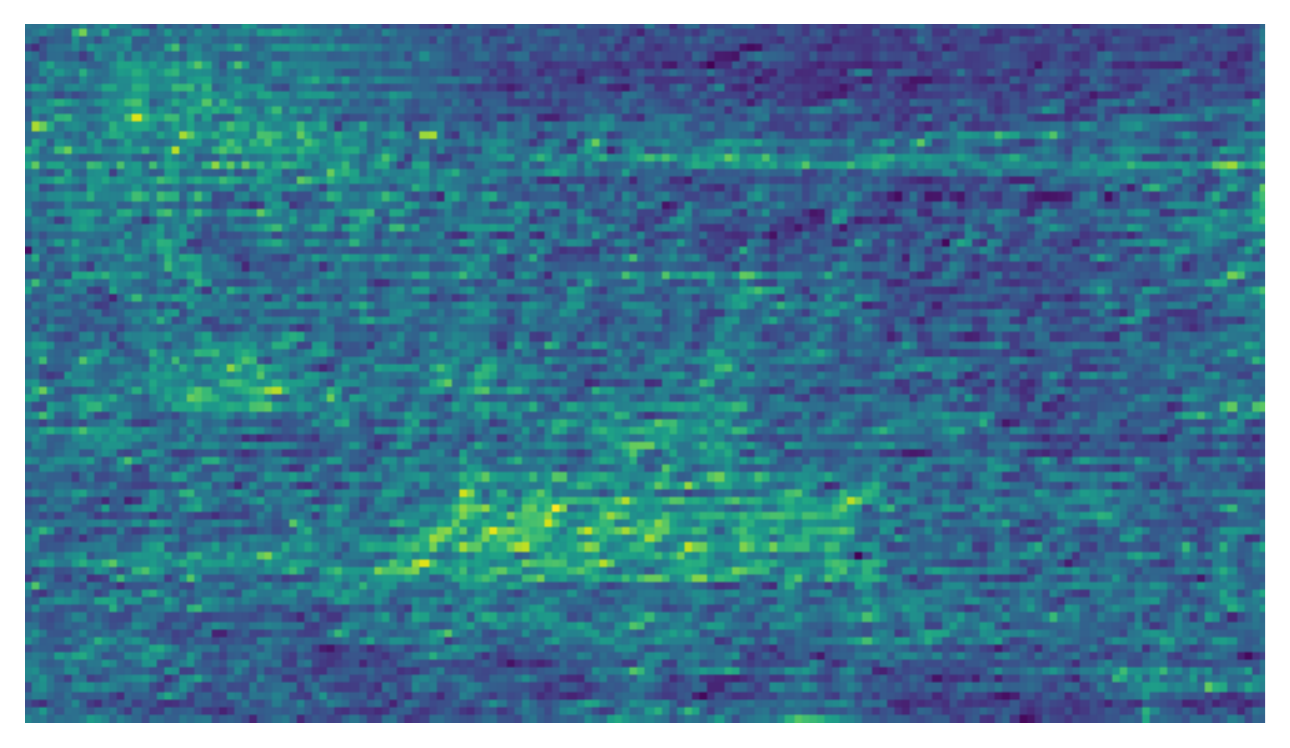} & % NeRV3
    \includegraphics[width=0.20\columnwidth, trim={0.1cm 0.4cm 0.1cm 0.1cm}, clip]{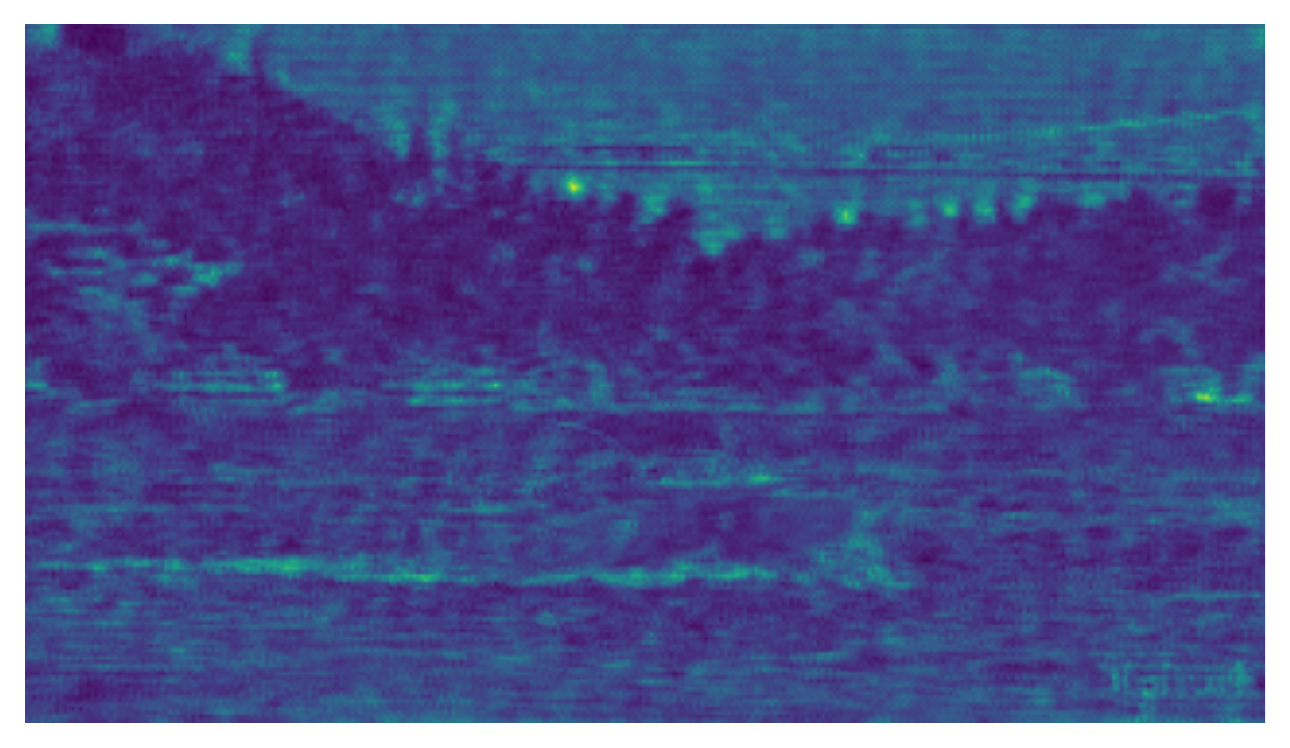} & % NeRV4
    \includegraphics[width=0.20\columnwidth, trim={0.1cm 0.4cm 0.1cm 0.1cm}, clip]{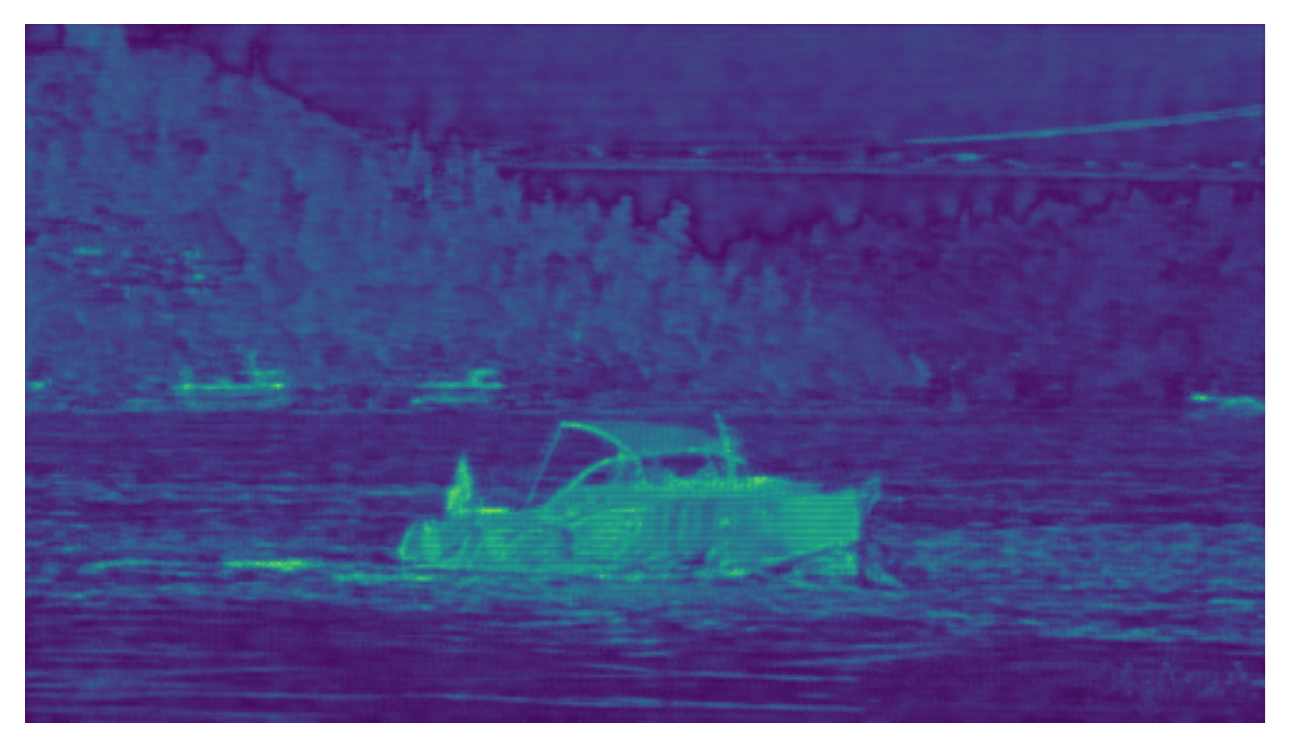} & % NeRV5
    \includegraphics[width=0.20\columnwidth, trim={0.1cm 0.4cm 0.1cm 0.1cm}, clip]{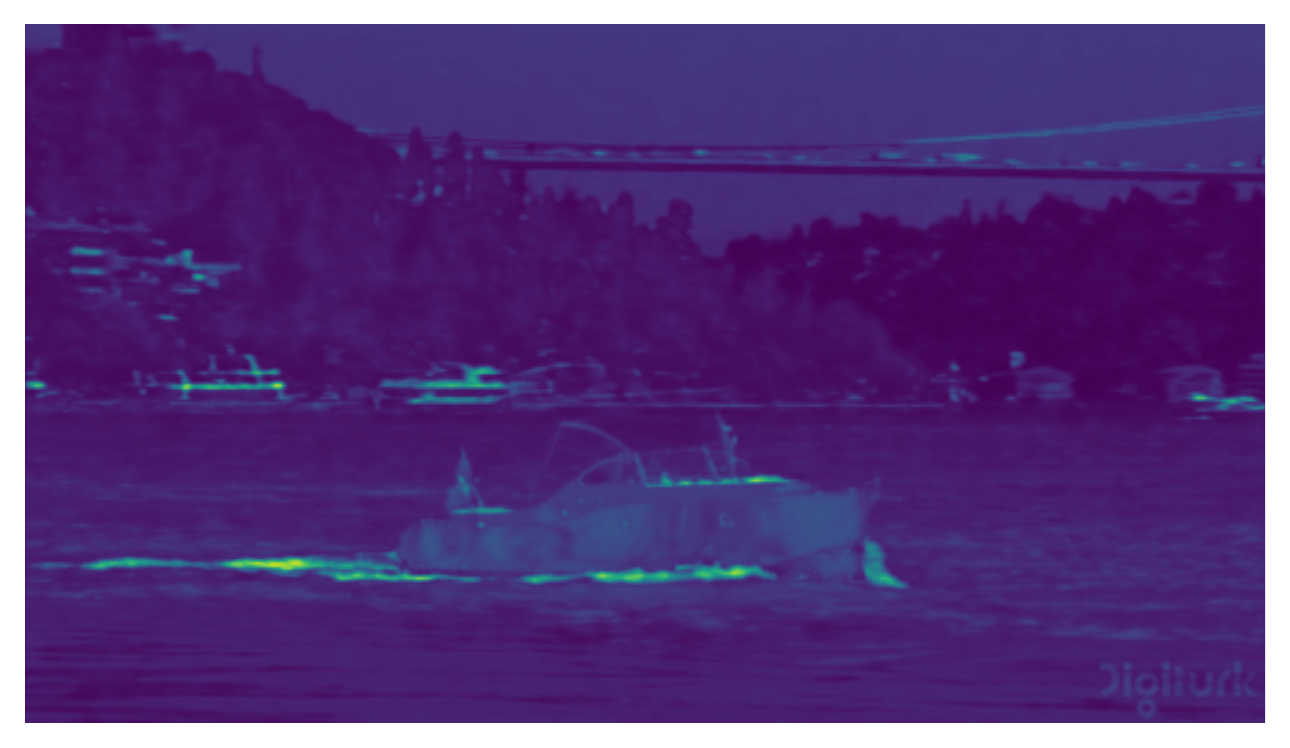} & % NeRV6
    \includegraphics[width=0.19\columnwidth]{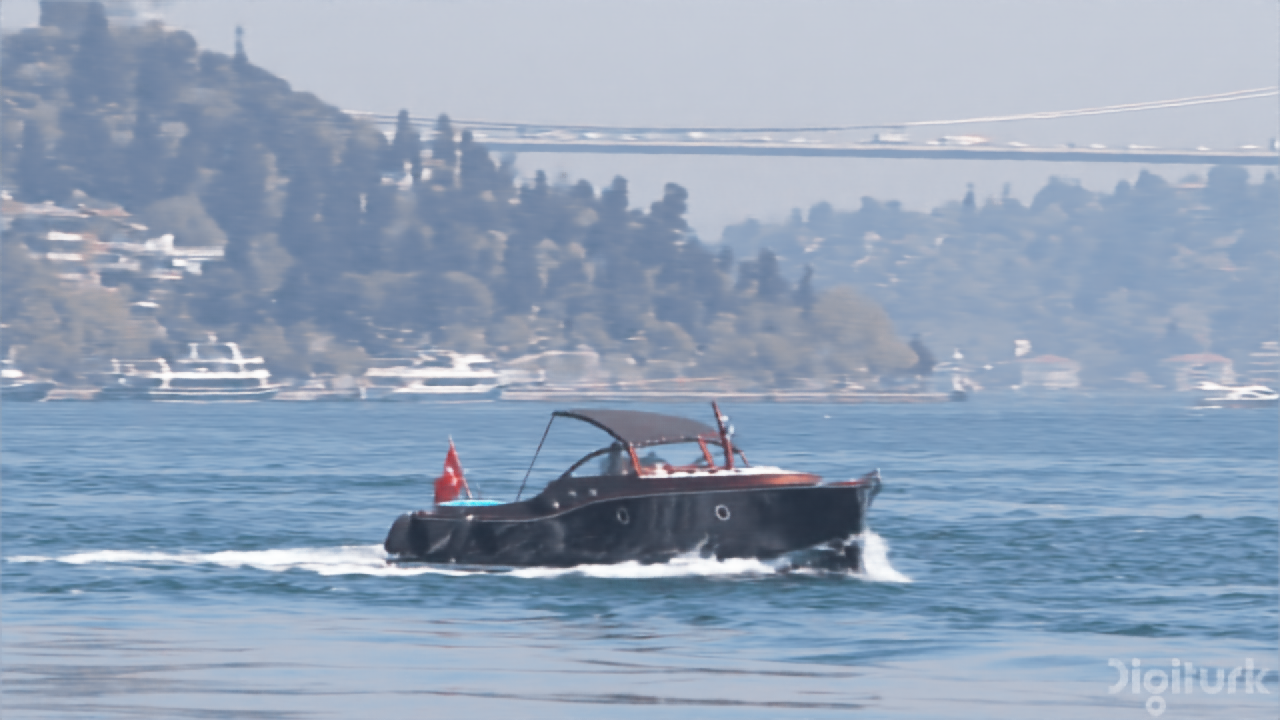} \\

    \multicolumn{5}{l}{\small PFNR, c=50.0\%, \textcolor{red}{$f$-NeRV3} (34.05, PSNR)} \\
   
    \end{tabular}
    }
    \vspace{-0.1in}
    \caption{PFNR's Representations of NeRV Blocks with $c = 50.0 \%$ on the UVG17 dataset.}
    \label{fig:fmap_uvg17}
    
\end{figure*}

%% file: materials/plot_main_video_mtl.tex
\begin{figure}[ht]
    \centering 
    \vspace{-0.1in}
    \setlength{\tabcolsep}{0pt}{%
    \begin{tabular}{cccc}
    
    % 2.city 
    t=0 & t=1 & t=2 &t=3 \\

    \includegraphics[width=0.22\columnwidth]{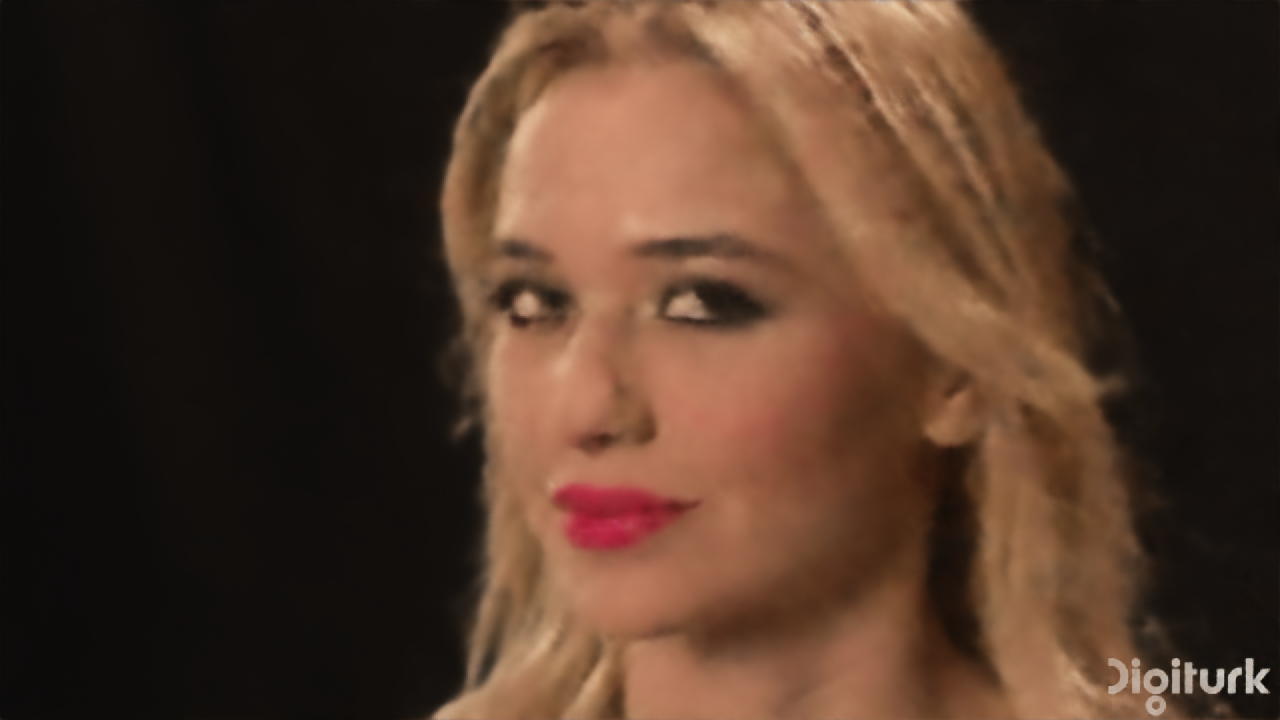} & 
    \includegraphics[width=0.22\columnwidth]{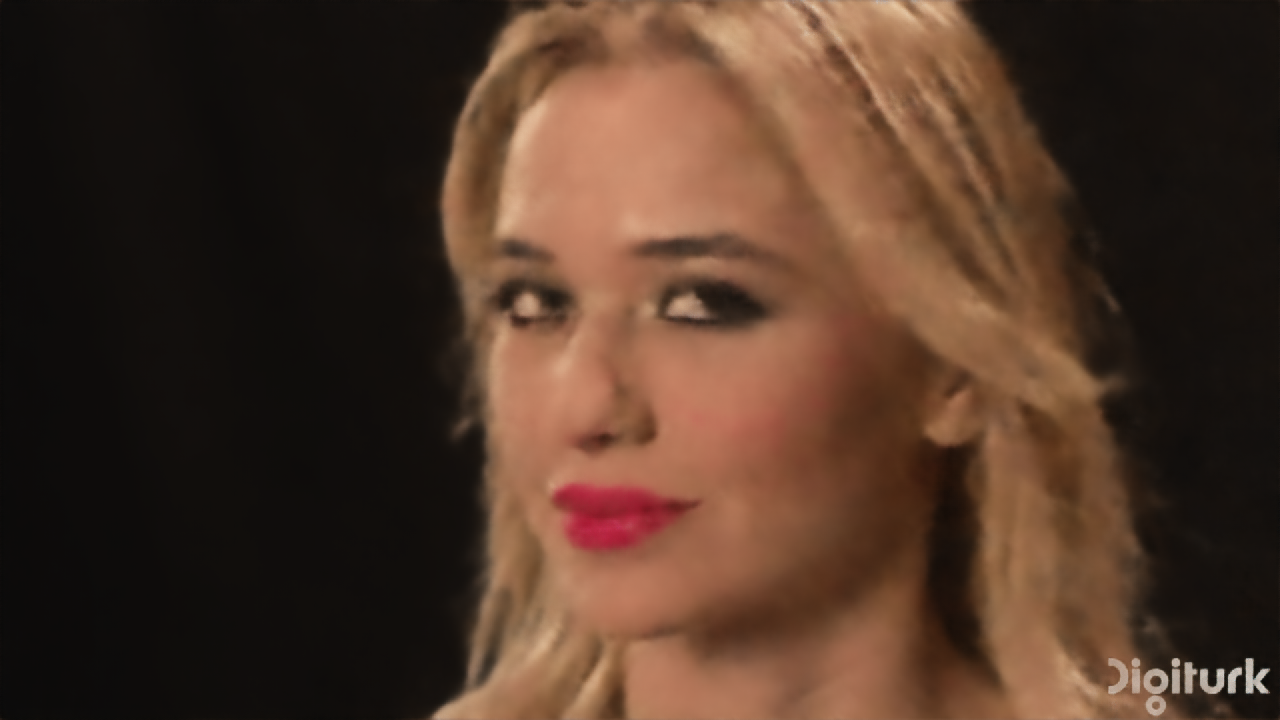} &
    \includegraphics[width=0.22\columnwidth]{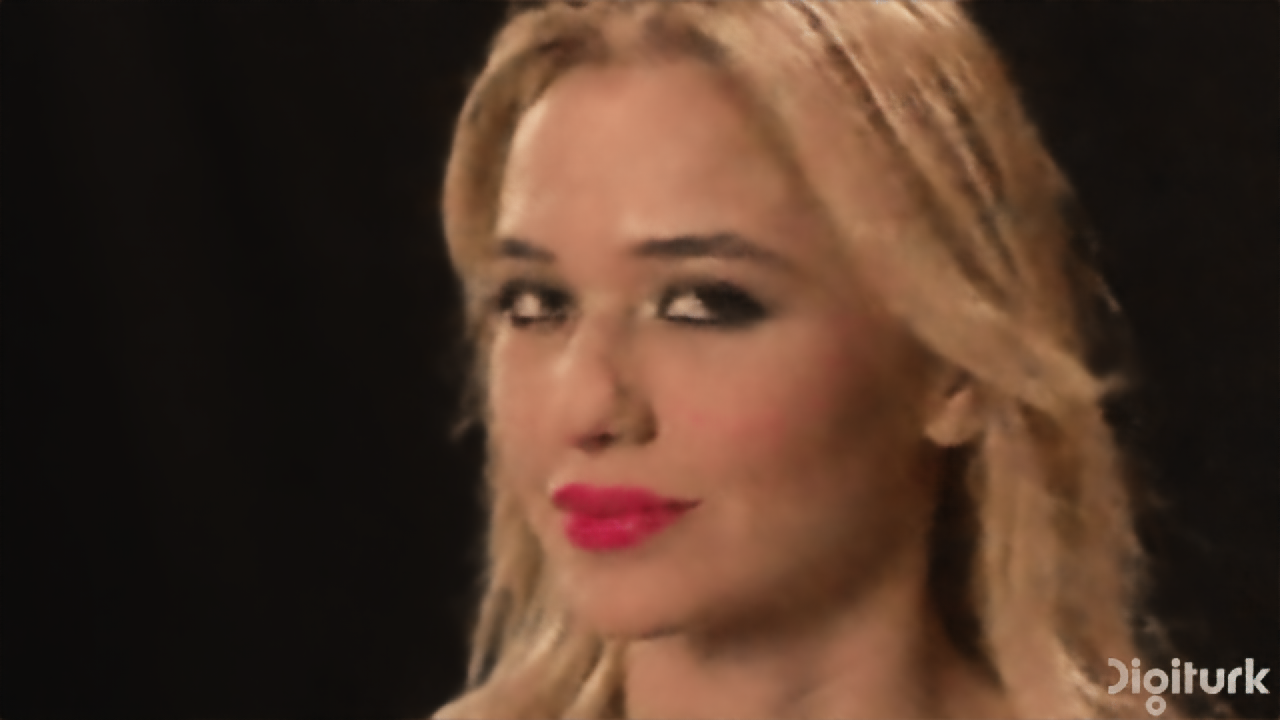} &
    \includegraphics[width=0.22\columnwidth]{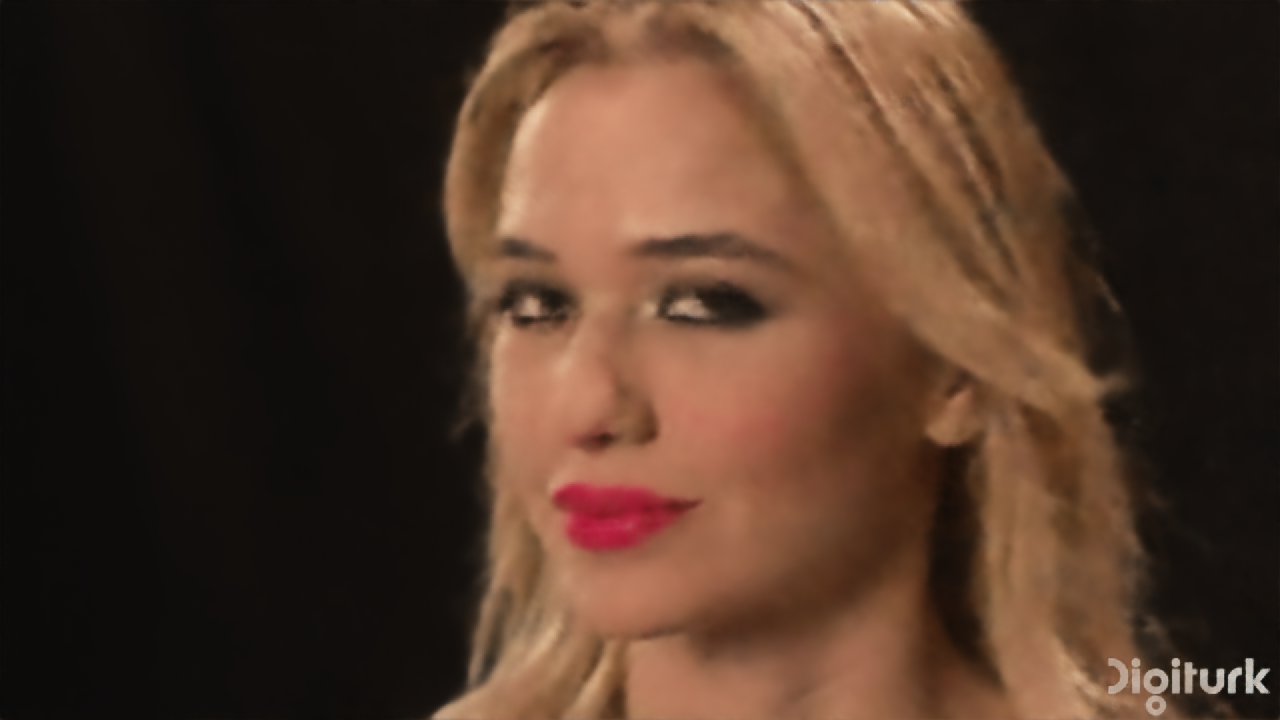} \\ 
    \multicolumn{4}{l}{\small WSN (31.00, PSNR)}\\
    
    %\includegraphics[width=0.22\columnwidth]{images/video_reinit/2/pred_0.png} & 
    %\includegraphics[width=0.22\columnwidth]{images/video_reinit/2/pred_1.png} &
    %\includegraphics[width=0.22\columnwidth]{images/video_reinit/2/pred_2.png} &
    %\includegraphics[width=0.22\columnwidth]{images/video_reinit/2/pred_3.png} \\ 
    %\multicolumn{4}{l}{\small PFNR, \textcolor{red}{$f$-NeRV2} (32.97, PSNR)}\\

    \includegraphics[width=0.22\columnwidth]{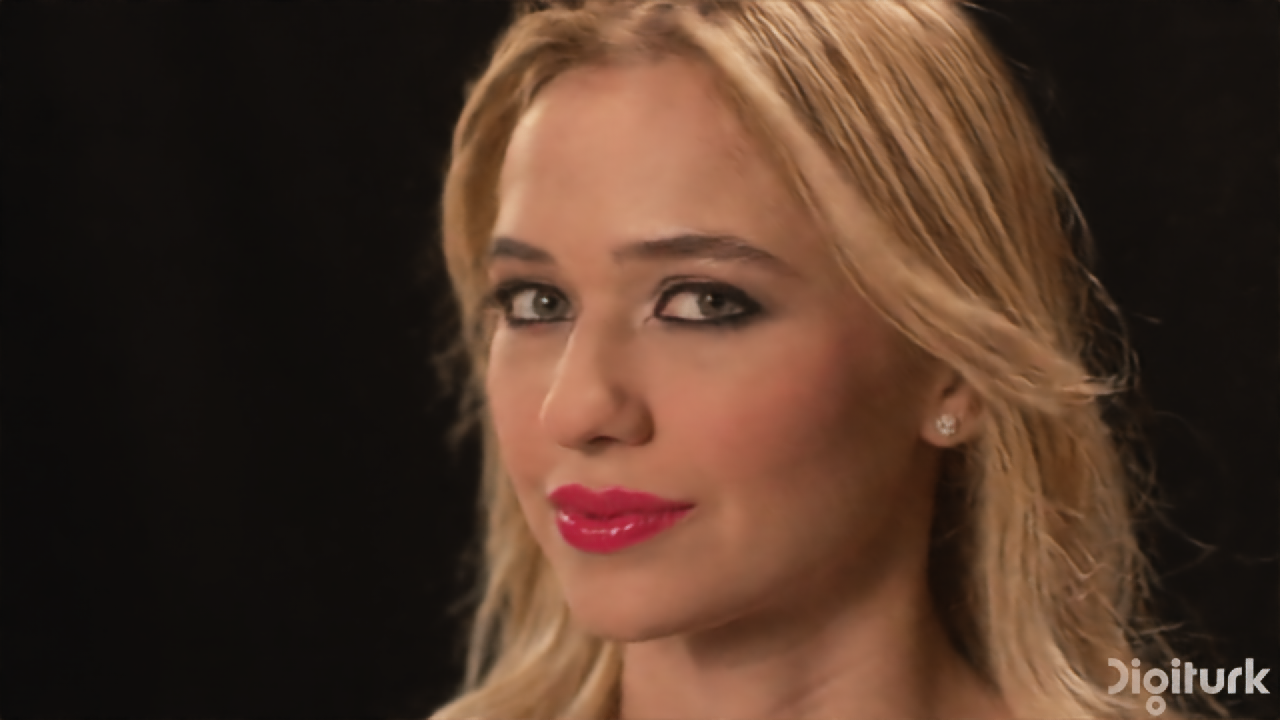} & 
    \includegraphics[width=0.22\columnwidth]{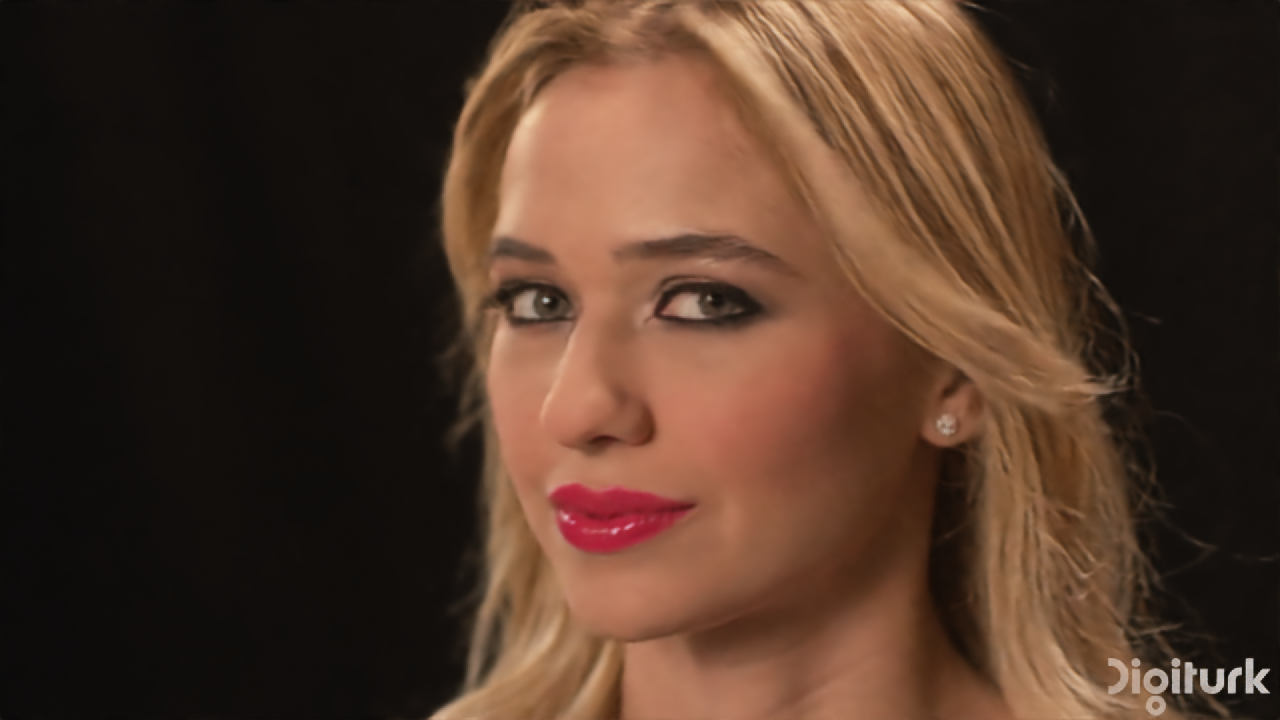} &
    \includegraphics[width=0.22\columnwidth]{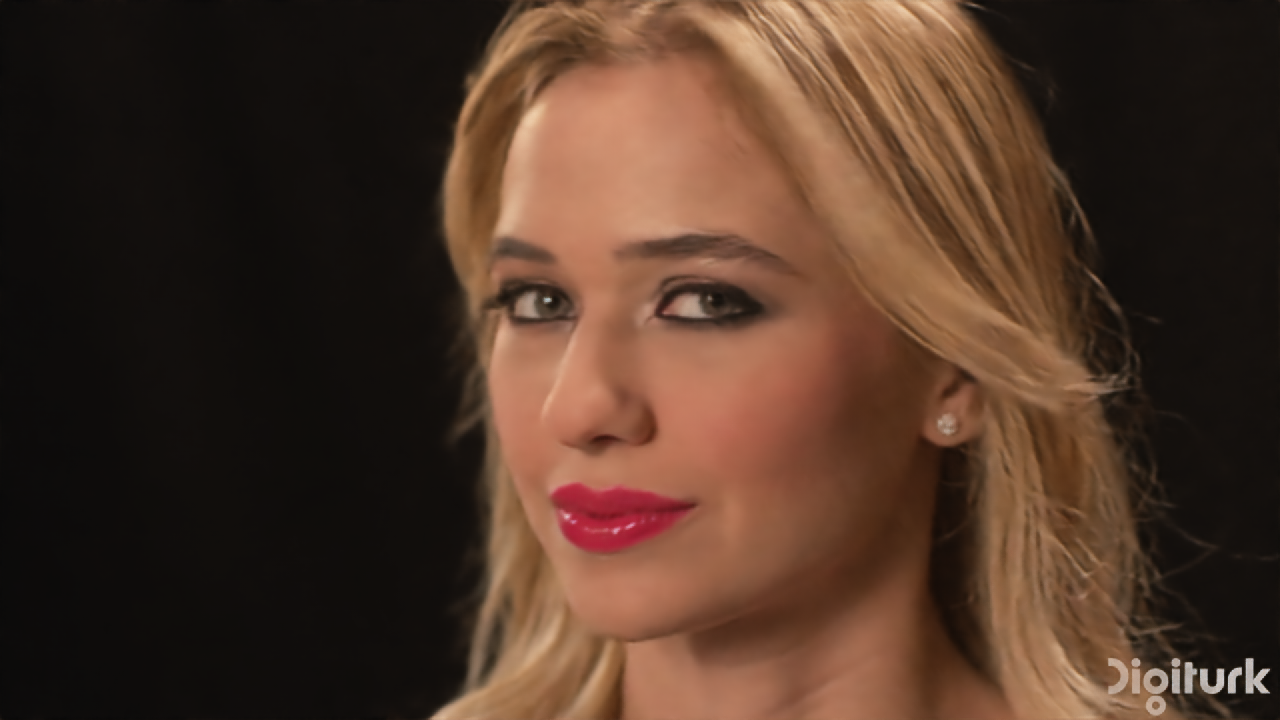} &
    \includegraphics[width=0.22\columnwidth]{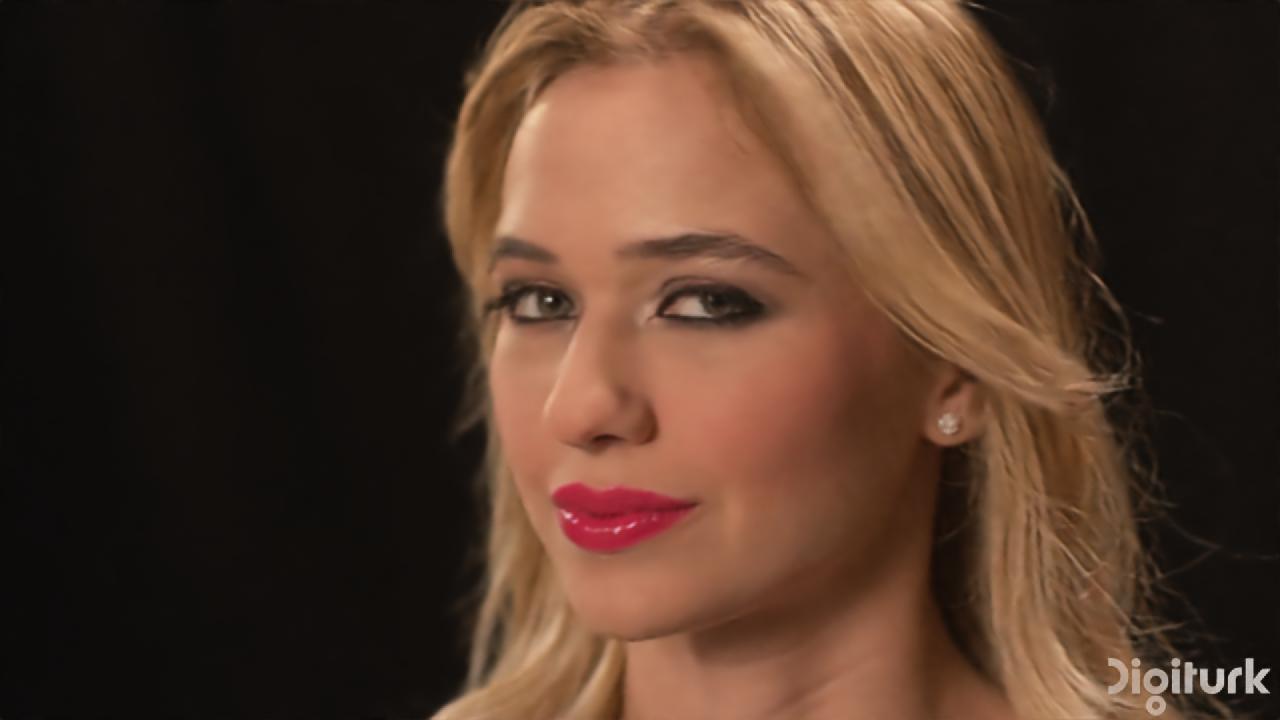} \\ 
    \multicolumn{4}{l}{\small PFNR, \textcolor{red}{$f$-NeRV3} (34.21, PSNR)}\\

    % 5.bosphorus
    %\includegraphics[width=0.22\columnwidth]{images/video_reinit/4/gt_0.png} & 
    %\includegraphics[width=0.22\columnwidth]{images/video_reinit/4/gt_1.png} &
    %\includegraphics[width=0.22\columnwidth]{images/video_reinit/4/gt_2.png} &
    %\includegraphics[width=0.22\columnwidth]{images/video_reinit/4/gt_3.png} \\ 
    %\multicolumn{4}{l}{\small Ground-truth of \textit{5.bosphorus}}\\

    %\includegraphics[width=0.22\columnwidth]{images/video_reinit/4/wsn/pred_0.png} & 
    %\includegraphics[width=0.22\columnwidth]{images/video_reinit/4/wsn/pred_1.png} &
    %\includegraphics[width=0.22\columnwidth]{images/video_reinit/4/wsn/pred_2.png} &
    %\includegraphics[width=0.22\columnwidth]{images/video_reinit/4/wsn/pred_3.png} \\ 
    %\multicolumn{4}{l}{\small WSN (29.26, PSNR)}\\
    
    %\includegraphics[width=0.22\columnwidth]{images/video_reinit/4/pred_0.png} & 
    %\includegraphics[width=0.22\columnwidth]{images/video_reinit/4/pred_1.png} &
    %\includegraphics[width=0.22\columnwidth]{images/video_reinit/4/pred_2.png} &
    %\includegraphics[width=0.22\columnwidth]{images/video_reinit/4/pred_3.png} \\ 
    %\multicolumn{4}{l}{\small PFNR, \textcolor{red}{$f$-NeRV2} (31.24, PSNR)}\\

    %\includegraphics[width=0.22\columnwidth]{images/video_reinit/4/pfnr-nerv3/pred_0.png} & 
    %\includegraphics[width=0.22\columnwidth]{images/video_reinit/4/pfnr-nerv3/pred_1.png} &
    %\includegraphics[width=0.22\columnwidth]{images/video_reinit/4/pfnr-nerv3/pred_2.png} &
    %\includegraphics[width=0.22\columnwidth]{images/video_reinit/4/pfnr-nerv3/pred_3.png} \\ 
    %\multicolumn{4}{l}{\small PFNR, \textcolor{red}{$f$-NeRV3} (34.05, PSNR)}\\
    
    \end{tabular}
    }
    \vspace{-0.12in}
    \caption{PFNR's Video Generation (from t=0 to t=3) with $c = 30.0 \%$ on the UVG17 dataset.}
    \label{fig:video_reinit_mtl}
\end{figure}

%% file: 6_conclusion.tex
\section{Conclusion}
Neural Implicit Representations (NIR) have gained significant attention recently due to their ability to represent complex and high-dimensional data. Unlike explicit representations, which require storing and manipulating individual data points, implicit representations capture information through a learned mapping function without explicitly representing the data points themselves. While they often compress neural networks substantially to accelerate encoding/decoding speed, yet existing methods fail to transfer learned representations to new videos. This work investigates the continuous expansion of implicit video representations as videos arrive sequentially over time, where the model can only access the videos from the current session. To tackle this problem, we propose a novel neural video representation, \emph{Progressive Fourier Neural Representation (PFNR)}, that finds an adaptive substructure from the supernet to the given video based on Lottery Ticket Hypothesis in a complex domain. At each training session, our PFNR transfers the learned knowledge of the previously obtained subnetworks to obtain the representation of the current video without modifying past subnetwork weights. Therefore, it can perfectly preserve the decoding ability (i.e., catastrophic forgetting) on previous videos. We demonstrate the effectiveness of our proposed PFNR over baselines on the novel UVG8/17 and DAVIS50 video sequence benchmark datasets.

%% file: 7_appendix.tex
\section{Appendix}

%%%%%%%%%%%%%%%%%%%%%%%%%%%%%%%%%%%%%
% Fourier Subneural Operator (FSO)
\subsection{Fourier Subneural Operator (FSO)} \label{app:fso}
To elucidate the Fourier Subneural Operator (FSO) as delineated in \Cref{eq:FSOper}, We'll delve into a comprehensive review of the Neural~\citep{li2020neural} Operator and Fourier Operator~\citep{li2020fourier}, focusing on its discretization methodology.

\subsubsection{Discretization for Neural/Fourier Operators}
 Let's consider $D_j = \{x_1, \cdots, x_n \} \subset D$ as an $n$-point discretization of the domain $D$. Given this setting, we have observation $a_{j | D_j} \in \mathbb{R}^{n \times d_a}$, $u_{j | D_j} \in \mathbb{R}^{n \times d_v}$, pertaining to a finite collection of input-output pairs indexed by $j$. The goal of achieving discretization-invariance with a neural operator implies that the operator is capable of generating a response $u(x)$ for any point $x$ in the domain $D$, even for instances where $x$ may not be an element of the discretized subset $D_j$. This characteristic ensures the neural operator maintains its predictive and functional integrity across the continuous domain $D$, notwithstanding the specific discretization points representing $D$. \\

%(Discretization for NIR): There is one difference between prior physical modeling for Neural/Fourier Operators and neural implicit representation for Fourier Subneural Operator. Following the previous definition~\citep{li2020fourier}, $\bm{v}_t$ is the temporal length. In the NIR setting, the function learns a time-specific continuous output (an image, $\bm{v}_t$) given a discrete-time index (a time).

%%%%%%%%%%%%%%%%%%%%%%%%%%%%%%%%%%%%%%
% Neural Operator
\subsubsection{Neural Operator}
The neural operator, as described by ~\cite{li2020neural} is formulated as an iterative architecture denoted by $v_0 \mapsto v_1 \mapsto, \cdots, \mapsto v_T$ where $v_j$ (for $j = 0, 1, \cdots, T-1$) represents a sequence of functions. Each function in this sequence yields values in the space $\mathbb{R}^{d_v}$. As the process iterates, the transformation from one state $v_t$ to the next state $v_{t+1}$ is defined by the interplay of two distinct types of operations: a non-local integral operator $\mathcal{K}$ and a local, nonlinear activation function $\sigma$. Specifically, the update from $v_t$ to $v_{t+1}$ in each iteration is articulated as the composition of these two operations, mathematically represented as follows:
\begin{equation}
    v_{t+1}(x) := \sigma(W v_t(x) + (\mathcal{K}(a; \phi)v_t(x)), \;\; \forall x \in D
    \label{eq:iter_update}
\end{equation}
In this formulation, a functional operator $\mathcal{K}: \mathcal{A} \times \Theta_{\mathcal{K}} \rightarrow \mathcal{L}(\mathcal{U}(D; \mathbb{R}^{d_v}), \mathcal{U}(D; \mathbb{R}^{d_v}))$ maps to bounded linear operators on the function space $\mathcal{U}(D; \mathbb{R}^{d_v})$. This mapping is parameterized by $\phi \in \Theta_{\mathcal{K}}$. Additionally, the function $W: \mathbb{R}^{d_v} \rightarrow \mathbb{R}^{d_v}$ is a linear transformation, and $\sigma: \mathbb{R} \rightarrow \mathbb{R}$ is a non-linear activation function whose action is defined-component-wise. 

For increasing integration in defining complex, flexible functional mappings, \cite{li2020fourier} choose $\mathcal{K}(a; \phi)$ to be a kernel integral transformation parameterized by a neural network. The kernel integral operator mapping in \Cref{eq:iter_update} is defined by 
\begin{equation}
    (\mathcal{K}(a; \phi) v_t) (x) := \int_D k(x, y, a(x), a(y); \phi) v_t(y) dy, \;\; \forall x \in D
    \label{eq:kernel}
\end{equation}
where $k_\phi : \mathbb{R}^{2(d + d_a)} \rightarrow \mathbb{R}^{d_v \times d_v}$ is a neural network parameterized by $\phi \in \Theta_{\mathcal{K}}$. Here, $k_\phi$ plays the role of a kernel function, which we learn from data. Together \Cref{eq:iter_update} and \Cref{eq:kernel} constitute a generalization of neural networks to infinite-dimensional spaces.

By removing the dependence on the function $a$ and enforcing a shift-invariance property in the kernel function $k_\phi(x, y) = k_\phi(x - y)$, the operator simplifies into a convolution operator. This transformation aligns the kernel integral operator with the principles of fundamental solutions and leverages the intrinsic properties of convolution operations.
\begin{equation}
    (\mathcal{K}(a; \phi) v_t) (x) := \int_D k_{\phi}(x - y) v_t(y) dy, \;\; \forall x \in D
    \label{eq:conv_op}
\end{equation}
This fact by parameterizing $K_\phi$ directly is exploited in Fourier space and used in the Fast Fourier Transform (FFT) to compute \Cref{eq:conv_op} efficiently.

%%%%%%%%%%%%%%%%%%%%%%%%%%%%%%%%%%%%%%
% Fourier Neural Operator
\subsubsection{Fourier Operator}
\cite{li2020fourier} suggest replacing the kernel integral operator in \Cref{eq:kernel}, by a convolution operator (see \Cref{eq:conv_op}) defined in Fourier space. Let $\mathcal{F}$ denote the Fourier transform of a function $f: D \rightarrow \mathbb{R}^{d_v}$ and $\mathcal{F}^{-1}$ its inverse then
$$
(\mathcal{F}f)_j(k) = \int_D f_j(x) e^{-2i\pi <x, k>} dx, \;\;\; (\mathcal{F}^{-1} f)_j (x) = \int_D f_j(k) e^{2i\pi <x, k>} dk
$$
for $j = 1, \cdots, d_v$ where $i = \sqrt{-1}$ is the imaginary unit. By letting $k_\phi(x, y, a(x), a(y)) = k_\phi(x - y)$ in \Cref{eq:conv_op} and applying the convolution theorem, the convolutional operation in Fourier space follows as:
$$
(\mathcal{K}(a; \phi) v_t) (x) = \mathcal{F}^{-1}(\mathcal{F}(k_\phi) \cdot \mathcal{F}(v_t))(x), \;\; \forall \in D.
$$
Furthermore, the Fourier integral operator is defined directly by parameterizing $k_\phi$ in Fourier space as follows:
\begin{equation}
    (\mathcal{K}(\phi)v_t)(x) = \mathcal{F}^{-1}(R_\phi \cdot (\mathcal{F} v_t)) (x) \;\; \forall x \in D
    \label{eq:f_op}
\end{equation}
where $R_\phi$ is the Fourier transform of a periodic function (i.e., sine and cosine functions in the time domain) $k : \bar{D} \rightarrow \mathbb{R}^{d_v \times d_v}$ parameterized by $\phi \in \Theta_\mathcal{K}$. 

%\subsubsection{Fourier Subnerual Operator}
For a given frequency mode $k \in D$, the Fourier transformation of $v_t$, denoted as $(\mathcal{F} v_t)(k) \in \mathbb{C}^{d_v}$. Similarly, $R_\phi (k) \in \mathbb{C}^{d_v \times d_v}$, representing a complex-valued matrix associated with each frequency mode $k$. Key aspects of this setup include:

\begin{itemize}[leftmargin=*]
    \item (Periodicity and Fourier Series Expansion): Given the periodic nature of $k$, it can be represented by a Fourier series. This allows for the analysis and computation to be conducted in terms of discrete modes $k \in \mathbb{Z}^d$. 

    \item (Truncation and Finite-dimensional Parameterization): To manage the complexity and ensure computational feasibility, the Fourier series is truncated at a maximal number of modes $k_{max}$. The truncation is quantified by the set $Z_{k_{max}}$, which includes all modes $k \in \mathbb{Z}^d$ that satisfy the condition $|k_j| \leq k_{max, j}$ for each dimension $j=1, \cdots, d$. The operator $R_\phi$ is parameterized as a complex-valued tensor of shape $(k_{max} \times d_v \times d_v)$, comprising the collection of truncated Fourier modes.

    \item (Conjugate Symmetry and Real-value of $k$): Due to the real-valued nature of $k$, conjugate symmetry is imposed on the Fourier coefficients. This is a fundamental property of the Fourier transform of real-valued functions, ensuring that the resulting inverse Fourier transform yields a real-valued function.
    
    \item (Choice of $Z_{k_{max}}$ and Efficiency Considerations): While the canonical choice for low-frequency modes typically involves an upper bound on the $\ell_1$-norm of $k \in \mathbb{Z}^d$, the set $Z_{k_{max}}$ is in this work chosen based on different criteria for efficient implementation. 
\end{itemize}

\subsubsection{Fourier Subnueral Operator (FSO)} 
(Discretization for VIL): One difference exists between prior physical modeling for Neural/Fourier Operators and neural implicit representation for Fourier Subneural Operator (FSO). Following the previous definition~\citep{li2020fourier}, $\bm{v}_t$ is the temporal length. In the VIL setting, the function learns a time-specific continuous hidden output (an implicit representation, $\tilde{\bm{v}}^s_t$) given discrete session and time indices (session $s$, time, $t$).

(Fourier Subneural Opeartor (FSO): Conventional continual learner (i.e., WSN) only uses a few learnable parameters in convolutional operations to represent complex sequential image streams. To capture more parameter-efficient, forget-free NIRs, the NIR model requires fine discretization and video-specific sub-parameters. This motivation leads us to propose a novel subnetwork operator in Fourier space, which provides it with various bandwidths. Following the previous definition of Fourier convolutional operator~\citep{li2020fourier}, we adapt and redefine this definition to better fit the needs of the NIR framework. We use the symbol $\mathcal{F}$ to represent the Fourier transform of a function $f$, which maps from an embedding space of dimension $d_{\bm{e}}=1 \times 160$ to a frame size denoted as $d_{\bm{v}}$. The inverse of this transformation is represented by $\mathcal{F}^{-1}$. In this context, we introduce our \textbf{F}ourier-integral \textbf{S}ubneural \textbf{O}perator (\textbf{FSO}), symbolized as $\mathcal{K}$, which is tailored to enhance the capabilities of our NIR system:
\begin{equation}
\left( \mathcal{K}(\phi) \tilde{\bm{v}}_t^s \right)(\bm{e}_{s,t}) = \mathcal{F}^{-1}(R_{\bm{\phi}} \cdot (\mathcal{F} \tilde{\bm{v}}_t^s))(\bm{e}_{s,t}),
\label{eq:FSOperApp}
\end{equation}
where $\tilde{\bm{v}}_t^s$ is a hidden representation, as shown in \Cref{fig:concept_CVRNet}; $R_{\bm \phi}$ is the Fourier transform of a periodic subnetwork function, $k_\phi$ stated in \Cref{eq:conv_op} which is parameterized by its subnetwork's parameters of real $(\bm{\theta}^{real} \odot \bm{m}_s^{real})$ and imaginary $(\bm{\theta}^{imag} \odot \bm{m}_s^{imag})$. We thus parameterize $R_{\bm \phi}$ separately as complex-valued tensors of real and imaginary $\bm{\phi}_{FSO} \in \{ \bm{\theta}^{real}, \bm{\theta}^{imag} \}$. One key aspect of the FSO is that its parameters grow with the depth of the layer and the input/output size since the operator $R_\phi$, as stated in \Cref{eq:f_op}, is parameterized as a complex-valued tensor of shape $(k_{max} \times d_{\tilde{\bm{v}}} \times d_{\tilde{\bm{v}}})$. However, through careful layer-wise inspection and adjustments for sparsity, we can find a balance that allows the FSO to describe neural implicit representations efficiently. In the experimental section, we will showcase the most efficient FSO structure and its performance. \Cref{fig:concept_CVRNet} shows one possible PFNR structure of a single FSO. %We describe the optimization in the following section.

%\textbf{Datasets.} 
\subsection{Datasets}
\textit{1) UVG of 8 Video Sessions}:
For "Big Buck Bunny" frames collected from the scikit-video, we use the frames provided with the NeRV official code. "Big Buck Bunny" comprises 132 frames of $720 \times 1080$ resolution. The frames for the other seven videos, collected from the UVG dataset, are extracted from YUV Y4M videos, and further information can be found in the implementation details. As shown in \Cref{table:uvg_8_dataset}, the seven videos have $1920 \times 1080$ resolution, with the shaking video comprising 300 frames and the other 6 videos containing 600 frames each. These videos are captured at 120 frames per second (FPS), and the duration of the shaking video is 2.5 seconds, while the duration of the other 6 videos is 5 seconds. For convenience, the video titles in the UVG of 8 Video Sessions are abbreviated, and their corresponding full titles in the UVG dataset are as follows:
\textit{
1.bunny : Big Buck Bunny,
2.beauty : Beauty,
3.bosphorus : Bosphorus,
4.bee : HoneyBee,
5.jockey : Jockey,
6.setgo : ReadySetGo,
7.shake : ShakeNDry,
8.yacht : YachtRide}.

\input{materials/main_uvg8_datasets}

\noindent 
\textit{2) UVG of 17 Video Sessions}: 
Compared to the UVG of 8 video sessions, the other nine videos are all collected from the UVG dataset. The frames for these videos are extracted from YUV RAW videos with a resolution of 1920x1080. Further information can be found in the implementation details. As shown in \Cref{table:uvg_17_dataset}, the sunbath video consists of 300 frames at 50 FPS and 12 seconds, the lips video consists of 600 frames at 120 FPS and 5 seconds, and the other seven videos comprised of 600 frames at 50 FPS and 12 seconds. The full names of each video in the UVG dataset are as follows:
\textit{
2.city : CityAlley,
4.focus : FlowerFocus,
6.kids : FlowerKids,
8.pan : FlowerPan,
10.lips : Lips,
12.race : RaceNight,
14.river : RiverBank,
16.sunbath : SunBath,
17.twilight : Twilight}.

\input{materials/main_uvg17_datasets}

\noindent
\textit{3) DAVIS (Densely Annotated VIdeo Segmentation) of 50 Video Sessions}: 
We prepare a large-scale sequential video dataset, the Densely Annotation Video Segmentation dataset (DAVIS)~\citep{Perazzi2016}. To validate our algorithm and investigate the limitations, we conducted the experiments on 50 video sequences with 3455 frames with a high-quality, high-resolution (1080p).

% Implementation Details.
\noindent 
%\textbf{Imprementation Details.}
\subsection{Imprementation Details}

\textit{1) 7 Videos in UVG of 8 Video Sessions}: To utilize the same video frame with NeRV, we downloaded 7 videos from the UVG dataset and employed the following commands to extract frames from the YUV Y4M videos.

>> Download file : \textbf{[title] 3840x2160 8bit YUV Y4M}

>> Command : \textbf{ffmpeg -i [file\_name] [path]/f\%05d.png}

\textit{2) 9 Videos in UVG of 17 Video Sessions}: To expand our usage of videos, we acquired an additional 9 videos from the UVG dataset that are exclusively available as YUV RAW videos with a resolution of 3840x2160. We then extracted and resized the frames using the following command.

>> Download file : \textbf{[title] 3840x2160 10bit YUV RAW}

>> Command : \textbf{ffmpeg -s 3840x2160 -pix\_fmt yuv420p10le -i [file\_name] -vf scale=1920:1080 -pix\_fmt rgb24 [path]/f\%05d.png}

Digiturk provides the video contents of the UVG dataset. The dataset videos are available online at \url{https://ultravideo.fi/#main}.

\input{materials/architecture_detail_table}

\textbf{EWC}~\citep{Kirkpatrick2017}. We trained EWC on the two novel benchmark datasets as a regularized baseline. When training with a new video, the EWC penalty was adopted as a regularization term to alleviate catastrophic forgetting. The importance of the parameter was calculated through the diagonal component of the Fisher Information matrix, and the EWC penalty increased as the difference in the vital parameter increased as follows:
\begin{equation}
\begin{split}
L_E(v_t^s) = L(v_t^s) + {\lambda \over 2}\sum_i F_i(\theta_i - \theta^*_{p,i})^2,
\end{split}
\label{eq:EWC}
\end{equation}
where $L_E$ is the total loss for EWC learning, $F$ represents the Fisher information matrix, $\lambda$ is a hyperparameter to determine the importance of the previous video, $i$ denotes each parameter, and $\theta_{p}^*$ denotes the parameter after training with the previous video. We randomly sampled 10 frames per video to compute the Fisher diagonal and stored them in a replay buffer. The hyperparameter $\lambda$ was experimentally set to 2e6 to scale the EWC penalty. 

%The results of our experiment are shown in Tables \ref{table:uvg8_psnr}, \ref{table:uvg8_msssim}, \ref{table:uvg17_psnr}, and \ref{table:uvg17_msssim}.

% iCaRL 저장되는 frame의 개수 100개 -> 800개로 수정되었습니다. (NeurIPS version은 100)
\textbf{iCaRL}~\citep{rebuffi2017icarl}. We trained iCaRL as a rehearsal-based baseline on the two novel benchmark datasets. To replay previous videos, we store a total of $m=800$ frames in the replay buffer as an exemplar set, and as the training progresses, we save $m/s$ frames per video. The exemplar management method is similar to that in iCaRL, where we compute the average feature map within the video and select $m/s$ video frames that approximate the average feature map. For knowledge distillation, we randomly sample frames from the exemplar set of previous videos at each learning step and performed training with the current video, as follows:
\begin{equation}
\begin{split}
 L_C(v_t^s) = L(v_t^s) + \lambda {1 \over t-1} \sum_i^{t-1} L(v_i^*),
\end{split}
\label{eq:iCaRL}
\end{equation}
where $v_i^*$ is a frame sampled from example set of $i$th video, and $\lambda$ is a hyperparameter experimentally set to 0.5. 

\textbf{ESMER}~\citep{sarfraz2023error}. We trained ESMER on the two novel benchmark datasets as a current strong rehearsal-based baseline. Error-Sensitive Reservoir Sampling (ESMER) maintains episodic memory, which leverages the error history to pre-select low-loss samples as candidates for the buffer of 800 samples. At this time, we didn't use any noisy labels when training ESMER. We observed that the ESMER could not reduce forgetting in representations. It seems better suited for retaining information in image classification tasks rather than neural implicit representations. Lastly, compared with iCaRL, ESMER replies buffer at each iteration, leading to ineffective training cost and performance as shown in \Cref{table:uvg8_psnr_str}.

\input{materials/app_table_uvg17_fps_psnr}

%\noindent 
%\textbf{Architecture.} 
\subsection{Architecture}
In this paper, we set NeRV as the baseline architecture, and we present a detailed overview of our architecture, as shown in \Cref{table:architecture_detail}. Our architecture has two differences compared to the previous NeRV. Firstly, to incorporate the outputs from the positional encoder of both the video index and frame index, we expand the input size of the first layer in the MLP from 80 to 160. Secondly, to enable Multi-Task Learning (MTL), we employ distinct head layers (multi-heads) for each video after the NeRV block. Apart from these modifications, our architecture remains consistent with the previous NeRV architecture. We stack five NeRV blocks with upscaling factors of 5, 2, 2, 2, and 2, respectively. The lowest channel width for output feature maps in the NeRV block is also set to 96. Comparing our architecture to the baseline NeRV, the number of parameters increases by 0.3433\% to 12,567,560 for eight videos. Similarly, for seventeen videos, the number of parameters increases by 0.3642\% to 12,570,179.

\subsection{Additional Results}
\noindent 
\textbf{Dataset statistic of FPS and PSNR.} As shown in \Cref{table:uvg_8_dataset} and \Cref{table:uvg_17_dataset}, we have extracted sample frames according to each video’s FPS and Length (sec) during training. There are two kinds of videos: 300 and 600 frames. We train each video for 150 epochs. Statistically, the average PSNR of 120 FPS was better than those of 50 FPS, as shown \Cref{table:uvg17_fps_psnr}. When considering 30 PSNR is known as a sufficient resolution, 50 FPS-based sequence training could be adequate using the proposed PFNR. Moreover, the number of frames does not seem to be a critical factor to PSNR when considering that the PSNR of video session 1 (bunny), with 132 ($720 \times 1080$) frames, is greater than the 30 PSNR score.

\noindent
\textbf{PFNR's Structure.} 
 We investigate the most expressive, effective, and efficient structure with Fourier Subneural Operator (FSO, \Cref{eq:FSOper}) for progressive neural implicit representations. To do so, we prepare the layer-wise FSO to maintain the output size of the baseline layer as shown in \Cref{table:architecture_detail}: the number of parameters of a spectral layer depends on input and output size, so as the layer increases, the parameters also increase. \Cref{table:uvg8_psnr_str} shows the effectiveness of PFNR with spectral layers in terms of PSNR, BWT, and CAP. To acquire neural implicit representations, PFNR explores more diverse parameters than WSN regarding the same capacity. 

\input{materials/main_table_uvg8_structure}

\noindent
\textbf{Large-scale Sequence.} We conducted the large-scale video sequence training to show its effectiveness and investigate the limitation of architecture parameters as shown in \Cref{table:app_davis_psnr}. The overall PFNR performances (STL, WSN, PFNR, and MTL) on the DAVIS50 dataset on the UVG8/17 dataset are lower than those on the UVG17 dataset. However, the performance trends of PFNR observed on the DAVIS50 dataset are consistent with the experimental results obtained from the UVG8/17 dataset. Regarding sequential and multi-task learning, the current performance falls short of achieving the target of 30 PSNR. This indicates a need for future work focused on designing model structures capable of continual learning on large-scale datasets.

\input{materials/app_table_davis50_psnr}

\input{materials/plot_transfer_matrix}
% Transfer Matrix
\noindent 
\textbf{Forget-free Transfer Matrix.} We prepare the transfer matrix to prove our PFNR's forget-freeness and to show video correlation among other videos, as shown in \Cref{fig:transf_matrix} on the UVG17 dataset; lower triangular estimated by each session subnetwork denotes that our PFNR is a forget-free method and upper triangular calculated by current session subnetwork denotes the video similarity between source and target. The PFNR proves the effectiveness from the lower triangular of \Cref{fig:transf_matrix} (a) and (b). Nothing special is observable from the upper triangular since they are not correlated, however, there might be some shared representations.

\noindent
\textbf{Ablation Study of FSO.} We prepare several ablation studies to prove the effectiveness of FSO. First, we show the performances of only real part (ignore an imaginary part) in f-NeRV2/3 as shown in \Cref{table:uvg8_fso_real}. The PSNR performances of only real part were lower than those of both real and imaginary parts in f-NeRV2/3. We infer that the imaginary part of the winning ticket improves the implicit neural representations. Second, we also investigate the effectiveness of only FSO without Conv. Layer in f-NeRV2/3, as shown in \Cref{table:uvg8_fso_conv}. The PSNR performances were lower than FSO with Conv block. Therefore, the ensemble of FSO and Conv improves the implicit representations. Lastly, we investigate the effectiveness of sparse FSO in STL, as shown in \Cref{table:uvg8_stl}. The sparse FSO boots the PSNR performances in STL. These ablation studies further strengthen the effectiveness of FSO for sequential neural implicit representations.

\input{materials/main_table_uvg8_psnr_fso}

\input{materials/main_table_uvg8_psnr_conv}

\input{materials/main_table_uvg8_psnr_stl}

\noindent
\textbf{Training time and Decoding FPS.} We train and test two baselines (NeRV, ESMER) with $f$-NeRV2 using one GPU (TITAN V, 12G) with a single batch size to investigate the computational expenses and decoding FPS on the UVG8 dataset, as shown in \Cref{table:uvg8_stl_time}. In STL, NeRV with $f$-NeRV2 costs more computational times than NeRV. In VCL, memory buffer-based methods, i.e., ESMER, cost more training time since they replay the memory buffer in sequential training. On the other hand, architecture-based methods, i.e., PFNR, provide parameter-efficient, faster, forget-free solutions in training while cost computation expenses in the decoding process. Considering these limitations and advantages, we would find a more parameter-efficient FSO algorithm in future work.

\input{materials/main_table_uvg8_stl_time}

\textbf{Video Generation.} We prepared some results of video generation as shown in \Cref{fig:video_app_reinit_mtl}. At the human recognition level, approximately 30 PSNR provides a little bit of blurred generated images as shown in WSN’s PSNR(29.26); if the PSNR score is greater than 30, we hardly distinguish the quality of sequential neural implicit representations (city session). Thus, our objective is to maintain the sequential neural implicit representation of the 30 PSNR score. We have achieved this target PSNR score on the UVG17 dataset. We also show the quantized and encoded PFNR's video generation results \Cref{fig:video_reinit_bits}. We demonstrate that a compressed sparse solution in FP8 (PFNR with $c=30.0 \%$, $f$-NeRV2) generates video representations sequentially without a significant quality drop. The results of the FP4 showed that the compressed model can not create image pixels in detail.

\input{materials/plot_video_reinit_mtl}

\input{materials/plot_video_reinit_bits}
\input{materials/app_table_uvg17_psnr}
\input{materials/app_table_uvg17_msssim}

\subsection{Layer-wise Representations}
To investigate the property of FSO, we prepare the layer-wise representations, as shown in \Cref{fig:fmap_app_uvg17}. The representations of PFNR ($f$-NeRV3) focus on textures rather than objects at the NeRV3 layer in the video session of bosphorus. 
In the video session of bee, PFNR ($f$-NeRV3) also tends to capture local textures broadly at the NeRV3 layer. This behavior of $f$-NeRV3 leads to better generalization at final prediction than others, such as WSN and PFNR ($f$-NeRV2). Moreover, we conducted an ablation study to inspect the best sparsity of $f$-NeRV3, as shown in \Cref{fig:fmap_sparsity_uvg17}. Specifically, the performances of PFNR, c=$50.0 \%$ depend on the sparsity of $f$-NeRV3. We observed that $f$-NeRV3 with c=$50.0 \%$ was the best sparsity.

\input{materials/plot_fmap_uvg17}
\input{materials/plot_fmap_sparsity_uvg17}

%\noindent 
%\textbf{Limitations and Future Work.}
\subsection{Current Limitations and Future Work}
Since the parameters of FSO depend on the input/output feature map size, in this task, the deeper the FSO layer, the larger the parameters increase. Nevertheless, we found the most parameter-efficient FSO structure through layer-wise inspection and layer sparsity to describe the best neural implicit representations. In future work, we will design a more parameter-efficient FSO layer in continual tasks such as neural implicit representation and task/class incremental learnings. 

%There are some limitations to NeRV's architecture, both in sequential and joint learning.  First, we could expect more precise video representation from our experimental observation if we use more discriminative input embedding for the architecture. Moreover, since the sequential and joint learning solution tends to decrease its ability to generate detailed (high-frequency) video representations in long sequential video training, we need to propose a new architecture to learn high-frequency video representation for high-quality video generation as future work.

%\noindent 
%\textbf{Broader Impacts.} 
\subsection{Broader Impacts}
As the most popular media format nowadays, videos are generally viewed as frames of sequences. Unlike that, our proposed PFNR is a novel way to represent sequential videos as a function of video session and time, parameterized by the neural network firstly in Fourier space, which is more efficient and might be used in many video-related tasks, such as sequential video compression, sequential video denoising, complex sequential physical modeling \citep{li2020fourier, li2020neural, kovachki2021neural, tran2021factorized}, and so on. Hopefully, this can save bandwidth and fasten media streaming, enriching entertainment potential. Unfortunately, like many advances in deep learning for videos, this approach could be used for various purposes beyond our control.

\input{materials/app_table_uvg17_psnr_bitwise}

\subsection{Public Codes of our PFNR}
We provide the core parts of PFNR to understand better. Please refer to the attached file. We will provide all training and inference codes soon.

%% file: materials/main_uvg8_datasets.tex
\begin{table}[ht]
\centering
\caption{the UVG8 Video Sessions.}
\resizebox{0.92\textwidth}{!}{
\begin{tabular}{lcccccccc}
\toprule 

& \multicolumn{8}{c}{\textbf{Video Sessions}}  \\
\cline{2-9}
& \textbf{1} & \textbf{2} & \textbf{3} & \textbf{4} & \textbf{5} & \textbf{6} & \textbf{7} & \textbf{8} \\ \midrule 

Categories &  bunny & beauty & bosphorus & bee & jockey & setgo & 
 shake & yacht \\

 FPS & - & 120 & 120 & 120 & 120 & 120 & 120 & 120 \\
 Length (sec) & - & 5 & 5 & 5 & 5 & 5 & 2.5 & 5 \\
Num. of Frames & 132 & 600 & 600 & 600 & 600 & 600 & 300 & 600 \\

Resolutions & 720 $\times$ 1080 &  1920 $\times$ 1080 & 1920 $\times$ 1080 & 1920 $\times$ 1080 & 1920 $\times$ 1080 & 1920 $\times$ 1080 & 1920 $\times$ 1080 & 1920 $\times$ 1080 \\

\bottomrule
\end{tabular}
}
\label{table:uvg_8_dataset}
\end{table}

%% file: materials/main_uvg17_datasets.tex
\begin{table}[ht]
\centering
\caption{the UVG17 Video Sessions.}
\resizebox{\textwidth}{!}{
\begin{tabular}{lccccccccc}
\toprule 

% split into 2 row
& \multicolumn{9}{c}{\textbf{Video Sessions}}  \\
\cline{2-10}
& \textbf{1} & \textbf{2} & \textbf{3} & \textbf{4} & \textbf{5} & \textbf{6} & \textbf{7} & \textbf{8} & \textbf{9} \\ \midrule 

Categories &  bunny & city & beauty & focus & bosphorus & kids & bee & pan & jockey  \\
FPS & - & 50 & 120 & 50 & 120 & 50 & 120 & 50 & 120  \\
Length (sec) & - & 12 & 5 & 12 & 5 & 12 & 5 & 12 & 5  \\
Num. of Frames & 132 & 600 & 600 & 600 & 600 & 600 & 600 & 600 & 600  \\

Resolutions & 720 $\times$ 1080 & 1920 $\times$ 1080 & 1920 $\times$ 1080 & 1920 $\times$ 1080 & 1920 $\times$ 1080 & 1920 $\times$ 1080 & 1920 $\times$ 1080 & 1920 $\times$ 1080 & 1920 $\times$ 1080 \\ \bottomrule

\toprule 
& \multicolumn{9}{c}{\textbf{Video Sessions}}  \\
\cline{2-10}
& \textbf{10} & \textbf{11} & \textbf{12} & \textbf{13} & \textbf{14} & \textbf{15} & \textbf{16} & \textbf{17} & \\ \midrule 

Categories & lips & setgo & race &  shake & river & yacht & sunbath & twilight &  \\
FPS & 120 & 120 & 50 & 120 & 50 & 120 & 50 & 50 &   \\
Length (sec) & 5 & 5 & 12 & 2.5 & 12 & 5 & 6 & 12 &   \\
Num. of Frames & 600 & 600 & 600 & 300 & 600 & 600 & 300 & 600 & \\

Resolutions & 1920 $\times$ 1080 & 1920 $\times$ 1080 & 1920 $\times$ 1080 & 1920 $\times$ 1080 & 1920 $\times$ 1080 & 1920 $\times$ 1080 & 1920 $\times$ 1080 & 1920 $\times$ 1080 & \\

\bottomrule
\end{tabular}
}
\label{table:uvg_17_dataset}
\end{table}

%% file: materials/architecture_detail_table.tex
\begin{table}[h]
\centering
\caption{PFNR's architecture of Fourier Subneural Operator (FSO), \textcolor{red}{$f$-NeRV$\ast$}. Note that PE denotes positional encoding.}
\resizebox{0.7\textwidth}{!}{
\begin{tabular}{clcrr}
\toprule 
layer & Module     & Upscale Factor & \begin{tabular}[c]{@{}c@{}}Output Size\\ ($C\times W\times H$)\end{tabular} & \begin{tabular}[c]{@{}c@{}}Number of \\ parameters\end{tabular}  \\ \midrule 
\multirow{2}{*}{0} & PE (w frame index)         & -              & $\text{80}\times  \text{1}\times  \text{1}$      & -                    \\
                   & PE (w video index)         & -              & $\text{80}\times  \text{1}\times  \text{1}$      & -                    \\ \midrule
\multirow{2}{*}{1} & STEM of fc1                & -   & \multirow{1}{*}{$\text{512} \times  \text{1} \times  \text{1}$}  & 81,920 \\
                   %& STEM of \textcolor{red}{$f$}-fc1  & -   &   & 163,840 \\ \cline{2-5}
                   & STEM of fc2                & -   & \multirow{1}{*}{$\text{112} \times  \text{16} \times  \text{9}$}  & 8,257,536  \\ \midrule
                   %& STEM of \textcolor{red}{$f$}-fc2  & -   &                                    & 16,515,072  \\ \midrule
                   
\multirow{2}{*}{2} & NeRV2 block of conv        & $\text{5} \times$    & $\text{112} \times  \text{80} \times  \text{45}$   & 2,825,200   \\
                   & \textcolor{red}{$f$-NeRV2}  & $\text{1} \times$    & $\text{112} \times  \text{80} \times  \text{45}$   & 1,605,632   \\
                   %& NeRV2 block of \textcolor{red}{$f$}-deconv  & $\text{1} \times$    & $\text{112} \times  \text{80} \times  \text{45}$   & 802,816   \\ 
                   \midrule 

\multirow{2}{*}{3}  & NeRV3 block of conv  & $\text{2} \times$ & $\text{96} \times  \text{160} \times  \text{90}$   & 387,456              \\
                    & \textcolor{red}{$f$-NeRV3}  &  $\text{1} \times$  & $\text{96} \times  \text{160} \times  \text{90}$  &  37,847,040 \\
                    %& NeRV3 block of \textcolor{red}{$f$}-deconv  &  $\text{1} \times$  & $\text{96} \times  \text{160} \times  \text{90}$  &  37,847,040 \\ 
                    \midrule
                    
4                   & NeRV4 block of conv  & $\text{2} \times$ & $\text{96} \times  \text{320} \times  \text{180}$  & 332,160   \\
5                   & NeRV5 block of conv  & $\text{2} \times$ & $\text{96} \times  \text{640} \times  \text{360}$  & 332,160   \\  
6                   & NeRV6 block of conv  & $\text{2} \times$ & $\text{96} \times  \text{1280} \times  \text{720}$ & 332,160   \\ \bottomrule
7                   & \begin{tabular}[c]{@{}c@{}}Multi-head layer \\ for a video session \end{tabular} & - & $\text{3} \times  \text{1280} \times  \text{720}$  & 291 \\ \bottomrule
\end{tabular}
}
\label{table:architecture_detail}
\end{table}

%% file: materials/app_table_uvg17_fps_psnr.tex
\begin{table*}[!ht]
\small
\centering
\caption{\small \textcolor{black}{Statistics of FPS and PSNR on the UVG17 Video Sessions}}
\resizebox{0.85\textwidth}{!}{
\begin{tabular}{lcccc}
\toprule 

\textbf{Method} & \textbf{FPS} / \textbf{Resolution} & \textbf{Avg. PSNR}  & \textbf{FPS} / \textbf{Resolution} & \textbf{Avg. PSNR} \\ \midrule 

STL, NeRV~\cite{chen2021nerv}$^{\ast}$  & 120 / 1920 $\times$ 1080 & \textbf{35.97} & 50 / 1920 $\times$ 1080 & 35.56 \\  \midrule 

%EWC~\cite{Kirkpatrick2017}$^{\ast}$    & 120 / 1920 $\times$ 1080 & ~~9.78 & 50 / 1920 $\times$ 1080 & 13.07 \\  %\midrule 
iCaRL~\cite{rebuffi2017icarl}$^{\ast}$ & 120 / 1920 $\times$ 1080 & \textbf{21.94} & 50 / 1920 $\times$ 1080 & 21.06 \\  %\midrule 
ESMER~\cite{sarfraz2023error}$^{\ast}$ & 120 / 1920 $\times$ 1080 & \textbf{21.43} & 50 / 1920 $\times$ 1080 & 19.25 \\  %\midrule 
\midrule

WSN$^{\ast}$                           & 120 / 1920 $\times$ 1080 & \textbf{27.01} & 50 / 1920 $\times$ 1080 & 25.95 \\  %\midrule 
PFNR, \textcolor{red}{$f$-NeRV2}       & 120 / 1920 $\times$ 1080 & \textbf{29.65} & 50 / 1920 $\times$ 1080 & 28.36 \\  %\midrule 
PFNR, \textcolor{red}{$f$-NeRV3}       & 120 / 1920 $\times$ 1080 & \textbf{31.53} & 50 / 1920 $\times$ 1080 & 31.47 \\  \midrule 

MTL (upper-bound)                      & 120 / 1920 $\times$ 1080 & \textbf{29.64} & 50 / 1920 $\times$ 1080 & 28.83 \\  %\midrule 

\bottomrule
\end{tabular}
}
\label{table:uvg17_fps_psnr}
\vspace{-0.12in}
\end{table*}

%% file: materials/main_table_uvg8_structure.tex
\begin{table*}[!ht]
\small
\centering
\caption{\small PSNR results with Fourier Subnueral Operator (FSO) layer (\textcolor{red}{$f$-NeRV$\ast$}) (NeRV block, \Cref{table:architecture_detail}) on UVG8 Video Sessions with average PSNR and Backward Transfer (BWT), and Capacity (CAP). Note that $\ast$ denotes our reproduced results.}
\resizebox{0.9\textwidth}{!}{
\begin{tabular}{lccccccccccc}
\toprule 

\multicolumn{1}{c}{\multirow{2}{*}{\textbf{Method}}} & \multicolumn{8}{c}{\textbf{Video Sessions}} & \multirow{2}{*}{\thead{\textbf{Avg. PSNR / } \\ \textbf{BWT}}} & \multirow{2}{*}{\textbf{CAP}}  \\

\cline{2-9}
& \textbf{1} & \textbf{2} & \textbf{3} & \textbf{4} & \textbf{5} & \textbf{6} & \textbf{7} & \textbf{8} &  \\ \midrule 
%STL, NeRV~\cite{chen2023hnerv} & 39.63 & 36.06 & 37.35 & 41.23 & 38.14 & 31.86 & 37.22 & 32.45 & 36.74 / ~~-~~ & ~~800.00 \% \\

STL, NeRV~\cite{chen2021nerv}$^{\ast}$  & 39.66 & 36.28 & 38.14 & 42.03 & 36.58 & 29.22 & 37.27 & 31.45 & 36.33 / ~~-~~ & ~~800.00 \% \\ \midrule 

EWC~\cite{Kirkpatrick2017}$^{\ast}$    & 10.19 & 11.15  & 14.47 & 8.39 & 12.21 & 10.27 & 9.97 & 23.98 & 12.58 / -17.59 & ~~100.00 \% \\  
iCaRL~\cite{rebuffi2017icarl}$^{\ast}$ & 30.84 & 26.30  & 27.28 & 34.48 & 20.90 & 17.28 & 30.33 & 24.64 & 26.51 / ~-3.90 & ~~100.00 \% \\ 
ESMER~\cite{sarfraz2023error}$^{\ast}$ & 31.71 & 23.09  & 24.15 & 28.03 & 17.30  & 13.81 &  12.45 &  24.57 & 21.92  / ~-9.99  & ~~100.00 \% \\
\midrule

WSN$^{\ast}$, c = 10.0 \% & 27.81 & 30.66 & 29.30 & 33.06 & 22.16 & 18.40 & 27.81 & 22.97 & 26.52 / 0.0 & ~~~~28.00 \% \\ 
WSN$^{\ast}$, c = 30.0 \% & 31.37 & 32.19  & 29.92 & {33.62} & {22.82}  & {18.96} & {28.43} & 23.40 & 27.59 / 0.0 & ~~~~59.00 \% \\ 
WSN$^{\ast}$, c = 50.0 \% & 34.05 & {32.28}  & {29.98} & 32.88 & 22.15  & 18.61 & 27.68 & {23.64} & {27.66} / 0.0 & ~~~~77.00 \%\\ 
WSN$^{\ast}$, c = 70.0 \% & {35.62} & 32.08 & 29.46 & 31.37 & 21.60  & 18.13 & 27.33 & 22.61 & 27.28 / 0.0 & ~~~~91.00 \% \\ 

% re-init
%WSN, c = 10.0 \% & 28.12 & 31.31 & 29.89 & 34.83 & 23.82 & 19.56 & 29.46 & 24.58 & 27.72 / 0.0 & ~~~~30.00 \%\\ 
%WSN, c = 30.0 \% & 31.36 & 32.91 & 31.42 & 36.39 & 24.93 & 20.58 & 30.78 & 25.25 & 29.20 / 0.0 & ~~~~68.00 \% \\ 
%WSN, c = 50.0 \% & 34.10 & 33.45 & 31.77 & 36.09 & 24.82 & 20.25 & 30.02 & 25.01 & 29.44 / 0.0 & ~~~~91.00 \% \\ 
%WSN, c = 70.0 \% & 35.55 & 33.04 & 30.44 & 32.11 & 23.00 & 19.02 & 28.09 & 23.52 & 28.10 / 0.0 & ~~~~99.00 \% \\ 
\midrule

PFNR, c = 10.0 \%, \textcolor{red}{$f$-NeRV2} & 28.49 & 32.30 & 30.30 & 35.12 & 24.10 & 19.82 & 29.89 & 24.76 & 28.10 / 0.0  &  ~~~~33.83 \% \\ 
PFNR, c = 30.0 \%, \textcolor{red}{$f$-NeRV2} & 31.99 & 33.56 & 31.82 & 36.61 & 25.28 & 20.97 & 31.07 & 25.73 & 29.63 / 0.0  &  ~~~~76.76 \% \\
PFNR, c = 50.0 \%, \textcolor{red}{$f$-NeRV2} & 34.46 & 33.91 & 32.17 & 36.43 & 25.26 & 20.74 & 30.18 & 25.45 & 29.82 / 0.0  &  ~~102.64 \% \\ 
PFNR, c = 70.0 \%, \textcolor{red}{$f$-NeRV2} & 36.04 & 33.46 & 31.05 & 32.57 & 23.40 & 19.41 & 28.31 & 24.31 & 28.57 / 0.0  &  ~~111.66 \% \\
\midrule 

%PFNR, c = 10.0 \%, \textcolor{red}{$f$-NeRV3} & 29.78 & 33.30 & 32.29 & 37.29 & 24.65 & 20.82 & 31.94 & 26.19 & 29.53 / 0.0 & ~~-.- \% \\ 
%PFNR, c = 30.0 \%, \textcolor{red}{$f$-NeRV3} & 33.69 & 34.76 & 34.57 & 38.50 & 27.09 & 23.16 & 33.10 & 27.94 & 31.36 / 0.0 & -.- \% \\ 
%PFNR, c = 50.0 \%, \textcolor{red}{$f$-NeRV3} & 36.45 & 35.15 & 35.10 & 38.57 & 28.07 & 23.06 & 32.83 & 27.70 & 32.12 / 0.0 & -.- \% \\ 
%PFNR, c = 70.0 \%, \textcolor{red}{$f$-NeRV3} & 38.15 & 34.91 & 34.05 & 35.94 & 24.32 & 20.37 & 30.58 & 25.69 & 30.50 / 0.0 & -.- \% \\ \midrule 

%PFNR, c = 50.0 \%, \textcolor{red}{$f$-NeRV3-FP32} & 36.45 & 35.15 & 35.10 & 38.57 & 28.07 & 23.06 & 32.83 & 27.70 & 32.12 / ~~0.0~~ & -.- \% \\
%PFNR, c = 50.0 \%, \textcolor{red}{$f$-NeRV3-FP16} & 36.45 &	35.15 &	35.10 &	38.57 &	28.07 &	23.06 &	32.83 &	27.70 & 32.12 / ~~0.0~~ & -.- \% \\ 
%PFNR, c = 50.0 \%, \textcolor{red}{$f$-NeRV3-FP8}  & 36.43 & 35.15 &	35.09 &	38.56 &	28.07 &	23.06 &	32.83 &	27.70 &	32.11 / -0.01 & ~~-.- \% \\ 
%PFNR, c = 50.0 \%, \textcolor{red}{$f$-NeRV3-FP4}  & 32.79 & 34.27 & 34.30 & 37.70 & 27.74 & 22.76 & 32.40 & 27.37 & 31.17 / -0.60 & ~~-.-\% \\  \midrule 

MTL (upper-bound) & 34.22 & 32.79 & 32.34 & 38.33 & 25.30 & 22.44 & 33.73 & 27.05 & 30.78 / -~~~~&  ~~100.00 \% \\ 
\bottomrule
\end{tabular}
}
\label{table:uvg8_psnr_str}
%\vspace{-0.12in}
\end{table*}

%% file: materials/app_table_davis50_psnr.tex
\begin{table*}[!ht]
\small
\centering
\caption{\small PSNR results with Fourier Subnueral Operator (FSO) layer (\textcolor{red}{$f$-NeRV$\ast$}) (NeRV block, \Cref{table:architecture_detail}) on DAVIS50 Video Sessions with average PSNR and Backward Transfer (BWT), and Capacity (CAP). Note that $\ast$ denotes our reproduced results.}
\resizebox{0.75\textwidth}{!}{
\begin{tabular}{lccccccc}
\toprule 

\multicolumn{1}{c}{\multirow{2}{*}{\textbf{Method}}} & \multicolumn{5}{c}{\textbf{Video Sessions}} & \multirow{2}{*}{\thead{\textbf{Avg. PSNR / } \\ \textbf{BWT}}} \\ %& \multirow{2}{*}{\textbf{CAP}}  \\

\cline{2-6}
& \textbf{10} & \textbf{20} & \textbf{30} & \textbf{40} & \textbf{50} &  \\ \midrule 
STL, NeRV~\cite{chen2021nerv}$^{\ast}$ & 28.92 & 31.10 & 34.96 & 29.35 &  28.87 &  31.41 / ~~-~~ \\ %& ~~5000.00 \% \\ 

EWC~\cite{Kirkpatrick2017}$^{\ast}$    &  10.57	& 12.87 & 15.22 & 11.25 & 10.45 &	12.24 / -18.23 \\ % & ~~100.00 \% \\  
iCaRL~\cite{rebuffi2017icarl}$^{\ast}$ & 15.22 & 18.55 & 17.59 & 13.47 & 15.89 &	16.54 / ~~~-8.48 \\ % & ~~100.00 \% \\ 
ESMER~\cite{sarfraz2023error}$^{\ast}$ & 13.54 & 15.46 & 16.78 & 12.48 & 13.22 &	14.78 / ~-15.95  \\ %& ~~100.00 \% \\
\midrule

WSN$^{\ast}$, c = 30.0 \% & 19.20 &	20.80 &	23.39 &	21.56 &	21.45 &	21.56 / 0.0 \\ %& ~~~~67.45 \% \\ 

PFNR, c = 30.0 \%, \textcolor{red}{$f$-NeRV2} & 23.14 & 23.14 &	23.14 &	23.14 &	23.08 &	24.22 / 0.0  \\ % &  ~~~~97.84 \% \\
PFNR, c = 30.0 \%, \textcolor{red}{$f$-NeRV3} & \textbf{25.58} & \textbf{27.79} &	\textbf{31.42} & \textbf{27.22} & \textbf{24.88} & \textbf{27.57} / \textbf{0.0} \\ % &  ~~108.64 \% (maximum, 400 %)\\ 

\midrule 
MTL (upper-bound) & 23.10 &	23.19 &	24.63 &  22.84 & 23.45 & 24.57 / - ~~~ \\ %&  ~~100.00 \% \\ 
\bottomrule
\end{tabular}
}
\label{table:app_davis_psnr}
\vspace{-0.12in}
\end{table*}

%% file: materials/plot_transfer_matrix.tex
\begin{figure}[ht]
    \centering
    \small
    %\vspace{-0.1in}
    \setlength{\tabcolsep}{0pt}{%
    \begin{tabular}{cc}
    % we prepare the figure 
    %\includegraphics[width=0.5\columnwidth]{images/transfer_matrix/UVG17B_CVRNet_epoch150_UVGB_sparsity0.30000000000000004_fc9_16_112_psnr.pdf} & 
    %\includegraphics[width=0.5\columnwidth]{images/transfer_matrix/UVG17B_CVRNet_epoch150_UVGB_sparsity0.30000000000000004_fc9_16_112_masssim.pdf} \\
    %\small (a) PSNR, $c = 30.0\%$ & \small (b) MS-SSIM, $c = 30.0\%$ \\

    \includegraphics[width=0.5\columnwidth]{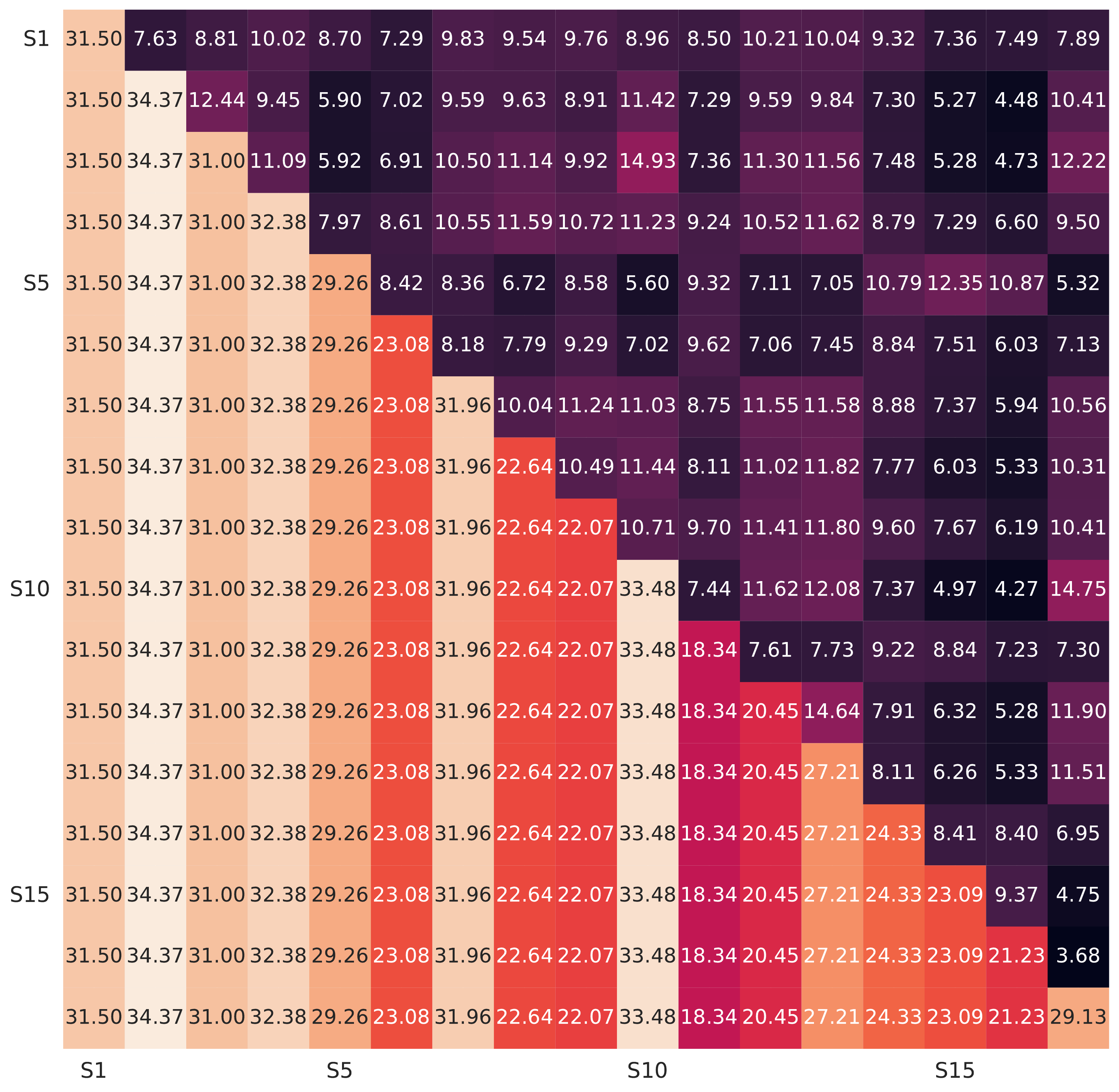} & 
    %\includegraphics[width=0.5\columnwidth]{images/transfer_matrix/UVG17B_CVRNet_epoch150_increase0.01_sparsity0.30000000000000004_fc9_16_112_Fusion6_reinit_psnr.pdf} \\
    %\small (a) WSN, $c = 30.0\%$ & \small (b) PFNR, $c = 30.0\%$, $f$-NeRV2 \\
    \includegraphics[width=0.5\columnwidth]{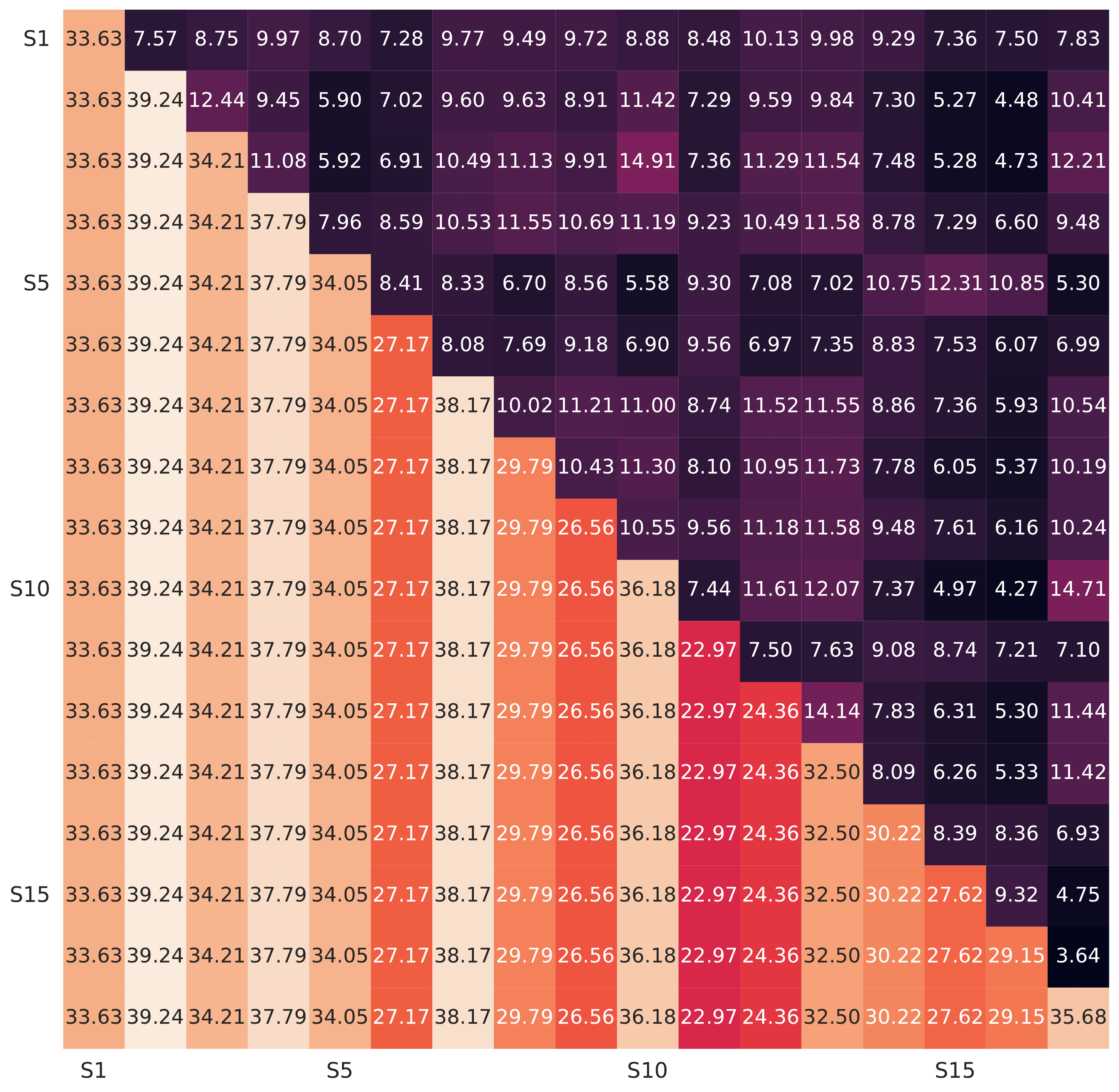} \\
    \small (a) WSN, $c = 30.0\%$ & \small (b) PFNR, $c = 30.0\%$, $f$-NeRV3 \\
    
    \end{tabular}
    }
    %\caption{ExNIR's Transfer Matrixes of PSNR and MS-SSIM on the UVG 17 dataset.}
    \caption{\small Transfer Matrixes of WSN v.s. PFNR on the UVG17 dataset measured by PSNR of source and target.}
    \label{fig:transf_matrix}
    \vspace{-0.05in}
\end{figure}

%% file: materials/main_table_uvg8_psnr_fso.tex
\begin{table*}[!ht]
\small
\centering
\caption{\small PSNR results with Fourier Subnueral Operator (FSO) layer (\textcolor{red}{$f$-NeRV$\ast$}) (detailed in \Cref{table:architecture_detail}) on UVG8 Video Sessions with average PSNR and Backward Transfer (BWT). Note that \textit{w/o imag.} ignores the imaginary part in $f$-NeRV$\ast$.}
\resizebox{0.83\textwidth}{!}{
\begin{tabular}{lcccccccccc}
\toprule 

\multicolumn{1}{c}{\multirow{2}{*}{\textbf{Method}}} & \multicolumn{8}{c}{\textbf{Video Sessions}} & \multirow{2}{*}{\thead{\textbf{Avg. PSNR / } \\ \textbf{BWT}}} \\ % & \multirow{2}{*}{\textbf{CAP}}  \\

\cline{2-9}
& \textbf{1} & \textbf{2} & \textbf{3} & \textbf{4} & \textbf{5} & \textbf{6} & \textbf{7} & \textbf{8} &  \\ \midrule

% real/image
%PFNR, c = 10.0 \%, \textcolor{red}{$f$-NeRV2} & 28.49 & 32.30 & 30.30 & 35.12 & 24.10 & 19.82 & 29.89 & 24.76 & 28.10 / 0.0  \\ % &  ~~~~31.91 \% \\ 
%PFNR, c = 30.0 \%, \textcolor{red}{$f$-NeRV2} & 31.99 & 33.56 & 31.82 & 36.61 & 25.28 & 20.97 & 31.07 & 25.73 & 29.63 / 0.0  \\ % &  ~~~~72.35 \% \\
PFNR, c = 50.0 \%, \textcolor{red}{$f$-NeRV2} & \textbf{34.46} & \textbf{33.91} & \textbf{32.17} & \textbf{36.43} & \textbf{25.26} & \textbf{20.74} & \textbf{30.18} & \textbf{25.45} & \textbf{29.82} / \textbf{0.0}  \\ % &  ~~96.82 \% \\ 
%PFNR, c = 70.0 \%, \textcolor{red}{$f$-NeRV2} & 36.04 & 33.46 & 31.05 & 32.57 & 23.40 & 19.41 & 28.31 & 24.31 & 28.57 / 0.0  \\ % &  ~~104.26 \% \\
%\midrule 

% only real part
%PFNR, c = 10.0 \%, \textcolor{red}{$f$-NeRV2} w/o imag & 28.58 & 32.11 & 30.00 & 35.05 & 23.99 & 19.82 & 29.85 & 24.78 & 28.02 / 0.0  \\ % &  ~~~~31.91 \% \\ 
%PFNR, c = 30.0 \%, \textcolor{red}{$f$-NeRV2} w/o imag & 31.90 & 33.42 & 31.66 & 36.59 & 25.21 & 30.84 & 30.98 & 25.46 & 29.51 / 0.0  \\ % &  ~~~~72.35 \% \\
PFNR, c = 50.0 \%, \textcolor{red}{$f$-NeRV2} \textbf{w/o imag.} & 34.34 & 33.79 & 32.04 & 36.40 & 25.11 & 20.59 & 30.17 & 25.27 & 29.71 / 0.0  \\ % &  ~~96.82 \% \\ 
%PFNR, c = 70.0 \%, \textcolor{red}{$f$-NeRV2} w/o imag & 35.89 & 33.35 & 30.94 & 31.94 & 23.34 & 19.06 & 27.96	& 24.10 & 28.32 / 0.0  \\ % &  ~~104.26 \% \\
\midrule

% real-image
%PFNR, c = 10.0 \%, \textcolor{red}{$f$-NeRV3} & 29.78 & 33.30 & 32.29 & 37.29 & 24.65 & 20.82 & 31.94 & 26.19 & 29.53 / 0.0 \\ % & ~~482.38 \% \\ 
%PFNR, c = 30.0 \%, \textcolor{red}{$f$-NeRV3} & 33.69 & 34.76 & 34.57 & 38.50 & 27.09 & 23.16 & 33.10 & 27.94 & 31.36 / 0.0 \\ % & 1093.04 \% \\ 
PFNR, c = 50.0 \%, \textcolor{red}{$f$-NeRV3} & \textbf{36.45} & \textbf{35.15} & \textbf{35.10} & \textbf{38.57} & \textbf{28.07} & \textbf{23.06} & \textbf{32.83} & \textbf{27.70} & \textbf{32.12} / \textbf{0.0} \\ % & 1463.60 \% \\ 
%PFNR, c = 70.0 \%, \textcolor{red}{$f$-NeRV3} & 38.15 & 34.91 & 34.05 & 35.94 & 24.32 & 20.37 & 30.58 & 25.69 & 30.50 / 0.0 \\ % & 1575.00 \% \\ 

% only real part
%PFNR, c = 10.0 \%, \textcolor{red}{$f$-NeRV3} w/o imag  & 29.02 &	32.68 &	31.08 &	36.90 &	24.15 &	20.37 &	31.51 &	25.46 &	28.90 / 0.0 \\ % & ~~482.38 \% \\ 
%PFNR, c = 30.0 \%, \textcolor{red}{$f$-NeRV3} w/o imag & 32.90 &	34.28 &	33.46 &	38.08 &	26.13 &	22.34 &	32.69 &	27.15 &	30.88 / 0.0 \\ % & 1093.04 \% \\ 
PFNR, c = 50.0 \%, \textcolor{red}{$f$-NeRV3} \textbf{w/o imag.} & 35.66 & 34.65 &	34.09 &	37.95 &	25.80 &	21.94 &	32.17 &	26.91 &	31.15 / 0.0 \\ % & 1463.60 \% \\ 
%PFNR, c = 70.0 \%, \textcolor{red}{$f$-NeRV3} w/o imag & 37.56 &	34.36 &	33.11 &	35.12 &	23.58 &	20.04 &	 29.83 & 25.39 & 29.88 / 0.0 \\ % & 1575.00 \% \\ 

\bottomrule
\end{tabular}
}
\label{table:uvg8_fso_real}
%\vspace{-0.12in}
\end{table*}

%% file: materials/main_table_uvg8_psnr_conv.tex
\begin{table*}[!ht]
\small
\centering
\caption{\small PSNR results with Fourier Subnueral Operator (FSO) layer (\textcolor{red}{$f$-NeRV$\ast$}) (detailed in \Cref{table:architecture_detail}) on UVG8 Video Sessions with average PSNR and Backward Transfer (BWT). Note that \textit{w/o conv.} ignores the conv. layer in $f$-NeRV$\ast$.}
\resizebox{0.83\textwidth}{!}{
\begin{tabular}{lcccccccccc}
\toprule 

\multicolumn{1}{c}{\multirow{2}{*}{\textbf{Method}}} & \multicolumn{8}{c}{\textbf{Video Sessions}} & \multirow{2}{*}{\thead{\textbf{Avg. PSNR / } \\ \textbf{BWT}}} \\ % & \multirow{2}{*}{\textbf{CAP}}  \\

\cline{2-9}
& \textbf{1} & \textbf{2} & \textbf{3} & \textbf{4} & \textbf{5} & \textbf{6} & \textbf{7} & \textbf{8} &  \\ \midrule

% real/image
%PFNR, c = 10.0 \%, \textcolor{red}{$f$-NeRV2} & 28.49 & 32.30 & 30.30 & 35.12 & 24.10 & 19.82 & 29.89 & 24.76 & 28.10 / 0.0  \\ % &  ~~~~31.91 \% \\ 
%PFNR, c = 30.0 \%, \textcolor{red}{$f$-NeRV2} & 31.99 & 33.56 & 31.82 & 36.61 & 25.28 & 20.97 & 31.07 & 25.73 & 29.63 / 0.0  \\ % &  ~~~~72.35 \% \\
PFNR, c = 50.0 \%, \textcolor{red}{$f$-NeRV2} & \textbf{34.46} & \textbf{33.91} & \textbf{32.17} & \textbf{36.43} & \textbf{25.26} & \textbf{20.74} & \textbf{30.18} & \textbf{25.45} & \textbf{29.82} / \textbf{0.0}  \\ % &  ~~96.82 \% \\ 
%PFNR, c = 70.0 \%, \textcolor{red}{$f$-NeRV2} & 36.04 & 33.46 & 31.05 & 32.57 & 23.40 & 19.41 & 28.31 & 24.31 & 28.57 / 0.0  \\ % &  ~~104.26 \% \\
%\midrule 

% only real part
PFNR, c = 50.0 \%, \textcolor{red}{$f$-NeRV2} \textbf{w/o conv.} & 30.05 &  32.10 & 30.12 &	31.82 &	24.00 &	19.60 &	28.21 & 	24.47 &	27.54 / 0.0  \\ % &  ~~96.82 \% \\ 
\midrule

% real-image
%PFNR, c = 10.0 \%, \textcolor{red}{$f$-NeRV3} & 29.78 & 33.30 & 32.29 & 37.29 & 24.65 & 20.82 & 31.94 & 26.19 & 29.53 / 0.0 \\ % & ~~482.38 \% \\ 
%PFNR, c = 30.0 \%, \textcolor{red}{$f$-NeRV3} & 33.69 & 34.76 & 34.57 & 38.50 & 27.09 & 23.16 & 33.10 & 27.94 & 31.36 / 0.0 \\ % & 1093.04 \% \\ 
PFNR, c = 50.0 \%, \textcolor{red}{$f$-NeRV3} & \textbf{36.45} & \textbf{35.15} & \textbf{35.10} & \textbf{38.57} & \textbf{28.07} & \textbf{23.06} & \textbf{32.83} & \textbf{27.70} & \textbf{32.12} / \textbf{0.0} \\ % & 1463.60 \% \\ 
%PFNR, c = 70.0 \%, \textcolor{red}{$f$-NeRV3} & 38.15 & 34.91 & 34.05 & 35.94 & 24.32 & 20.37 & 30.58 & 25.69 & 30.50 / 0.0 \\ % & 1575.00 \% \\ 

% only real part
PFNR, c = 50.0 \%, \textcolor{red}{$f$-NeRV3} \textbf{w/o conv.} & 35.46 & 35.06 & 34.98 & 38.23 & 28.00 & 22.98 & 32.57 & 27.45 & 31.84 / 0.0 \\ % & 1463.60 \% \\ 

\bottomrule
\end{tabular}
}
\label{table:uvg8_fso_conv}
\vspace{-0.12in}
\end{table*}

%% file: materials/main_table_uvg8_psnr_stl.tex
\begin{table*}[!ht]
\small
\centering
\caption{\small PSNR results of STL with Fourier Subnueral Operator (FSO) layer (\textcolor{red}{$f$-NeRV$\ast$}) (detailed in \Cref{table:architecture_detail}) on UVG8 Video Sessions with average PSNR and Backward Transfer (BWT).}
\resizebox{0.83\textwidth}{!}{
\begin{tabular}{lcccccccccc}
\toprule 

\multicolumn{1}{c}{\multirow{2}{*}{\textbf{Method}}} & \multicolumn{8}{c}{\textbf{Video Sessions}} & \multirow{2}{*}{\thead{\textbf{Avg. PSNR / } \\ \textbf{BWT}}} \\ % & \multirow{2}{*}{\textbf{CAP}}  \\

\cline{2-9}
& \textbf{1} & \textbf{2} & \textbf{3} & \textbf{4} & \textbf{5} & \textbf{6} & \textbf{7} & \textbf{8} &  \\ \midrule 
%STL, NeRV~\cite{chen2023hnerv} & 39.63 & 36.06 & 37.35 & 41.23 & 38.14 & 31.86 & 37.22 & 32.45 & 36.74 / ~~-~~ \\ % & ~~800.00 \% \\

STL, NeRV~\cite{chen2021nerv}$^{\ast}$  & 39.66 & 36.28 & 38.14 & 42.03 & 36.58 & 29.22 & 37.27 & 31.45 & 36.33 / ~~-~~ \\ %& ~~800.00 \% \\ 
\midrule 

STL, NeRV , \textcolor{red}{$f$-NeRV2}            & {39.73} & {36.30} & {38.29} & {42.03} & {36.64} & {29.25} & {37.35} & {31.65} & {36.40} / ~~-~~ \\ %& ~~800.00 \% \\ 

%STL, NeRV , \textcolor{red}{$f$-NeRV2}, c=50.0\%  & 39.68 &	36.29 &	38.20 &	42.03 &	36.60 &	29.25 &	37.29 &	31.53 &	36.36 / ~~-~~ \\ %& ~~800.00 \% \\ 
%\midrule 

STL, NeRV , \textcolor{red}{$f$-NeRV3}  & \textbf{42.75}  & \textbf{37.65}  & \textbf{42.05} & \textbf{42.36} & \textbf{40.01} & \textbf{34.21} & \textbf{40.15}  & \textbf{36.15} & \textbf{39.41}  / ~~-~~ \\ %& ~~800.00 \% \\ 
%STL, NeRV , \textcolor{red}{$f$-NeRV3}, c=50.0\%  &  &  &  &  &  &  &  &  &  / ~~-~~ \\ %& ~~800.00 \% \\ 

%MLT (upper-bound) & 34.22 & 32.79 & 32.34 & 38.33 & 25.30 & 22.44 & 33.73 & 27.05 & 30.78 / -~~~~ \\ % &  ~~100.00 \% \\ 
\bottomrule
\end{tabular}
}
\label{table:uvg8_stl}
\vspace{-0.12in}
\end{table*}

%% file: materials/main_table_uvg8_stl_time.tex
\begin{table*}[!ht]
\small
\centering
\caption{\small Training time and Decoding FPS with Fourier Subnueral Operator (FSO) layer (\textcolor{red}{$f$-NeRV2}) (detailed in \Cref{table:architecture_detail}) on UVG8 Video Sessions}
\resizebox{0.6\textwidth}{!}{
\begin{tabular}{lcc}
\toprule 

\textbf{Method} & \textbf{Training time [hours] } & \textbf{Decoding FPS}  \\ \midrule 

STL, NeRV~\cite{chen2021nerv}$^{\ast}$  & ~~13.10 & 56.88  \\  %\midrule 
STL, NeRV, \textcolor{red}{$f$-NeRV2}  & ~~25.66  & 31.72 \\ \midrule 
%STL, NeRV , \textcolor{red}{$f$-NeRV3}  & 48.22 & 5.66  \\ 

ESMER~\cite{sarfraz2023error}$^{\ast}$  & 101.71 & 56.93  \\
PFNR, \textcolor{red}{$f$-NeRV2}       & ~~31.66  & 31.55 \\ 
%PFNR, \textcolor{red}{$f$-NeRV3}       & 53.73  & 5.53 \\ 

\bottomrule
\end{tabular}
}
\label{table:uvg8_stl_time}
\vspace{-0.12in}
\end{table*}

%% file: materials/plot_video_reinit_mtl.tex
\begin{figure}[ht]
    \centering 
    %\vspace{-0.1in}
    \setlength{\tabcolsep}{0pt}{%
    \begin{tabular}{ccc}
    
    % 2.city 
    t=0 & t=1 & t=2 \\ %&t=3 \\
    %\includegraphics[width=0.22\columnwidth]{images/video_reinit/1/gt_0.png} & 
    %\includegraphics[width=0.22\columnwidth]{images/video_reinit/1/gt_1.png} &
    %\includegraphics[width=0.22\columnwidth]{images/video_reinit/1/gt_2.png} &
    %\includegraphics[width=0.22\columnwidth]{images/video_reinit/1/gt_3.png} \\ 
    %\multicolumn{4}{l}{\small Ground-truth of \textit{2.city}}\\

    % wsn
    \includegraphics[width=0.3\columnwidth]{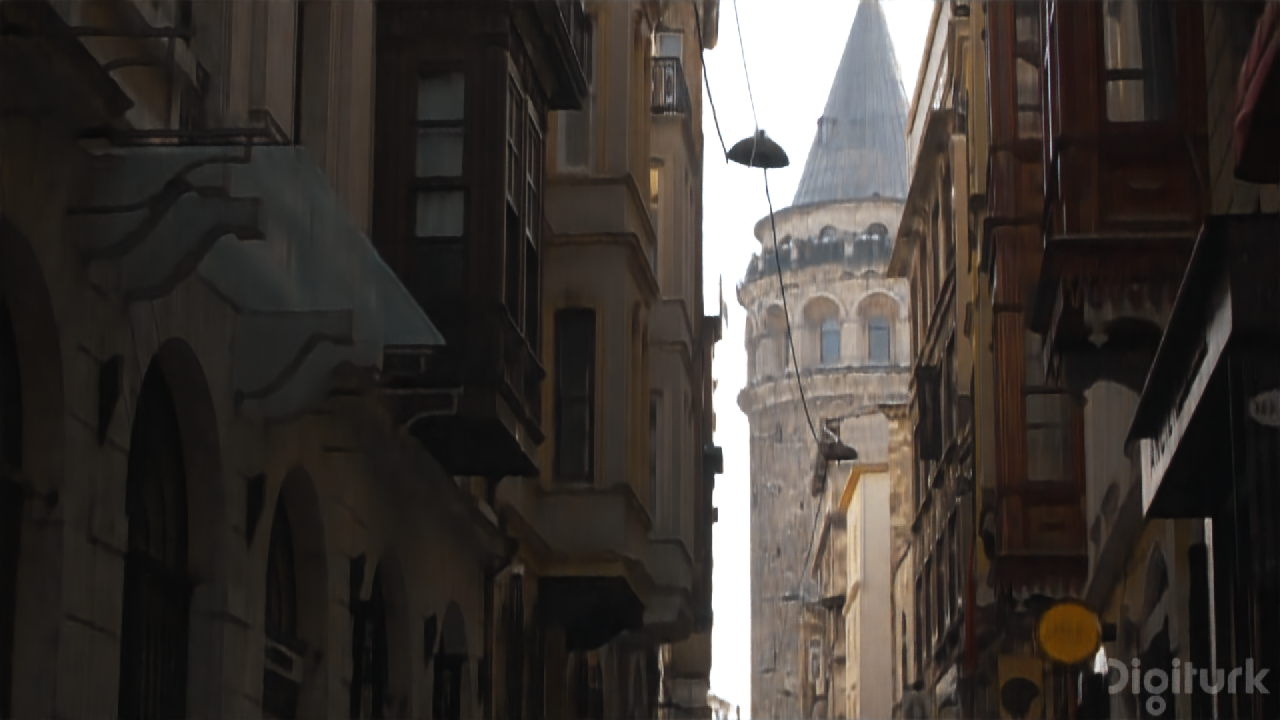} & 
    \includegraphics[width=0.3\columnwidth]{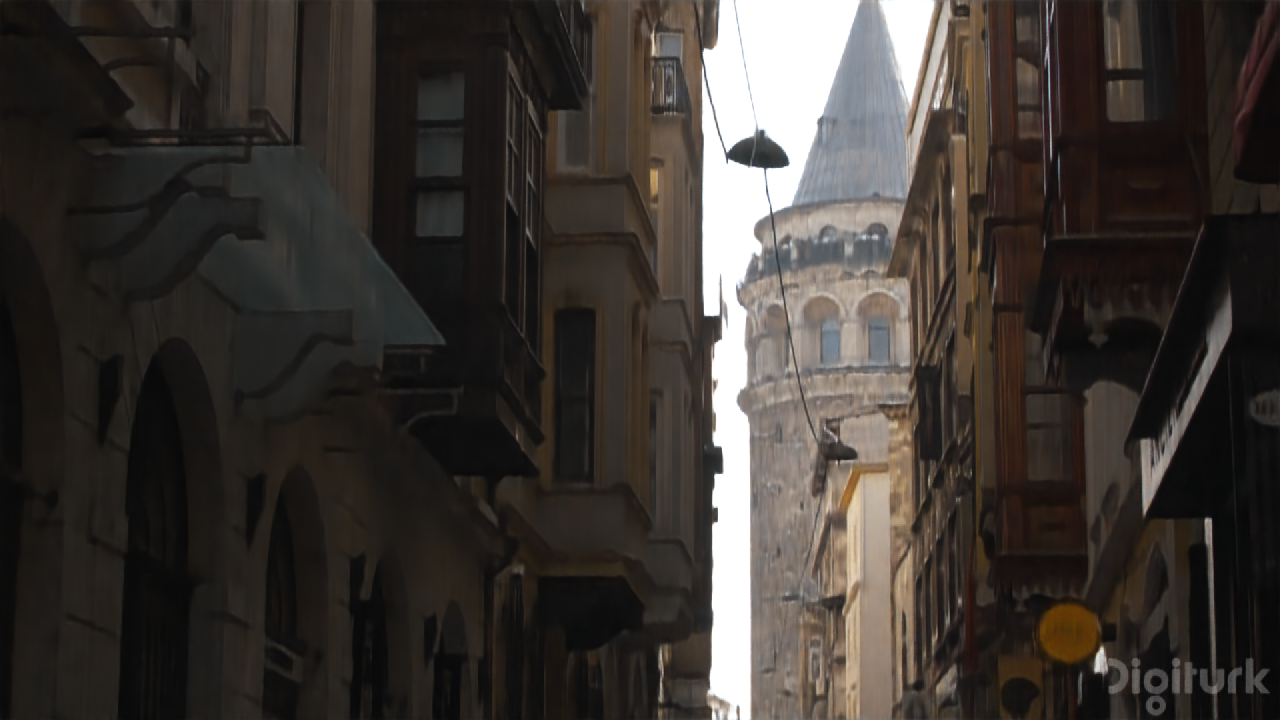} &
    \includegraphics[width=0.3\columnwidth]{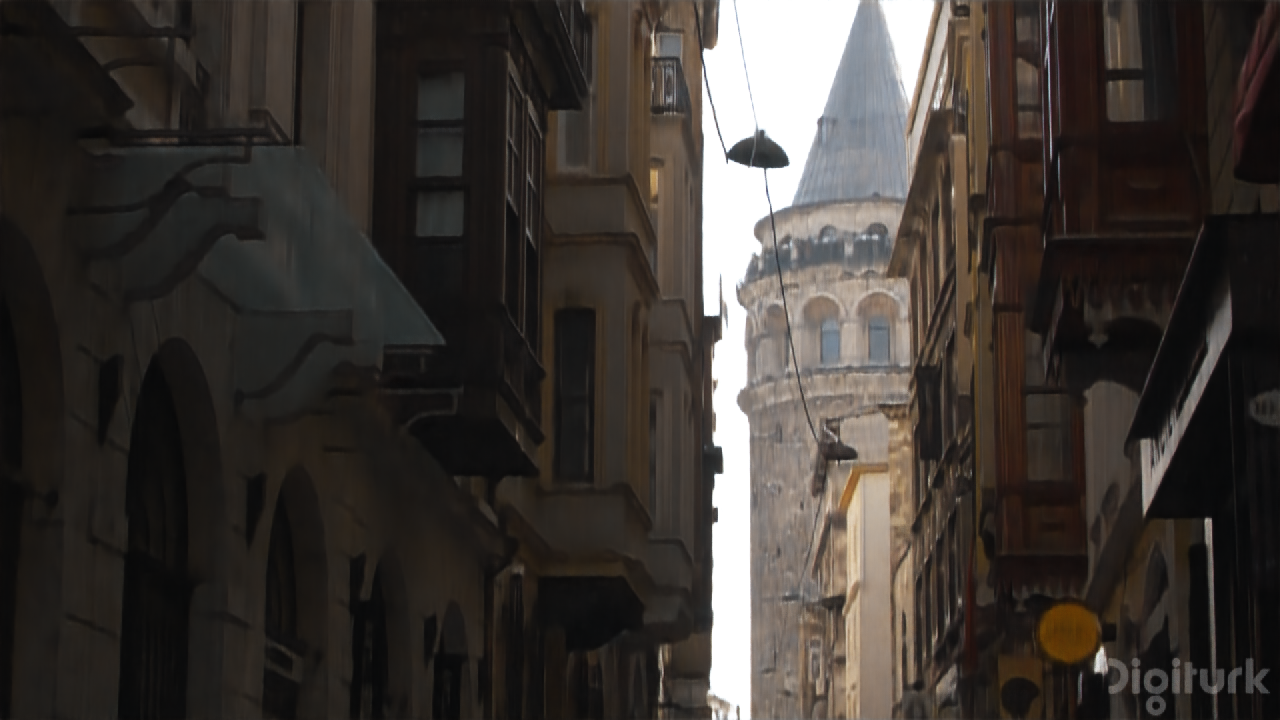} \\ %&
    \multicolumn{3}{l}{\small WSN (34.37, PSNR)}\\
    
    \includegraphics[width=0.3\columnwidth]{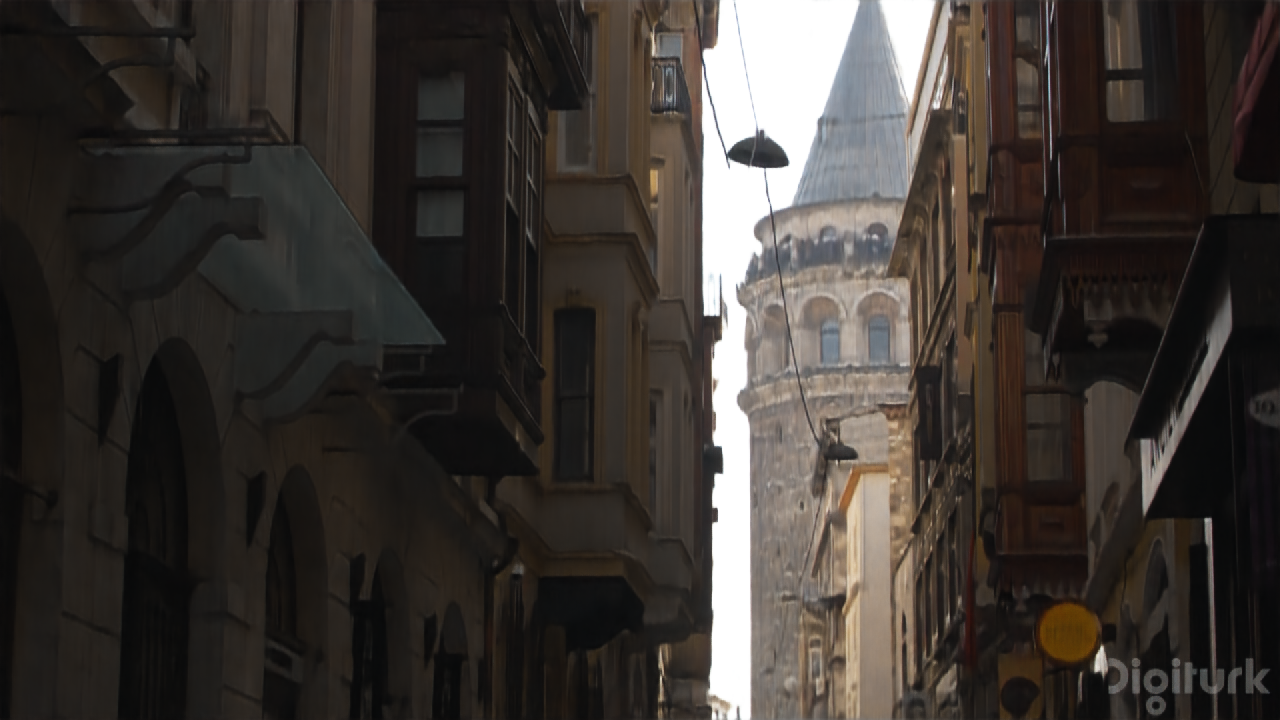} & 
    \includegraphics[width=0.3\columnwidth]{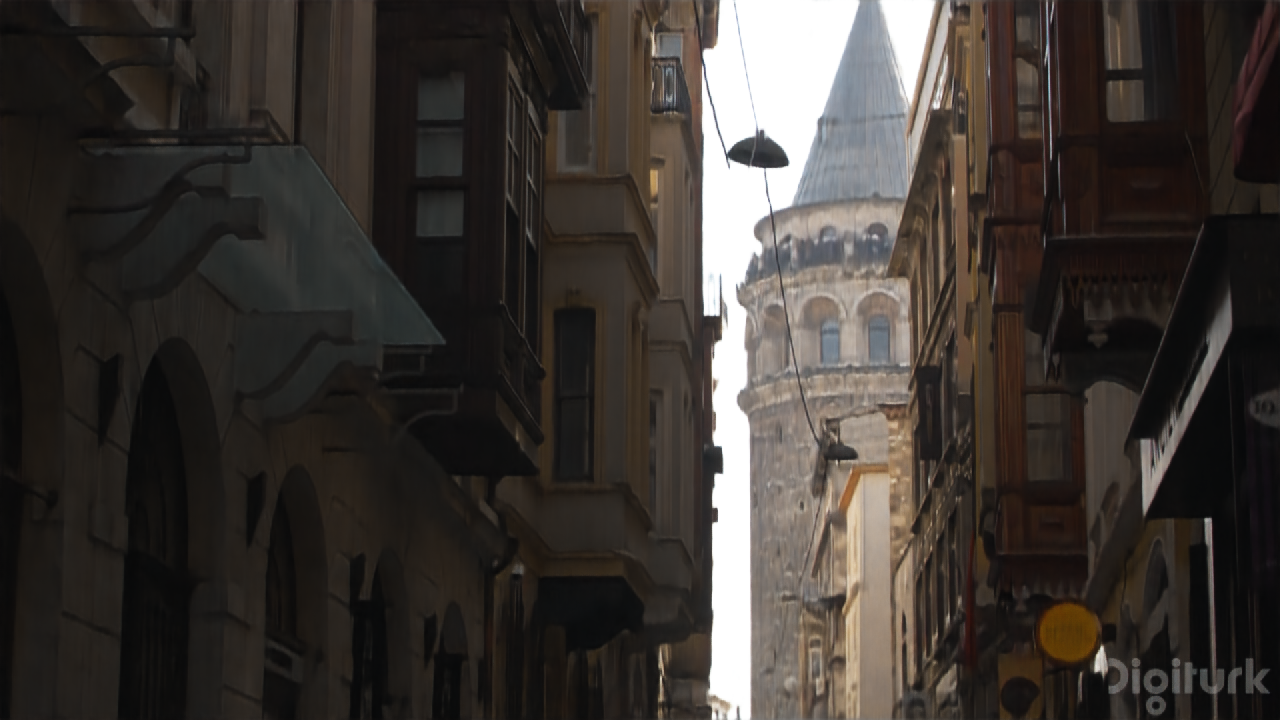} &
    \includegraphics[width=0.3\columnwidth]{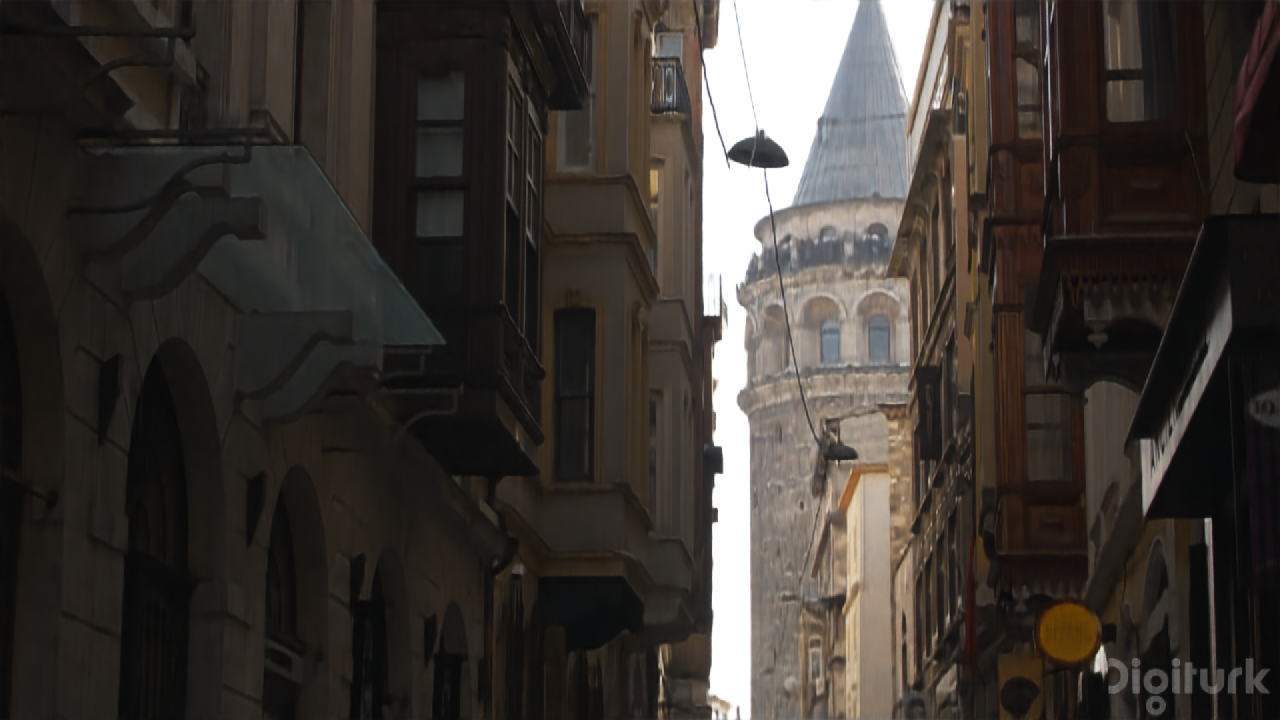} \\ % &
    \multicolumn{3}{l}{\small PFNR, \textcolor{red}{$f$-NeRV2} (35.84, PSNR)}\\

    \includegraphics[width=0.3\columnwidth]{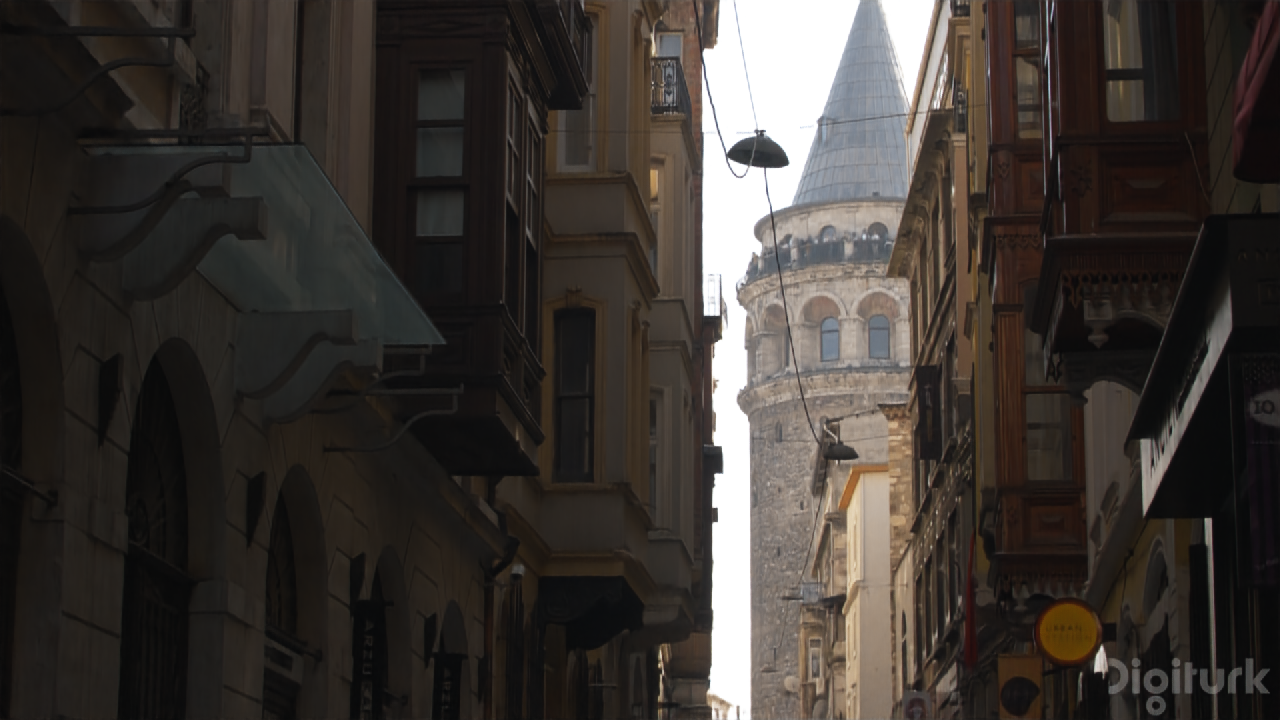} & 
    \includegraphics[width=0.3\columnwidth]{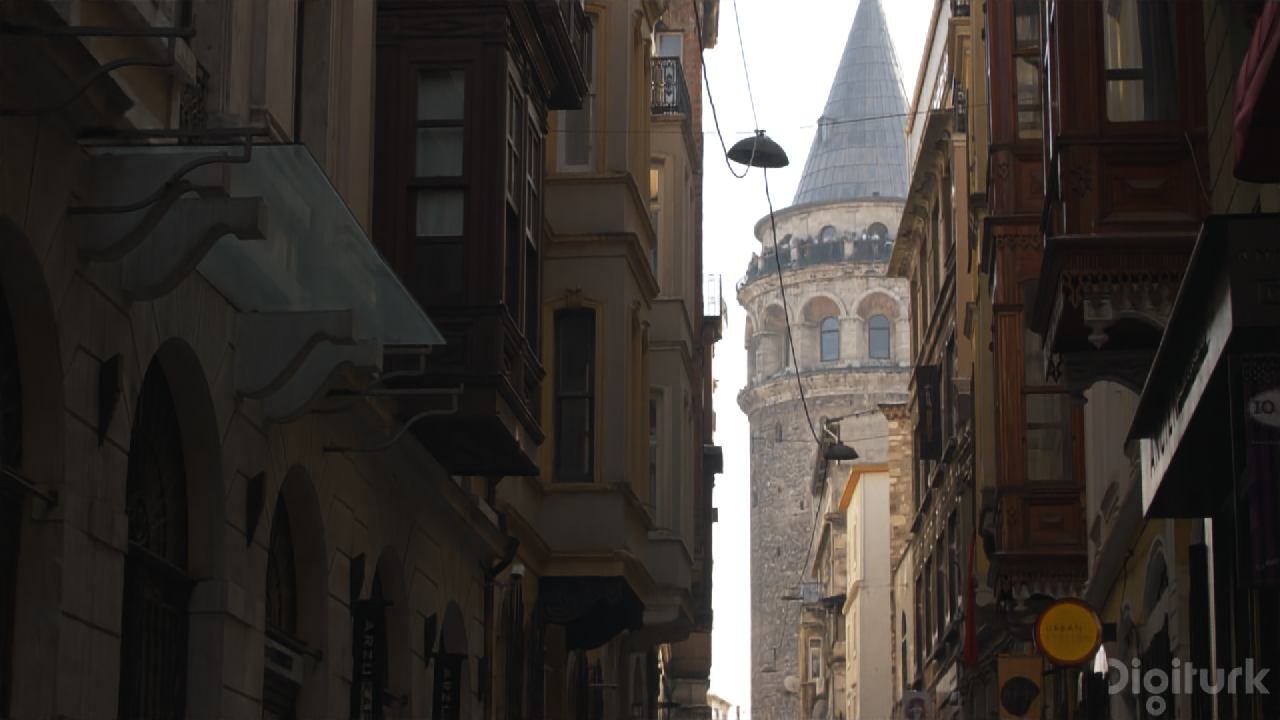} &
    \includegraphics[width=0.3\columnwidth]{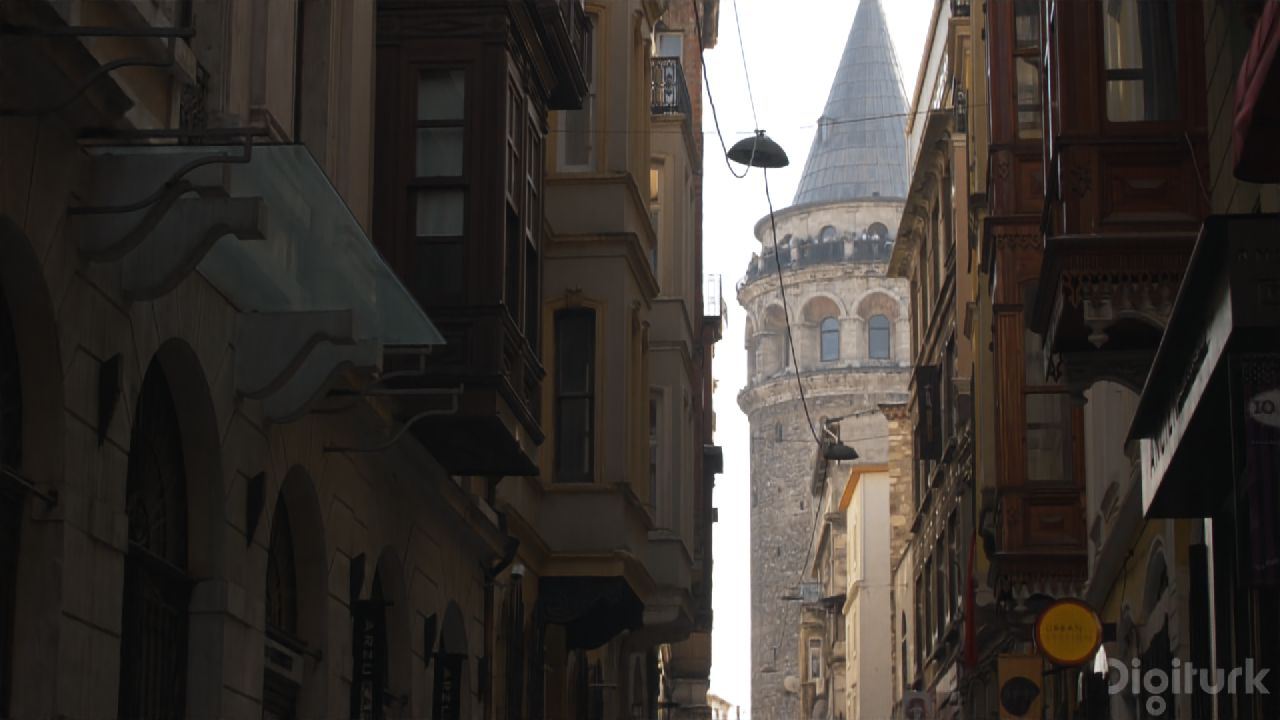} \\ % &
    \multicolumn{3}{l}{\small PFNR, \textcolor{red}{$f$-NeRV3} (39.24, PSNR)}\\

    \includegraphics[width=0.3\columnwidth]{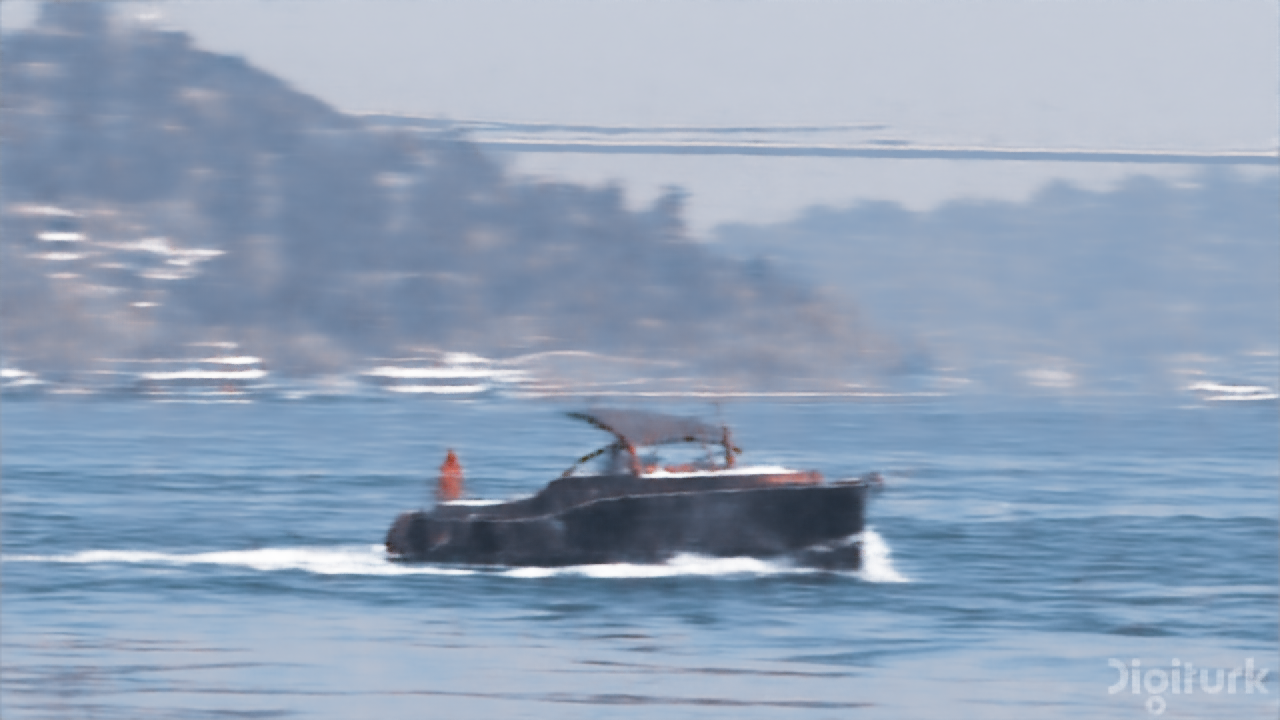} & 
    \includegraphics[width=0.3\columnwidth]{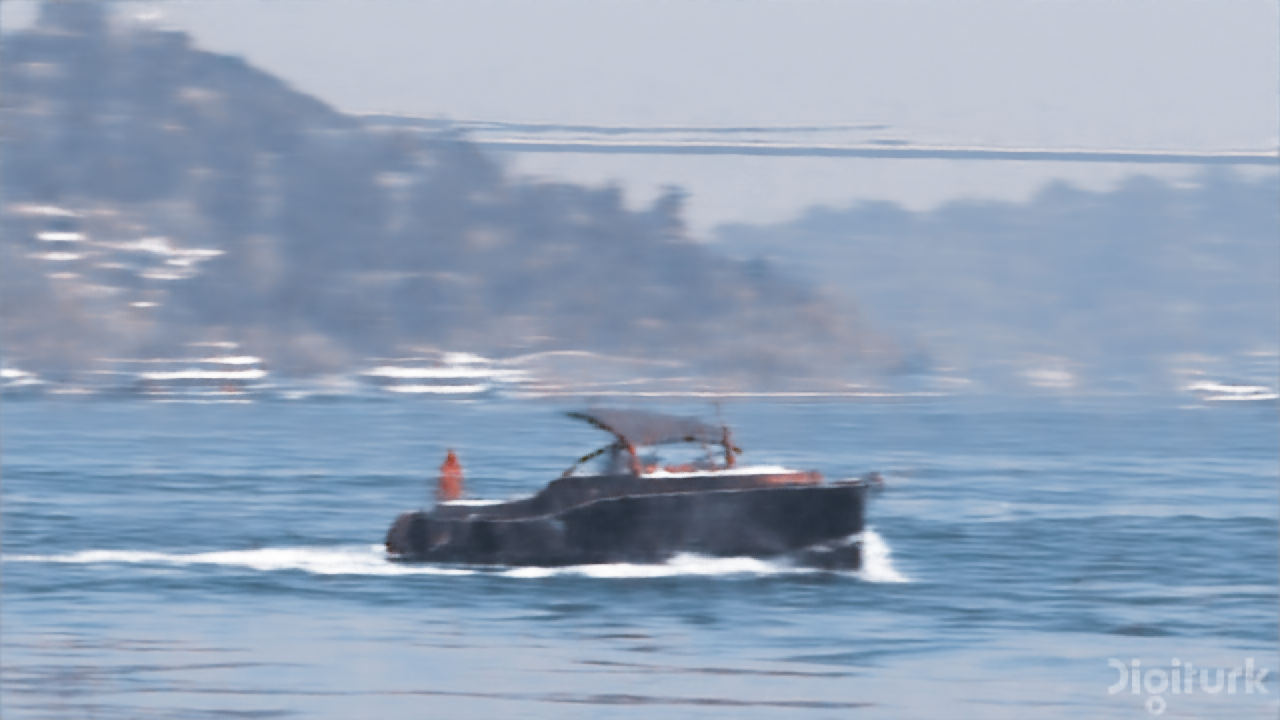} &
    \includegraphics[width=0.3\columnwidth]{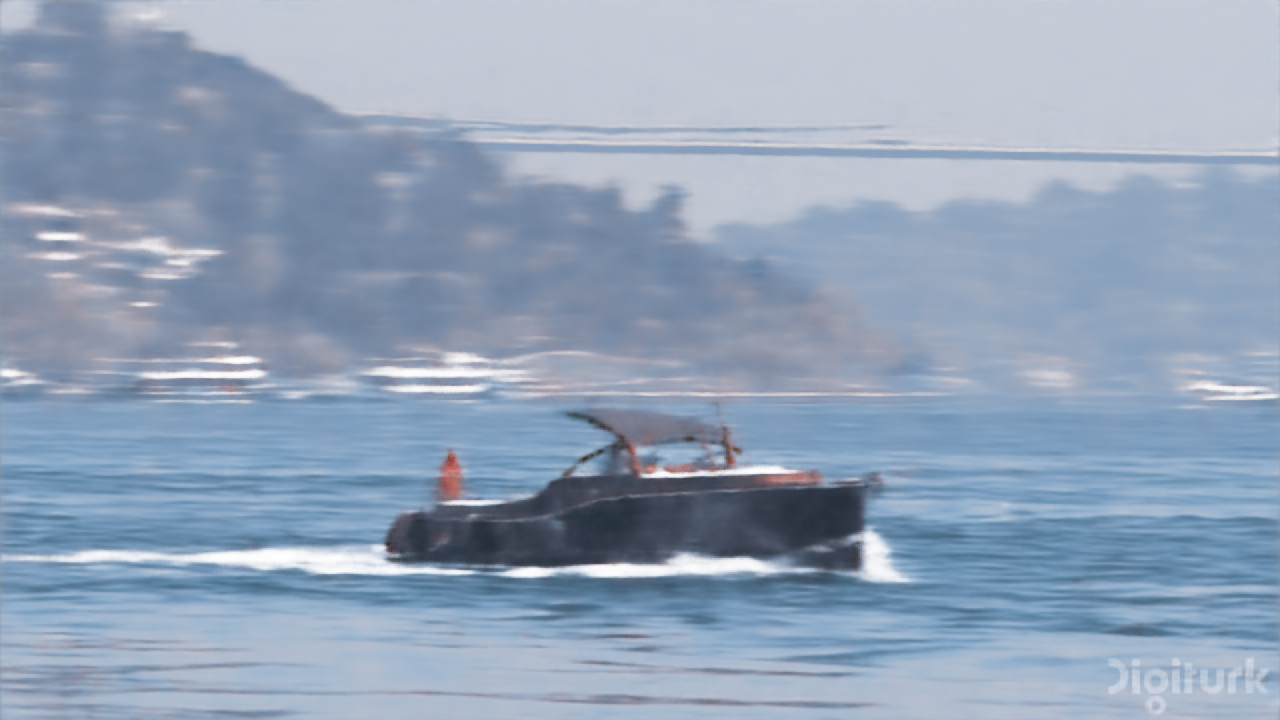} \\ %&
    \multicolumn{3}{l}{\small WSN (29.26, PSNR)}\\
    
    \includegraphics[width=0.3\columnwidth]{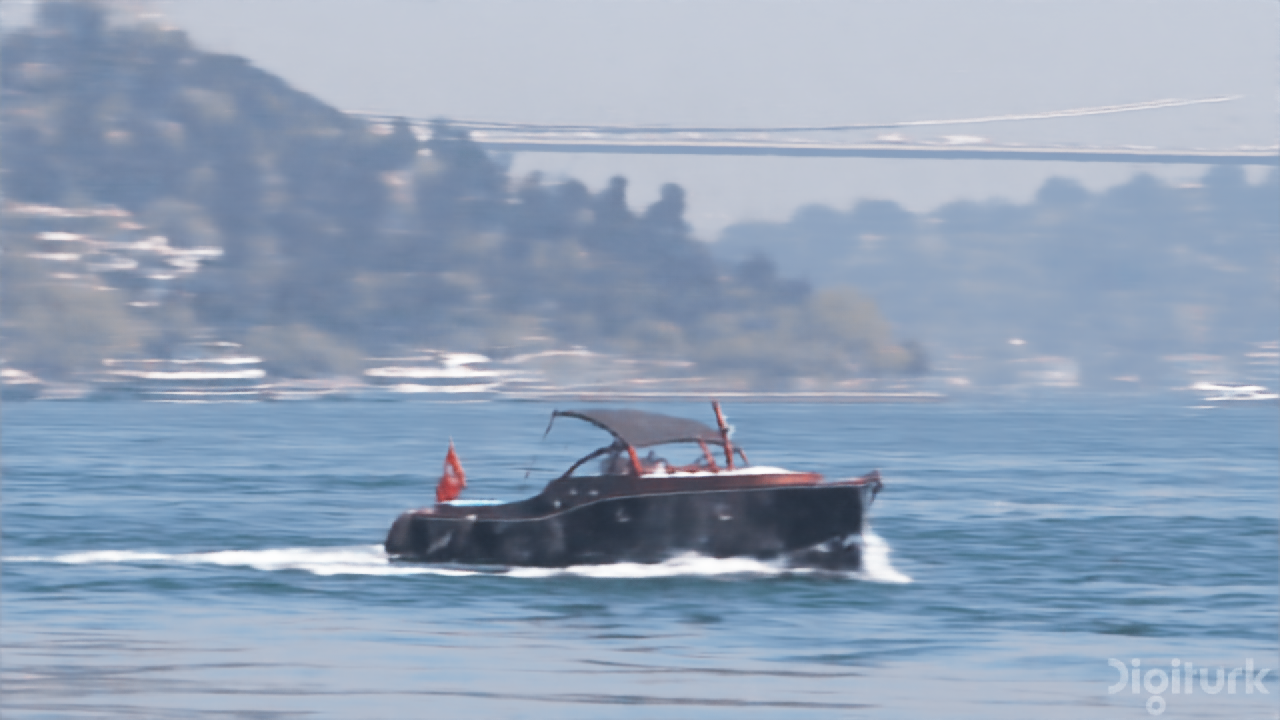} & 
    \includegraphics[width=0.3\columnwidth]{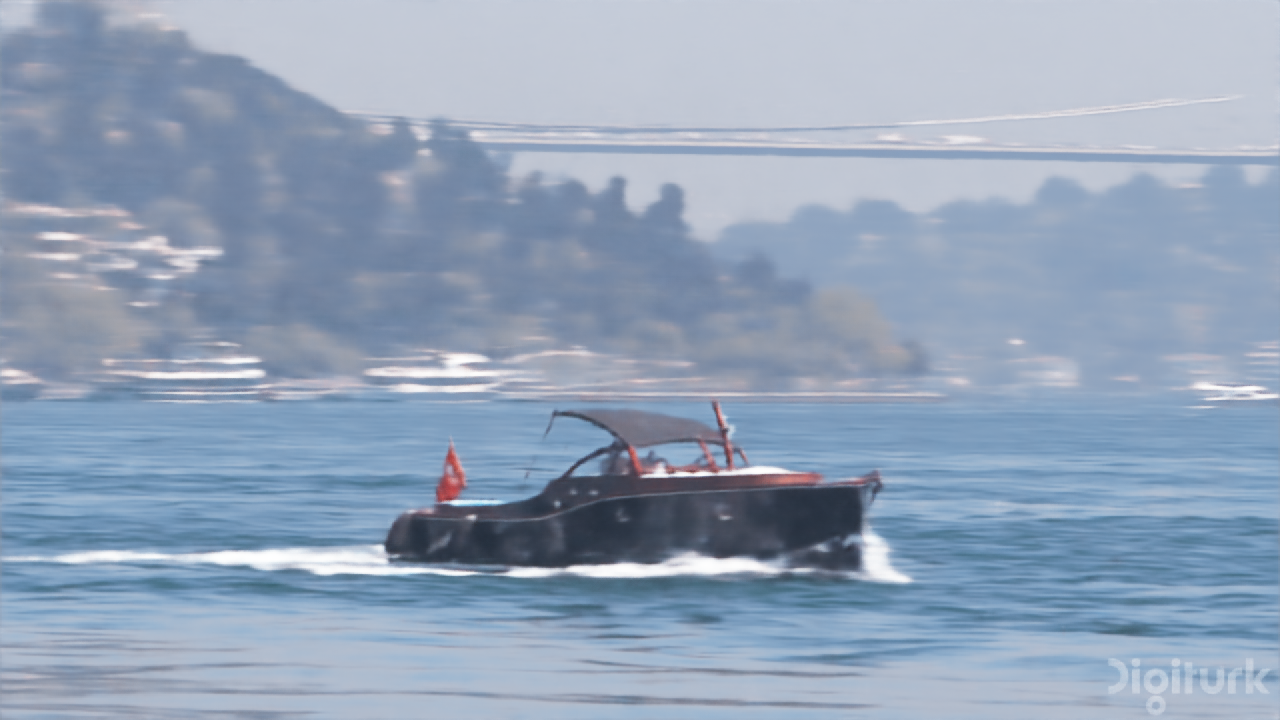} &
    \includegraphics[width=0.3\columnwidth]{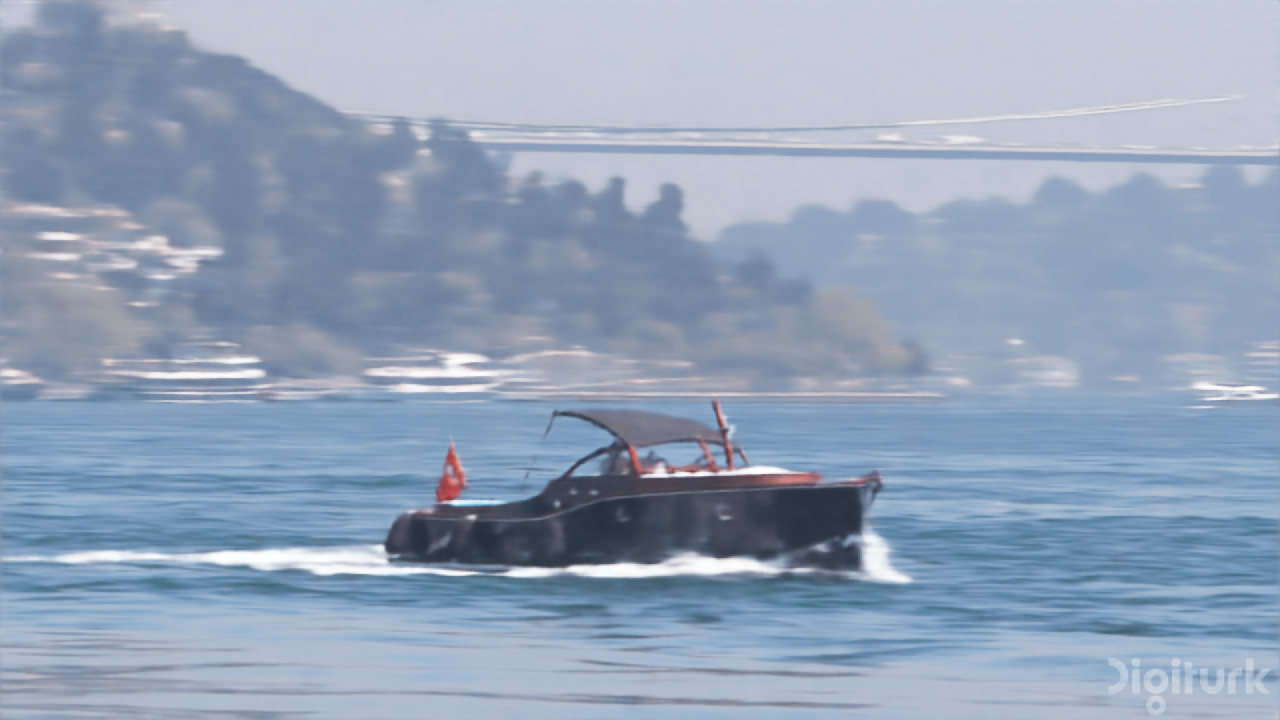} \\ %&
    \multicolumn{3}{l}{\small PFNR, \textcolor{red}{$f$-NeRV2} (31.24, PSNR)}\\

    \includegraphics[width=0.3\columnwidth]{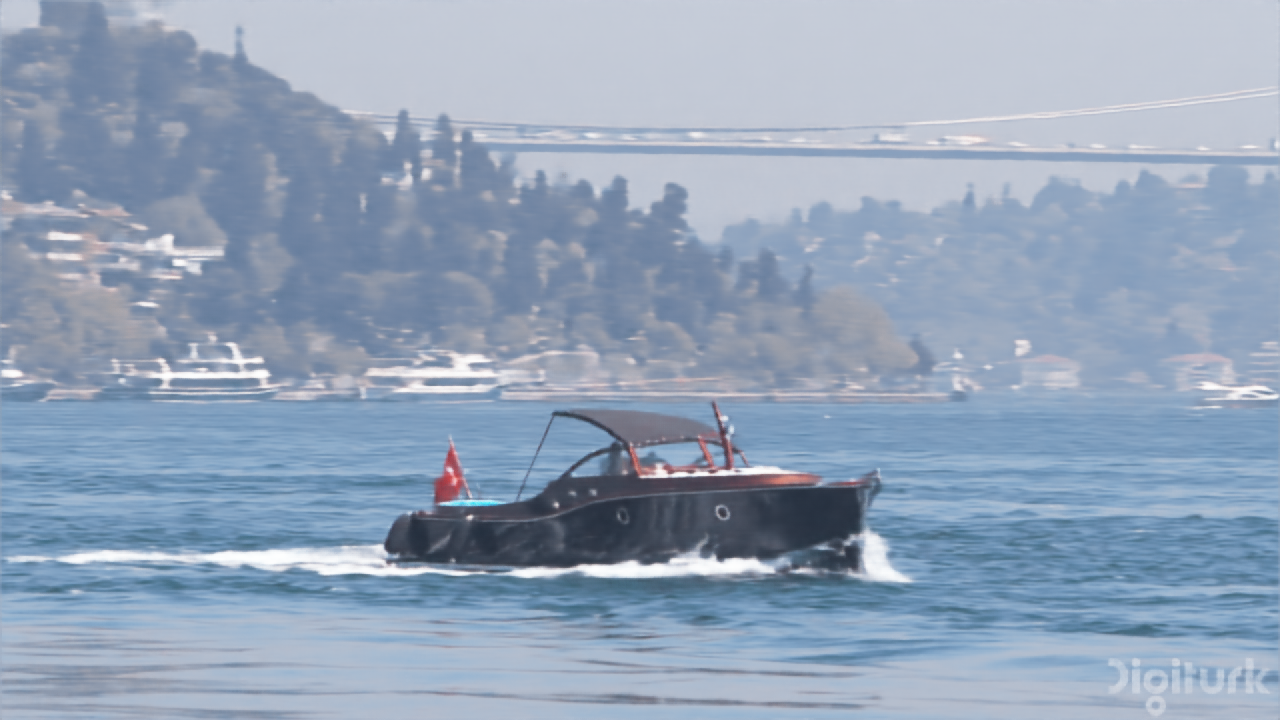} & 
    \includegraphics[width=0.3\columnwidth]{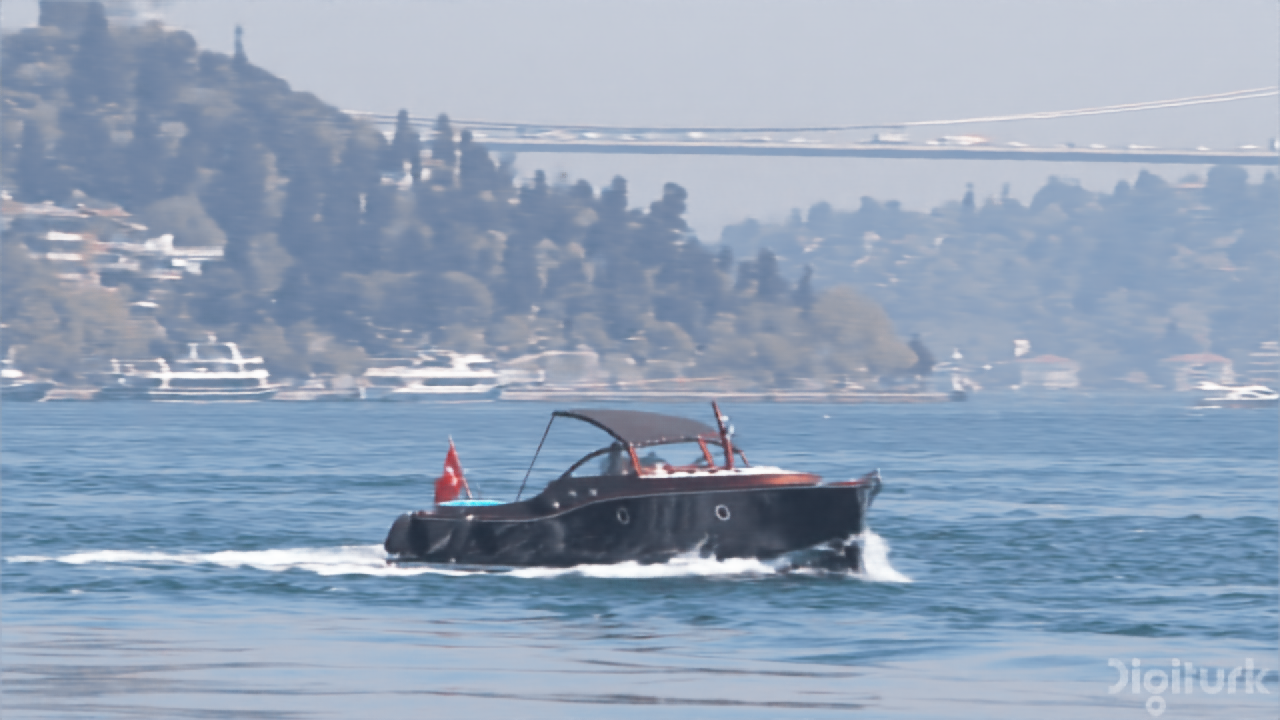} &
    \includegraphics[width=0.3\columnwidth]{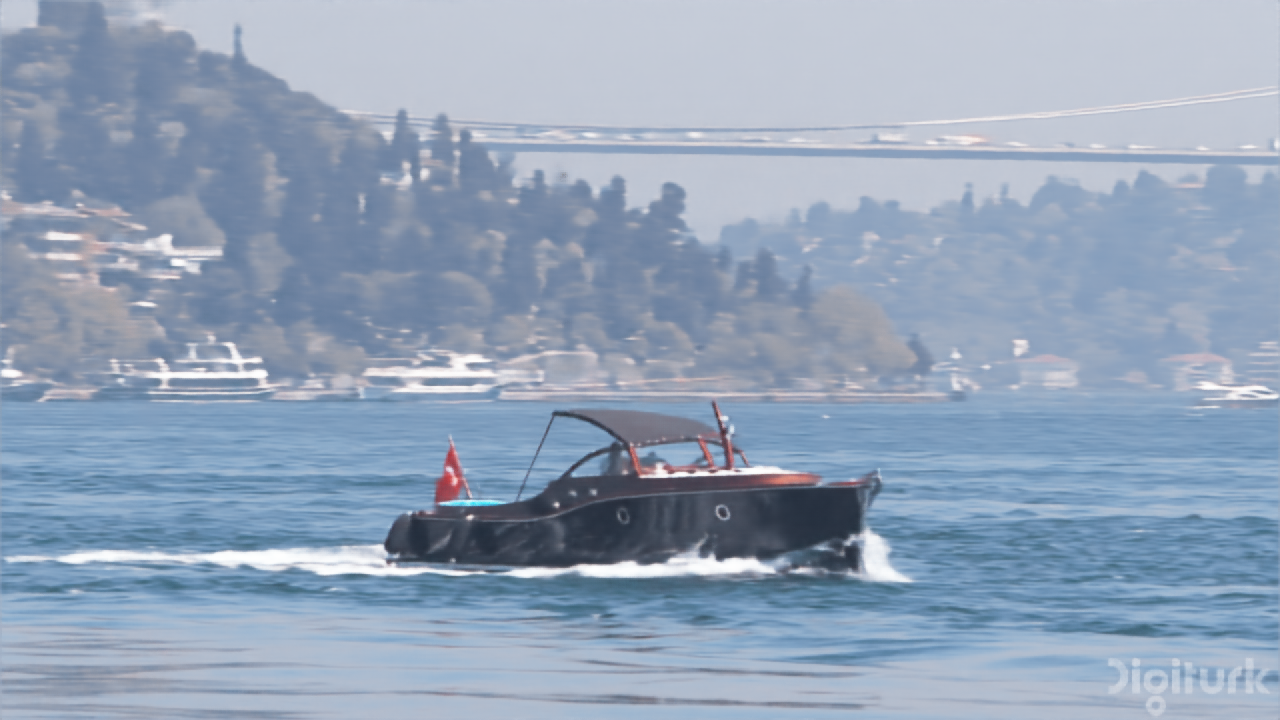} \\ % &
    \multicolumn{3}{l}{\small PFNR, \textcolor{red}{$f$-NeRV3} (34.05, PSNR)}\\
    
    \end{tabular}
    }
    \caption{PFNR's Video Generation (from t=0 to t=2) with $c = 30.0 \%$ on the UVG17 dataset.}
    \label{fig:video_app_reinit_mtl}
\end{figure}

%% file: materials/plot_video_reinit_bits.tex
\begin{figure}[ht]
    \centering 
    %\vspace{-0.1in}
    \setlength{\tabcolsep}{0pt}{%
    \begin{tabular}{ccc}
    
    % 2.city 
    \textit{2.city} & \textit{3.beauty} & \textit{7.bee} \\ 
    %\includegraphics[width=0.3\columnwidth]{images/video_reinit_bits/1/FP32/gt_0.png} & 
    %\includegraphics[width=0.3\columnwidth]{images/video_reinit_bits/2/FP32/gt_0.png} &
    %\includegraphics[width=0.3\columnwidth]{images/video_reinit_bits/6/FP32/gt_0.png} \\
    %\multicolumn{1}{l}{\small \textcolor{blue}{GT}} & & \\
    
    %\includegraphics[width=0.3\columnwidth]{images/video_bits/MTL/city_0.png} & 
    %\includegraphics[width=0.3\columnwidth]{images/video_bits/MTL/beauty_0.png} &
    %\includegraphics[width=0.3\columnwidth]{images/video_bits/MTL/bee_0.png} \\ 
    %\multicolumn{1}{l}{\small \textcolor{red}{PRED} of MLT (35.35, PSNR)}, & (31.45 PSNR), & (37.13, PSNR) in FP32.  \\

   \includegraphics[width=0.3\columnwidth]{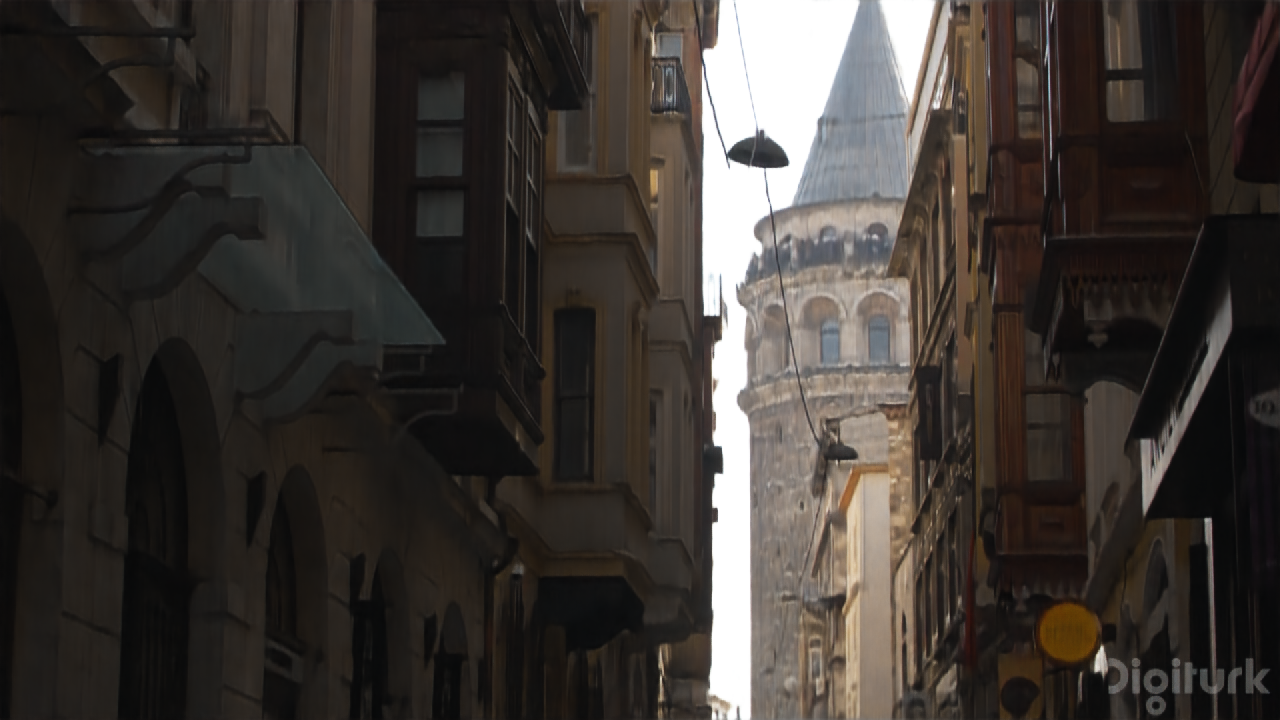} & 
    \includegraphics[width=0.3\columnwidth]{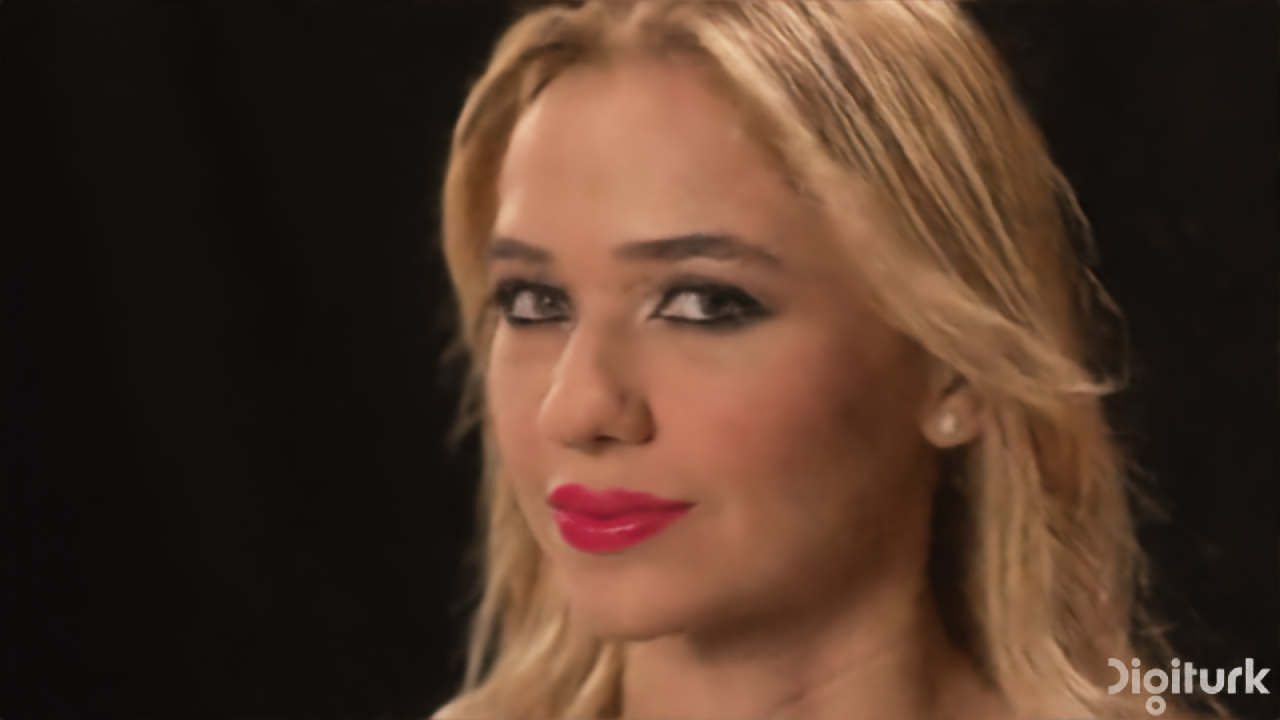} &
    \includegraphics[width=0.3\columnwidth]{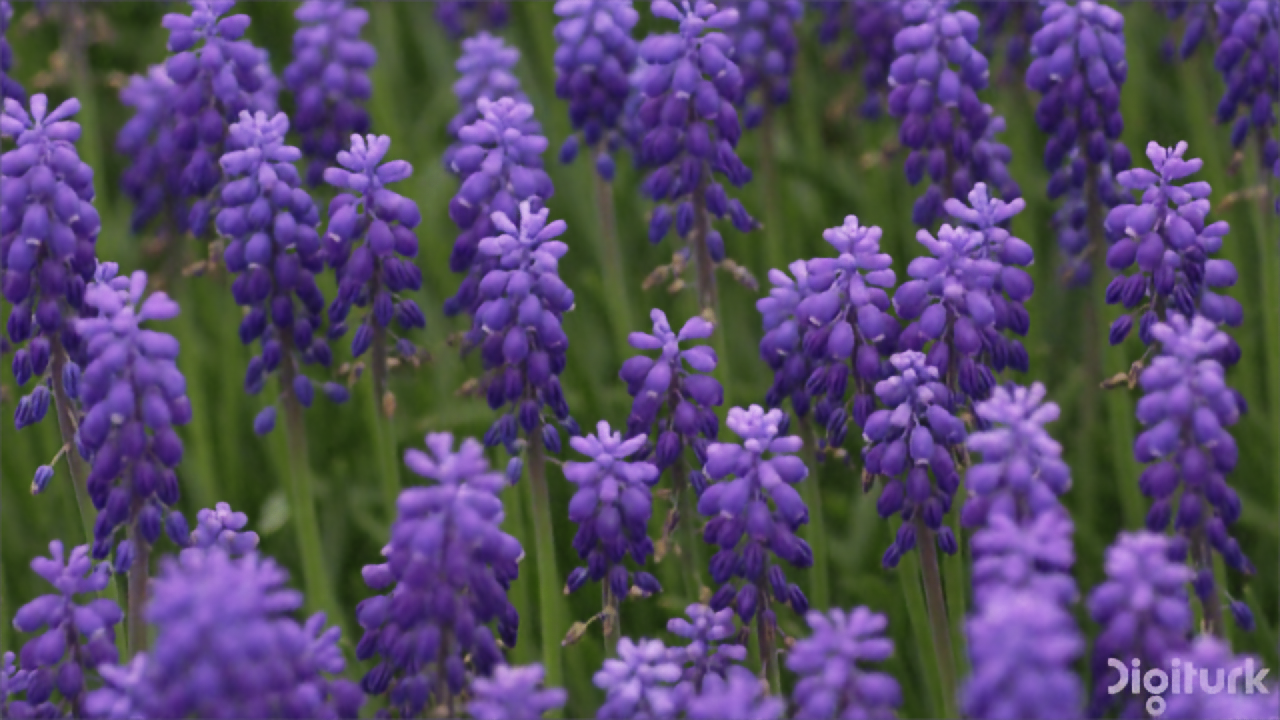} \\ 
    \multicolumn{1}{l}{\small PFNR (35.84, PSNR),} & {\small(32.97, PSNR),} & {\small(36.01, PSNR) in FP32.}  \\
    
   \includegraphics[width=0.3\columnwidth]{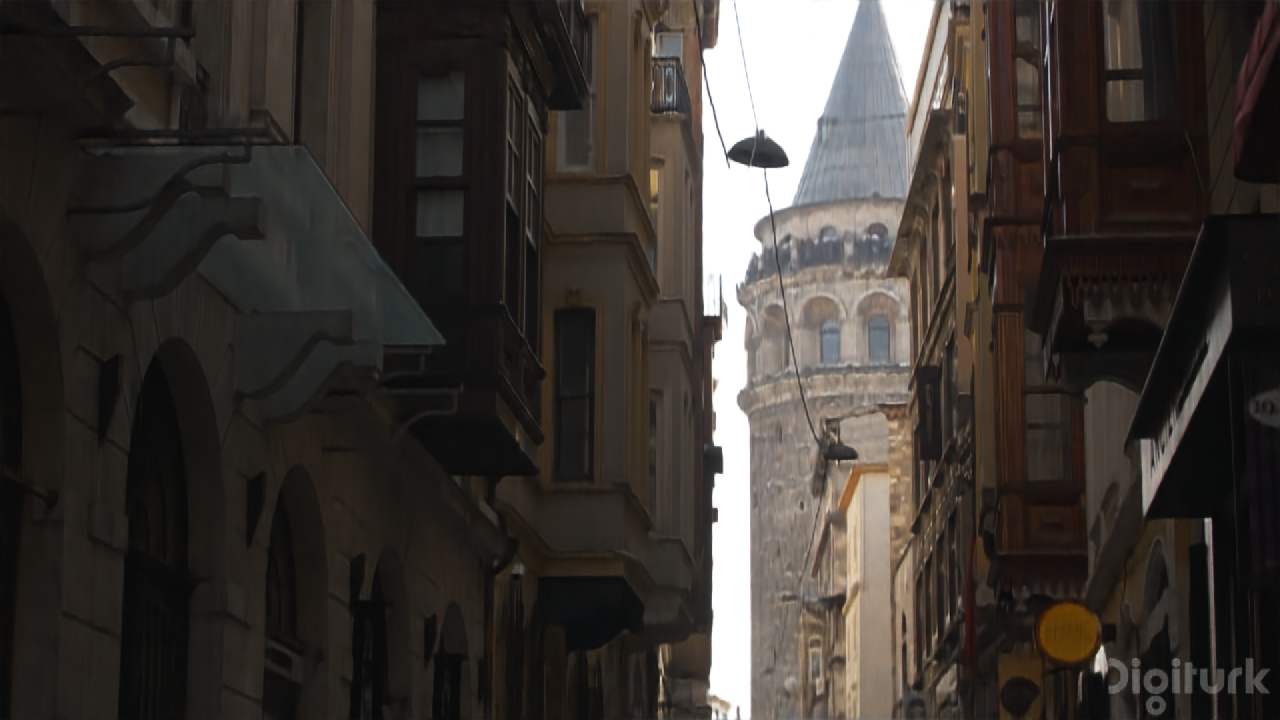} & 
    \includegraphics[width=0.3\columnwidth]{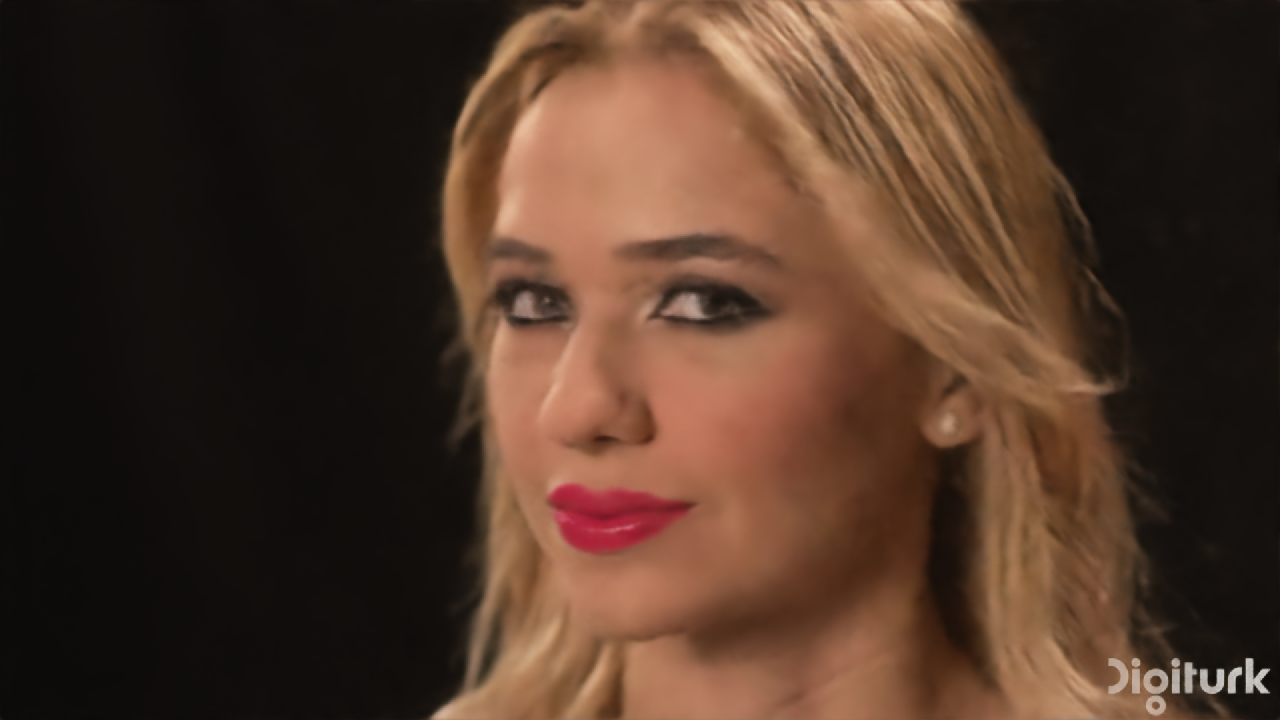} &
    \includegraphics[width=0.3\columnwidth]{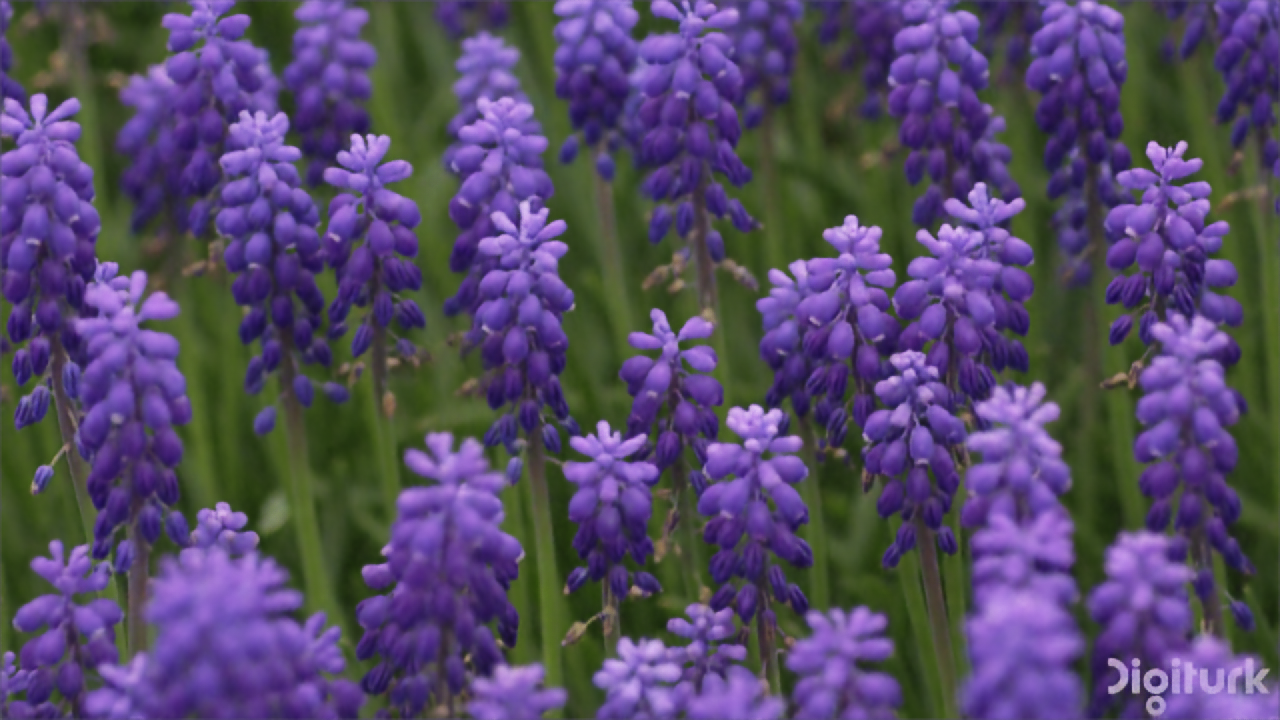} \\ 
    \multicolumn{1}{l}{\small PFNR (35.84, PSNR),} & {\small(32.97, PSNR),} & {\small(36.01, PSNR) in FP16.} \\

    \includegraphics[width=0.3\columnwidth]{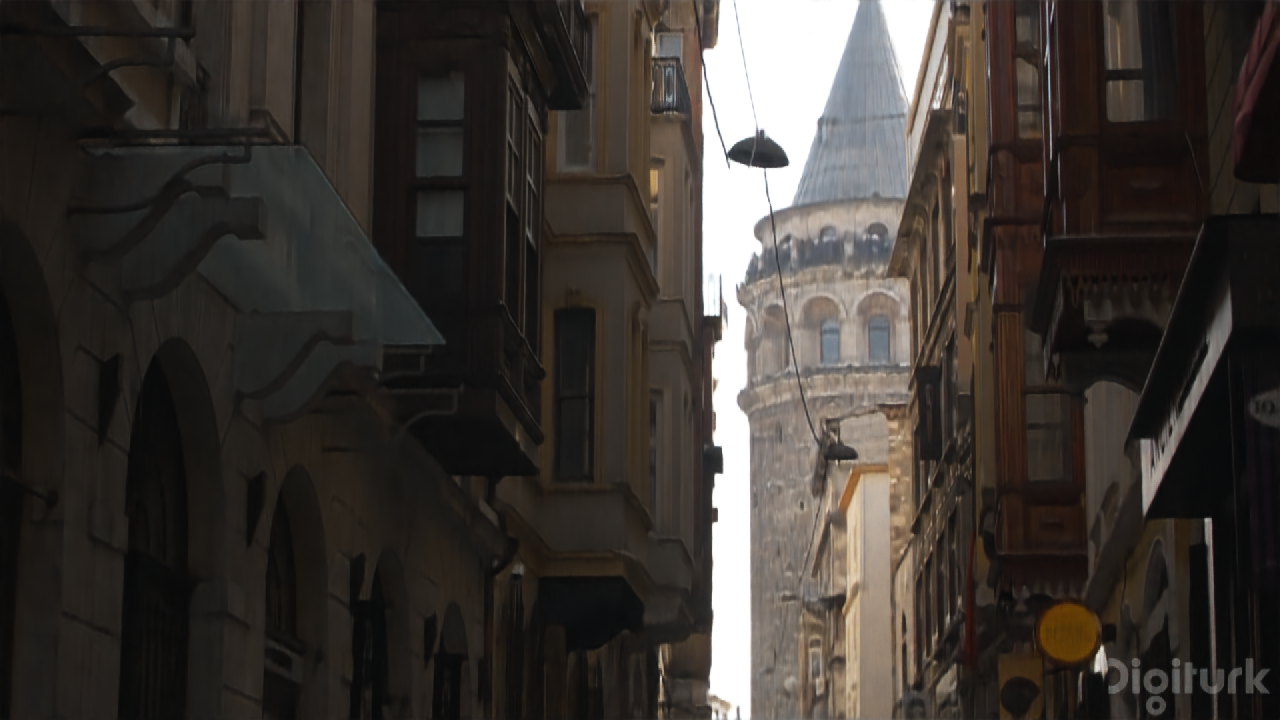} & 
    \includegraphics[width=0.3\columnwidth]{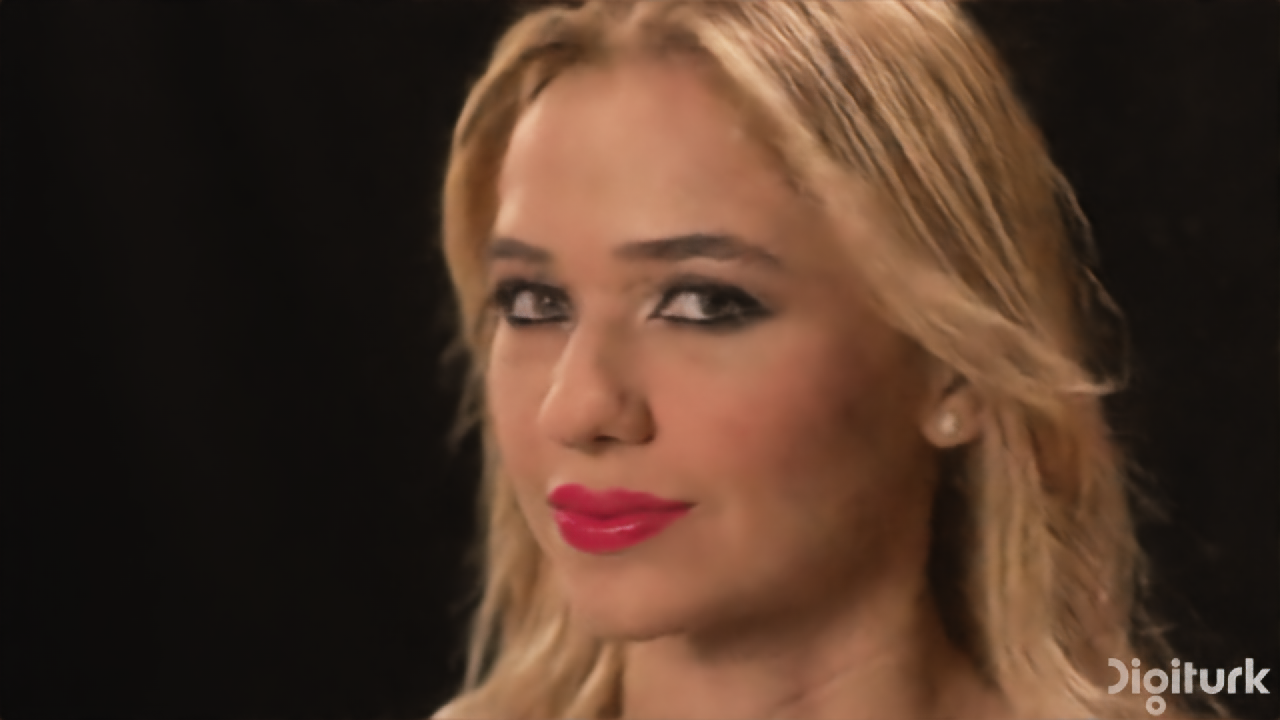} &
    \includegraphics[width=0.3\columnwidth]{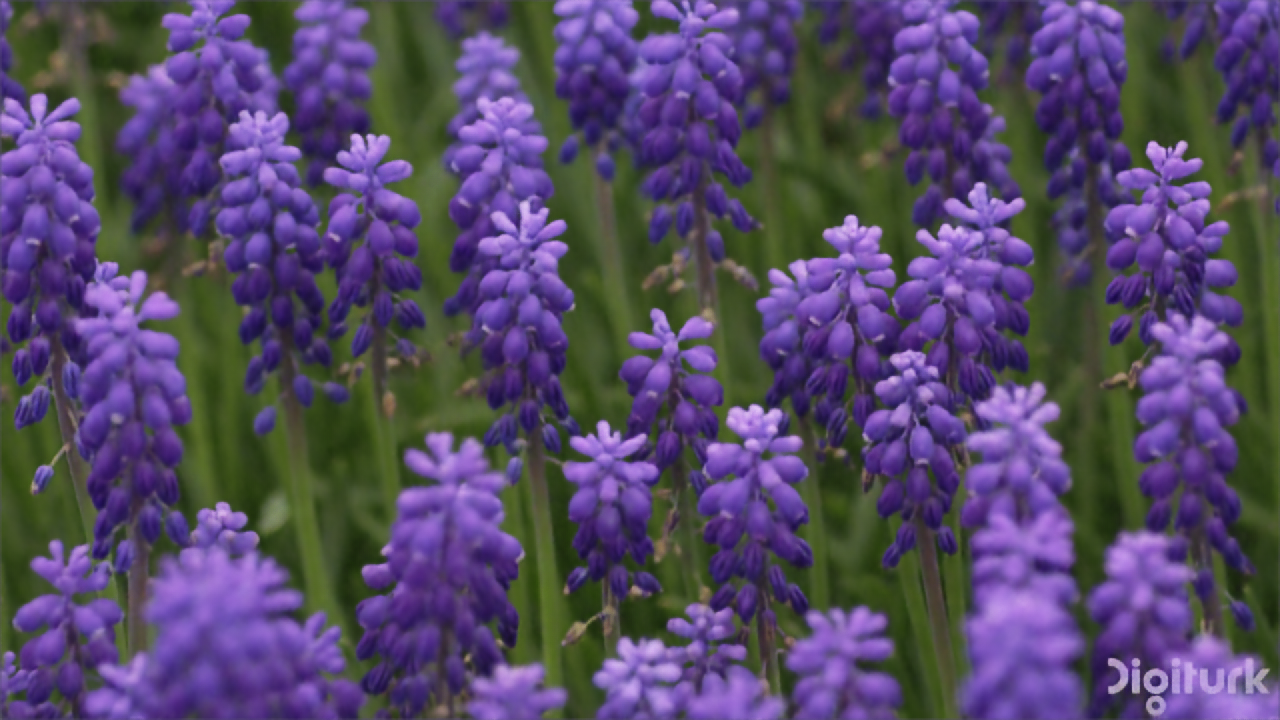} \\ 
    \multicolumn{1}{l}{\small PFNR (35.81, PSNR),} & {\small(32.95, PSNR),} & {\small(35.82, PSNR) in FP8.} \\
    
    \includegraphics[width=0.3\columnwidth]{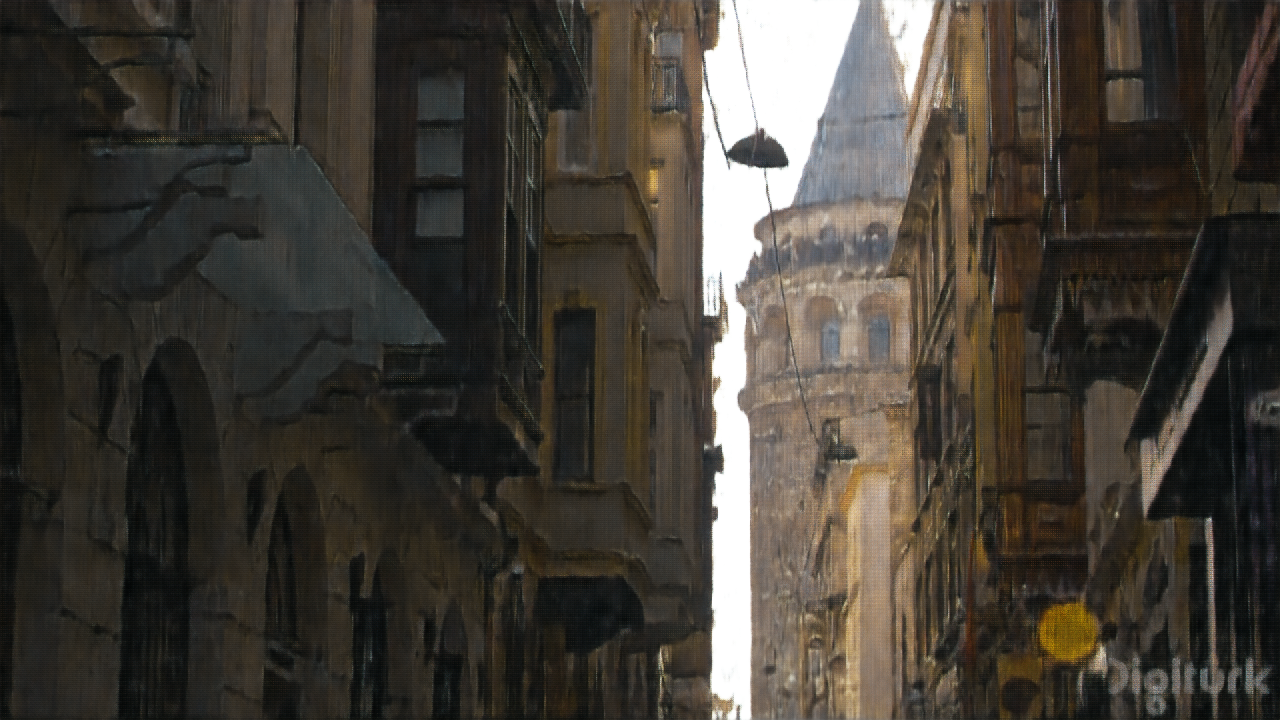} & 
    \includegraphics[width=0.3\columnwidth]{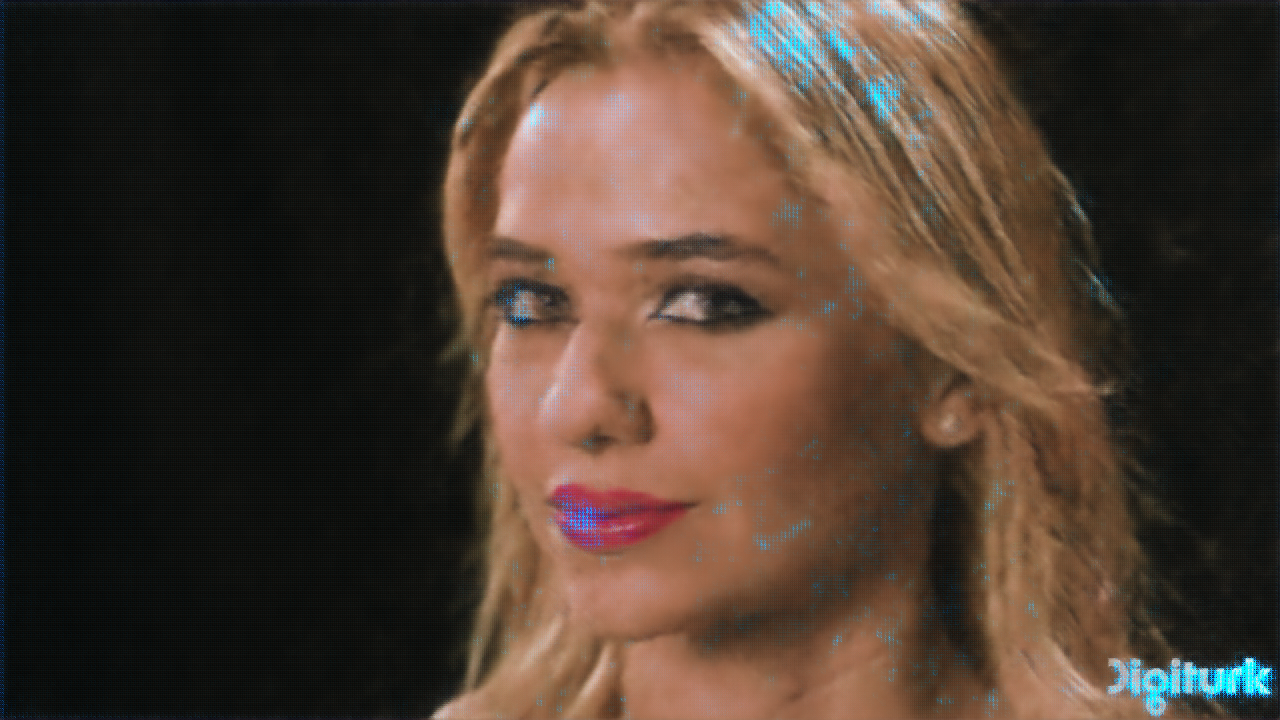} &
    \includegraphics[width=0.3\columnwidth]{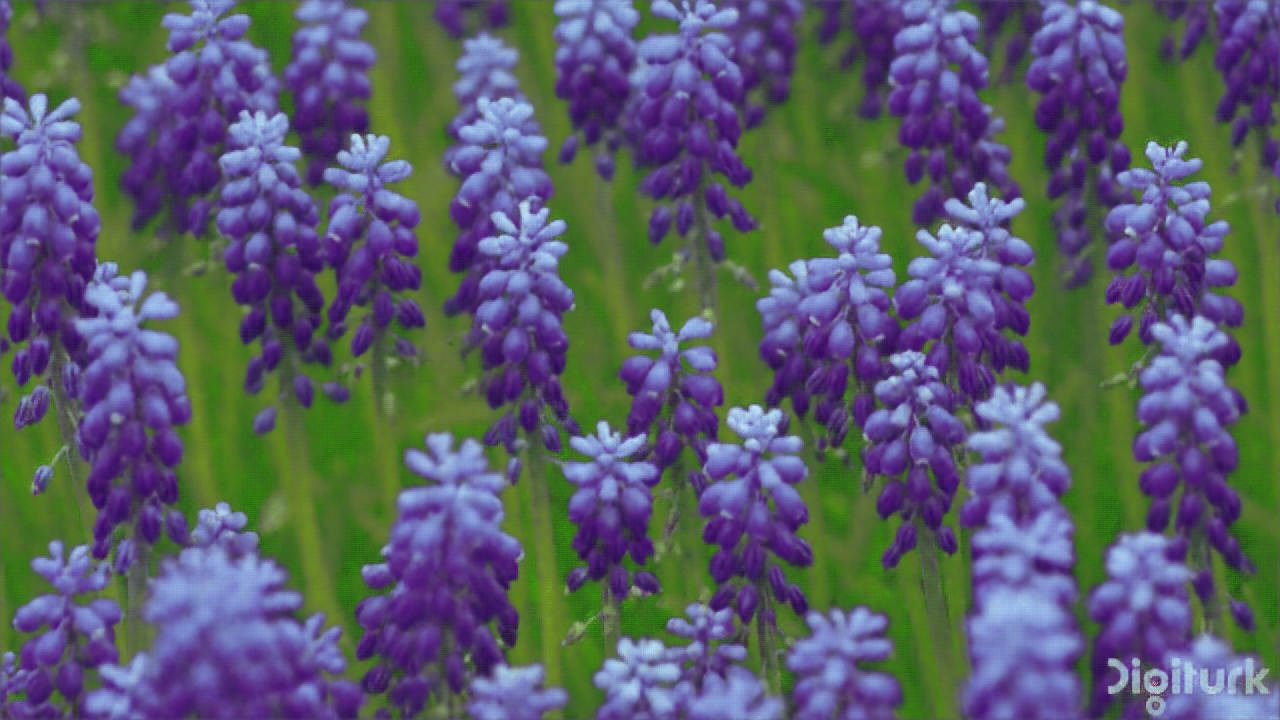} \\ 
    \multicolumn{1}{l}{\small PFNR (27.43, PSNR),} & {\small(24.65, PSNR),} & {\small(16.24, PSNR) in FP4.} \\
    \end{tabular}
    }
    \caption{PFNR's Qunatizationed and Compresssed Video Generation (t=0) with $c = 30.0 \%$, $f$-NeRV2 on the UVG17 dataset. Note that PFNR's prediction in FP32, 16, 8, and 4.}
    \label{fig:video_reinit_bits}
\end{figure}

%% file: materials/app_table_uvg17_psnr.tex
\begin{table}[ht]
\small\centering
\caption{\small PSNR results with Fourier Subnueral Operator (FSO) layer (\textcolor{red}{$f$-NeRV$\ast$}) (detailed in \Cref{table:architecture_detail}) on UVG17 Video Sessions with average PSNR, Backward Transfer (BWT) of PSNR, Model Capacity (CAP). Note that $\ast$ denotes our reproduced results.}
\resizebox{\textwidth}{!}{
\renewcommand{\arraystretch}{1.2}
\begin{tabular}{lcccccccccccccccccccc}
\toprule 

\multicolumn{1}{c}{\multirow{2}{*}{\textbf{Method}}} & \multicolumn{17}{c}{\textbf{Video Sessions}} & \multirow{2}{*}{\thead{\textbf{Avg. PSNR} \\ \textbf{BWT}}}  & \multirow{2}{*}{\thead{\textbf{CAP}}} \\

\cline{2-18}

& \textbf{1} & \textbf{2} & \textbf{3} & \textbf{4} & \textbf{5} & \textbf{6} & \textbf{7} & \textbf{8} & \textbf{9} & \textbf{10} & \textbf{11} & \textbf{12} & \textbf{13} & \textbf{14} & \textbf{15} & \textbf{16} & \textbf{17} \\ \midrule 
%STL, NeRV~\cite{chen2023hnerv} & 39.63 & - & 36.06 & - & 37.35 & - & 41.23 & - & 38.14 & - & 31.86 & - & 37.22 & - & 32.45 & - & - & - / - \\ % & 1700.00 \% \\

STL, NeRV~\cite{chen2021nerv}$^{\ast}$ & 39.66 & 44.89 & 36.28 & 41.13 & 38.14 & 31.53 & 42.03 & 34.74 & 36.58 & 36.85 & 29.22 & 31.81 & 37.27 & 34.18 & 31.45 & 38.41 & 43.86 & 36.94 / - & 1700.00 \% \\ 
\midrule 

EWC~\cite{Kirkpatrick2017}$^{\ast}$ & 11.15 & 9.21  & 12.71 & 11.40 & 15.58  & 9.25 & 7.06 & 12.96 & 6.34 & 10.31 & 9.55 & 13.39 & 5.76 & 8.67 & 10.93 & 10.92 & 28.29 & 11.38 / -16.13  & 100.00 \% \\
iCaRL~\cite{rebuffi2017icarl}$^{\ast}$ & 24.31 & 28.25  & 22.19 & 22.74 & 22.84  & 16.55 & 29.37 & 17.92 & 16.65 & 27.43 & 13.64 & 16.42 & 24.02 & 21.60 & 19.40 & 18.60 & 26.46 & 21.67 / ~-6.23  & 100.00 \% \\ 
ESMER~\cite{sarfraz2023error}$^{\ast}$ & 30.77 & 26.33 & 22.79 & 21.35 & 23.76 & 13.64 & 28.25 & 15.22 & 16.71 & 23.78 &	13.35 & 15.23 & 18.21 & 19.22 & 24.59 & 20.61 & 22.42 &  20.95 / -15.23  & 100.00 \% \\
\midrule

WSN$^{\ast}$, c = 10.0 \% & 27.68 & 31.31 & 30.29 & 31.63 & 28.66 & 22.57 & 31.62 & 22.04 & 21.05 & 32.71 & 17.85 & 20.09 & 27.07 & 23.84 & 22.98 & 20.50 & 28.56 & 25.91 / ~0.0 & 53.25 \% \\ 
WSN$^{\ast}$, c = 30.0 \% & 31.50 & 34.37 & 31.00 & {{32.38}} & {{29.26}} & {{23.08}} & {{31.96}} & {{22.64}} & {{22.07}} & {{33.48}} & {{18.34}} & {{20.45}} & {{27.21}} & {{24.33}} & {{23.09}} & {{21.23}} & {{29.13}} & {{26.80}} / ~0.0 & 91.10 \% \\ 
WSN$^{\ast}$, c = 50.0 \% & {{34.02}} & {{34.93}} & {{31.04}} & 31.74 & 28.95 & 23.07 & 31.26 & 22.32 & 21.93 & 33.35 & 18.22 & 20.34 & 26.88 & 24.22 & 22.72 & 21.30 & 28.86 & 26.77 / ~0.0 & 97.23 \% \\ 
WSN$^{\ast}$, c = 70.0 \% & 35.64 & 34.36 & 30.26 & 30.27 & 27.99  & 22.55 & 29.88 & 21.46 & 20.79 & 32.37 & 17.63 & 20.00 & 26.68 & 23.79 & 22.34 & 20.69 & 28.68 &  26.20 / ~0.0 & 99.01 \% \\

% reinit
%WSN, c = 10.0 \% & 28.02 & 32.68 & 31.38 & 33.13 & 29.54 & 23.75 & 33.91 & 24.49 & 23.63 & 34.91 & 19.42 & 21.71 & 29.09  & 26.14 & 24.47 & 23.57 & 31.34 &  27.72 / ~~0.0 \\ % & 56.00 \% \\ 
%WSN, c = 30.0 \% & 31.47 & 35.42 & 32.51 & 34.73 & {{30.70}} & {{24.53}} & {35.63} & {{25.49}} & {{24.50}} & {35.59} & {{20.24}} & {{22.49}} & {30.22} & {{27.03}} & {{25.14}} & {24.86} & {{32.16}} &  {{28.98}} / ~~0.0 \\ % & 94.00 \% \\ 
%WSN, c = 50.0 \% & {{34.05}} & {{36.61}} & {{32.73}} & {{34.99}} & 30.43 & 24.37 & 33.98 & 24.53 & 24.31 & 35.51 & 19.99 & 22.31 & 29.65 & 26.77 & 24.94 & 24.76 & 31.84 & 28.93 / ~~0.0 \\ % & 99.00 \% \\ 
%WSN, c = 70.0 \% & 35.57 & 35.93 & 31.61 & 31.39 & 28.32 & 23.19 & 30.40 & 22.69 & 22.70 & 34.69 & 18.82 & 21.09 & 27.75 & 25.43 & 23.47 &  23.24 & 30.30 &  27.45 / ~~0.0 \\ % & 99.00 \% \\  

\midrule

% real
PFNR, c = 10.0 \%, \textcolor{red}{$f$-NeRV2} &  28.31 & 33.57 & 31.92 & 33.67 & 29.98 & 23.99 &	34.39 & 24.8 & 23.94 & 35.08 & 19.70 & 22.03 & 29.56 & 26.57 & 24.79 & 24.10 &	31.35 & 28.10 / ~0.0  & 59.58 \% \\
PFNR, c = 30.0 \%, \textcolor{red}{$f$-NeRV2} & 32.01 &	35.84 &	32.97 &	35.17 &	31.24 &	24.82 &	36.01 &	25.85 &	24.83 &	35.76 &	20.50 &	22.79 &	30.40 &	27.37 &	25.52 &	25.40 &	32.70 &	29.36 / ~0.0  & 100.01 \% \\
PFNR, c = 50.0 \%, \textcolor{red}{$f$-NeRV2} & 34.49 &	37.13 &	33.21 &	35.50 &	30.87 &	24.72 &	34.36 &	24.79 &	24.73 &	35.65 &	20.33 &	22.65 &	29.78 &	27.05 &	25.18 &	25.18 &	32.39 &	29.29 / ~0.0  & 105.33 \% \\
PFNR, c = 70.0 \%, \textcolor{red}{$f$-NeRV2} & 36.02 &	36.50 &	32.09 &	32.15 &	28.67 &	23.35 &	30.63 &	22.86 &	23.18 &	34.90 &	19.08 &	21.30 &	27.87 &	25.86 &	24.12 &	23.47 &	30.34 &	27.79 / ~0.0  & 112.33 \% \\

\midrule

MTL (upper-bound) & 32.39 & 34.35  & 31.45 & 34.03 & 30.70  & 24.53 & 37.13 & 27.83 & 23.80 & 34.69 & 20.77 & 22.37 & 32.71 & 28.00 & 25.89 & 26.40 & 33.16 & 29.42 / - ~~~  & 100.00 \% \\ 
\bottomrule
\end{tabular}
}
\label{table:uvg17_fso_psnr_app}
\end{table}

%% file: materials/app_table_uvg17_msssim.tex
\begin{table}[ht]
\small\centering
\caption{\small MS-SSIM results with Fourier Subnueral Operator (FSO) layer (\textcolor{red}{$f$-NeRV$\ast$}) (detailed in \Cref{table:architecture_detail}) on UVG17 Video Sessions with average MS-SSIM and Backward Transfer (BWT) of MS-SSIM. Note that $\ast$ denotes our reproduced results.}
\resizebox{\textwidth}{!}{
\renewcommand{\arraystretch}{1.2}
\begin{tabular}{lccccccccccccccccccc}
\toprule 

\multicolumn{1}{c}{\multirow{2}{*}{\textbf{Method}}} & \multicolumn{17}{c}{\textbf{Video Sessions}} & \multirow{2}{*}{\thead{\textbf{Avg. MS-SSIM} \\ \textbf{BWT}}} \\

\cline{2-18}

& \textbf{1} & \textbf{2} & \textbf{3} & \textbf{4} & \textbf{5} & \textbf{6} & \textbf{7} & \textbf{8} & \textbf{9} & \textbf{10} & \textbf{11} & \textbf{12} & \textbf{13} & \textbf{14} & \textbf{15} & \textbf{16} & \textbf{17} \\ \midrule 
STL, NeRV$^{\ast}$~\cite{chen2021nerv}   & 0.99 & 0.99 & 0.95 & 0.98 & 0.98 & 0.96 & 0.99 & 0.98 & 0.97 & 0.95 & 0.96 & 0.96 & 0.98 & 0.98 & 0.96 & 0.99 & 0.99 & 0.97 / - ~~~~ \\ \midrule 

%& 99.61 & 98.21 & 96.10 & 98.37 & 94.57 & 96.27 & 98.03 & 99.13 & 99.37 & 97.74/- \\ \midrule 

EWC~\cite{Kirkpatrick2017}$^{\ast}$ & 0.26 & 0.24  & 0.44 & 0.24 & 0.40  & 0.29 & 0.15 & 0.17 & 0.26 & 0.26 & 0.17 & 0.34 & 0.04 & 0.30 & 0.33 & 0.31 & 0.91 & 0.30 / -0.55  \\

iCaRL~\cite{rebuffi2017icarl}$^{\ast}$ & 0.74 & 0.88  & 0.67 & 0.67 & 0.64  & 0.48 & 0.91 & 0.53 & 0.37 & 0.82 & 0.35 & 0.53 & 0.75 & 0.70 & 0.61 & 0.60 & 0.87 & 0.65 / -0.20  \\ 
ESMER~\cite{sarfraz2023error}$^{\ast}$ &  0.85 & 0.86 & 0.64 & 0.63 & 0.66 & 0.46 &	0.89 & 0.51 & 0.42 & 0.79 &	0.30 & 0.51 &	0.43 &	0.68 &  0.82 & 0.6 & 0.63 & 0.62 / -0.37  \\
\midrule

WSN$^{\ast}$, c = 10.0 \% & 0.90 & 0.94 & 0.88 & 0.92 & 0.87 & 0.75 & 0.96 & 0.74 & 0.69 & 0.91 & 0.57 & 0.72 & 0.86 & 0.80 & 0.74 & 0.69 & 0.92 & 0.82 / 0.0  \\ 
WSN$^{\ast}$, c = 30.0 \% & 0.96 & 0.97 & 0.89 & {{0.93}} & {{0.88}} & {{0.77}} & {{0.97}} & {{0.77}} & {{0.73}} & {{0.91}} & {{0.60}} & {{0.74}} & {{0.86}} & {{0.81}} & {{0.76}} & {{0.72}} & {{0.93}} & {{0.84}} / 0.0  \\ 
WSN$^{\ast}$, c = 50.0 \% & {{0.98}} & {{0.97}} & {{0.89}} & 0.92 & 0.88 & 0.77 & 0.96 & 0.75 & 0.73 & 0.91 & 0.60 & 0.74 & 0.85 & 0.80 & 0.75 & 0.73 & 0.92 & 0.83 / 0.0  \\ 
WSN$^{\ast}$, c = 70.0 \% & 0.98 & 0.97 & 0.88 & 0.91 & 0.85 & 0.75 & 0.95 & 0.70 & 0.68 & 0.91 & 0.55 & 0.72 & 0.85 & 0.80 & 0.73 & 0.70 & 0.92 & 0.82 / 0.0  \\

% reinit
% WSN, c = 10.0 \%  & 0.91 & 0.96 & 0.90 & 0.94 & 0.89 & 0.80 & 0.98 & 0.84 & 0.78 & 0.93 & 0.68 & 0.78 & 0.91 & 0.86 & 0.81 & 0.82 & 0.96 & 0.87 / 0.0 \\ 
% WSN, c = 30.0 \%  & 0.96 & 0.98 & 0.91 & 0.95 & 0.91 & 0.83 & 0.98 & 0.87 & 0.81 & 0.93 & 0.72 & 0.81 & 0.92 & 0.88 & 0.83 & 0.86 & 0.97 & 0.89 / 0.0 \\ 
% WSN, c = 50.0 \% & 0.98 & 0.98 & 0.91 & 0.95 & 0.91 & 0.82 & 0.98 & 0.85 & 0.80 & 0.93 & 0.71 & 0.80 & 0.91 &  0.88 & 0.83 & 0.86 & 0.96 & 0.89 / 0.0  \\ 
% WSN, c = 70.0 \% & 0.98 & 0.98 & 0.90 & 0.92 & 0.87 & 0.78 & 0.95 & 0.77 & 0.76 & 0.92 & 0.64 & 0.77 & 0.88 & 0.84 & 0.79 & 0.81 & 0.95 &  0.85 / 0.0  \\ 
\midrule

PFNR, c = 10.0 \%, \textcolor{red}{$f$-NeRV2} & 0.92 &	0.97 &	0.90 &	0.94 &	0.90 &	0.81 &	0.98 &	0.86 &	0.79 &	0.93 &	0.69 &	0.79 &	0.91 &	0.87 &	0.82 &	0.84 &	0.96 &	0.88  / 0.0  \\
PFNR, c = 30.0 \%, \textcolor{red}{$f$-NeRV2} & 0.97 &	0.98 &	0.92 &	0.95 &	0.92 &	0.84 &	0.98 &	0.88 &	0.82 &	0.93 &	0.73 &	0.82 &	0.92 &	0.89 &	0.84 &	0.87 &	0.97 &	0.90 / 0.0 \\
PFNR, c = 50.0 \%, \textcolor{red}{$f$-NeRV2} & 0.98 &	0.99 &	0.92 &	0.95 &	0.92 &	0.83 &	0.98 &	0.86 &	0.81 &	0.93 &	0.72 &	0.82 &	 0.91 &	0.88 &	0.84 &	0.87 &	0.97 &	0.89 / 0.0 \\
PFNR, c = 70.0 \%, \textcolor{red}{$f$-NeRV2} & 0.99 & 0.98 & 0.91 & 0.93 & 0.87 & 0.78 & 0.96 & 0.77 & 0.77 & 0.93 & 0.66	& 0.77 & 0.89 &	0.85 & 0.80 &	0.82 & 0.95 & 0.86 / 0.0  \\
\midrule

PFNR, c = 10.0 \%, \textcolor{red}{$f$-NeRV3} & 0.96 &	0.99 &	0.92 &	0.96 &	0.94 &	0.86 &	0.99 &	0.94 &	0.82 &	0.93 &	0.76 &	0.83 &	0.93 &	0.92 &	0.87 &	0.90 &	0.98 &	0.91 / 0.0 \\
PFNR, c = 30.0 \%, \textcolor{red}{$f$-NeRV3} & 0.98 &	0.99 &	0.93 &	0.97 &	0.96 &	0.91 &	0.99 &	0.96 &	0.87 &	0.94 &	0.84 &	0.87 &	0.94 &	0.94 &	0.90 & 	0.94 &	0.98	&  0.94  / 0.0 \\
PFNR, c = 50.0 \%, \textcolor{red}{$f$-NeRV3} & 0.99 &	0.99 &	0.93 &	0.97 &	0.96 &	0.91 &	0.99 &	0.95 &	0.87 &	0.94 &	0.83 &	0.88 &	 0.94 &	0.94 &	0.9 &	0.95 &	0.98 &	0.94 / 0.0 \\
PFNR, c = 70.0 \%, \textcolor{red}{$f$-NeRV3} & 0.99 & 0.99 & 0.93 & 0.96 & 0.93 & 0.87 & 0.98 & 0.90 & 0.81 & 0.93	 & 0.76 & 0.83 & 0.93 & 0.92 & 0.87 &	0.91  & 0.97 & 0.91 / 0.0 \\
\midrule

MTL (upper-bound) & 0.97 & 0.97  & 0.90 & 0.94 & 0.91  & 0.82 & 0.99 & 0.92 & 0.80 & 0.92 & 0.75 & 0.81 & 0.94 & 0.90 & 0.85 & 0.89 & 0.97 & 0.90 / - ~~~  \\ 
\bottomrule
\end{tabular}
}
\label{table:uvg17_fso_msssim_app}
\end{table}

%% file: materials/plot_fmap_uvg17.tex
\begin{figure*}[ht]
    \centering 
    %\vspace{-0.1in}
    \setlength{\tabcolsep}{0pt}{%
    \begin{tabular}{ccccc}
    
    % 2.city 
    %\textit{2.city} & \textit{3.beauty} & \textit{7.bee} \\ 
     NeRV3 & NeRV4 & NeRV5 & NeRV6 & Head \\ 

    %\makecell{\small WSN \\ \tiny c=50\% } & 
    %\includegraphics[width=0.2\columnwidth]{images/fmaps/5/nerv3/gt_0.png} &  
    %\includegraphics[width=0.3\columnwidth]{images/fmaps/5/wsn/pred_0_l1.pdf} & % NeRV2
    \includegraphics[width=0.20\columnwidth, trim={0.1cm 0.4cm 0.1cm 0.1cm}, clip]{images/fmaps/5/wsn/pred_0_l2.pdf} & % NeRV3
    \includegraphics[width=0.20\columnwidth, trim={0.1cm 0.4cm 0.1cm 0.1cm}, clip]{images/fmaps/5/wsn/pred_0_l3.pdf} & % NeRV4
    \includegraphics[width=0.20\columnwidth, trim={0.1cm 0.4cm 0.1cm 0.1cm}, clip]{images/fmaps/5/wsn/pred_0_l4.pdf} & % NeRV5
    \includegraphics[width=0.20\columnwidth, trim={0.1cm 0.4cm 0.1cm 0.1cm}, clip]{images/fmaps/5/wsn/pred_0_l5.pdf} & % NeRV6
    \includegraphics[width=0.19\columnwidth]{images/fmaps/5/wsn/pred_0.png} \\

    \includegraphics[width=0.20\columnwidth, trim={0.1cm 0.4cm 0.1cm 0.1cm}, clip]{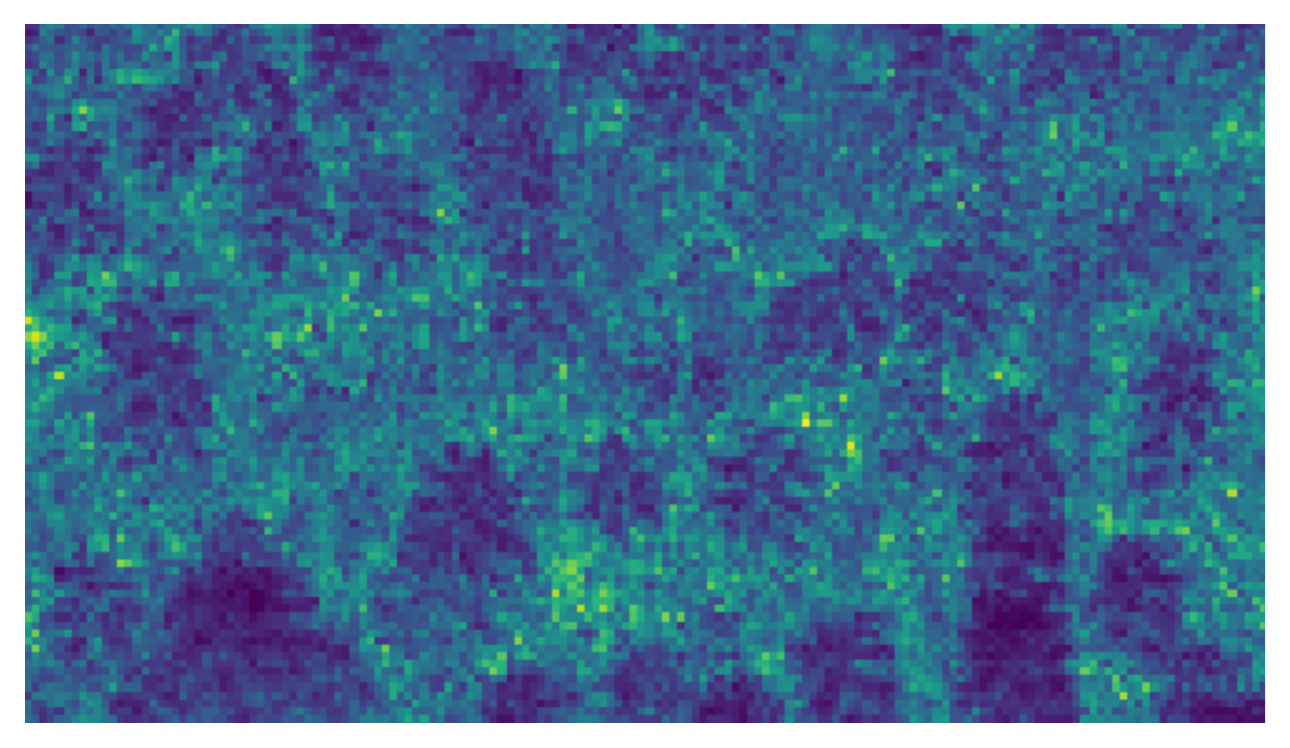} & % NeRV3
    \includegraphics[width=0.20\columnwidth, trim={0.1cm 0.4cm 0.1cm 0.1cm}, clip]{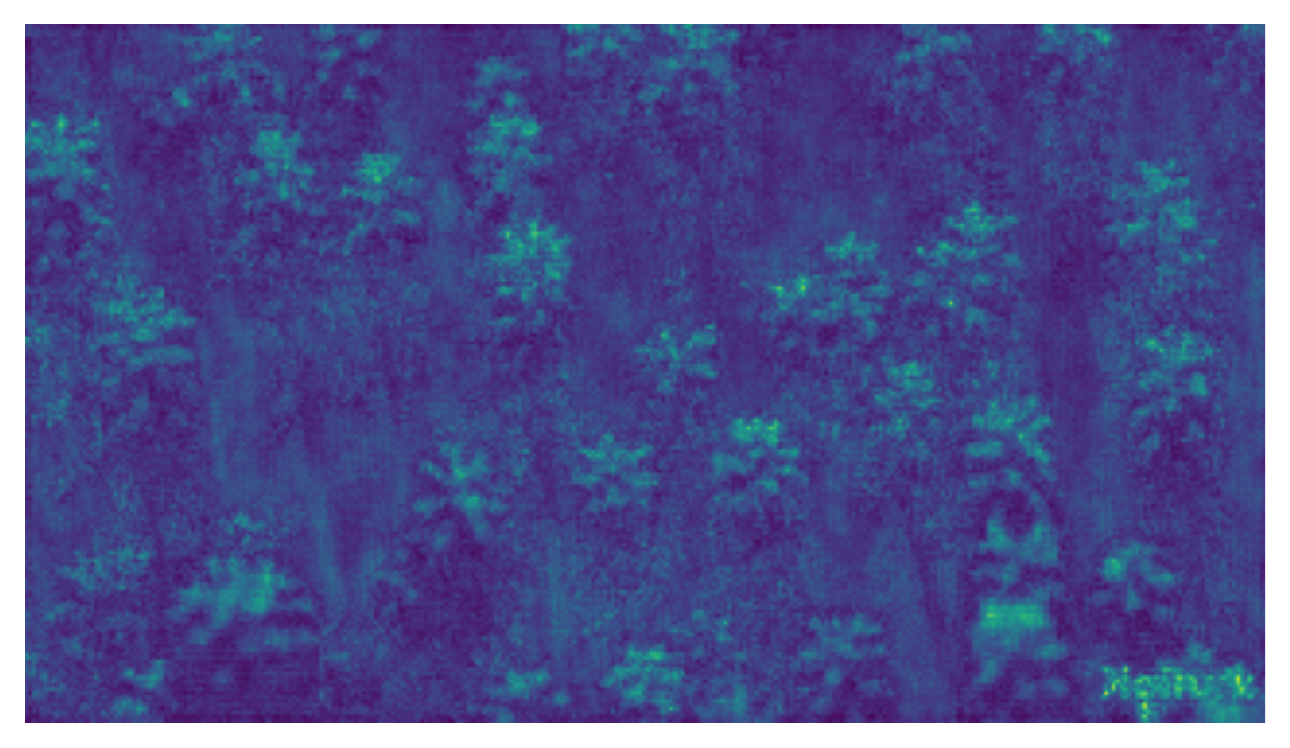} & % NeRV4
    \includegraphics[width=0.20\columnwidth, trim={0.1cm 0.4cm 0.1cm 0.1cm}, clip]{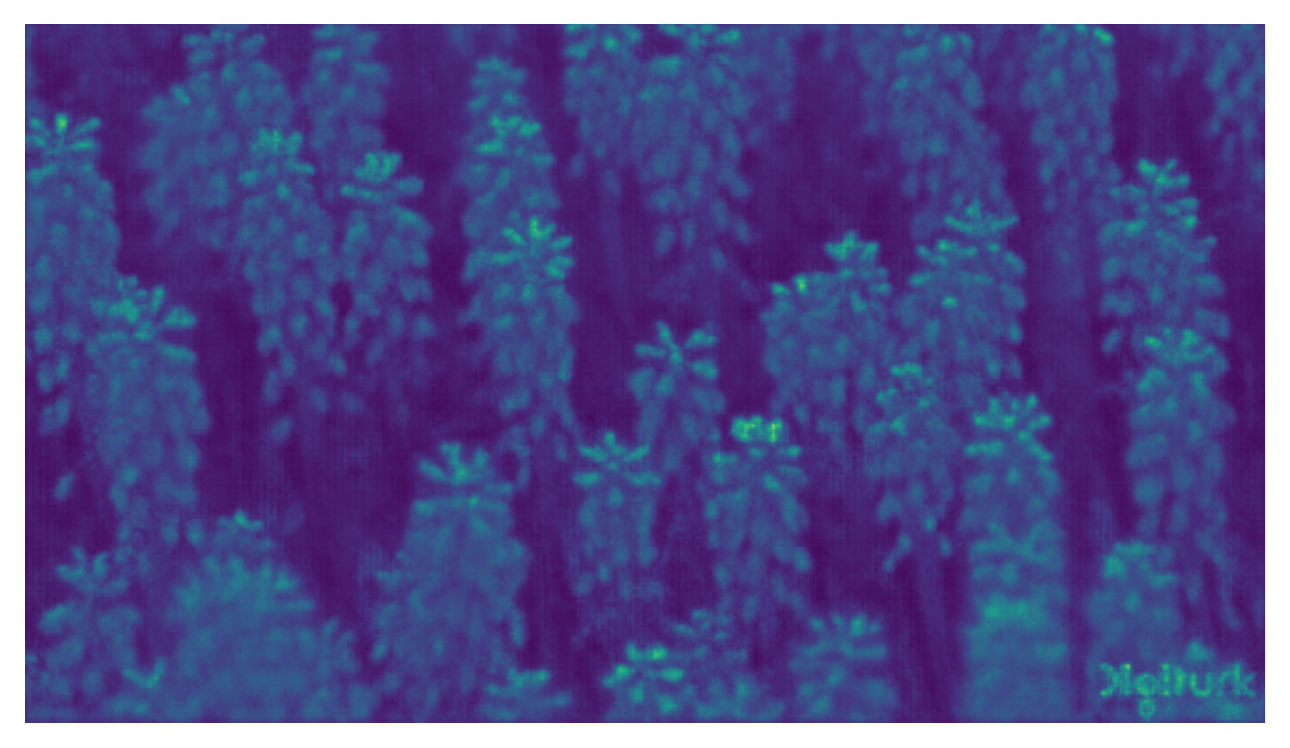} & % NeRV5
    \includegraphics[width=0.20\columnwidth, trim={0.1cm 0.4cm 0.1cm 0.1cm}, clip]{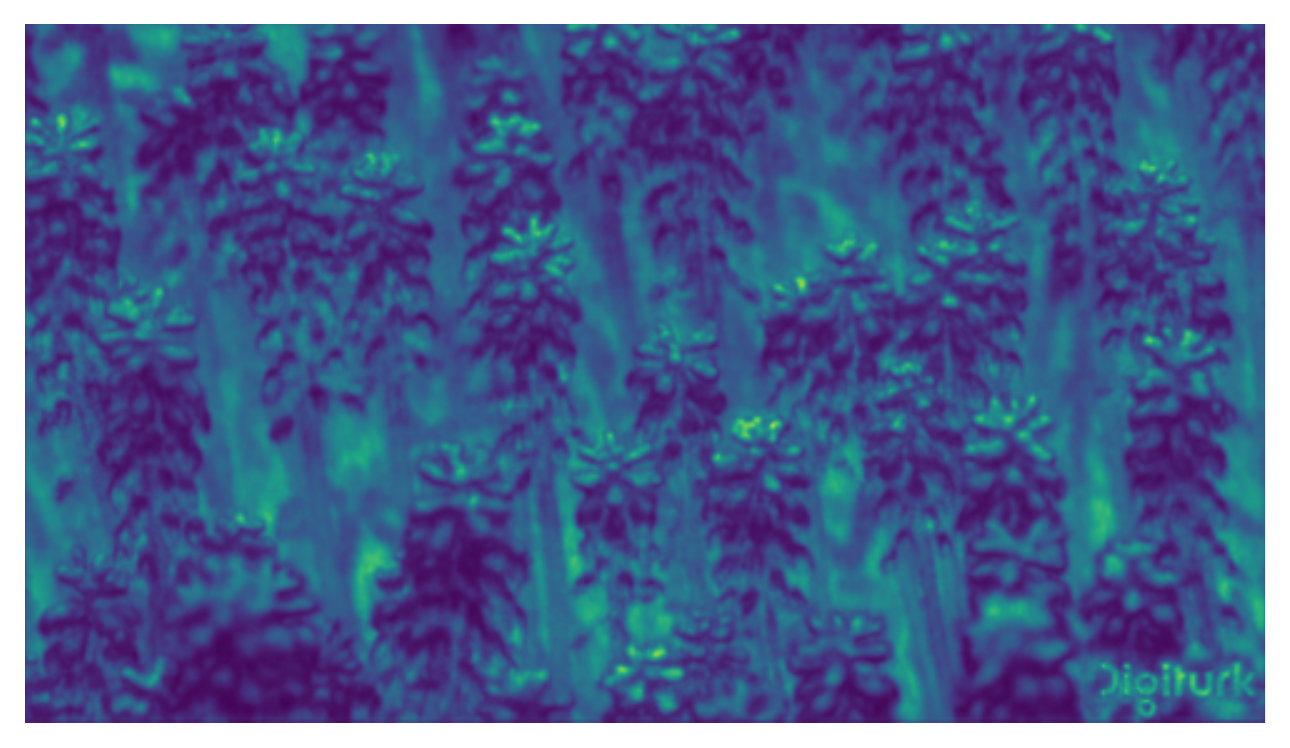} & % NeRV6
    \includegraphics[width=0.19\columnwidth]{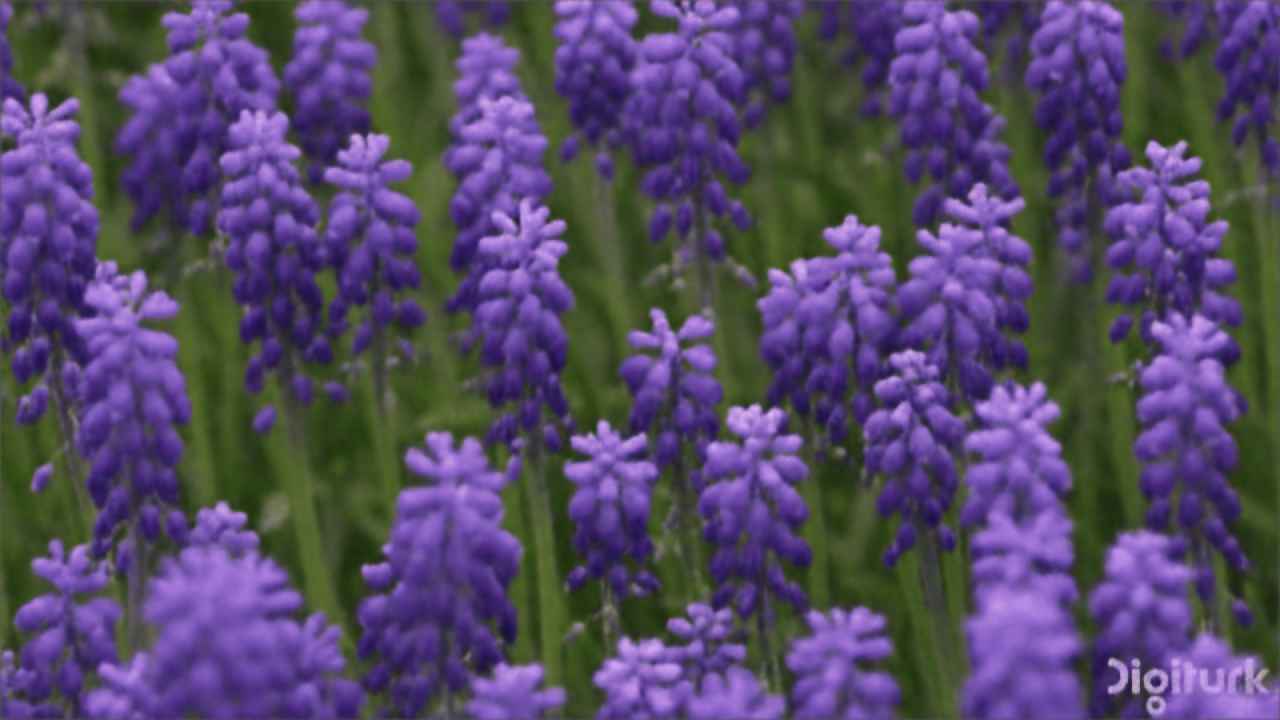} \\
    
    \multicolumn{5}{l}{\small WSN, c=50.0\%} \\
    
    %\makecell{\small PFNR \\ \tiny c=50\% \\ \tiny $f$-NeRV2} & 
    %\includegraphics[width=0.2\columnwidth]{images/fmaps/5/nerv3/gt_0.png} &  
    %\includegraphics[width=0.3\columnwidth]{images/fmaps/5/nerv3/pred_0_l1.pdf} & % NeRV2
    \includegraphics[width=0.20\columnwidth, trim={0.1cm 0.4cm 0.1cm 0.1cm}, clip]{images/fmaps/5/nerv2/pred_0_l2.pdf} & % NeRV3
    \includegraphics[width=0.20\columnwidth, trim={0.1cm 0.4cm 0.1cm 0.1cm}, clip]{images/fmaps/5/nerv2/pred_0_l3.pdf} & % NeRV4
    \includegraphics[width=0.20\columnwidth, trim={0.1cm 0.4cm 0.1cm 0.1cm}, clip]{images/fmaps/5/nerv2/pred_0_l4.pdf} & % NeRV5
    \includegraphics[width=0.20\columnwidth, trim={0.1cm 0.4cm 0.1cm 0.1cm}, clip]{images/fmaps/5/nerv2/pred_0_l5.pdf} & % NeRV6
    \includegraphics[width=0.19\columnwidth]{images/fmaps/5/nerv2/pred_0.png} \\

    \includegraphics[width=0.20\columnwidth, trim={0.1cm 0.4cm 0.1cm 0.1cm}, clip]{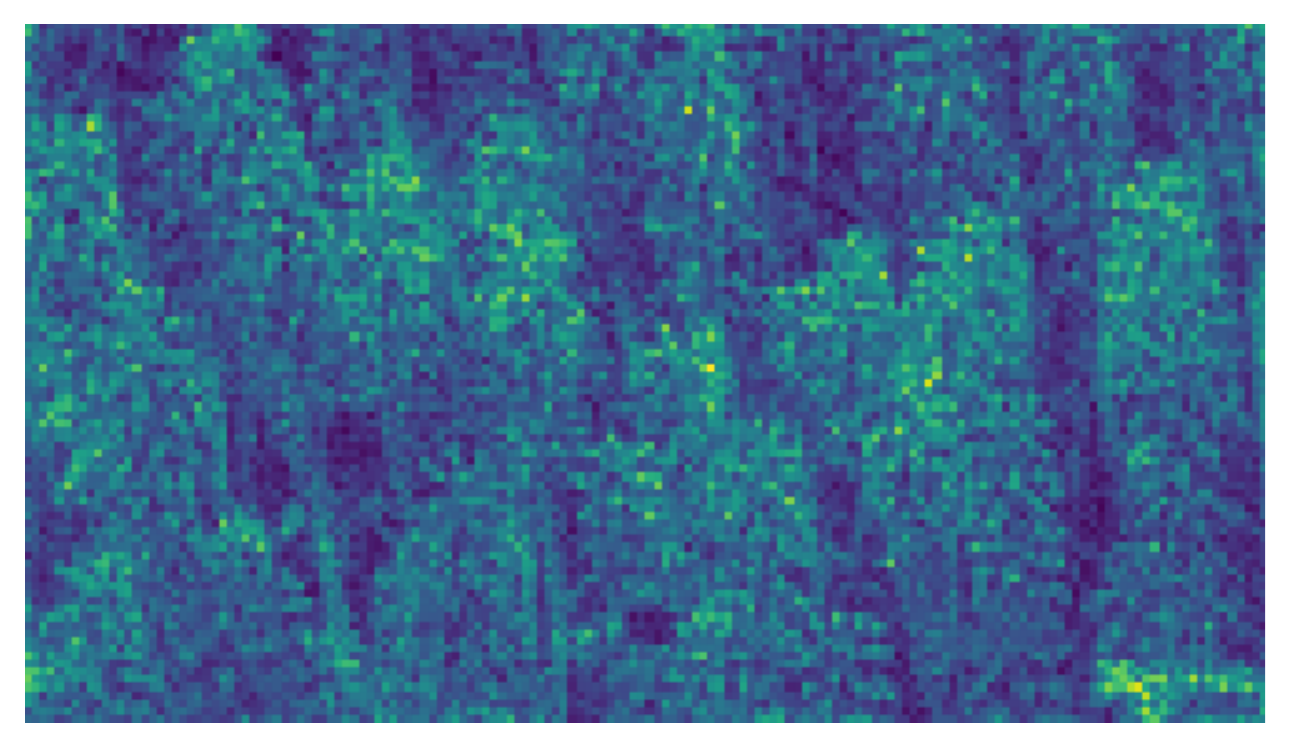} & % NeRV3
    \includegraphics[width=0.20\columnwidth, trim={0.1cm 0.4cm 0.1cm 0.1cm}, clip]{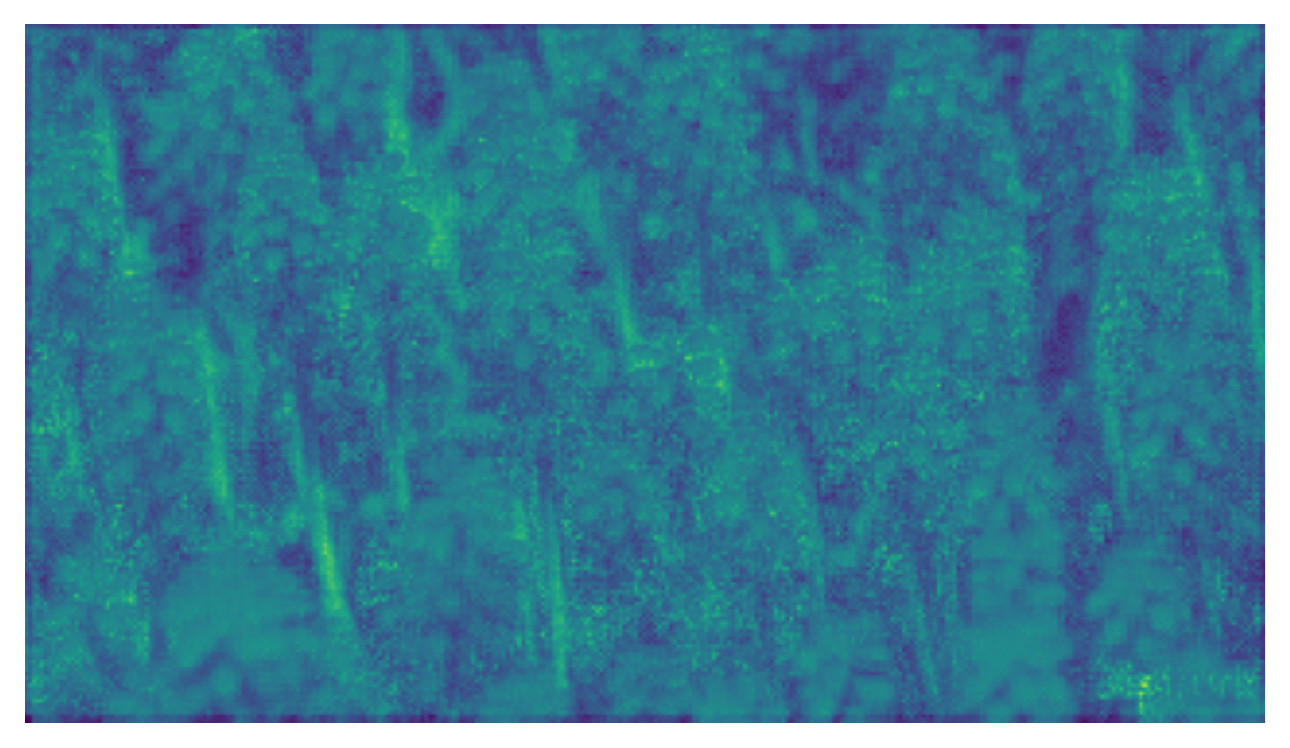} & % NeRV4
    \includegraphics[width=0.20\columnwidth, trim={0.1cm 0.4cm 0.1cm 0.1cm}, clip]{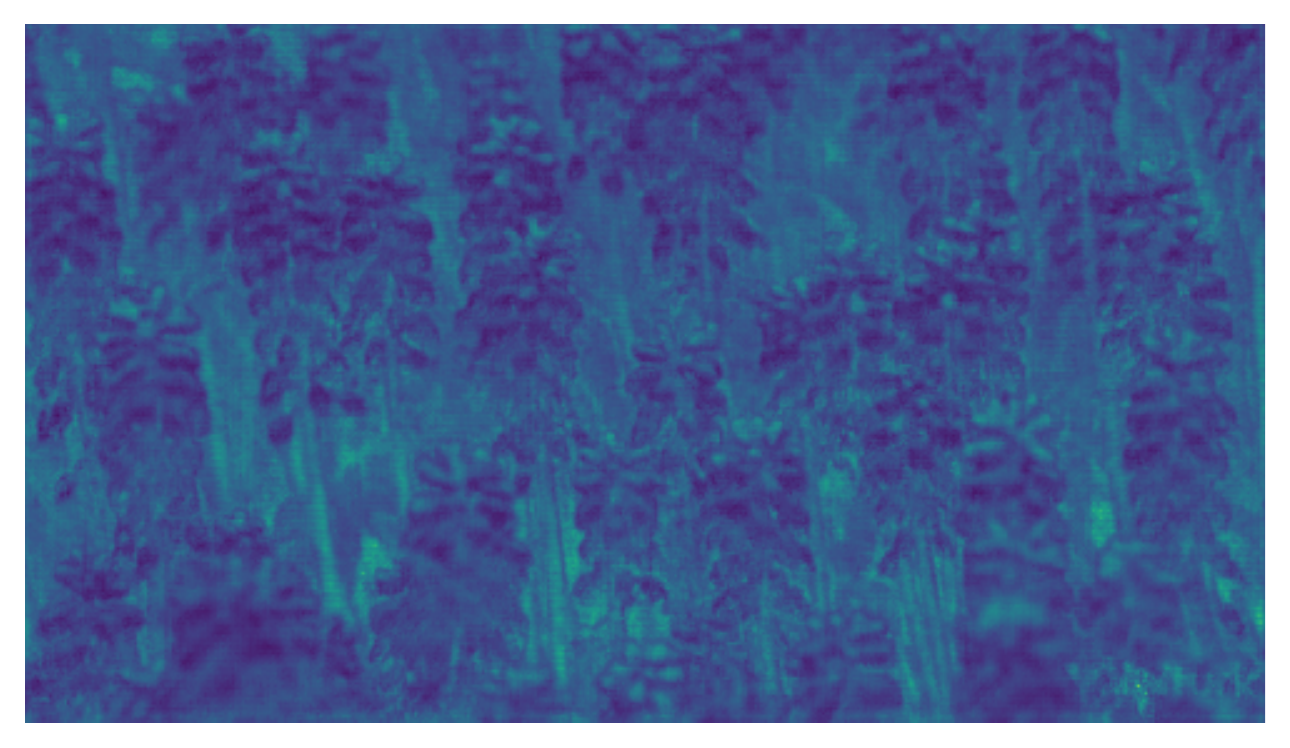} & % NeRV5
    \includegraphics[width=0.20\columnwidth, trim={0.1cm 0.4cm 0.1cm 0.1cm}, clip]{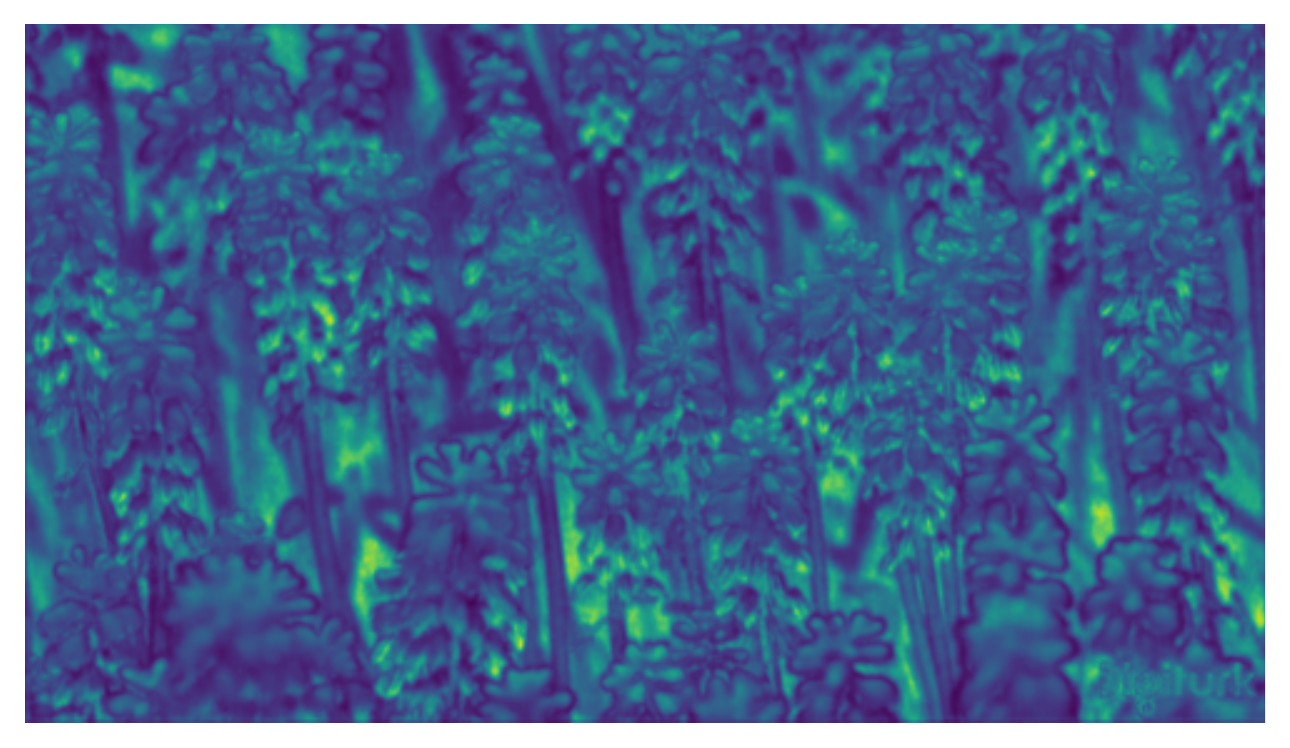} & % NeRV6
    \includegraphics[width=0.19\columnwidth]{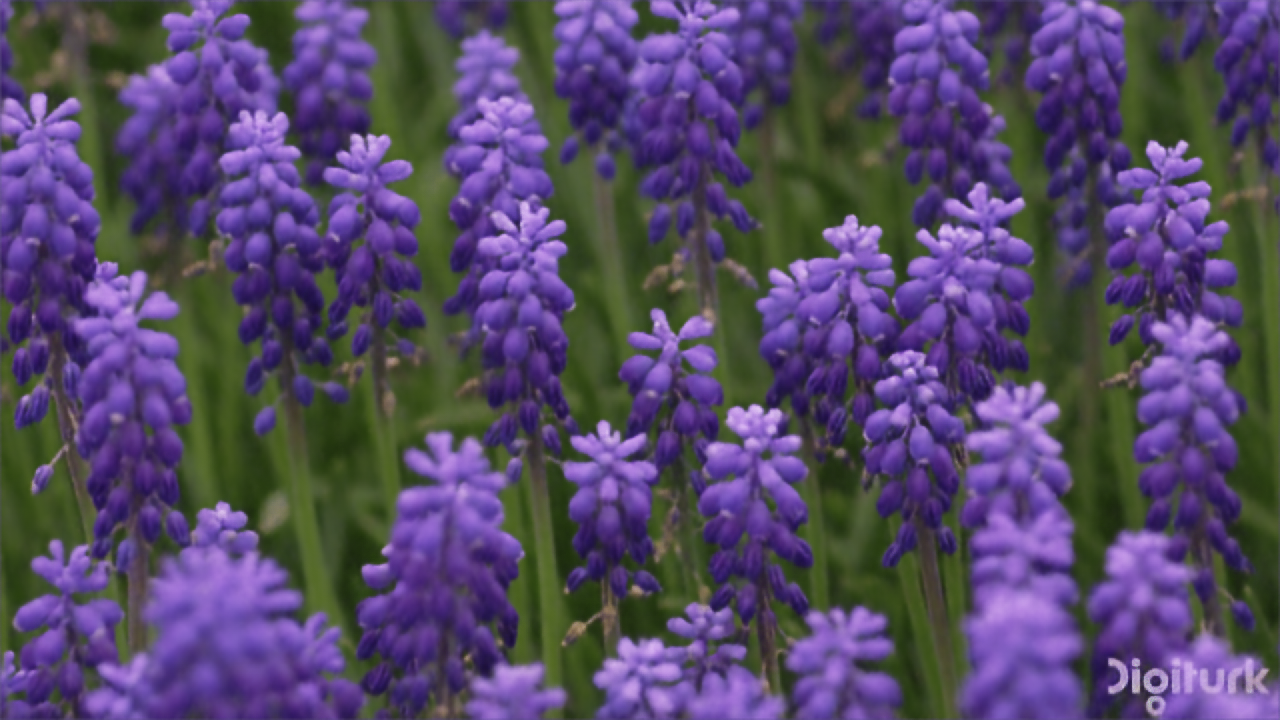} \\
    
    \multicolumn{5}{l}{\small PFNR, c=50.0\%, \textcolor{red}{$f$-NeRV2}} \\

    %\makecell{\small PFNR \\ \tiny c=50\% \\ \tiny $f$-NeRV3} & 
    %\includegraphics[width=0.2\columnwidth]{images/fmaps/5/nerv3/gt_0.png} &  
    %\includegraphics[width=0.3\columnwidth]{images/fmaps/5/nerv3/pred_0_l1.pdf} & % NeRV2
    \includegraphics[width=0.20\columnwidth, trim={0.1cm 0.4cm 0.1cm 0.1cm}, clip]{images/fmaps/5/nerv3/pred_0_l2.pdf} & % NeRV3
    \includegraphics[width=0.20\columnwidth, trim={0.1cm 0.4cm 0.1cm 0.1cm}, clip]{images/fmaps/5/nerv3/pred_0_l3.pdf} & % NeRV4
    \includegraphics[width=0.20\columnwidth, trim={0.1cm 0.4cm 0.1cm 0.1cm}, clip]{images/fmaps/5/nerv3/pred_0_l4.pdf} & % NeRV5
    \includegraphics[width=0.20\columnwidth, trim={0.1cm 0.4cm 0.1cm 0.1cm}, clip]{images/fmaps/5/nerv3/pred_0_l5.pdf} & % NeRV6
    \includegraphics[width=0.19\columnwidth]{images/fmaps/5/nerv3/pred_0.png} \\

    \includegraphics[width=0.20\columnwidth, trim={0.1cm 0.4cm 0.1cm 0.1cm}, clip]{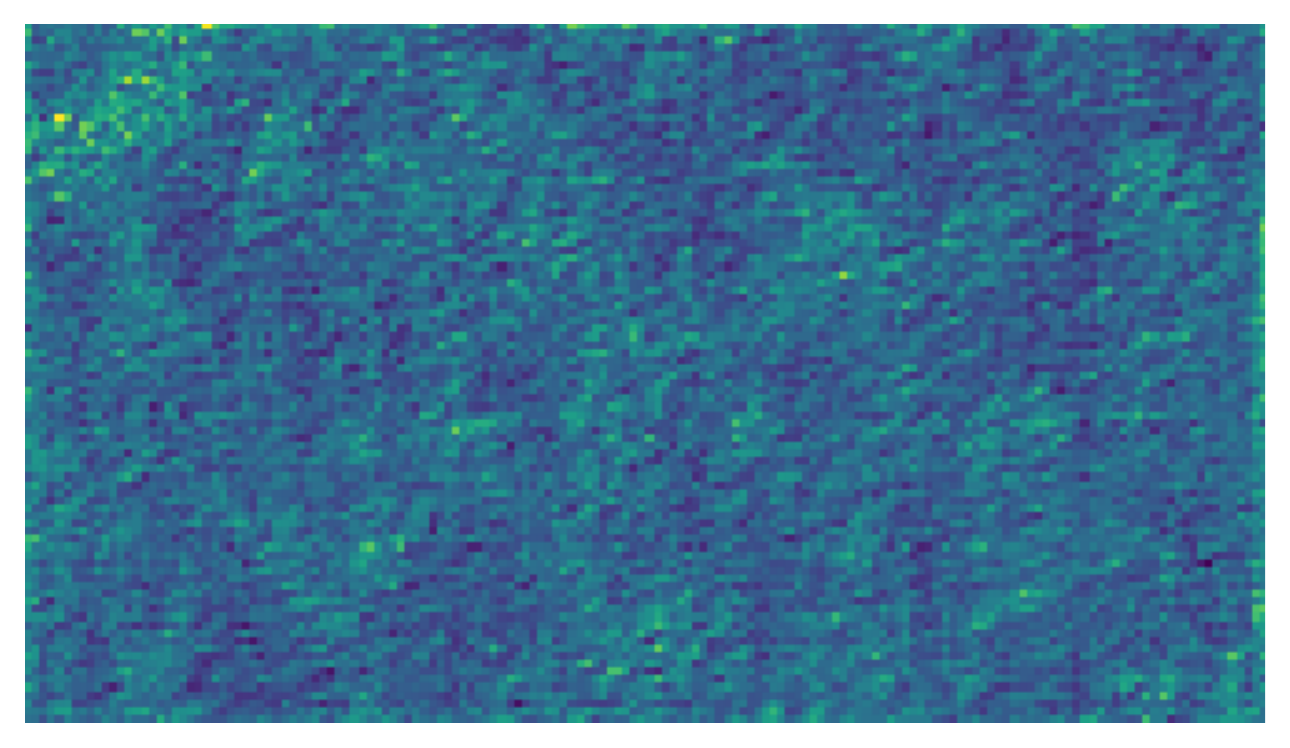} & % NeRV3
    \includegraphics[width=0.20\columnwidth, trim={0.1cm 0.4cm 0.1cm 0.1cm}, clip]{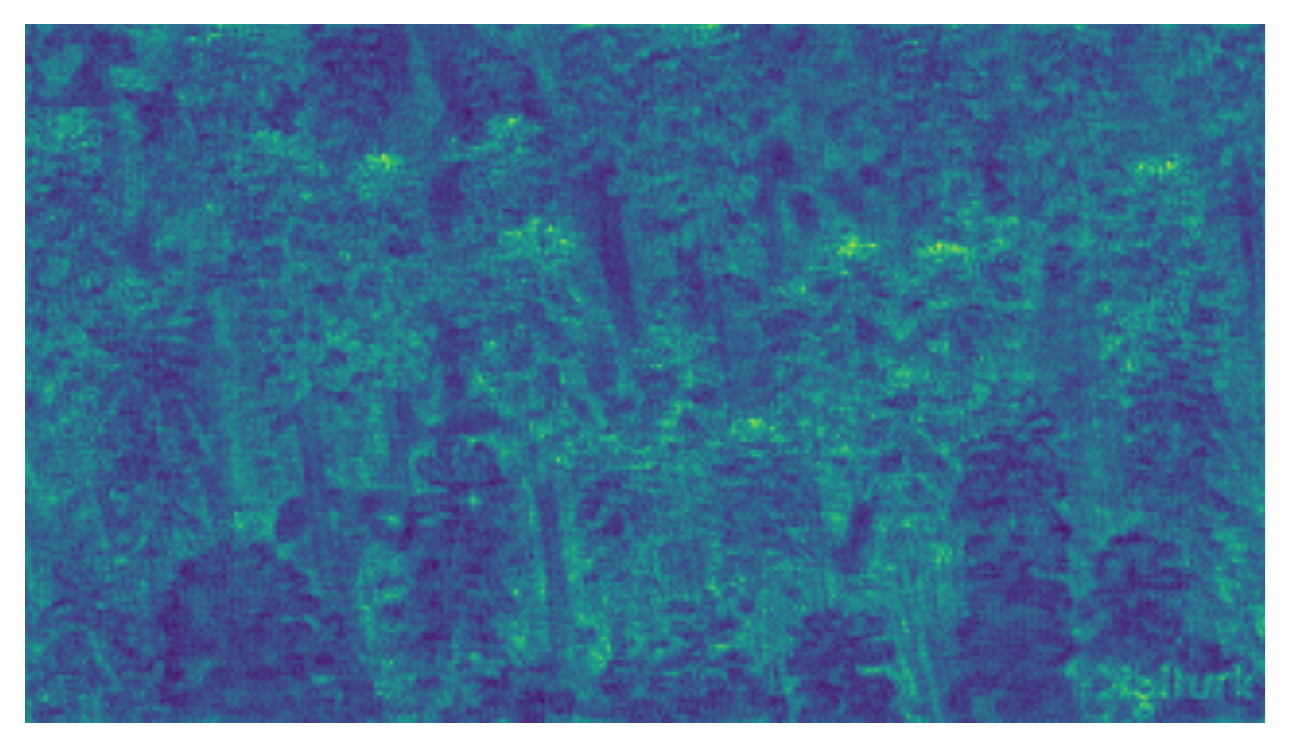} & % NeRV4
    \includegraphics[width=0.20\columnwidth, trim={0.1cm 0.4cm 0.1cm 0.1cm}, clip]{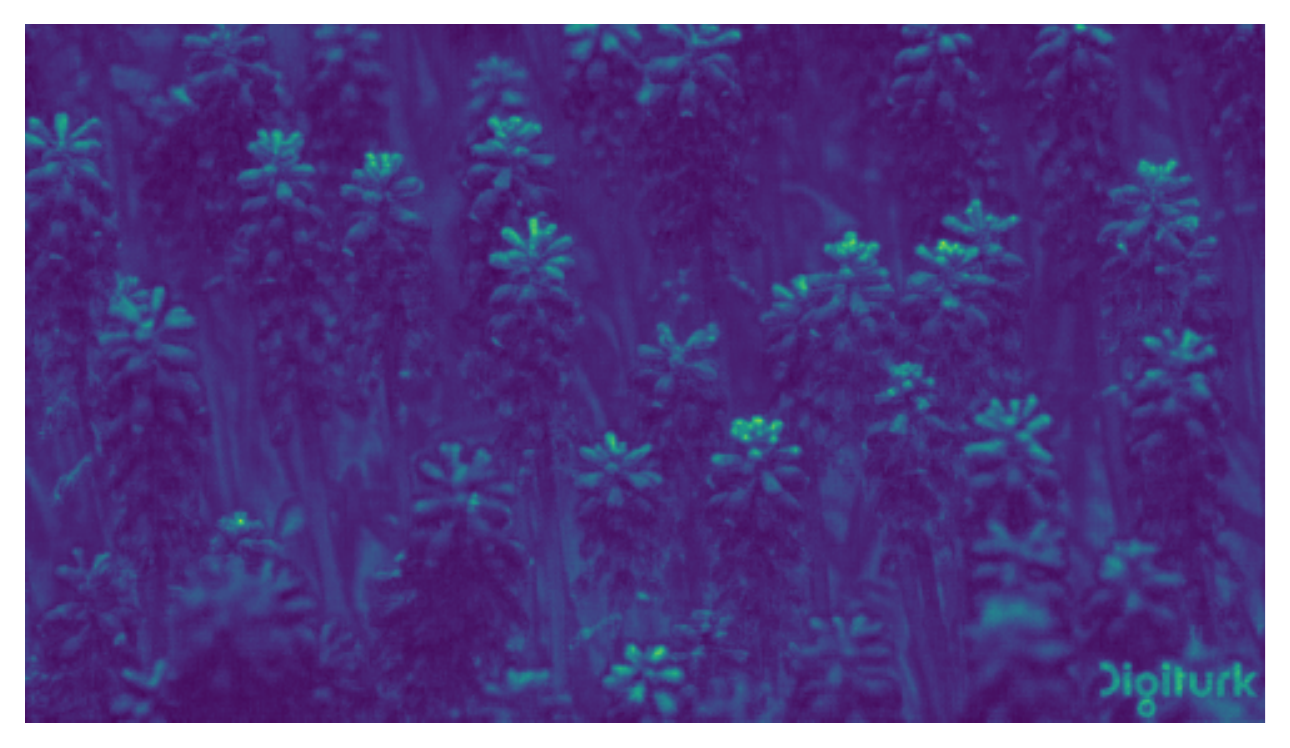} & % NeRV5
    \includegraphics[width=0.20\columnwidth, trim={0.1cm 0.4cm 0.1cm 0.1cm}, clip]{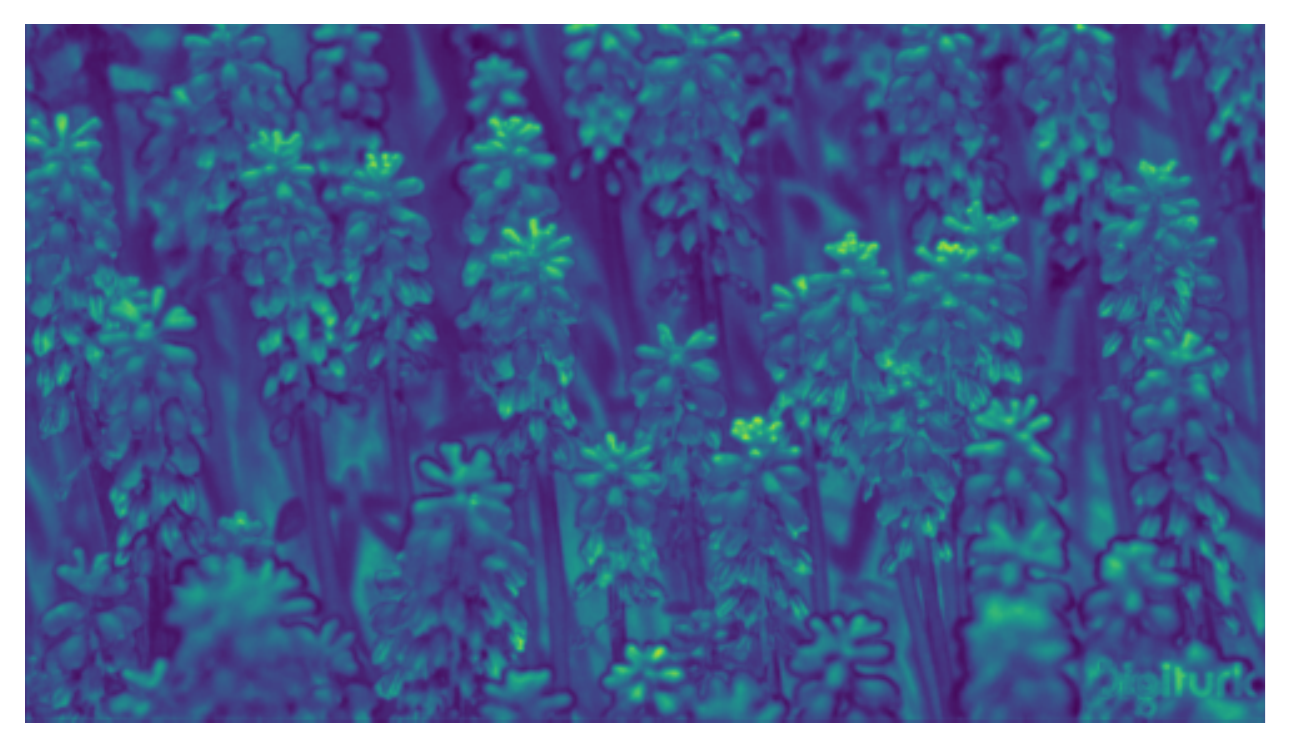} & % NeRV6
    \includegraphics[width=0.19\columnwidth]{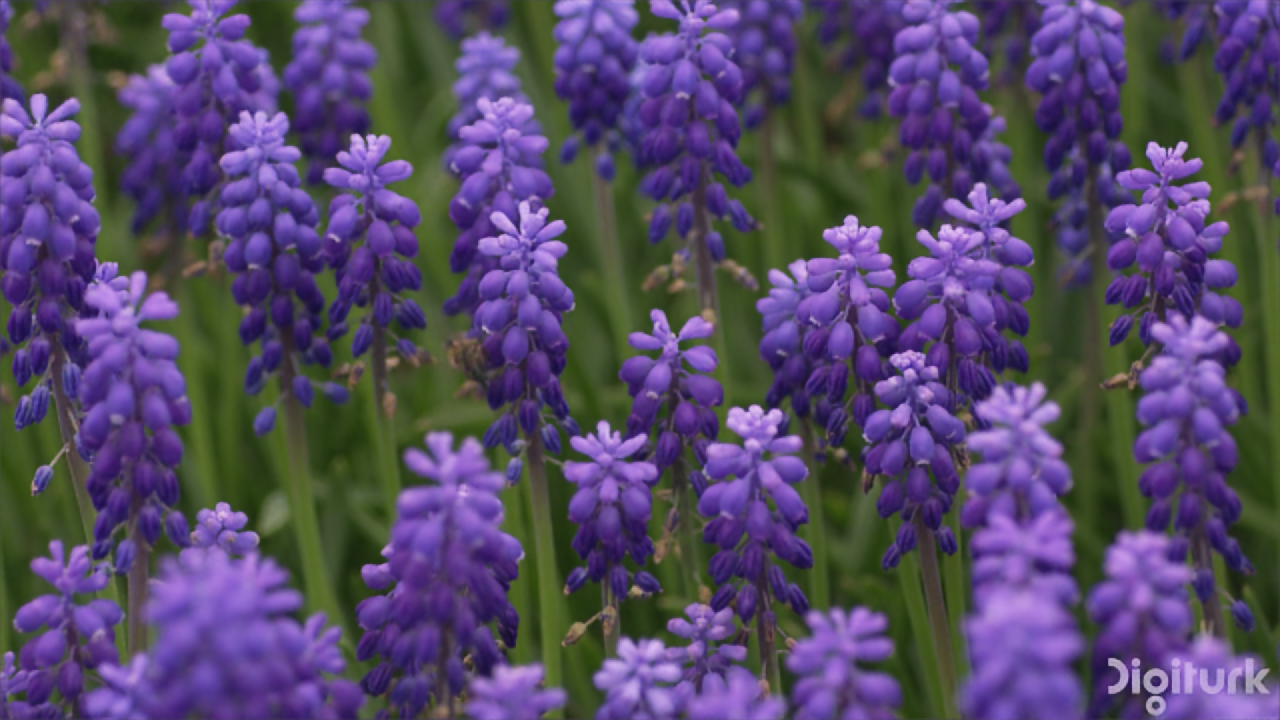} \\
    
    \multicolumn{5}{l}{\small PFNR, c=50.0\%, \textcolor{red}{$f$-NeRV3}} \\
   
    \end{tabular}
    }
    \caption{PFNR's Representations of NeRV Blocks with $c = 50.0 \%$ on the UVG17 dataset.}
    \label{fig:fmap_app_uvg17}
\end{figure*}

%% file: materials/plot_fmap_sparsity_uvg17.tex
\begin{figure*}[ht]
    \centering 
    %\vspace{-0.1in}
    \setlength{\tabcolsep}{0pt}{%
    \begin{tabular}{ccccc}
    
    % 2.city 
    %\textit{2.city} & \textit{3.beauty} & \textit{7.bee} \\ 
     NeRV3 & NeRV4 & NeRV5 & NeRV6 & Head \\ 

    \includegraphics[width=0.20\columnwidth, trim={0.1cm 0.4cm 0.1cm 0.1cm}, clip]{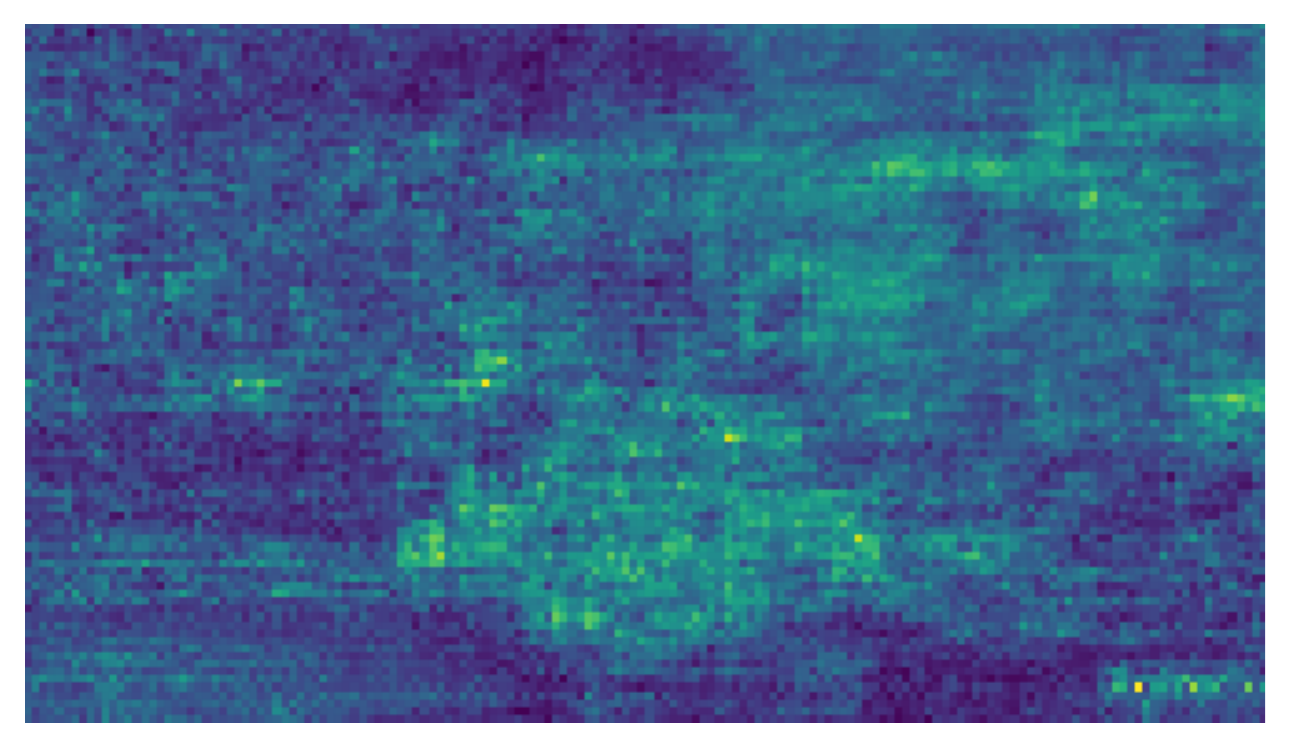} & % NeRV3
    \includegraphics[width=0.20\columnwidth, trim={0.1cm 0.4cm 0.1cm 0.1cm}, clip]{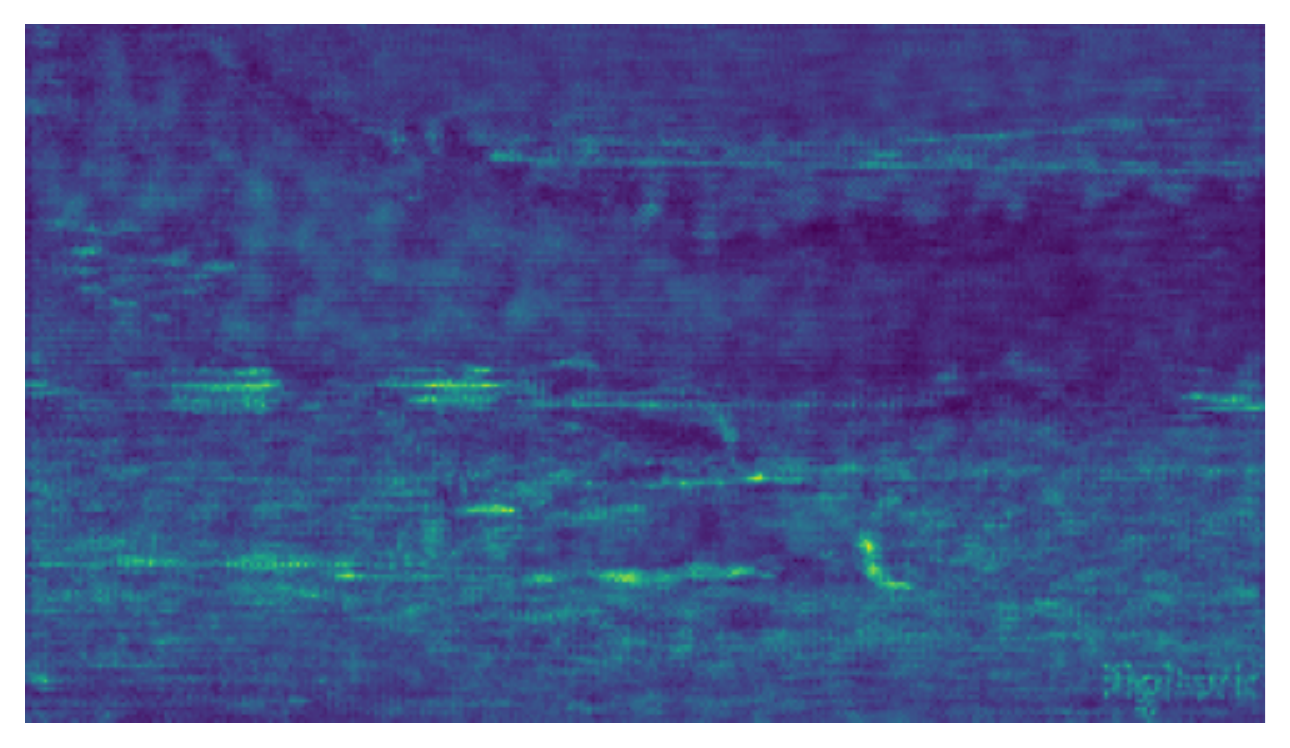} & % NeRV4
    \includegraphics[width=0.20\columnwidth, trim={0.1cm 0.4cm 0.1cm 0.1cm}, clip]{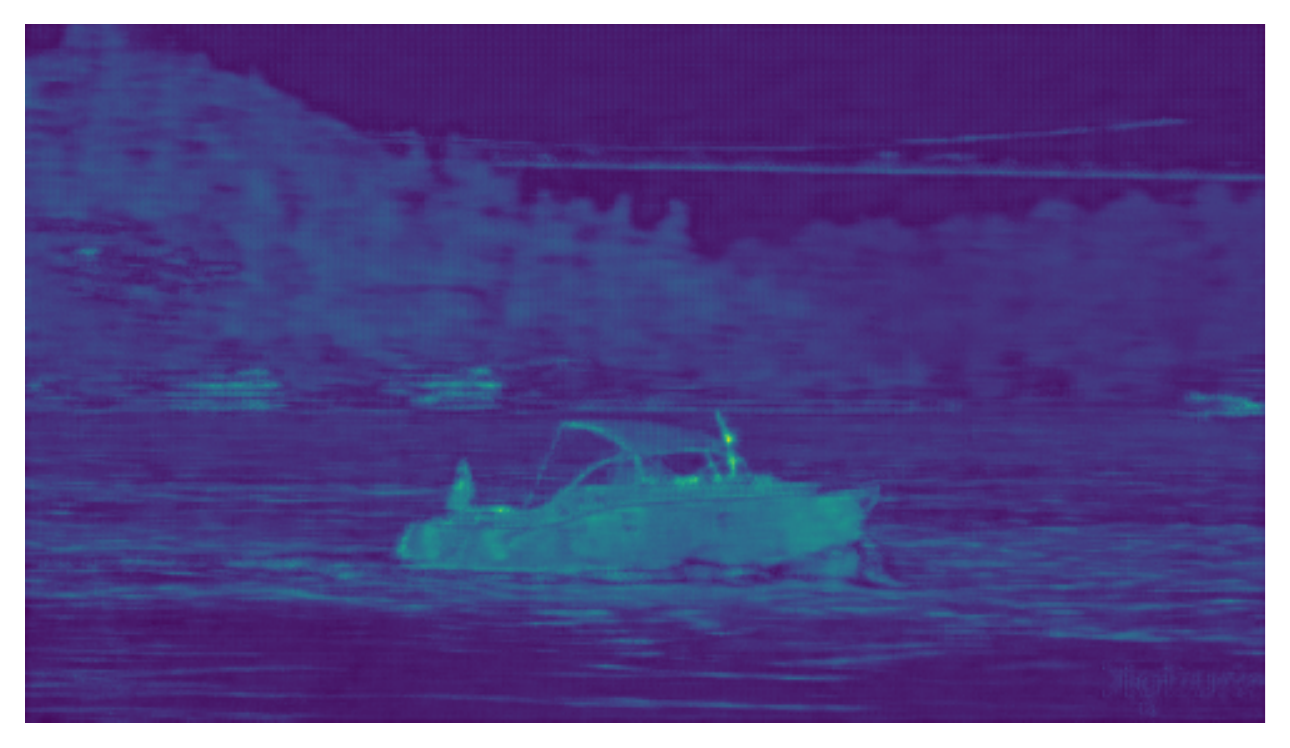} & % NeRV5
    \includegraphics[width=0.20\columnwidth, trim={0.1cm 0.4cm 0.1cm 0.1cm}, clip]{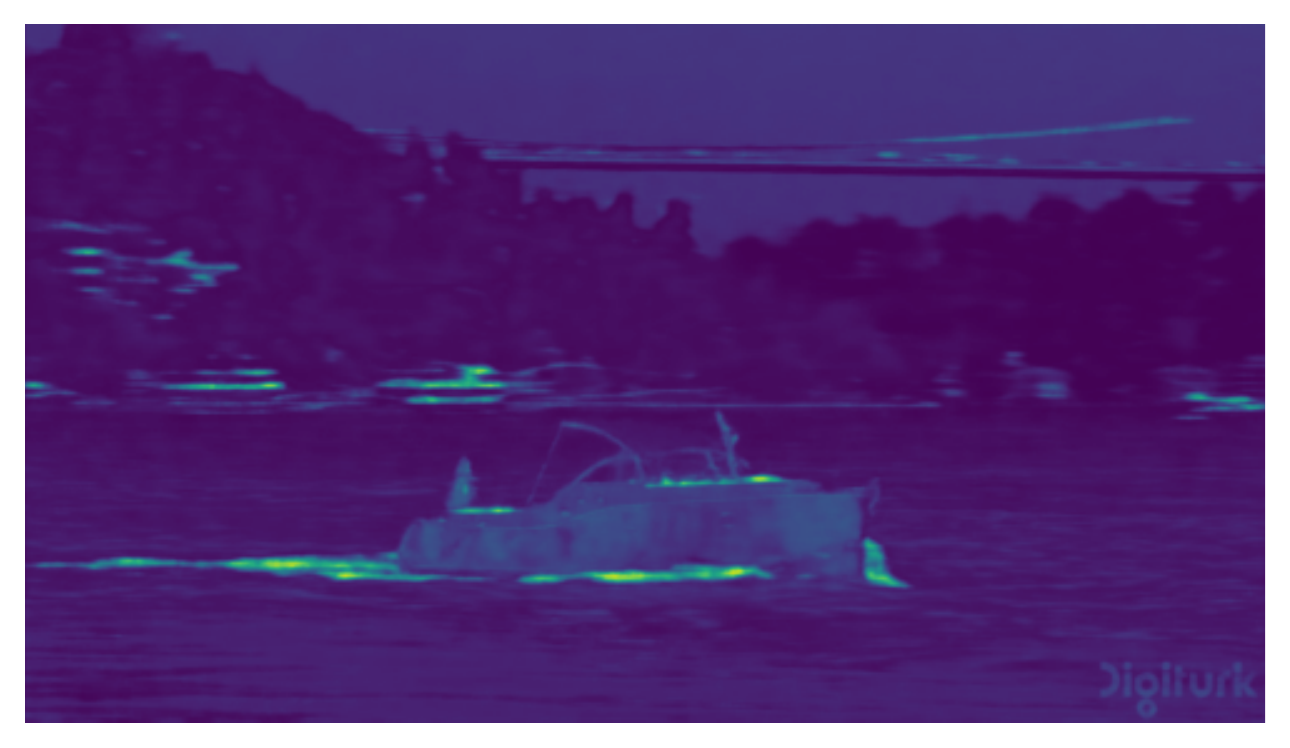} & % NeRV6
    \includegraphics[width=0.19\columnwidth]{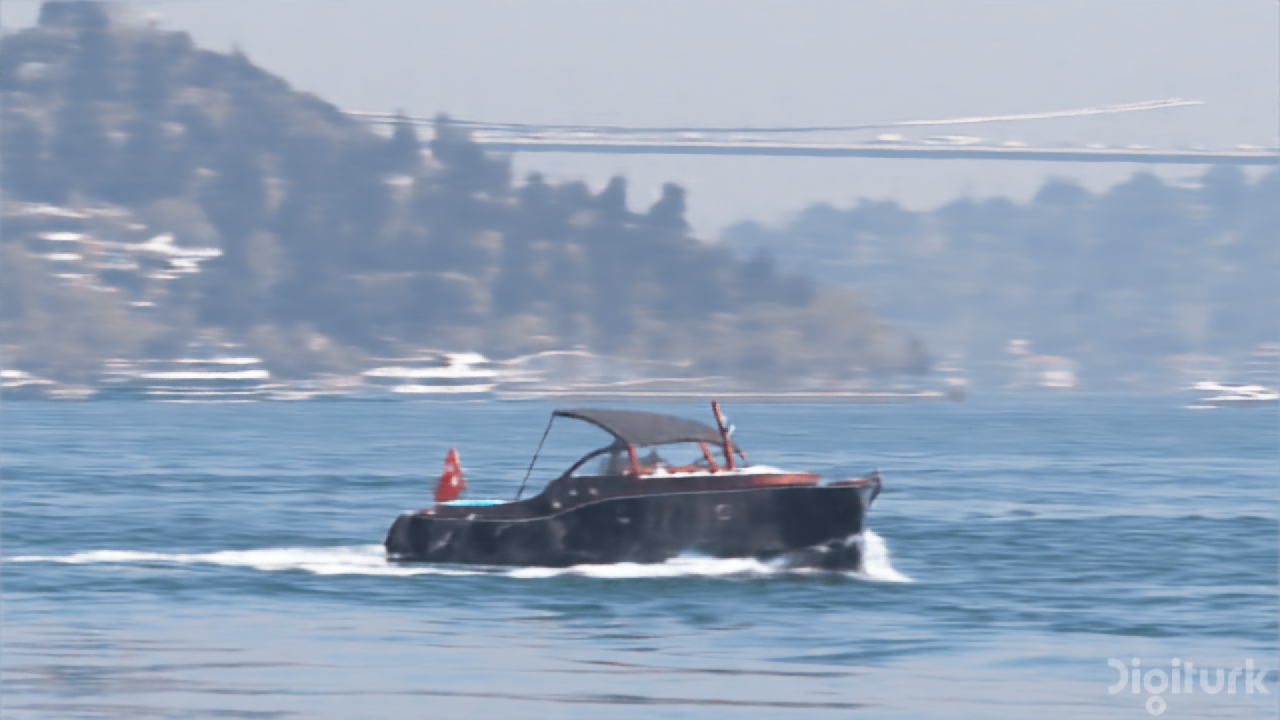} \\
    \multicolumn{5}{l}{\small PFNR, c=50.0\%, sparsity of \textcolor{red}{$f$-NeRV3=0.5\%}.} \\

    \includegraphics[width=0.20\columnwidth, trim={0.1cm 0.4cm 0.1cm 0.1cm}, clip]{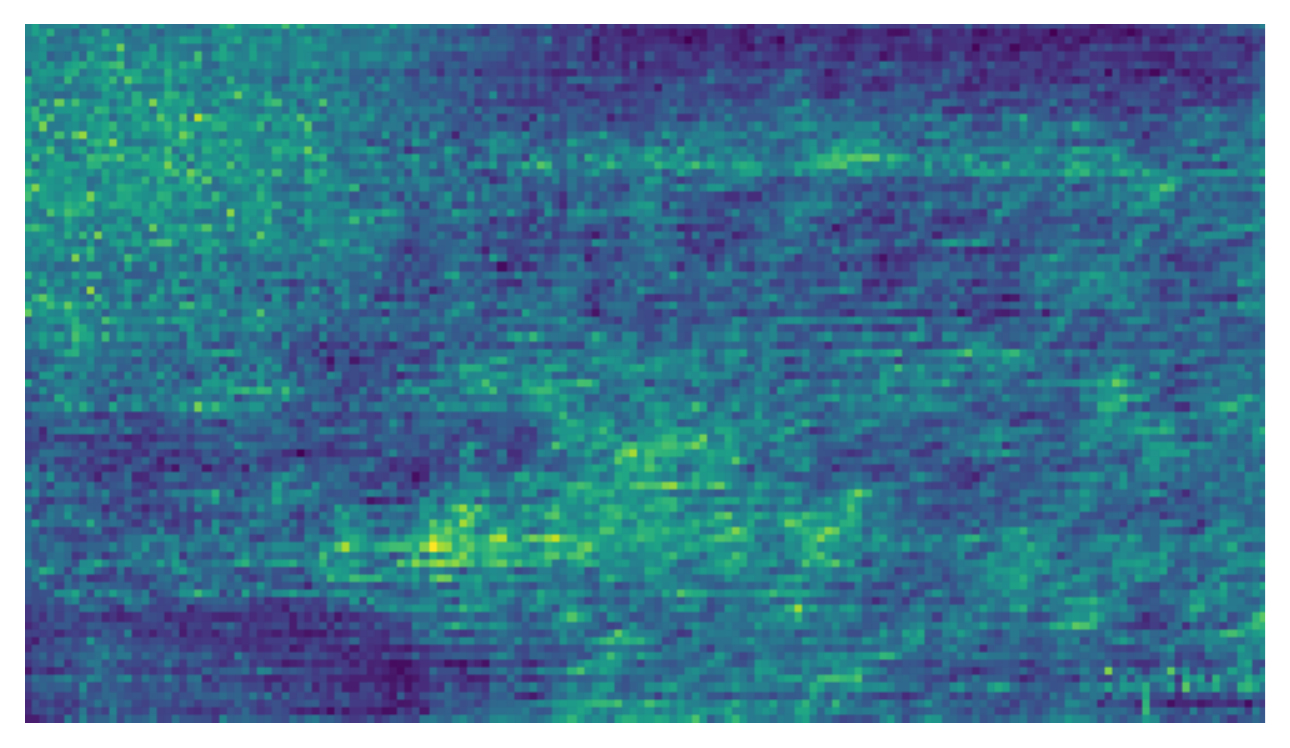} & % NeRV3
    \includegraphics[width=0.20\columnwidth, trim={0.1cm 0.4cm 0.1cm 0.1cm}, clip]{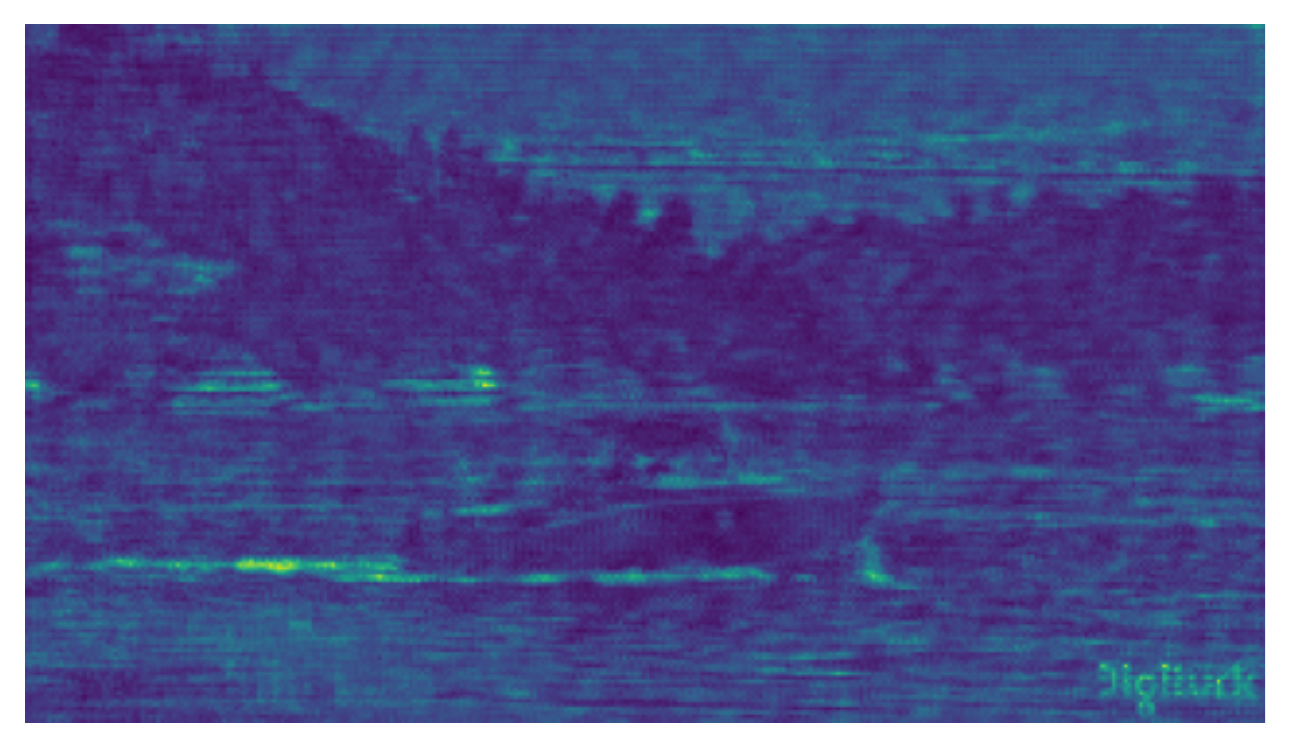} & % NeRV4
    \includegraphics[width=0.20\columnwidth, trim={0.1cm 0.4cm 0.1cm 0.1cm}, clip]{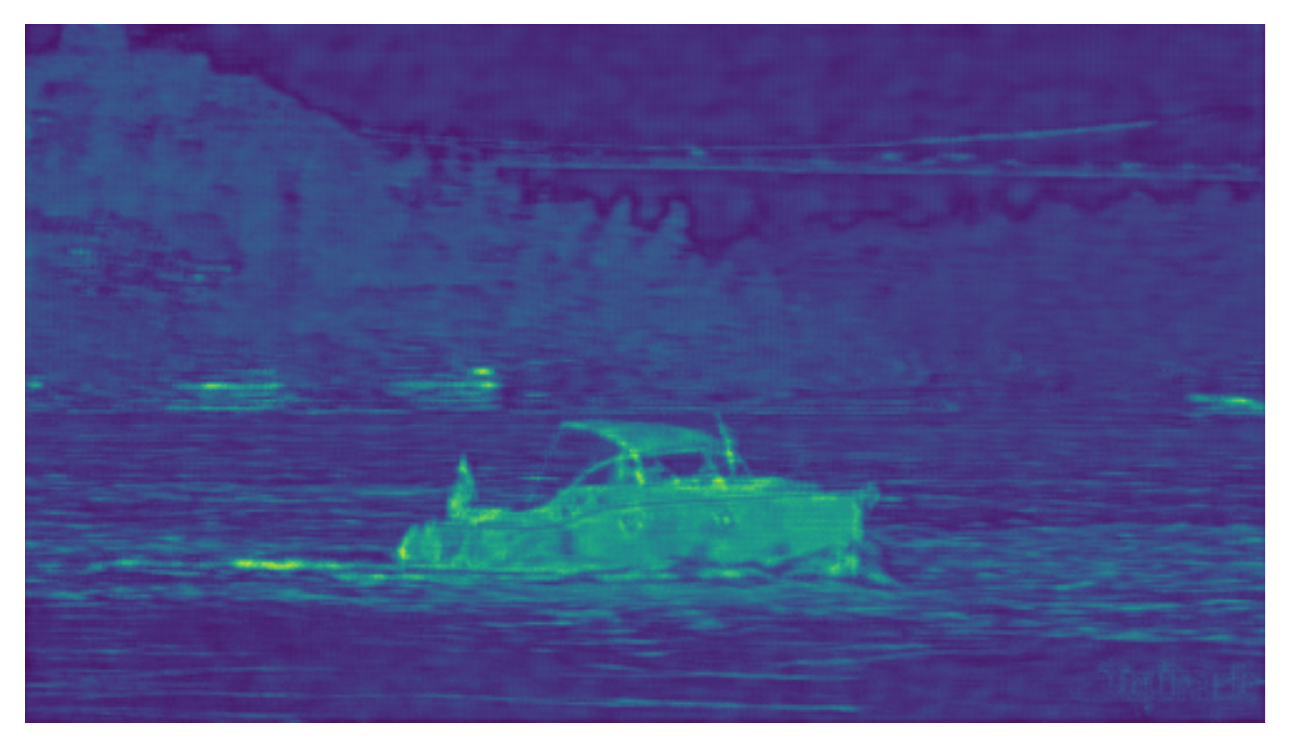} & % NeRV5
    \includegraphics[width=0.20\columnwidth, trim={0.1cm 0.4cm 0.1cm 0.1cm}, clip]{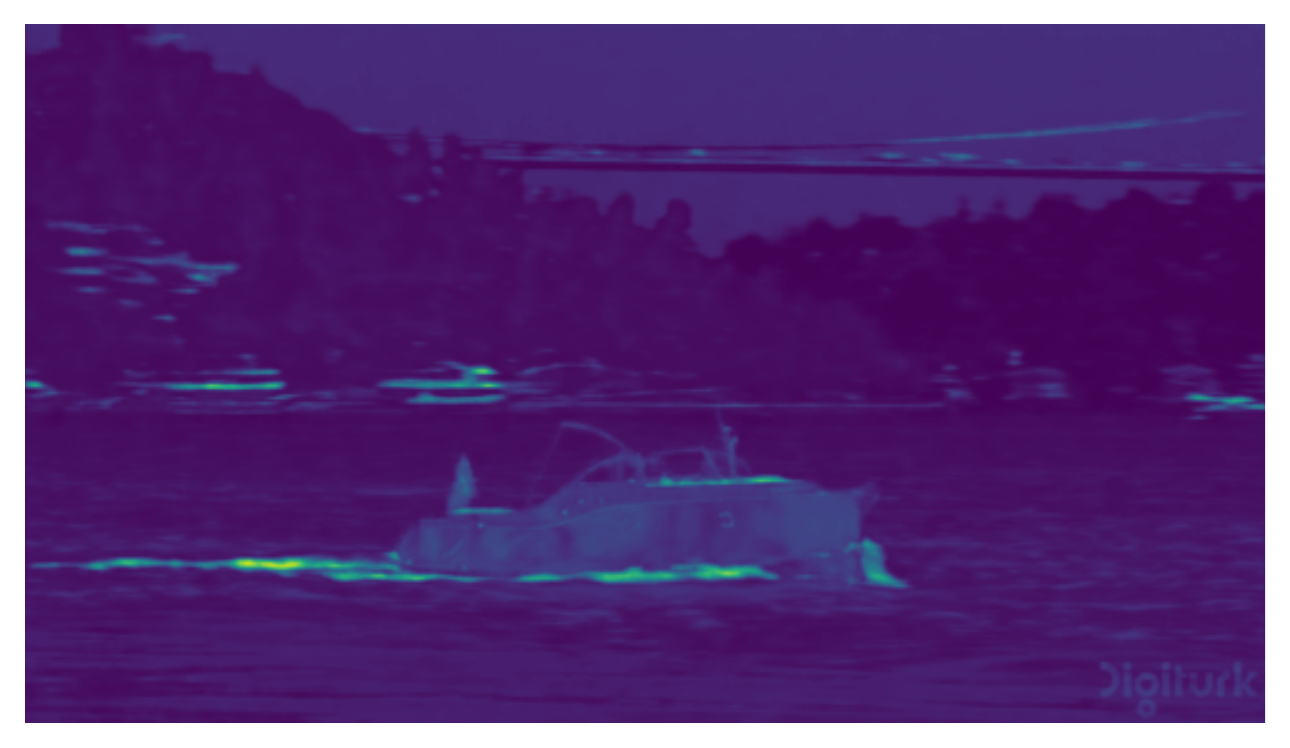} & % NeRV6
    \includegraphics[width=0.19\columnwidth]{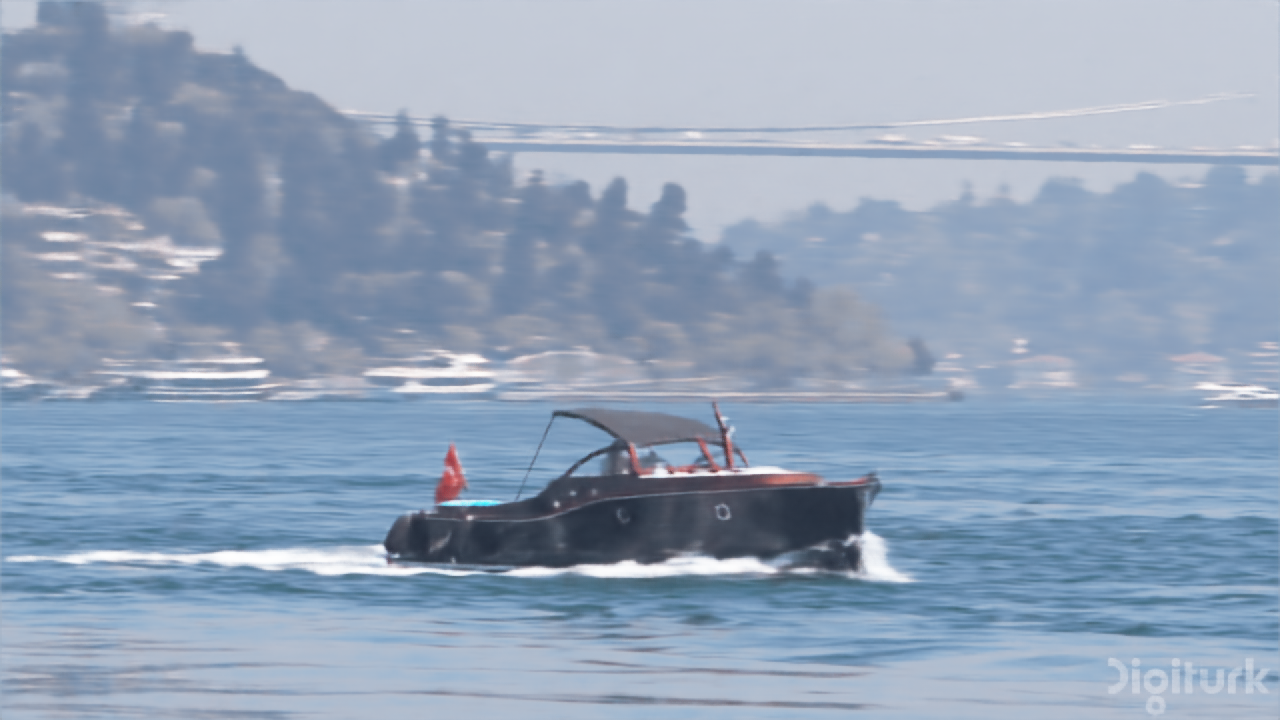} \\
    \multicolumn{5}{l}{\small PFNR, c=50.0\%, sparsity of \textcolor{red}{$f$-NeRV3=2.5\%}.} \\

    \includegraphics[width=0.20\columnwidth, trim={0.1cm 0.4cm 0.1cm 0.1cm}, clip]{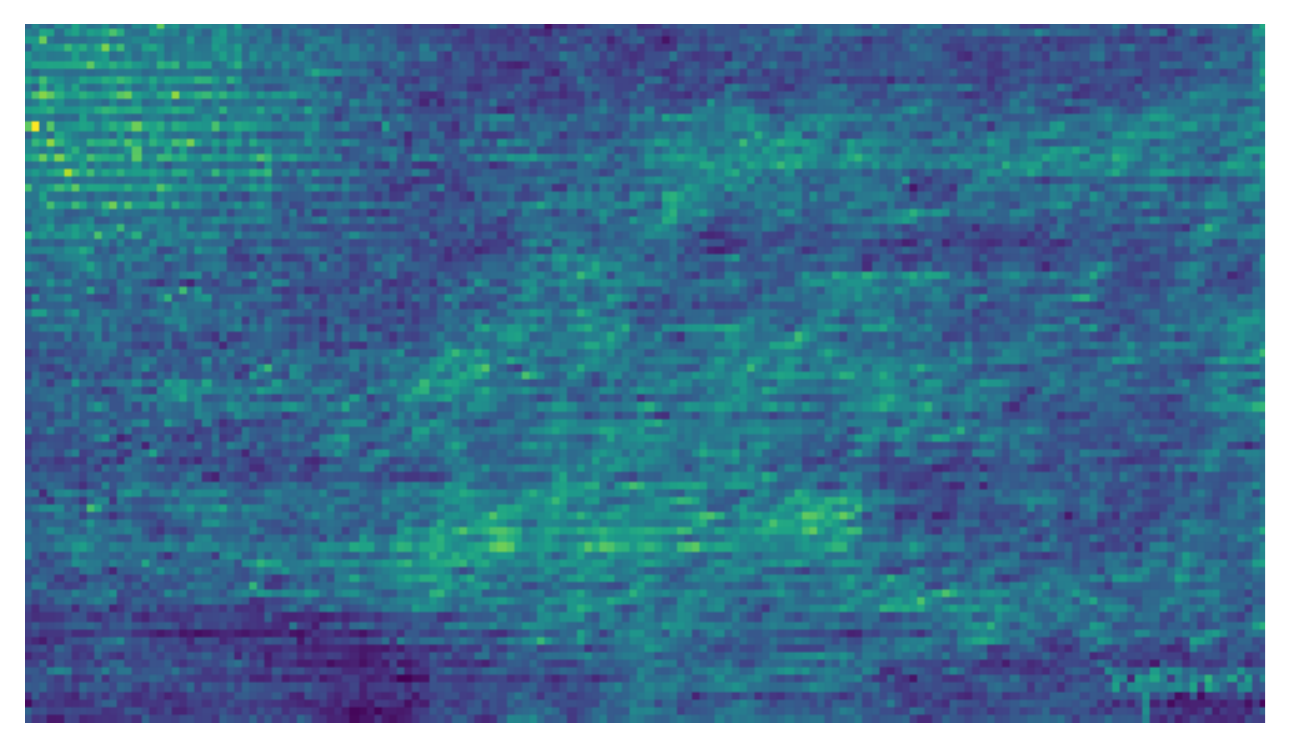} & % NeRV3
    \includegraphics[width=0.20\columnwidth, trim={0.1cm 0.4cm 0.1cm 0.1cm}, clip]{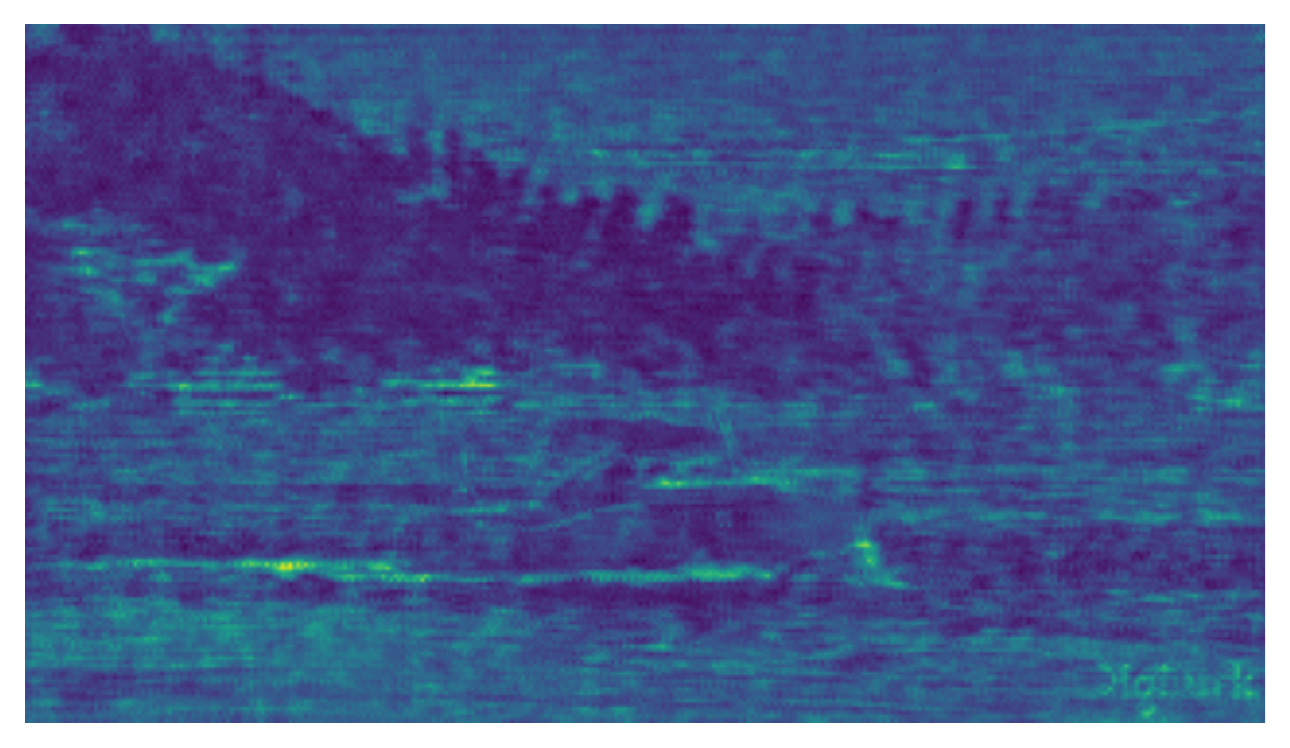} & % NeRV4
    \includegraphics[width=0.20\columnwidth, trim={0.1cm 0.4cm 0.1cm 0.1cm}, clip]{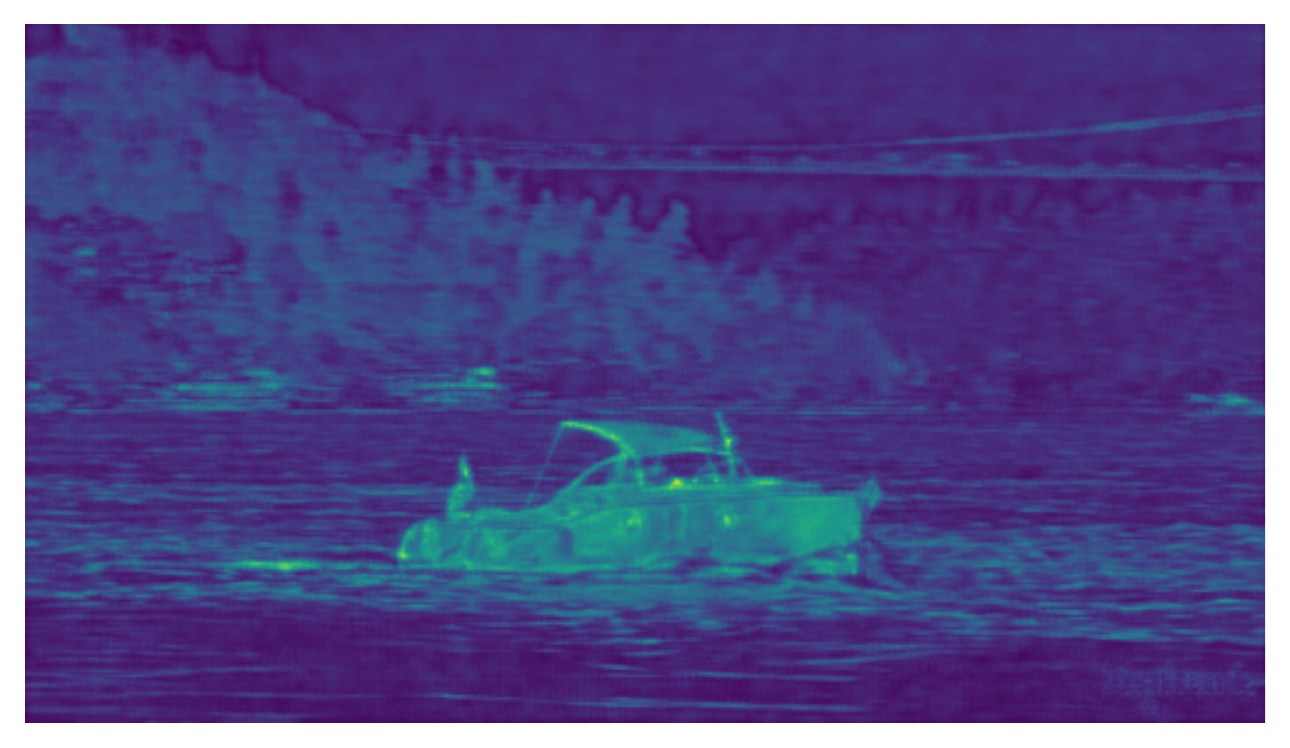} & % NeRV5
    \includegraphics[width=0.20\columnwidth, trim={0.1cm 0.4cm 0.1cm 0.1cm}, clip]{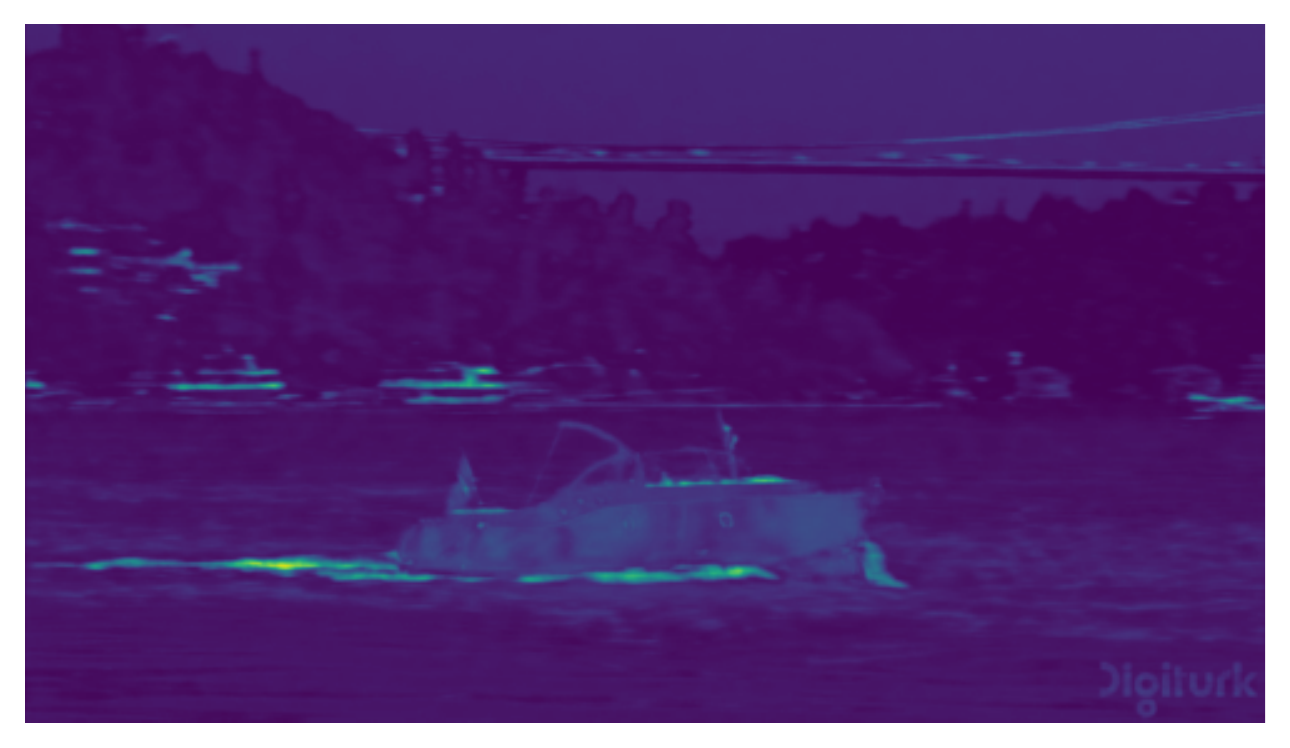} & % NeRV6
    \includegraphics[width=0.19\columnwidth]{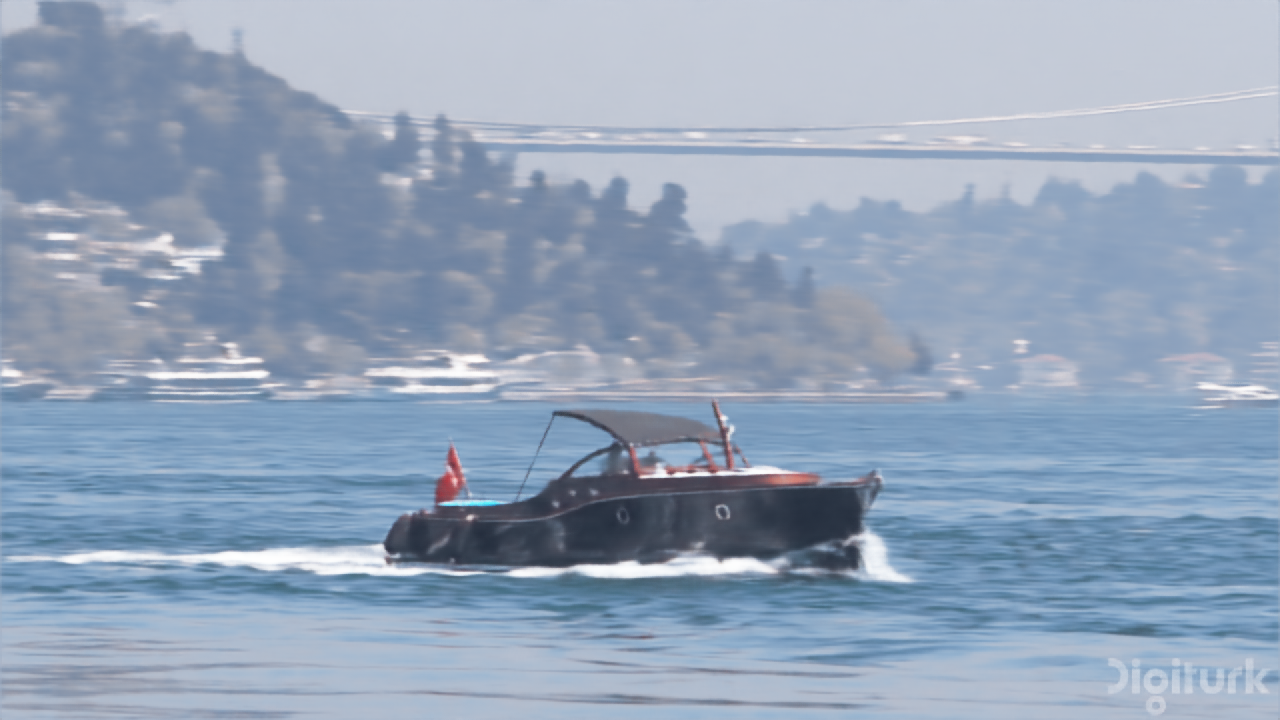} \\
    \multicolumn{5}{l}{\small PFNR, c=50.0\%, sparsity of \textcolor{red}{$f$-NeRV3=5.0\%}.} \\

    \includegraphics[width=0.20\columnwidth, trim={0.1cm 0.4cm 0.1cm 0.1cm}, clip]{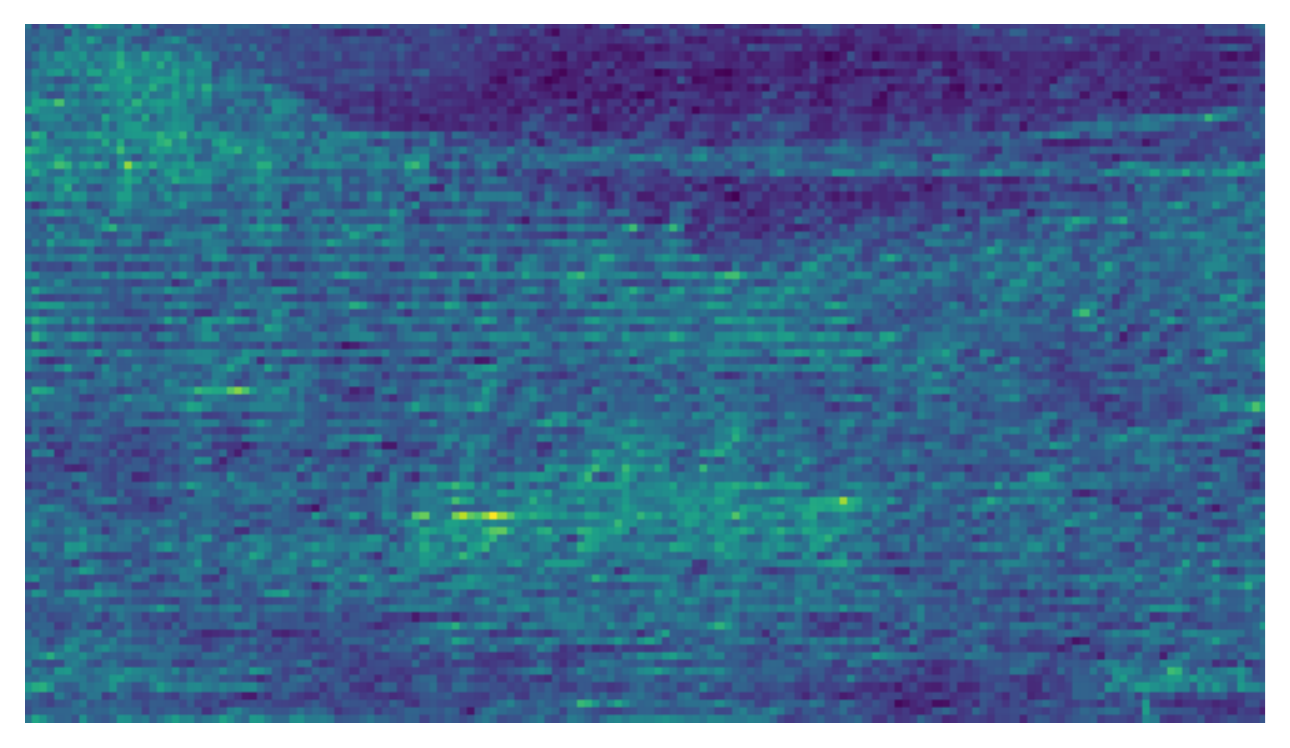} & % NeRV3
    \includegraphics[width=0.20\columnwidth, trim={0.1cm 0.4cm 0.1cm 0.1cm}, clip]{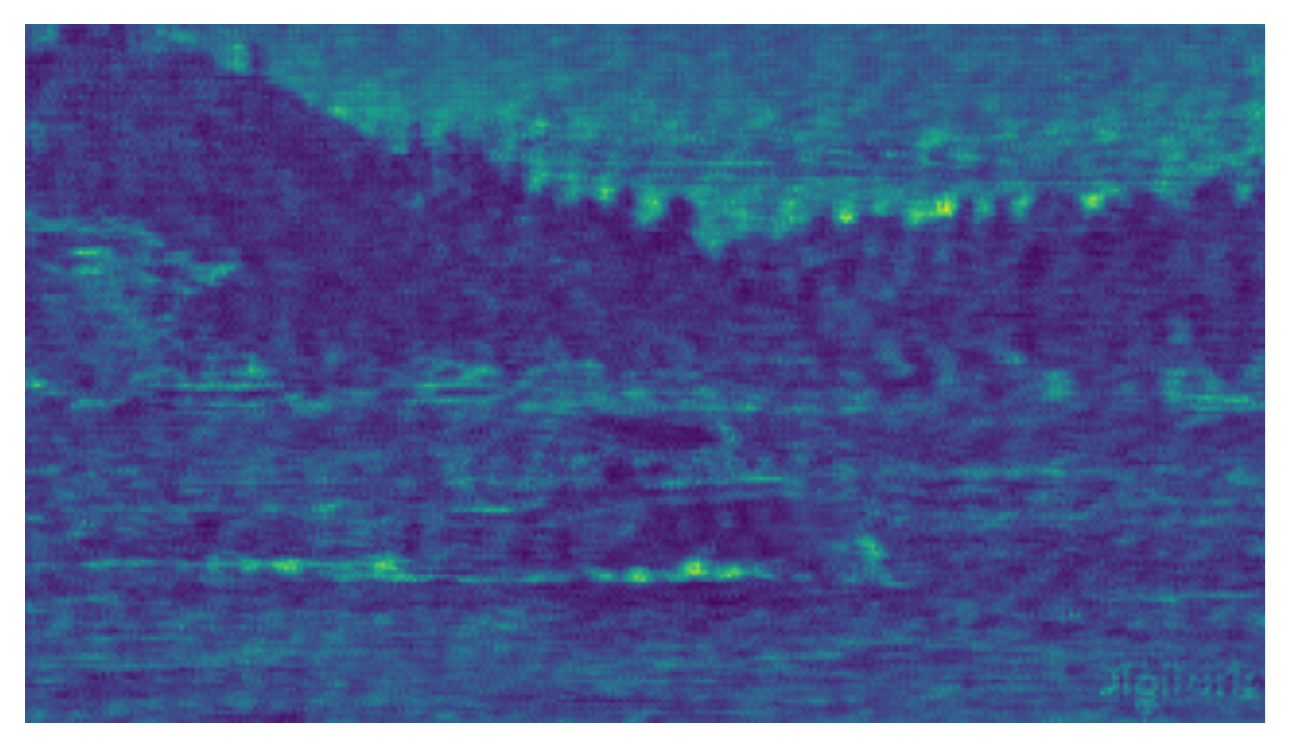} & % NeRV4
    \includegraphics[width=0.20\columnwidth, trim={0.1cm 0.4cm 0.1cm 0.1cm}, clip]{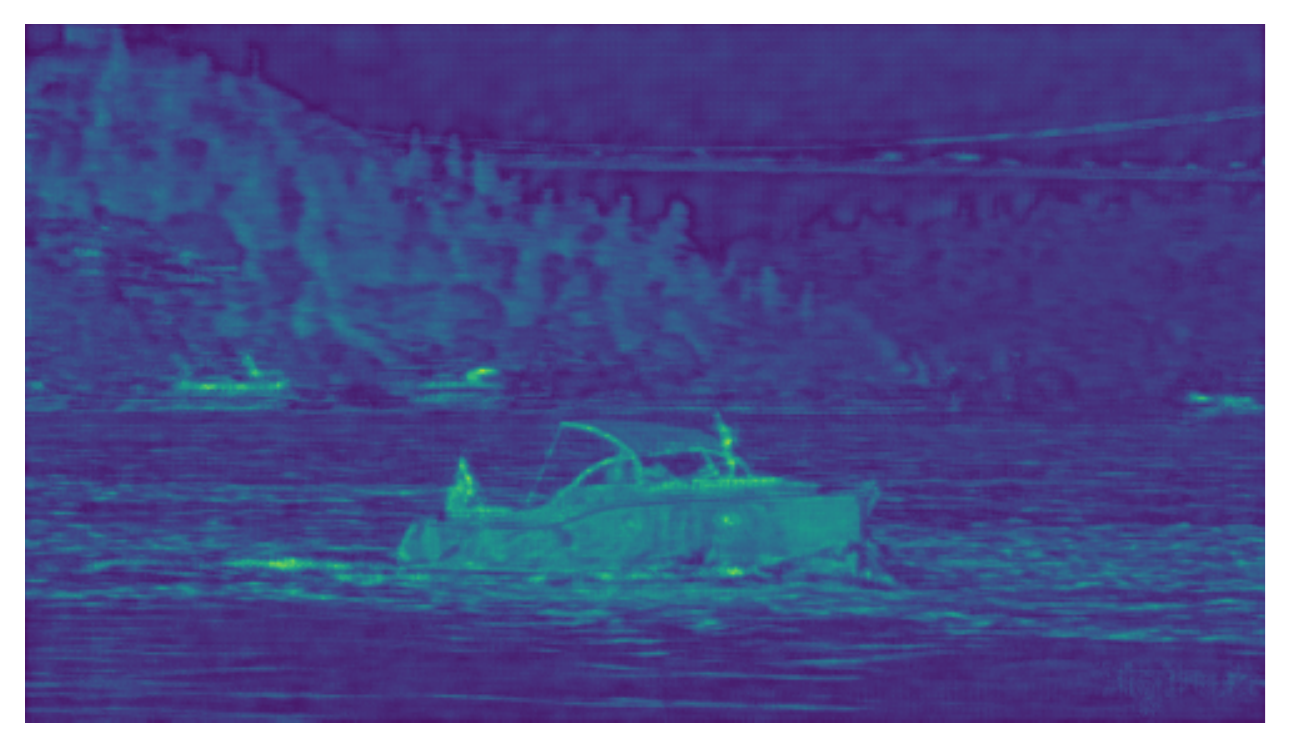} & % NeRV5
    \includegraphics[width=0.20\columnwidth, trim={0.1cm 0.4cm 0.1cm 0.1cm}, clip]{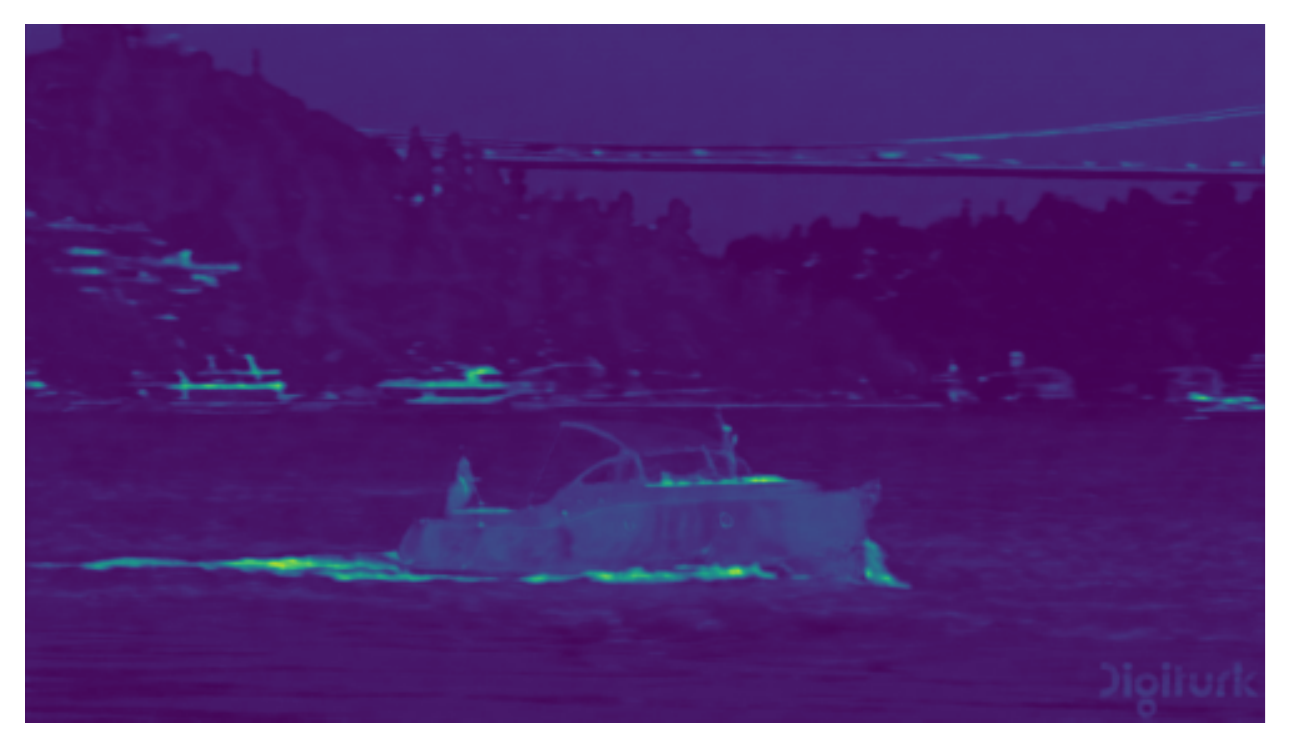} & % NeRV6
    \includegraphics[width=0.19\columnwidth]{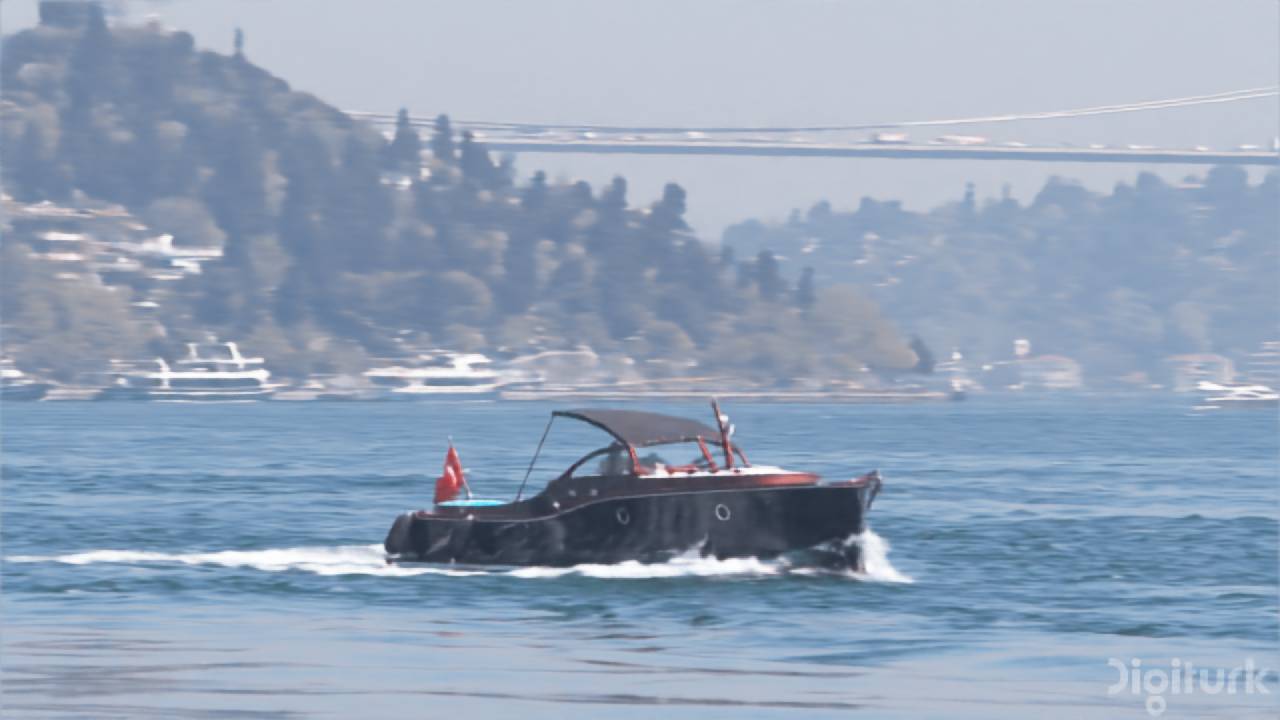} \\
    \multicolumn{5}{l}{\small PFNR, c=50.0\%, sparsity of \textcolor{red}{$f$-NeRV3=15.0\%}.} \\

    %\midrule
    \includegraphics[width=0.20\columnwidth, trim={0.1cm 0.4cm 0.1cm 0.1cm}, clip]{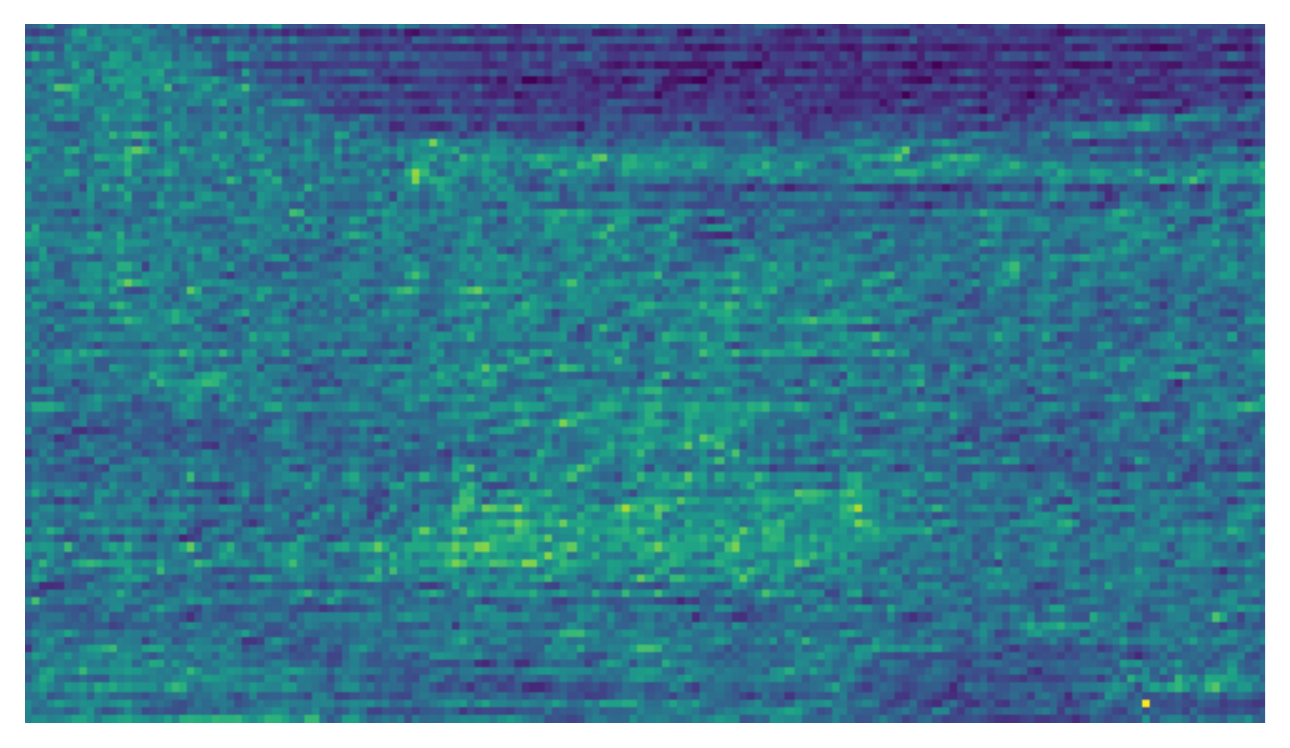} & % NeRV3
    \includegraphics[width=0.20\columnwidth, trim={0.1cm 0.4cm 0.1cm 0.1cm}, clip]{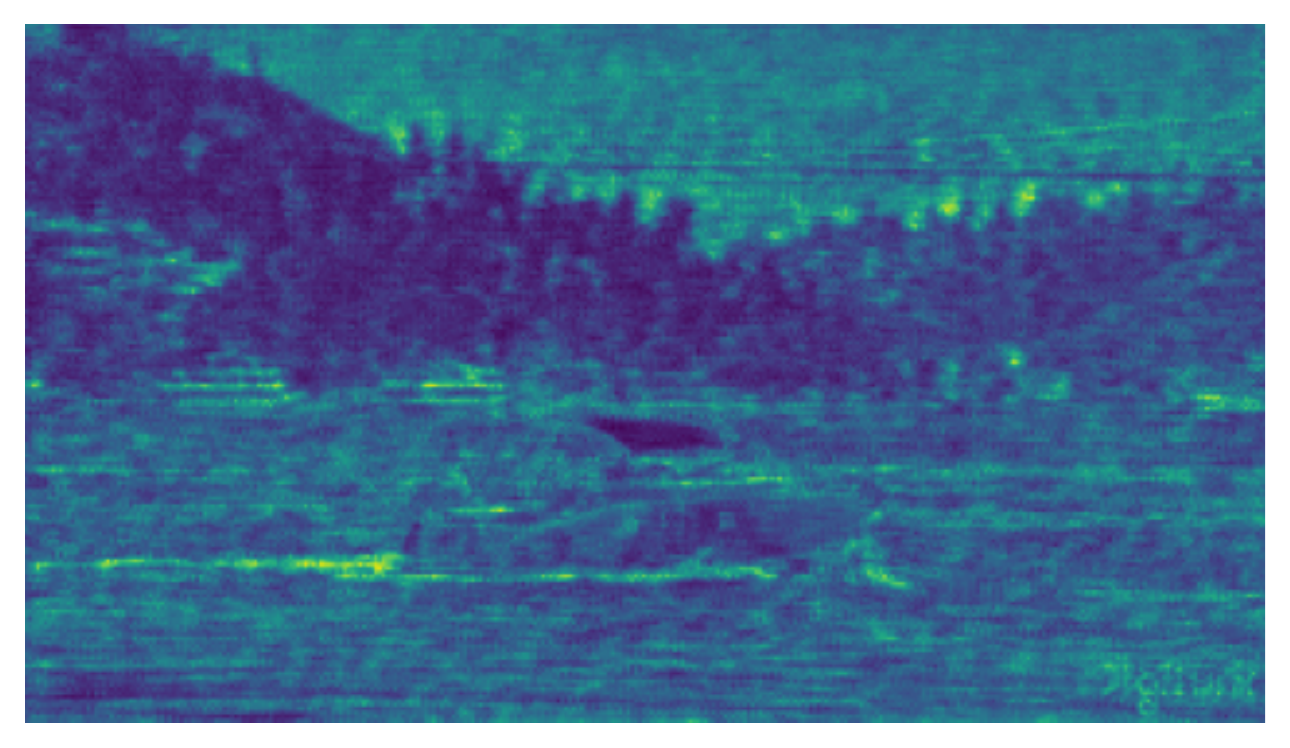} & % NeRV4
    \includegraphics[width=0.20\columnwidth, trim={0.1cm 0.4cm 0.1cm 0.1cm}, clip]{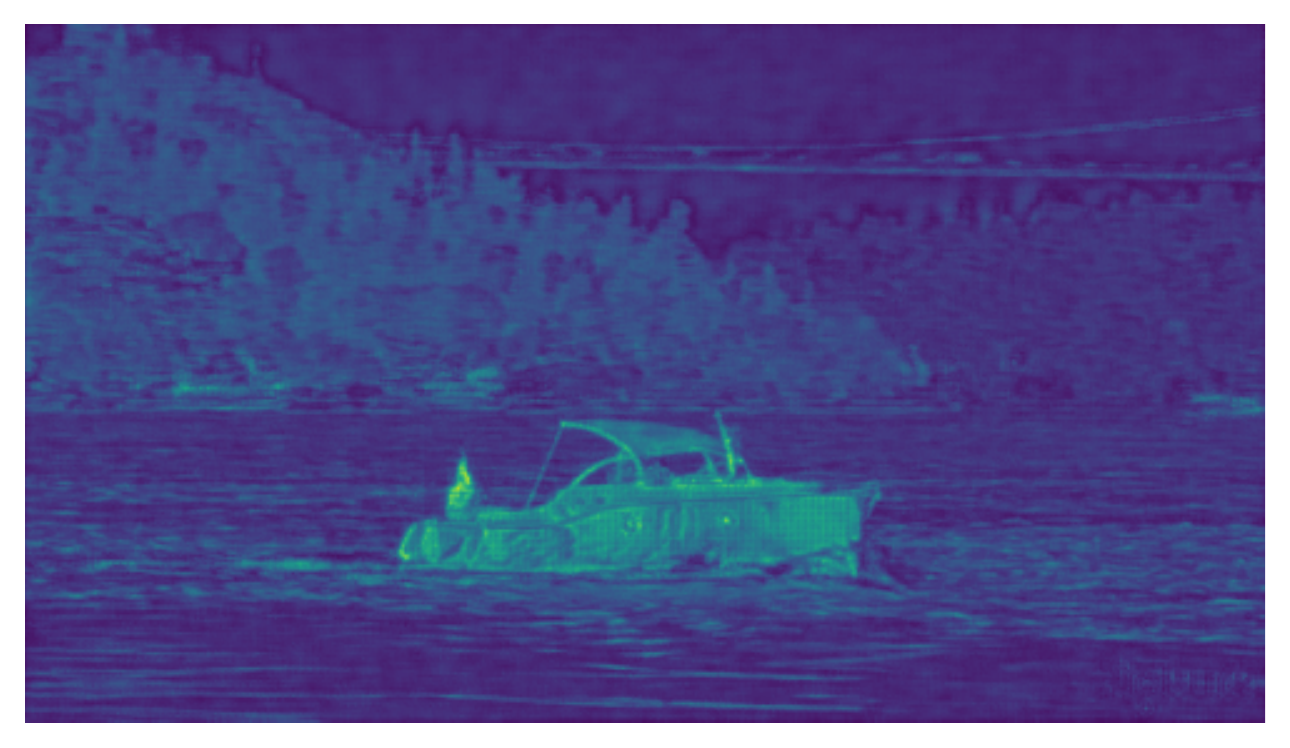} & % NeRV5
    \includegraphics[width=0.20\columnwidth, trim={0.1cm 0.4cm 0.1cm 0.1cm}, clip]{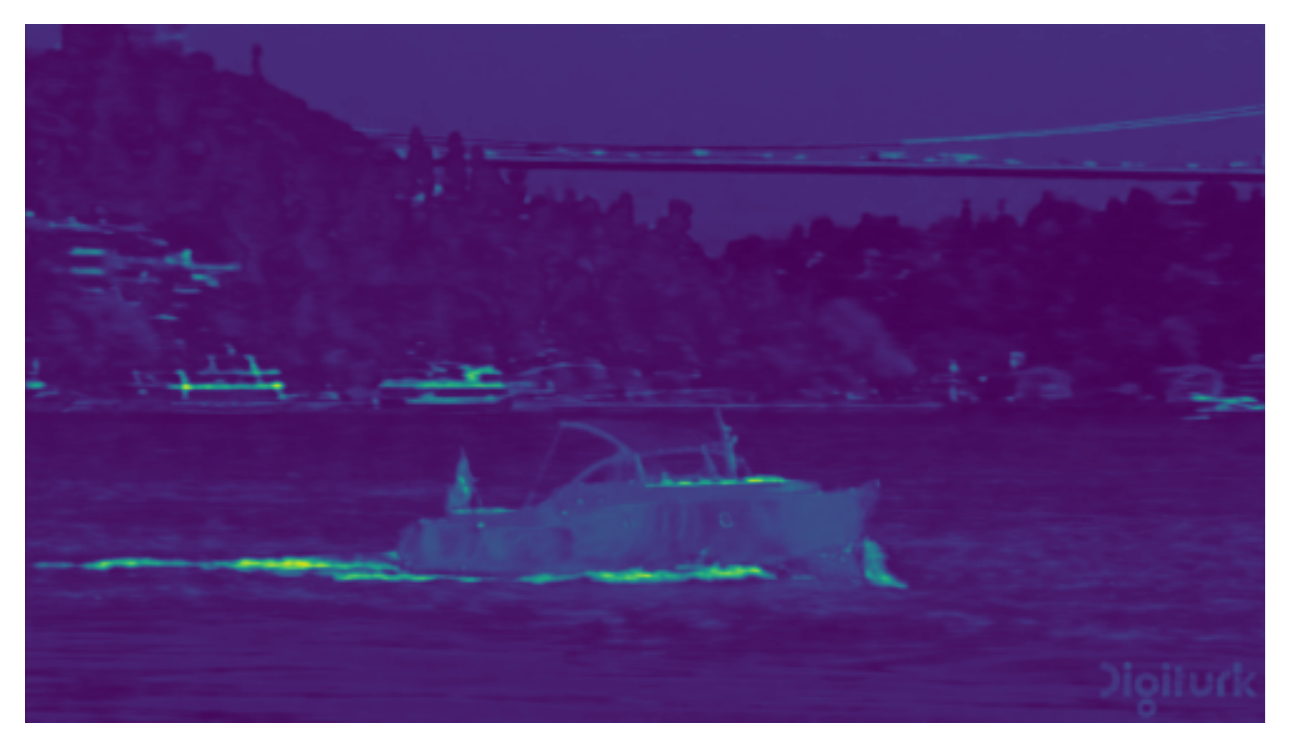} & % NeRV6
    \includegraphics[width=0.19\columnwidth]{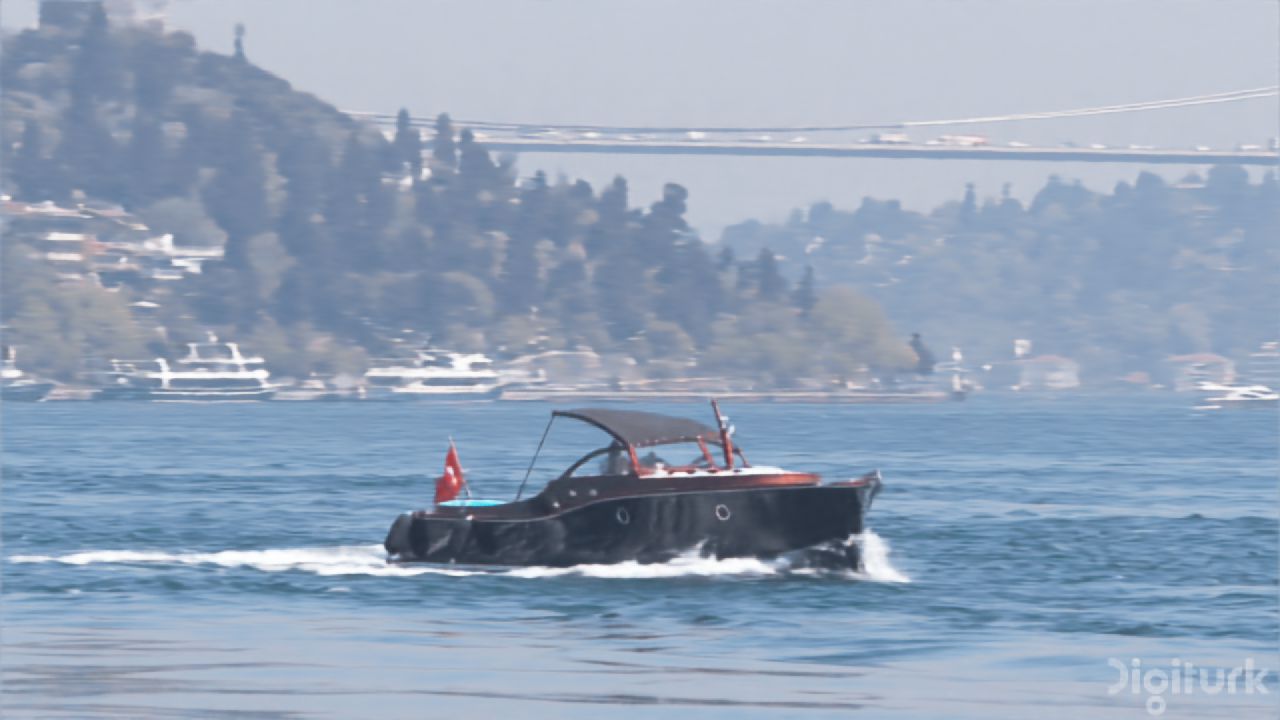} \\
    \multicolumn{5}{l}{\small PFNR, c=50.0\%, sparsity of \textcolor{red}{$f$-NeRV3=25.0\%}.} \\
    %\midrule

    \includegraphics[width=0.20\columnwidth, trim={0.1cm 0.4cm 0.1cm 0.1cm}, clip]{images/fmaps/5/nerv3/pred_0_l2.pdf} & % NeRV3
    \includegraphics[width=0.20\columnwidth, trim={0.1cm 0.4cm 0.1cm 0.1cm}, clip]{images/fmaps/5/nerv3/pred_0_l3.pdf} & % NeRV4
    \includegraphics[width=0.20\columnwidth, trim={0.1cm 0.4cm 0.1cm 0.1cm}, clip]{images/fmaps/5/nerv3/pred_0_l4.pdf} & % NeRV5
    \includegraphics[width=0.20\columnwidth, trim={0.1cm 0.4cm 0.1cm 0.1cm}, clip]{images/fmaps/5/nerv3/pred_0_l5.pdf} & % NeRV6
    \includegraphics[width=0.19\columnwidth]{images/fmaps/5/nerv3/pred_0.png} \\
    
    %\multicolumn{5}{l}{\small PFNR, c=50.0\%, various sparsity of \textcolor{red}{$f$-NeRV3} ranging from 5\% (top row) to 50 \% (bottom row).} \\
    \multicolumn{5}{l}{\small PFNR, c=50.0\%, sparsity of \textcolor{red}{$f$-NeRV3=50.0\%}.} \\
   
    \end{tabular}
    }
    \caption{Various sparsity of $f$-NeRV3 ranging from 0.05 \% (top row) to 50.0 \% (bottom row) on the UVG17 dataset.}
    \label{fig:fmap_sparsity_uvg17}
\end{figure*}

%% file: materials/app_table_uvg17_psnr_bitwise.tex
\begin{table}[ht]
\centering
\small
\caption{\small PFNR ($f$-NeRV2) of Quantization + Compression on the UVG17 Video Sessions with average PSNR/MS-SSIM and Backward Transfer (BTW). Note that bit is the bit length used to represent parameter value.}

\resizebox{\textwidth}{!}{
\renewcommand{\arraystretch}{1.2}
\begin{tabular}{lccccccccccccccccccc}
\toprule 

\multicolumn{1}{c}{\multirow{2}{*}{\textbf{Method}}} & \multicolumn{17}{c}{\textbf{Video Sessions}} & \multirow{2}{*}{\thead{\textbf{Avg. * / } \\ \textbf{BWT}}}  \\
\cline{2-18}
& \textbf{1} & \textbf{2} & \textbf{3} & \textbf{4} & \textbf{5} & \textbf{6} & \textbf{7} & \textbf{8} & \textbf{9} & \textbf{10} & \textbf{11} & \textbf{12} & \textbf{13} & \textbf{14} & \textbf{15} & \textbf{16} & \textbf{17} \\ \midrule 

% c = 10.0 %, init
\textbf{c = $10.0 \%$}, & & PSNR & & & & & & & & & & & & & & & &  \\ 
\midrule 
bit=4  & 17.73 & 30.45 & 30.74 & 30.56 & 26.15 & 23.29 & 23.16 & 19.75 & 23.80 & 26.19 & 19.53 & 20.36 & 24.99 & 26.28 & 24.61 & 23.86 & 29.95 & 24.79 / -0.80   \\ 
bit=8  & 28.21 & 33.56 & 31.92 & 33.66 & 29.97 & 23.98 & 34.35 & 24.78 & 23.95 & 35.07 & 19.70 & 22.02 & 29.57 & 26.58 & 24.79 & 24.09 & 31.34 & 28.09 / -0.01   \\ 
bit=16 & 28.31 & 33.57 & 31.92 & 33.67 & 29.98 & 23.99 & 34.39 & 24.80 & 23.94 & 35.08 & 19.70 & 22.03 & 29.56 & 26.57 & 24.79 & 24.10 & 31.35 & 28.10 / ~0.00 \\ 
bit=32 & 28.31 & 33.57 & 31.92 & 33.67 & 29.98 & 23.99 & 34.39 & 24.80 & 23.94 & 35.08 & 19.70 & 22.03 & 29.56 & 26.57 & 24.79 & 24.10 & 31.35 & 28.10 / ~0.00 \\ 

\midrule 
 & & MS-SSIM &  & & & & & & & & & & & & & & &  \\ \midrule 
bit=4  & 0.60 &	0.94 & 0.90 & 0.93 & 0.87 &	0.80 & 0.89 & 0.73 & 0.79 &	0.89 & 0.69 & 0.76 & 0.87 & 0.86 & 0.82 & 0.83 & 0.95 &	0.83 / -0.02  \\ 
bit=8  & 0.92 &	0.97 & 0.90 & 0.94 & 0.90 &	0.81 & 0.98 & 0.86 & 0.79 &	0.93 & 0.69 & 0.79 & 0.91 &	0.87 & 0.82 & 0.84 & 0.96 &	0.88 / ~0.00 \\ 
bit=16 & 0.92 &	0.97 & 0.90 & 0.94 & 0.90 &	0.81 & 0.98 & 0.86 & 0.79 &	0.93 & 0.69 & 0.79 & 0.91 &	0.87 & 0.82 & 0.84 & 0.96 &	0.88 / ~0.00  \\ 
bit=32 & 0.92 &	0.97 & 0.90 & 0.94 & 0.90 &	0.81 & 0.98 & 0.86 & 0.79 &	0.93 & 0.69 & 0.79 & 0.91 &	0.87 & 0.82 & 0.84 & 0.96 &	0.88 / ~0.00  \\ 
%\midrule 
\bottomrule 

& & & & & & & & & & & & & & & & & &  \\ 
& & & & & & & & & & & & & & & & & &  \\ 

% c = 30.0 %, init
\toprule  
\textbf{c = $30.0 \%$}, & & PSNR &  & & & & & & & & & & & & & & &  \\ 
\midrule 
bit=4  & 10.96 & 27.43 & 24.65 & 25.04 & 24.05 & 22.92 & 16.24 & 23.08 & 23.91 & 33.51 & 20.10 & 22.47 & 28.79 & 26.58 & 24.94 & 24.87 & 31.61 & 24.19 / -2.13 \\ 
\textbf{bit=8}  & \textbf{31.72} & \textbf{35.81} & \textbf{32.95} & \textbf{35.11} & \textbf{31.22} & \textbf{24.81} & \textbf{35.82} & \textbf{25.84} & \textbf{24.84} & \textbf{35.76} & \textbf{20.49} & \textbf{22.79} & \textbf{30.40} & \textbf{27.37} & \textbf{25.52} & \textbf{25.40} & \textbf{32.69} & \textbf{29.33} / \textbf{-0.02} \\ 
bit=16 & 32.01 & 35.84 & 32.97 & 35.17 & 31.24 & 24.82 & 36.01 & 25.85 & 24.83 & 35.76 & 20.50 & 22.79 & 30.40 & 27.37 & 25.52 & 25.40 & 32.70 & 29.36 / ~0.00 \\
bit=32 & 32.01 & 35.84 & 32.97 & 35.17 & 31.24 & 24.82 & 36.01 & 25.85 & 24.83 & 35.76 & 20.50 & 22.79 & 30.40 & 27.37 & 25.52 & 25.40 & 32.70 & 29.36 / ~0.00 \\
\midrule 
 & & MS-SSIM &  & & & & & & & & & & & & & & &  \\  \midrule 
bit=4  & 0.47 &	0.92 & 0.85 & 0.91 & 0.84 &	0.81 & 0.74 & 0.82 & 0.80 &	0.93 & 0.72 & 0.81 & 0.91 &	0.88 & 0.83 & 0.86 & 0.96 &	0.83 / -0.04  \\ 
bit=8  & 0.97 & 0.98 & 0.92 & 0.95 & 0.92 &	0.84 & 0.98 & 0.88 & 0.82 &	0.93 & 0.73 & 0.82 & 0.92 &	0.89 & 0.84 & 0.87 & 0.97 &	0.90 / ~0.00  \\ 
bit=16 & 0.97 & 0.98 & 0.92 & 0.95 & 0.92 &	0.84 & 0.98 & 0.88 & 0.82 &	0.93 & 0.73 & 0.82 & 0.92 &	0.89 & 0.84 & 0.87 & 0.97 &	0.90 / ~0.00  \\ 
bit=32 & 0.97 & 0.98 & 0.92 & 0.95 & 0.92 &	0.84 & 0.98 & 0.88 & 0.82 &	0.93 & 0.73 & 0.82 & 0.92 &	0.89 & 0.84 & 0.87 & 0.97 &	0.90 / ~0.00  \\ 
%\midrule 
\bottomrule

& & & & & & & & & & & & & & & & & &  \\ 
& & & & & & & & & & & & & & & & & &  \\ 

% c = 50.0 %, init
\toprule  
\textbf{c = $50.0 \%$}, & & PSNR & & & & & & & & & & & & & & & &  \\ 
\midrule 
bit=4  & 7.19  & 23.92 & 20.83 & 24.71 & 24.77 & 21.91 & 28.55 & 23.03 & 23.33 & 32.96 & 19.22 & 21.80 & 25.86 & 22.00 & 23.85 & 22.82 & 29.59 & 22.92 / -3.91 \\ 
bit=8  & 34.03 & 37.08 & 33.20 & 35.47 & 30.86 & 24.71 & 34.34 & 24.78 & 24.73 & 35.64 & 20.33 & 22.65 & 29.78 & 27.04 & 25.17 & 25.17 & 32.39 & 29.26 / -0.03 \\ 
bit=16 & 34.49 & 37.13 & 33.21 & 35.50 & 30.87 & 24.72 & 34.36 & 24.79 & 24.73 & 35.65 & 20.33 & 22.65 & 29.78 & 27.05 & 25.18 & 25.18 & 32.39 & 29.29 / ~0.00 \\
bit=32 & 34.49 & 37.13 & 33.21 & 35.50 & 30.87 & 24.72 & 34.36 & 24.79 & 24.73 & 35.65 & 20.33 & 22.65 & 29.78 & 27.05 & 25.18 & 25.18 & 32.39 & 29.29 / ~0.00 \\
\midrule 

 & & MS-SSIM & & & & & & & & & & & & & & & &  \\ \midrule 
bit=4  & 0.25 &	0.87 & 0.79 & 0.87 & 0.85 &	0.78 & 0.95 & 0.81 & 0.79 &	0.92 & 0.68 & 0.78 & 0.88 &	0.80 & 0.81 & 0.82 & 0.94 &	0.80 / -0.07   \\ 
bit=8  & 0.98 &	0.99 & 0.92 & 0.95 & 0.92 &	0.83 & 0.98 & 0.86 & 0.81 &	0.93 & 0.72 & 0.82 & 0.91 &	0.88 & 0.84 & 0.87 & 0.97 &	0.89 / ~0.00  \\ 
bit=16 & 0.98 & 0.99 & 0.92 & 0.95 & 0.92 &	0.83 & 0.98 & 0.86 & 0.81 &	0.93 & 0.72 & 0.82 & 0.91 &	0.88 & 0.84 & 0.87 & 0.97 &	0.89 / ~0.00  \\ 
bit=32 & 0.98 & 0.99 & 0.92 & 0.95 & 0.92 &	0.83 & 0.98 & 0.86 & 0.81 &	0.93 & 0.72 & 0.82 & 0.91 &	0.88 & 0.84 & 0.87 & 0.97 &	0.89 / ~0.00  \\ 
%\midrule 
\bottomrule

& & & & & & & & & & & & & & & & & &  \\ 
& & & & & & & & & & & & & & & & & &  \\ 

% c = 70.0 %, init
\toprule  
\textbf{c = $70.0 \%$}, & & PSNR & & & & & & & & & & & & & & & &  \\ 
\midrule 
bit=4 & 7.05 &	15.54 &	16.27 &	18.14 &	25.63 &	21.95 &	24.75 &	20.50 &	20.97 &	17.84 &	18.34 &	19.18 &	23.49 &	14.20 &	22.55 &	22.04 &	17.19 &	19.15/ -3.48  \\ 
bit=8 & 35.52 &	36.43 &	32.08 &	32.14 &	28.66 &	23.35 & 30.61 &	22.86 &	23.18 &	34.89 &	19.08 &	21.29 &	21.86 &	25.86 &	24.13 &	23.47 &	30.30 &	27.75 /	-0.03 \\ 
bit=16 & 36.02 & 36.50 & 32.09 & 32.15 & 28.67 & 23.35 & 30.63 & 22.86 & 23.18 & 34.90 & 19.08 & 21.30 & 27.87 & 25.86 & 24.12 & 23.47 & 30.34 & 27.79 / ~0.00 \\ 
bit=32 & 36.02 & 36.50 & 32.09 & 32.15 & 28.67 & 23.35 & 30.63 & 22.86 & 23.18 & 34.90 & 19.08 & 21.30 & 27.87 & 25.86 & 24.12 & 23.47 & 30.34 & 27.79 / ~0.00 \\

\midrule 
 & & MS-SSIM &  & & & & & & & & & & & & & & &  \\ \midrule 
bit=4  & 0.28 &	0.56 & 0.68 & 0.76 & 0.82 &	0.74 & 0.88 & 0.69 & 0.71 &	0.75 & 0.61 & 0.71 & 0.79 &	0.55 & 0.75 & 0.78 & 0.88 &	0.70 / -0.08 \\ 
bit=8  & 0.98 & 0.98 & 0.91 & 0.93 & 0.87 &	0.78 & 0.96 & 0.77 & 0.77 &	0.93 & 0.65 & 0.77 & 0.89 &	0.85 & 0.80 & 0.82 & 0.95 &	0.86 / ~0.00  \\ 
bit=16 & 0.99 & 0.98 & 0.91 & 0.93 & 0.87 &	0.78 & 0.96 & 0.77 & 0.77 &	0.93 & 0.66 & 0.77 & 0.89 &	0.85 & 0.80	& 0.82 & 0.95 &	0.86 / ~0.00  \\  
bit=32 & 0.99 & 0.98 & 0.91 & 0.93 & 0.87 &	0.78 & 0.96 & 0.77 & 0.77 &	0.93 & 0.66 & 0.77 & 0.89 &	0.85 & 0.80	& 0.82 & 0.95 &	0.86 / ~0.00  \\ 
%\midrule 
\bottomrule

\end{tabular}
}
\label{table:uvg17_quant_app}
\end{table}